\documentclass[a4paper, 12pt, twoside, oldfontcommands]{memoir}
\usepackage[inner=37mm, outer=30mm, bottom=37mm, top=37mm]{geometry}
\usepackage[T1]{fontenc}
\usepackage{mytitle}
\usepackage{amsmath}
\usepackage{amssymb}
\usepackage{amsthm}
\usepackage{bm} 
\usepackage[dvipsnames]{xcolor}
\usepackage{makecell}
\usepackage{multirow}
\usepackage{adjustbox}
\usepackage{booktabs}
\usepackage{pdflscape}

\usepackage{imakeidx}
\makeindex[columns=2, title=Index, intoc]

\usepackage{xurl}
\usepackage[linktocpage]{hyperref}
\hypersetup{
colorlinks=true,
linkcolor=Green,
citecolor=NavyBlue
}

\usepackage{natbib}
\usepackage{bibentry}

\usepackage{todonotes}  
\usepackage{emptypage}
\usepackage{soul}
\usepackage[acronyms,symbols,nogroupskip]{glossaries-extra}

\usepackage{mathtools}
\usepackage{enumitem}

\usepackage[super]{nth}

\usepackage{caption}
\usepackage{subcaption}
\usepackage{tikz}
\usepackage[customcolors]{hf-tikz}
\usepackage[most]{tcolorbox}

\cftpagenumbersoff{part}

\usepackage{fancyhdr}
\renewcommand{\headrulewidth}{0.1pt} 

\newtheorem{theorem}{Theorem}
\newtheorem{definition}{Definition}
\newtheorem{proposition}{Proposition}
\newtheorem{lemma}{Lemma}

\newcommand{\bfA}{{\bf A}}

\newcommand{\bfD}{{\bf D}}
\newcommand{\bfE}{{\bf E}}
\newcommand{\bfF}{{\bf F}}

\newcommand{\bfH}{{\bf H}}
\newcommand{\bfI}{{\bf I}}
\newcommand{\bfJ}{{\bf J}}
\newcommand{\bfK}{{\bf K}}
\newcommand{\bfL}{{\bf L}}
\newcommand{\bfM}{{\bf M}}

\newcommand{\bfU}{{\bf U}}
\newcommand{\bfV}{{\bf V}}
\newcommand{\bfW}{{\bf W}}
\newcommand{\bfX}{{\bf X}}

\newcommand{\bfZ}{{\bf Z}}

\newcommand{\bfb}{{\bf b}}

\newcommand{\bfe}{{\bf e}}
\newcommand{\bfg}{{\bf g}}
\newcommand{\bfh}{{\bf h}}

\newcommand{\bfx}{{\bf x}}
\newcommand{\bfy}{{\bf y}}

\newcommand{\bfq}{{\bf q}}
\newcommand{\bfp}{{\bf p}}

\newcommand{\bfz}{{\bf z}}

\newcommand{\myodv}[2]{\frac{d #1}{d #2}}
\newcommand{\pdv}[2]{\frac{\partial #1}{\partial #2}}

\newcommand{\xk}{\mathbf{x}_u}

\newcommand{\Jskew}{\mathcal{J}}
\newcommand{\Jkskew}{\mathcal{J}_u}

\newcommand{\dsmth}{\textnormal{d}}

\newcommand{\ie}{i.e., }
\newcommand{\eg}{e.g., }
\usepackage{pifont}
\newcommand{\cmark}{\ding{51}}
\newcommand{\xmark}{\ding{55}}
 
\newcommand\gobbleone[1]{}

\definecolor{first}{HTML}{00A64F}  
\definecolor{second}{HTML}{006EB8} 
\newcommand{\one}[1]{\textcolor{first}{\bf#1}}
\newcommand{\onenobf}[1]{\textcolor{first}{#1}}
\newcommand{\two}[1]{\textcolor{second}{\bf#1}}
\newcommand{\three}[1]{{\bf#1}}
\newcommand{\four}[1]{{\bf#1}}

\definecolor{seqcolor}{HTML}{d55e00} 
\definecolor{pascalcolor}{HTML}{00A64F}
\definecolor{linkcolor}{HTML}{006EB8}
\definecolor{comcolor}{HTML}{FFC107}

\newcommand{\myparagraph}[1]{\paragraph*{#1.}}

\definecolor{neighcolor}{HTML}{F2798D}
\tikzset{style green/.style={
    set fill color=green!50!lime!60,
    set border color=white,
  },
  style red/.style={
    set fill color=white, 
    set border color=neighcolor,
    dashed,
    ultra thick
  },
  hor/.style={
    above left offset={-0.15,0.31},
    below right offset={0.15,-0.125},
    #1
  },
  ver/.style={
    above left offset={-0.15,0.4},
    below right offset={0.15,-0.15},
    #1
  }
}

\definecolor{verylightgray}{HTML}{f3f3f3} 
\newenvironment{boxedproof}[1][\proofname]{%
  \begin{tcolorbox}[breakable, enhanced, left=0pt, right=0pt, top=0pt, bottom=0pt, colback=verylightgray, colframe=verylightgray, boxrule=0pt, sharp corners]%
  \begin{proof}[#1]%
}{%
  \end{proof}%
  \end{tcolorbox}%
}

\let\origjobname\jobname
\def\jobname{refs}
\nobibliography{refs}
\let\jobname\origjobname

\newcommand{\listref}[2]{
\noindent\cite{#1}
\begin{adjustwidth}{2.5em}{0pt}
\bibentry{#1}#2
\end{adjustwidth}}

\newcommand{\listrefU}[2]{
\noindent\underline{\cite{#1}}
\begin{adjustwidth}{2.5em}{0pt}
\bibentry{#1}#2
\end{adjustwidth}}

\usepackage[cal=cm,scr=euler]{mathalpha}
\usepackage[ruled, linesnumbered]{algorithm2e}

\author{Alessio Gravina}
\title{Information propagation dynamics in Deep Graph Networks} 

\makenoidxglossaries

\newacronym{DTDG}{D-TDG}{discrete-time dynamic graph}
\newacronym{CTDG}{C-TDG}{continuous-time dynamic graph}
\newacronym{DGN}{DGN}{deep graph network}
\newacronym{GCL}{GCL}{graph convolutional layer}
\newacronym{GCN}{GCN}{graph convolutional network}
\newacronym{MPNN}{MPNN}{message passing neural network}
\newacronym{GAT}{GAT}{graph attention network}
\newacronym{GIN}{GIN}{graph isomorphism network}
\newacronym{MLP}{MLP}{multi-layer perceptron}
\newacronym{ODE}{ODE}{ordinary differential equation}
\newacronym{PDE}{PDE}{partial differential equation}
\newacronym{PE}{PE}{positional encoding}
\newacronym{SE}{SE}{structural encoding}
\newacronym{RNN}{RNN}{recurrent neural network}
\newacronym{ResNet}{ResNet}{residual neural network}
\newacronym{GRU}{GRU}{gated recurrent unit}
\newacronym{TGN}{TGN}{Temporal Graph Network}
\newacronym{MAE}{MAE}{mean absolute error}
\newacronym{MSE}{MSE}{mean squared error}
\newacronym{AUC}{AUC}{area under the ROC curve}
\newacronym{B-Acc}{B-Acc}{balanced accuracy}
\newacronym{DE-DGN}{DE-DGN}{differential-equations inspired DGN}
\newacronym{A-DGN}{A-DGN}{Antisymmetric Deep Graph Network}
\newacronym{SWAN}{SWAN}{Space-Weight ANtisymmetric Deep Graph Network}
\newacronym{H-DGN}{H-DGN}{Hamiltonian Deep Graph Network}
\newacronym{PH-DGN}{PH-DGN}{port-Hamiltonian Deep Graph Network}
\newacronym{TG-ODE}{TG-ODE}{Temporal Graph Ordinary Differential Equation}
\newacronym{CTAN}{CTAN}{Continuous-Time Graph Antisymmetric Network}
\newacronym{HMM4G}{HMM4G}{Hidden Markov Model for Dynamic Graphs}

\glsxtrnewsymbol[description={graph}]{G}{\ensuremath{\mathcal{G}}}
\glsxtrnewsymbol[description={set of nodes}]{V}{\ensuremath{\mathcal{V}}}
\glsxtrnewsymbol[description={set of edges}]{E}{\ensuremath{\mathcal{E}}}
\glsxtrnewsymbol[description={node features}]{Xf}{\ensuremath{\mathbf{X}}}
\glsxtrnewsymbol[description={edge features}]{Ef}{\ensuremath{\mathbf{E}}}
\glsxtrnewsymbol[description={number of node features}]{dn}{\ensuremath{d_n}}
\glsxtrnewsymbol[description={number of edge features}]{de}{\ensuremath{d_e}}
\glsxtrnewsymbol[description={features of node $u$}]{xu}{\ensuremath{\mathbf{x}_u}}
\glsxtrnewsymbol[description={features of the edge that links nodes $u$ and $v$}]{euv}{\ensuremath{\mathbf{e}_{uv}}}
\glsxtrnewsymbol[description={adjacency matrix}]{A}{\ensuremath{\mathbf{A}}}
\glsxtrnewsymbol[description={directed edge from $u$ to $v$}]{diruv}{\ensuremath{(u,v)}}
\glsxtrnewsymbol[description={undirected edge from $u$ to $v$}]{undiruv}{\ensuremath{\{u,v\}}}
\glsxtrnewsymbol[description={neighborhood of the node $u$}]{Nu}{\ensuremath{\mathcal{N}_u}}
\glsxtrnewsymbol[description={degree of the node $u$}]{degu}{\ensuremath{\mathbf{deg}(u)}}
\glsxtrnewsymbol[description={degree matrix}]{D}{\ensuremath{\mathbf{D}}}
\glsxtrnewsymbol[description={identity matrix}]{I}{\ensuremath{\mathbf{I}}}
\glsxtrnewsymbol[description={graph Laplacian}]{L}{\ensuremath{\mathbf{L}}}
\glsxtrnewsymbol[description={symmetric normalized graph Laplacian}]{Lsym}{\ensuremath{\mathbf{L}^{sym}}}
\glsxtrnewsymbol[description={random-walk normalized graph Laplacian}]{Lrw}{\ensuremath{\mathbf{L}^{rw}}}
\glsxtrnewsymbol[description={dynamic graph}]{G(t)}{\ensuremath{\mathcal{G}(t)}}
\glsxtrnewsymbol[description={set of nodes present in the graph at time $t$}]{V(t)}{\ensuremath{\mathcal{V}(t)}}
\glsxtrnewsymbol[description={set of edges present in the graph at time $t$}]{E(t)}{\ensuremath{\mathcal{E}(t)}}
\glsxtrnewsymbol[description={node features present in the graph at time $t$}]{Xf(t)}{\ensuremath{\mathbf{X}(t)}}
\glsxtrnewsymbol[description={edge features present in the graph at time $t$}]{Ef(t)}{\ensuremath{\mathbf{E}(t)}}
\glsxtrnewsymbol[description={graph snapshot at time $t$}]{Gt}{\ensuremath{\mathcal{G}_t}}
\glsxtrnewsymbol[description={temporal neighborhood of a node $u$ at time $t$}]{Nut}{\ensuremath{\mathcal{N}_u^t}}
\glsxtrnewsymbol[description={layer index}]{l}{\ensuremath{\ell}}
\glsxtrnewsymbol[description={number of layers}]{Layers}{\ensuremath{L}}
\glsxtrnewsymbol[description={temporal index}]{t}{\ensuremath{t}}
\glsxtrnewsymbol[description={embedding of (static) node $u$ at layer $\ell+1$}]{xul1}{\ensuremath{\bfx^{\ell+1}_u}}
\glsxtrnewsymbol[description={weight matrix}]{W}{\ensuremath{\bfW}}
\glsxtrnewsymbol[description={weight matrix}]{Vw}{\ensuremath{\bfV}}
\glsxtrnewsymbol[description={number of hidden features}]{d}{\ensuremath{d}}
\glsxtrnewsymbol[description={activation function}]{sigma}{\ensuremath{\sigma}}
\glsxtrnewsymbol[description={attention score between node $u$ and $v$}]{auv}{\ensuremath{\alpha_{uv}}}
\glsxtrnewsymbol[description={constant}]{beta}{\ensuremath{\beta}}
\glsxtrnewsymbol[description={constant}]{gamma}{\ensuremath{\gamma}}
\glsxtrnewsymbol[description={aggregation invariant function}]{bigoplus}{\ensuremath{\bigoplus_{v\in\mathcal{N}_u}}}

\glsxtrnewsymbol[description={update function}]{rhou}{\ensuremath{\rho_U}}
\glsxtrnewsymbol[description={message function}]{rhom}{\ensuremath{\rho_M}}
\glsxtrnewsymbol[description={function that combines the output of the previous Cauchy sub-problem with the current input}]{eta}{\ensuremath{\eta}}

\glsxtrnewsymbol[description={permutation invariant neighborhood aggregation function}]{Phi}{\ensuremath{\Phi}}
\glsxtrnewsymbol[description={permutation invariant  antisymmetric neighborhood aggregation function}]{Psi}{\ensuremath{\Psi}}

\glsxtrnewsymbol[description={dirichlet energy}]{energy}{\ensuremath{E}}
\glsxtrnewsymbol[description={Hadamard product}]{hadamard}{\ensuremath{\odot}}
\glsxtrnewsymbol[description={temporal embedding matrix of the nodes at time $t$}]{Ht}{\ensuremath{\bfH_t}}
\glsxtrnewsymbol[description={embedding matrix of the (static) nodes at layer $\ell$}]{Xl}{\ensuremath{\bfX^\ell}}
\glsxtrnewsymbol[description={observed node feautures associated to the graph snapshot at time step $t$}]{Xft}{\ensuremath{\overline{\bfX}_t}}
\glsxtrnewsymbol[description={the edge feautures associated to the graph snapshot at time step $t$}]{Eft}{\ensuremath{\bfE_t}}
\glsxtrnewsymbol[description={set of nodes associated to the graph snapshot at time step $t$}]{Vt}{\ensuremath{\mathcal{V}_t}}
\glsxtrnewsymbol[description={set of edges associated to the graph snapshot at time step $t$}]{Et}{\ensuremath{\mathcal{E}_t}}
\glsxtrnewsymbol[description={adjacency matrix associated to the graph snapshot at time step $t$}]{At}{\ensuremath{\bfA_t}}
\glsxtrnewsymbol[description={temporal embedding of node $u$ associated to the graph snapshot at time step $t$}]{ht}{\ensuremath{\bfh^{t}_u}}
\glsxtrnewsymbol[description={observed temporal node features of node $v$ associated to the graph snapshot at time step $t-1$}]{xvt-1}{\ensuremath{\overline{\bfx}^{t-1}_v}}
\glsxtrnewsymbol[description={temporal embedding matrix of the nodes associated to the graph snapshot at time step $t$ and layer $\ell$}]{Htell}{\ensuremath{\bfH_t^\ell}}
\glsxtrnewsymbol[description={concatenation operator}]{concatenation}{\ensuremath{\big\|}}
\glsxtrnewsymbol[description={time at which an event occurred before the current timestamp}]{t-}{\ensuremath{t^-}}
\glsxtrnewsymbol[description={temporal embedding of node $u$ at 
continuous time $t$ and layer $\ell$}]{hult}{\ensuremath{\bfh^\ell_u(t)}}
\glsxtrnewsymbol[description={observed temporal node features of node $u$ at 
continuous time $t^-_u$}]{xult}{\ensuremath{\overline{\bfx}_u(t^-_u)}}
\glsxtrnewsymbol[description={state of node $u$ at time $t$ in the DE-DGN}]{xut}{\ensuremath{\bfx_u(t)}}
\glsxtrnewsymbol[description={learnable parameters}]{theta}{\ensuremath{\theta}}

\makeindex

\begin{document}
\pagenumbering{gobble}

\setcounter{tocdepth}{3}

\maketitle
\clearpage\thispagestyle{empty}\null\newpage
\begin{flushright} 
    \thispagestyle{empty}
    \vspace*{\stretch{1}}
  \large \textit{``I still have a long way to go, but I'm already so far \\from where I used to be, and I am proud of that.''}\\ -- Node $u$ in graph $\mathcal{G}$
  {\par 
   \vspace{\stretch{3}} 
   \clearpage           
  }
\end{flushright}
\clearpage\thispagestyle{empty}\null\newpage
\chapter*{\centering Abstract}
Graphs are a highly expressive abstraction for modeling entities and their relations, such as molecular structures, social networks, and traffic networks. Deep Graph Networks (DGNs) have recently emerged as a family of deep learning models that can effectively process and learn such structured information. However, learning effective information propagation patterns within DGNs remains a critical challenge that heavily influences the model capabilities, both in the static domain and in the temporal domain (where features and/or topology evolve). 
This thesis investigates the dynamics of information propagation within DGNs for static and dynamic graphs, focusing on their design as dynamical systems.

With the aim of fostering research in this domain, at first, we review the principles underlying DGNs and their limitations in information propagation, followed by a survey of recent advantages in learning both temporal and spatial information, providing a fair performance comparison among the most popular proposed approaches.
The main challenge addressed in this thesis is the limited ability of DGNs to propagate and preserve long-term dependencies between nodes. To tackle this problem, we design principled approaches bridging non-dissipative dynamical systems with DGNs. We leverage properties of global and local non-dissipativity in both temporal and static domain, enabling  maintaining a constant information flow rate between nodes. We first exploit dynamical systems with antisymmetric constraints on both spatial and weight domains to achieve graph- and node-wise non-dissipativity. Then, we introduce a DGN that exploits port-Hamiltonian dynamics, thus defining a new message-passing scheme that balances non-dissipative long-range propagation and non-conservative behaviors.
We then tackle the task of learning complex spatio-temporal patterns from irregular and sparsely sampled data. 
Throughout this work, we provide theoretical and empirical evidence to demonstrate the effectiveness of our proposed architectures. In summary, this thesis provides a comprehensive exploration of the intersection between graphs, deep learning, and dynamical systems, providing insights and advancements for the field of graph representation learning and paving the way for more effective and versatile graph-based learning models.

\clearpage\thispagestyle{empty}\null\newpage
\chapter*{\centering List of Publications}
The following is a comprehensive list of publications (ordered by acceptance date) in which the author has contributed throughout the course of the doctoral studies. Underlined works indicate their use in this thesis.
\\

\nobibliography*
\listref{gravina_schizophrenia}{\\\textbf{Code}: \url{https://github.com/gravins/DGNs-for-schizophrenia}\\}
\listref{gravina_covid}{\\\textbf{Code}: \url{https://github.com/gravins/covid19-drug-repurposing-with-DGNs}\\}
\listrefU{gravina_adgn}{\\\textbf{Code}: \url{https://github.com/gravins/Anti-SymmetricDGN}\\\textbf{Preliminary version accepted at} DLG-AAAI’23 workshop (\url{https://deep-learning-graphs.bitbucket.io/dlg-aaai23/publications.html}) and awarded of the \textbf{Best Student Paper Award}\\}    
\listrefU{gravina_randomized_adgn}{\\\textbf{Code}: \url{https://github.com/gravins/Anti-SymmetricDGN}\\}
\listrefU{gravina_hmm4g}{\\\textbf{Code}: \url{https://github.com/nec-research/hidden_markov_model_temporal_graphs}\\}
\listref{gravina_anomaly}{\\\textbf{Code}: \url{https://github.com/JakubReha/ProvCTDG}\\}
\listrefU{gravina_dynamic_survey}{\\\textbf{Code}: \url{https://github.com/gravins/dynamic_graph_benchmark}\\}
\listrefU{gravina_tgode}{\\\textbf{Code}: \url{https://github.com/gravins/TG-ODE}\\}
\listrefU{gravina_ctan}{\\\textbf{Code}: \url{https://github.com/gravins/non-dissipative-propagation-CTDGs}\\\textbf{Preliminary version accepted at} Temporal Graph Learning Workshop @ NeurIPS 2023 (\url{https://openreview.net/forum?id=zAHFC2LNEe})}

\section*{List of Papers Under a Peer Review Process (at the time of thesis submission)}
\listrefU{gravina_swan}{\\\textbf{Code}:\url{https://github.com/gravins/SWAN}\\\textbf{Now accepted at AAAI 2025:}\\\bibentry{gravina_swan_accepted}\\}
\listrefU{gravina_phdgn}{\\\textbf{Code}:\url{https://github.com/simonheilig/porthamiltonian-dgn}\\\textbf{Now accepted at ICLR 2025:}\\\bibentry{gravina_phdgn_accepted}}

\clearpage\thispagestyle{empty}\null\newpage
\begin{KeepFromToc}
  \tableofcontents
\end{KeepFromToc}
\clearpage\thispagestyle{empty}\null\newpage
\printnoidxglossary[type=acronym,style=long3col,sort=use,title={Acronyms}]
\clearpage\thispagestyle{empty}\null\newpage
\printnoidxglossary[type=symbols,style=long3col,sort=use,title={Notation}]
\clearpage\thispagestyle{empty}\null\newpage

\pagenumbering{arabic}

\clearpage\thispagestyle{empty}\addtocounter{page}{-1}\null\clearpage
\chapter{Introduction}
Graphs, as mathematical structures, have a rich history dating back to the 18th century when Leonhard Euler laid the foundation with his solution to the K{\"o}nigsberg bridge problem\footnote{The problem consists in devising a walk through the city of K{\"o}nigsberg that would cross each of the seven bridges in the city once and only once.}~\citep{euler1741solutio,biggs1986graph}, giving birth to graph theory~\citep{graph_theory}. 
Graphs are composed of nodes (vertices), edges (links), and features (attributes). They can be defined in both the static
domain and in the temporal domain (where features and/or topology evolve). In the temporal case, graphs are referred to as dynamic graphs. Graphs, in general, serve as a powerful abstraction for representing relationships and interactions in a wide array of systems. From molecular structures and biological networks to social networks and infrastructure grids, graphs are invaluable for modeling complex systems where entities and their connections are of primary interest.

The development of graph theory over the centuries has led to significant advancements in various fields. In computer science, for example, algorithms for traversing, searching, and optimizing graphs are fundamental to data structures and network analysis~\citep{cormen}. In sociology, graph theory aids in understanding social dynamics and influence patterns~\citep{Wasserman_Faust_1994}. In biology, it helps in analyzing protein interactions networks~\citep{Barabsi2004NetworkBU}. As data becomes increasingly interconnected, the importance of graphs in representing and understanding this complexity continues to grow.

Parallel to the development of graph theory, the study of dynamical systems has been crucial in understanding how processes evolve over time. Dynamical systems' theory provides a framework for modeling and analyzing the behavior of complex systems, which can be deterministic or stochastic, linear or nonlinear. This theory is pivotal in various scientific disciplines, including physics, biology, economics, and engineering, offering insights into the stability, chaos, and long-term behavior of systems~\citep{glendinning_1994}.

In more recent years, the intersection of graph theory, dynamical systems, and machine learning has gained significant attention. Deep Graph Networks (DGNs) have emerged as a powerful paradigm for learning from graph-structured data. DGNs leverage the expressiveness of graphs to capture relationships and dependencies within data, enabling advanced applications in biology, social science, human mobility, and more~\citep{MPNN, bioinformatics, gravina_schizophrenia, gravina_covid, social_network, google_maps}.
Similarly, differential-equation based neural architectures have become a powerful paradigm for learning in different fields, proving that neural networks and differential equations are two sides of the same coin \citep{HaberRuthotto2017, neuralODE, chang2018antisymmetricrnn}. 

This thesis investigates novel dynamics of information propagation in DGNs for  both static and dynamic graphs, with a particular focus on integrating concepts from dynamical systems into DGNs. By merging the rich histories and powerful tools of 
dynamical systems and graphs, this research aims to advance the state-of-the-art in graph representation learning. 
Indeed, by framing DGNs as dynamical systems, we can utilize the mathematical rigor and insights from this field to enhance information flow and propagation stability within the graphs, uncovering complex information propagation patterns that are beyond the reach of current literature methodologies.
As an example, a critical challenge in DGNs lies in the effective propagation of information through the graph. Traditional models often struggle with long-range dependencies, where information needs to travel across distant nodes. This is where the principles of dynamical systems can play a transformative role. 

With the aim of advancing the state-of-the-art in graph representation learning, we propose methodologies not only to enhance the theoretical understanding of DGNs but also demonstrate practical improvements through empirical studies. This thesis sets the stage for future innovations in creating more effective and versatile graph-based learning models, capable of tackling the complexities of real-world data.

\section{Objectives}

Recent progress in research on DGNs has led to a maturation of the domain of learning on graphs. As a result, the last few years have witnessed a surge of works, especially on dynamic graphs, leading to a fragmented and scattered literature with respect to model formalization, empirical setups and performance benchmarks.
This aspect very much motivated us to look into a systematization of the literature which does not only look at surveying the existing works, but also actively promotes the identification of shared benchmarks and empirical protocols for the fair evaluation of dynamic graph models
, with the ultimate goal of fostering research in the field of graph representation learning.
\\

Despite the growth of this research field, there are still important challenges that are yet unsolved. 
Effective information diffusion within graphs is a critical open issue that heavily influences graph representation learning.
Therefore, the main objective of this thesis is to investigate novel dynamics of information propagation in DGNs in a principled manner by integrating concepts from dynamical systems into DGNs. Additionally, we aim to propose general solutions that offer an inductive bias easily applicable across the complexities of real-world data, rather than specialized solutions limited to specific scenarios.  
Inspired by the premises in the previous section, this dissertation aims to explore the potential of differential-equations inspired DGNs (DE-DGNs), particularly regarding their abilities to learn \emph{long-term dependencies} between nodes and complex spatio-temporal patterns from \emph{irregular and sparsely} sampled data. 

We first address the limited ability to facilitate effective information flow between distant nodes. This calls for principled approaches that control and regulate the degree of propagation and dissipation of information throughout the neural flow. Indeed, classical DGNs are typically limited in their ability to propagate and preserve long-term dependencies between nodes, 
which reduces their effectiveness, especially for predictive problems that require capturing interactions at various, potentially large, radii. Afterward, we move to the problem of effective learning of irregularly sampled dynamic graphs, since modern deep learning approaches for dynamic graphs generally assume regularly sampled temporal data, which is far from realistic. 
Real-world complex problems necessitate novel methods that transcend this common assumption, addressing mutable relational information and dealing with irregularly and severely under-sampled data.

In conclusion, with these objectives, we aim to propose novel architectural biases to deepen the theoretical understanding of DGNs, thereby developing more effective and versatile graph-based learning models.

\section{Thesis Outline}
In the following, we provide a brief outline regarding the content of this thesis.

In the first part of this thesis, we present background concepts to establish a clear understanding of the thesis's domain:
\begin{itemize}
    \item In Chapter~\ref{ch:preliminaries}, we introduce fundamental definitions and concepts pivotal to this thesis. We review the core principles of graph theory and propose a unified formalization for dynamic graphs, we explore the world of differential equations and dynamical systems, and, finally, we provide essential background on Deep Graph Networks (DGNs) and their challenges.
    
    \item In Chapter~\ref{ch:learning_dyn_graphs}, we survey state-of-the-art approaches in representation learning for dynamic graphs, building on the unified formalism defined in the previous chapter. We present a fair performance comparison of popular DGNs for dynamic graphs using a standardized, reproducible experimental setup. Additionally, we offer a curated selection of datasets as benchmarks for future research in the graph learning community.
\end{itemize}

In the second part of this thesis, we tackle the problem of non-dissipative propagation for static graphs:
\begin{itemize}
    \item In Chapter~\ref{ch:antisymmetry}, we address the primary challenge of \emph{long-range propagation} in graph representation learning. We draw on the concepts of neural differential equations to develop differential-equations inspired DGNs capable of non-dissipative propagation between nodes, through the use of antisymmetric constraints. Thus, we introduce an antisymmetric weight parameterization which allows for node-wise non-dissipative behavior. Then, we extend such concept to achieve both graph- and node-wise non-dissipative behaviors thanks to space and weight antisymmetric parameterization, thus guaranteeing a constant information flow rate. While doing so, we also introduce new synthetic benchmarks for assessing long-range propagation capabilities. 
    
    \item In Chapter~\ref{ch:phdgn}, we design the information flow within a static graph as a port-Hamiltonian system. Thus, we introduce a new message-passing scheme capable of balancing non-dissipative long-range propagation and non-conservative behaviors for improved effectiveness in specific tasks. We provide theoretical guarantees that ensure information conservation over time when pure Hamiltonian dynamic is employed.  
\end{itemize}

In the third part of this thesis, we discuss space and time propagation for dynamic graphs:
\begin{itemize}
    \item In Chapter~\ref{ch:tgode}, we tackle the problem of learning complex the spatio-temporal patterns of dynamic graphs under the real-world assumption of \emph{irregularly and severely under-sampled} data, thus overcoming the common assumption of dealing with regularly sampled temporal graph snapshots. Therefore, we introduce a general framework designed through the lens of neural differential equations for graphs. While doing so, we also introduce new benchmarks of synthetic and real-world scenarios for evaluating forecasting models on irregularly sampled dynamic graphs.
    
    This contribution has been developed during a visiting period at the Swiss AI Lab IDSIA (Istituto Dalle Molle di Studi sull'Intelligenza Artificiale) in Lugano, Switzerland.
    
    \item In Chapter~\ref{ch:ctan}, we focus on the problem of long-range propagation within dynamic graphs. To address this, we introduce a novel differential equation method for scalable long-range propagation. We establish theoretical conditions for achieving stability and non-dissipation by employing antisymmetric weight parameterization, which is the key factor for modeling long-range spatio-temporal interactions. We also present novel benchmark datasets specifically designed to assess the ability of DGNs to propagate information over long spatio-temporal distances within dynamic graphs.

    This contribution has been developed during an internship at Huawei Technologies, Munich Research Center, in Munich, Germany.
\end{itemize}

In Chapter~\ref{ch:conclusions}, we summarize the content of this dissertation and discuss future research directions.
    
Finally, in Appendix~\ref{app:suppl_ch3}, \ref{app:suppl_ch4}, \ref{app:suppl_ch5}, \ref{app:suppl_ch6}, and \ref{app:suppl_ch7} we provide additional details such as datasets description and statistics and the explored hyperparameter spaces. 
In Appendix~\ref{app:suppl_ch8}, we propose a fully probabilistic approach for modeling information propagation within dynamic graphs, thereby challenging the prevailing notion that only neural architectures are suitable for this task. This last contribution has been developed in collaboration with NEC Laboratories Europe, Heidelberg, Germany.

\section{Origin of the Chapters}
Part of the work presented in this thesis resulted in the following papers, either published or in the peer review process.

\nobibliography*
\begin{itemize}
    \item Section~\ref{sec:graphs}, Section~\ref{sec:static_dgn_fundamentals}, and Chapter~\ref{ch:learning_dyn_graphs}:
    \\\bibentry{gravina_dynamic_survey}
    \item Section~\ref{sec:ADGN}:
    \\ \bibentry{gravina_adgn}
    \\ \bibentry{gravina_randomized_adgn}
    \item Section~\ref{sec:SWAN}:
    \\ \bibentry{gravina_swan} 
        \\\textbf{Now accepted at AAAI 2025:} \bibentry{gravina_swan_accepted}
    \item Chapter~\ref{ch:phdgn}:
    \\ \bibentry{gravina_phdgn} 
        \\\textbf{Now accepted at ICLR 2025:} \bibentry{gravina_phdgn_accepted}
    \item Chapter~\ref{ch:tgode}:
    \\ \bibentry{gravina_tgode}
    \item Chapter~\ref{ch:ctan}:
    \\ \bibentry{gravina_ctan}
    \item Appendix~\ref{sec:hmm4g}:
    \\ \bibentry{gravina_hmm4g}
\end{itemize}

\part{Background}
\chapter{Preliminaries}\label{ch:preliminaries}
In this chapter, we delineate basic definitions and techniques that will be used throughout the rest of the manuscript.  In doing so, it is assumed that the reader is
familiar with linear algebra and the fundamental machine learning concepts, including hyperparameters, multi-layer perceptrons, and activation functions.

We first explore the fundamental principles and definitions of graph theory as well as we propose a coherent formalization of the dynamic graph domain, unifying different definitions and formalism gathered from the literature.

Subsequently, we delve into the realm of differential equations and dynamical systems. We explore the concept of initial valued problems and discuss their link with neural architectures. Furthermore, we define dynamical systems governed by Hamilton's equations and provide an overview of various discretization techniques for solving differential equations
. 

Lastly, we survey the domain of representation learning for graphs under our unified formalism, providing the necessary background for understanding deep graph networks
and their plights.
\section{Graphs}\label{sec:graphs}

In this section, we introduce fundamental concepts about graphs, taken from graph theory~\citep{graph_theory} and deep learning for dynamic graphs~\citep{gravina_dynamic_survey}
, which will be used throughout the rest of this thesis. 
\\

From a general perspective, a \textbf{graph} is a highly flexible mathematical structure for representing systems of relationships and interactions among some entities of interest. Such flexibility allows graphs as the primary choice for modelling complex systems across various fields. 
In biological sciences, graphs represent concepts such as molecules and proteins by using nodes to represent atoms and edges to represent chemical bonds. In biology, graphs have been used to model protein-protein interaction networks, thus leading to improved understanding of functional relationships between proteins and effective therapies. In social sciences, graphs model social networks, with nodes representing individuals and edges representing social connections between them, providing insights about social structure and dynamics. As a last example, graphs have been successfully applied to model road networks and their traffic. 
\\

As demonstrated by the examples provided, graphs can be categorized as either \emph{static} or \emph{dynamic}. Static graphs remain fixed and unchanged, as is common for biological sciences. Dynamic graphs, on the other hand, involve the evolution over time of the entities and their relations, such as the continual activities and interactions between members of social networks.

\subsection{Static Graphs}\label{sec:static_graph_notation}
We start by formally defining the concept of a graph.
\begin{definition}[Graph]
A \textbf{(static) graph}\index{graph} is a tuple $\gls*{G}=(\gls*{V}, \gls*{E}, \gls*{Xf}, \gls*{Ef})$ defined by the nonempty set $\mathcal{V}$ of \textbf{nodes}\index{node} (also referred to as \textbf{vertices}\index{vertex|see {node}}), and by the set $\mathcal{E}$ of \textbf{edges}\index{edge} (also called \textbf{links}\index{link|see {edge}} or \textbf{arcs}\index{arc|see {edge}}). \textbf{Node features}\index{node features} (also known as node representation\index{node representation|see {node features}}) are represented as a matrix $\mathbf{X} \in \mathbb{R}^{|\mathcal{V}|\times \gls*{dn}}$, where $d_n$ is the number of available features. The $u$-th row of $\mathbf{X}$ is denoted as $\gls*{xu}$ and represents a single node features. Similarly, \textbf{edge features}\index{edge features} are represented as the matrix $\mathbf{E} \in \mathbb{R}^{|\mathcal{E}| \times \gls*{de}}$, where $d_e$ is the number of edge features, and $\gls*{euv}$ denotes the feature vector of the edge that link node $u$ and $v$. 
\end{definition}

In general terms, nodes represent interacting entities, whereas edges denote connections between pairs of nodes. In many practical scenarios, nodes and edges are often enriched with additional attributes, here identified by $\mathbf{X}$ and $\mathbf{E}$. The topological arrangement of nodes and edges is called \emph{network topology}\index{network topology} (or \emph{topology}\index{topology|see {network topology}}). 

The structural information expressed by $\mathcal{E}$ can also be encoded into an \textbf{adjacency matrix}\index{adjacency matrix}, $\gls*{A}$, which is a square $|\mathcal{V}| \times |\mathcal{V}|$ matrix where each element $\mathbf{A}_{uv} \in \{0,1\}$ is $1$ if an edge connects the nodes $u$ and $v$, and it is $0$ otherwise. Depending on the structure of the adjacency matrix (consequently the structure of $\mathcal{E}$), a graph is \textit{directed} or \textit{undirected}.

\begin{definition}[Directed/Undirected graph]
    A graph $\mathcal{G}=(\mathcal{V}, \mathcal{E}, \mathbf{X}, \mathbf{E})$ is \textbf{directed}\index{directed graph|see {graph}}\index{graph!directed} when node pairs are ordered, \ie $\mathcal{E} \subseteq \{\gls*{diruv} \, | \, u,v \in \mathcal{V}\}$. 
    Inversely, a graph is \textbf{undirected}\index{undirected graph|see {graph}}\index{graph!undirected} when edges are unordered, \ie $\mathcal{E} \subseteq \{\gls*{undiruv} \, | \, u,v \in \mathcal{V}\}$.
\end{definition}

\begin{figure}[ht]
\centering 
\includegraphics[width=0.28\textwidth]{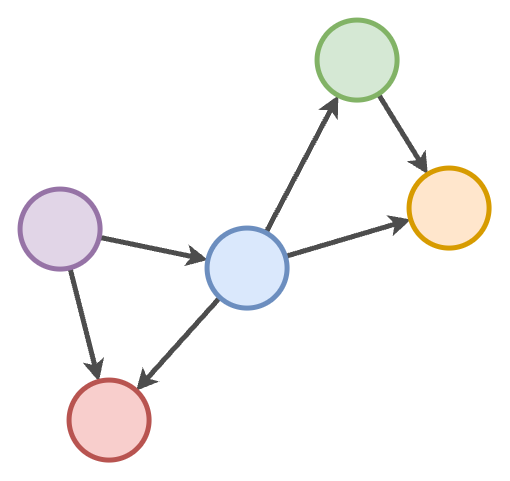}
\hspace{0.1cm}
\includegraphics[width=0.28\textwidth]{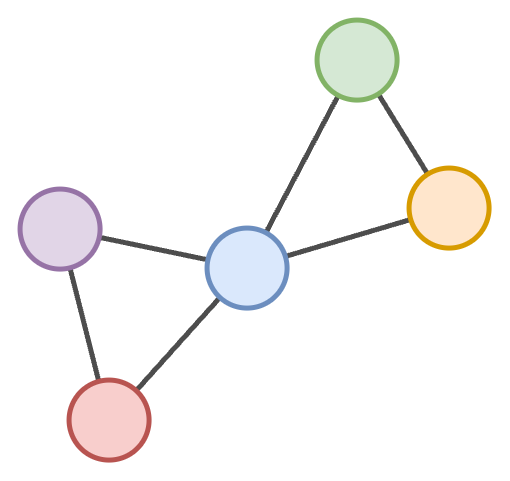}
\hspace{0.1cm}
\includegraphics[width=0.28\textwidth]{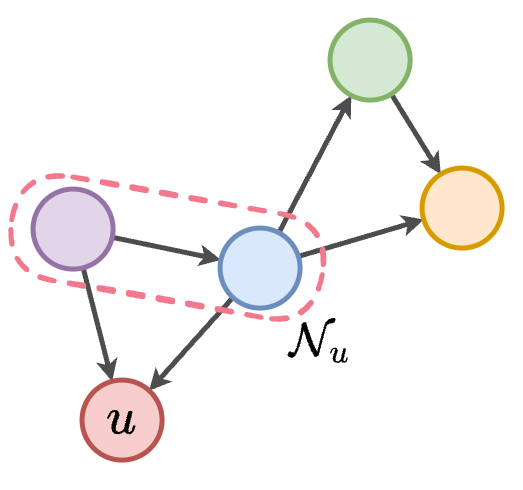}
\\
\subfloat[]{$\begin{pmatrix}
    0 & 0 & 0 & 0 & 0\\
    1 & 0 & 1 & 0 & 0\\
    1 & 0 & 0 & 1 & 1\\
    0 & 0 & 0 & 0 & 1\\
    0 & 0 & 0 & 0 & 0\\
\end{pmatrix}$}
\hspace{1.5cm}
\subfloat[]{$\begin{pmatrix}
    0 & 1 & 1 & 0 & 0\\
    1 & 0 & 1 & 0 & 0\\
    1 & 1 & 0 & 1 & 1\\
    0 & 0 & 1 & 0 & 1\\
    0 & 0 & 1 & 1 & 0\\
\end{pmatrix}$}
\hspace{1.5cm}
\subfloat[]{$\left(\begin{array}{ccccc}    \tikzmarkin[ver=style red]{col 1}    0 & 0 & 0 & 0 & 0\\  1 & 0 & 1 & 0 & 0\\   1 & 0 & 0 & 1 & 1\\    0 & 0 & 0 & 0 & 1\\    0\tikzmarkend{col 1} & 0 & 0 & 0 & 0     \end{array}\right)$}    

\caption{(a) A directed graph. (b) An undirected graph. (c) The neighborhood of node $u$. At the bottom, the adjacency matrix of the corresponding graph. The neighborhood of the node $u$ is depicted with a dotted line.} \label{fig:dir_undir_neigh}
\end{figure}

In other words, a graph is directed when the adjacency matrix is symmetric, undirected otherwise. We visually exemplify this concept in Figure~\ref{fig:dir_undir_neigh}. Undirected graphs are useful for representing mutual connections, such as chemical bonds in molecules and mutual friendships in social networks. Conversely, directed graphs are apt for situations where direction conveys extra information, such as indicating the flow of traffic in traffic networks.
\\

Another fundamental concept is that of the neighborhood of a node. We denote the neighborhood (or adjacency set) of a node $u \in \mathcal{V}$ as the set of nodes that have at least an ordered edge with tail $u$. Formally,
\begin{definition}[Neighborhood]
Let $\mathcal{G}=(\mathcal{V}, \mathcal{E}, \mathbf{X}, \mathbf{E})$  be a directed graph. The \textbf{neighborhood}\index{neighborhood} of a node $u \in \mathcal{V}$ is the set $\gls*{Nu} = \{v\in\mathcal{V} \mid (v, u) \in \mathcal{E}\}$.    
\end{definition}

Therefore, 
the $u$-th column of the adjacency matrix indicates the neighbors of node $u$, as it contains the set of ordered edges with node $u$ as the destination (\ie the incoming edges). Similarly, the $u$-th row of the adjacency matrix identifies the nodes to which $u$ appears as a neighbor, as it comprises the set of ordered edges with $u$ as the starting node (\ie the outgoing edges).
A visual representation of the neighborhood of a node is shown in Figure~\ref{fig:dir_undir_neigh}.

The number of incoming and outgoing edges of a node $u$ define the \emph{in-degree} and \emph{out-degree} of the node, respectively, \ie
\begin{definition}[In-degree/Out-degree]
    Let $\mathcal{G}=(\mathcal{V}, \mathcal{E}, \mathbf{X}, \mathbf{E})$  be a directed graph. The \textbf{in-degree}\index{in-degree} of a node $u \in \mathcal{V}$ is $$deg_{in}(u)=|\{v\in\mathcal{V} \mid (v, u) \in \mathcal{E}\}|= \sum_{v\in\mathcal{V}}\mathbf{A}_{uv}.$$ The \textbf{out-degree}\index{out-degree} is $$deg_{out}(u)=|\{v\in\mathcal{V} \mid (u, v) \in \mathcal{E}\}|=\sum_{v\in\mathcal{V}}\mathbf{A}_{vu}.$$
\end{definition}

Accordingly, the in-degree of a node $u$ can be computed by summing the values of the $u$-th column from the adjacency matrix, and it corresponds to the dimension of the neighborhood of $u$. The out-degree can be computed by summing the $u$-th row. In undirected graphs, since the $u$-th row and column are identical, the in-degree and out-degree are equal. In the following, we will primarily focus on the in-degree of nodes. For simplicity, we will refer to the \emph{degree}\index{degree|see {in-degree}} as the in-degree, denoted as $\gls*{degu}$. The \textbf{degree matrix}\index{degree matrix} is a diagonal matrix whose entries are $\mathbf{D}_{uu} = \mathbf{deg}(u)$.

From the notions of adjacency matrix and degree of a node it is possible to derive another matrix representation of a graph, which is the one of \emph{Laplacian matrix} (or graph Laplacian).

\begin{definition}[Graph Laplacian]
Let $\mathcal{G}=(\mathcal{V}, \mathcal{E}, \mathbf{X}, \mathbf{E})$  be a graph with adjacency matrix $\mathbf{A}$ and degree matrix $\gls*{D}$. The \textbf{graph Laplacian}\index{graph laplacian} is defined as $$\gls*{L}=\mathbf{D} - \mathbf{A}.$$    
\end{definition}

The Laplacian matrix provides insights into the connectivity and behavior of nodes within the graph. It is particularly valuable in spectral graph theory~\citep{chung1997spectral}, where eigenvalues and eigenvectors of the Laplacian encode information about graph topology and connectivity patterns. 

A node with a large degree results in a large diagonal entry in the Laplacian matrix, thus dominating the matrix properties. To mitigate this, normalization techniques can be applied to balance the influence of such nodes with that of others in the graph. We report two normalization strategies, which are symmetric and random walk normalizations.

\begin{definition}[Symmetric normalized graph Laplacian]\label{def:sym_laplacian}
Let $\mathcal{G}=(\mathcal{V}, \mathcal{E}, \mathbf{X}, \mathbf{E})$  be a graph with adjacency matrix $\mathbf{A}$ and degree matrix $\mathbf{D}$. The \textbf{symmetric normalized graph Laplacian}\index{symmetric normalized laplacian|see {graph laplacian}}\index{graph laplacian!symmetric normalized} is defined as 
\begin{align*}
\gls*{Lsym} &=\mathbf{D}^{-\frac{1}{2}}\mathbf{L}\mathbf{D}^{-\frac{1}{2}} \\
                &= \mathbf{I} -  \mathbf{D}^{-\frac{1}{2}}\mathbf{A}\mathbf{D}^{-\frac{1}{2}},
\end{align*}
with $\gls*{I}$ the identity matrix.
\end{definition}

\begin{definition}[Random-Walk normalized graph Laplacian]
Let $\mathcal{G}=(\mathcal{V}, \mathcal{E}, \mathbf{X}, \mathbf{E})$  be a graph with adjacency matrix $\mathbf{A}$ and degree matrix $\mathbf{D}$. The \textbf{Random-Walk normalized graph Laplacian}\index{random-walk normalized laplacian|see {graph laplacian}}\index{graph laplacian!random-walk normalized} is defined as 
\begin{align*}
    \gls*{Lrw} &=\mathbf{D}^{-1}\mathbf{L} \\
                    &= \mathbf{I} -  \mathbf{D}^{-1}\mathbf{A},
\end{align*}
with $\mathbf{I}$ the identity matrix.
\end{definition}

We observe that by replacing $\mathbf{L}$ with $\mathbf{A}$, we can apply such techniques to the adjacency matrix, with the effect of scaling the matrix eigenvalues. 
\\

Now that we have provided the reader with some useful definitions about graphs, we observe that, depending on the constraints imposed on the set of edges, it is possible to derive several families of graph structures. In the following, we describe six families of graphs.

\begin{definition}[Path graph]
    A  graph $\mathcal{G}=(\mathcal{V}, \mathcal{E}, \mathbf{X}, \mathbf{E})$ with $n>1$ nodes is called \textbf{path graph}\index{path graph} (or \textbf{line graph}\index{line graph|see {path graph}}) if its nodes can be arranged in an order $u_1, u_2, ..., u_n$ such that the edges $\{u_i, u_{i+1}\}\in \mathcal{E}$, $\forall i\in [1, n-1]$. 
\end{definition}

Path graphs are often important in their role as subgraphs of other graphs, in which case they are called \emph{paths}\index{path} and identify sequences of distinct edges and nodes in that graph. One notable application of path graphs is in the context of the \emph{shortest path problem}, extensively studied in graph theory and algorithmic \citep{Dijkstra, Bellman}. This problem involves determining the most efficient route between two nodes in a graph, and it finds practical applications in various real-world scenarios, including navigation. When nodes and/or edges are not distinct, the path is called \emph{walk}\index{walk}.

\begin{definition}[Ring graph]
A  graph $\mathcal{G}=(\mathcal{V}, \mathcal{E}, \mathbf{X}, \mathbf{E})$ with $n>2$ nodes is termed a \textbf{ring graph}\index{ring graph} (or \textbf{cycle graph}\index{cycle graph|see {ring graph}}) if it is a line graph where the last node is connected to the first node, forming the closed loop $\{u_1, u_2\}, \{u_2, u_3\}, ..., \{u_n, u_1\}$. 
\end{definition}

\begin{definition}[Crossed-ring graph]
Let $\mathcal{G}=(\mathcal{V}, \mathcal{E}, \mathbf{X}, \mathbf{E})$ be a graph with nodes $u_1, u_2, ..., u_n$, where $n=2k$ and $k\geq3$. $\mathcal{G}$ is called a \textbf{crossed-ring graph}\index{crossed-ring graph} if it exhibits a ring structure with additional connections between intermediate nodes, specifically, $\{u_i, u_{n-i+1}\}\in \mathcal{E}$ for $i\in [2, k)$ and $\{u_{n-j}, u_{3+j}\}\in \mathcal{E}$ for $j\in [0, k-2)$. 
\end{definition}

Even in this context, ring graphs remain relevant as subgraphs, known as \emph{cycles}\index{cycle}, with significance beyond graph theory. In chemistry, cycles represent closed chains of atoms, serving as the structural foundation for diverse organic compounds. Particularly prevalent in aromatic compounds, such as benzene, these cyclic arrangements are essential for delineating the properties and characteristics of the compound.

\begin{definition}[Grid graph]
A graph $\mathcal{G}=(\mathcal{V}, \mathcal{E}, \mathbf{X}, \mathbf{E})$ with $m \times n$ nodes arranged in a rectangular grid is termed a \textbf{grid graph}\index{grid graph}. Each node $u$ is connected to its adjacent nodes in the grid.
\end{definition}

A more complex family of graphs is that of \emph{random} graphs~\citep{Bollobás_2001}, where each pair of nodes is connected with a certain probability $p$. Depending on the definition of the probability, $p$, we obtain graphs with different characteristics.
The Erd\H{o}s–R\'{e}nyi model is one of the pioneering works on random graphs.

\begin{definition}[Erd\H{o}s–R\'{e}nyi graph]
    An \textbf{Erd\H{o}s–R\'{e}nyi graph}\index{Erd\H{o}s–R\'{e}nyi graph}, $\mathcal{G}$, is a random graph with $n$ nodes where each edge is sampled with probability $p$ such that $$P(\mathbf{deg}(u)=k)=\binom{n-1}{k}p^k(1-p)^{n-1-k}.$$
\end{definition}

The Erd\H{o}s–R\'{e}nyi's assumption, \ie edges are sampled with equal probability, may be inappropriate for modeling certain real-world phenomena. Such an assumption has the effect that nodes do not tend to cluster, making it difficult to model social networks, for example. An alternative to this scenario is that of Barabasi-Albert model.

\begin{definition}[Barabasi-Albert graph]
    A \textbf{Barabasi-Albert graph}\index{Barabasi-Albert graph}, $\mathcal{G}$, is a random graph with $n$ nodes defined by a growth process and preferential attachment mechanism. Starting with a small initial graph, nodes are added iteratively. Each new node $u$ is connected with $k$ other nodes already in the graph with a preferential attachment mechanism, such that $$p_{uv}=\frac{\mathbf{deg}(v)}{\sum_{v\leq v'} \mathbf{deg}(v')},$$ where $p_{uv}$ is the probability of the edge between nodes $u$ and $v$ to exist.
\end{definition}

\begin{figure}
     \centering
     \begin{subfigure}[b]{0.32\textwidth}
         \centering
         \includegraphics[width=0.8\textwidth]{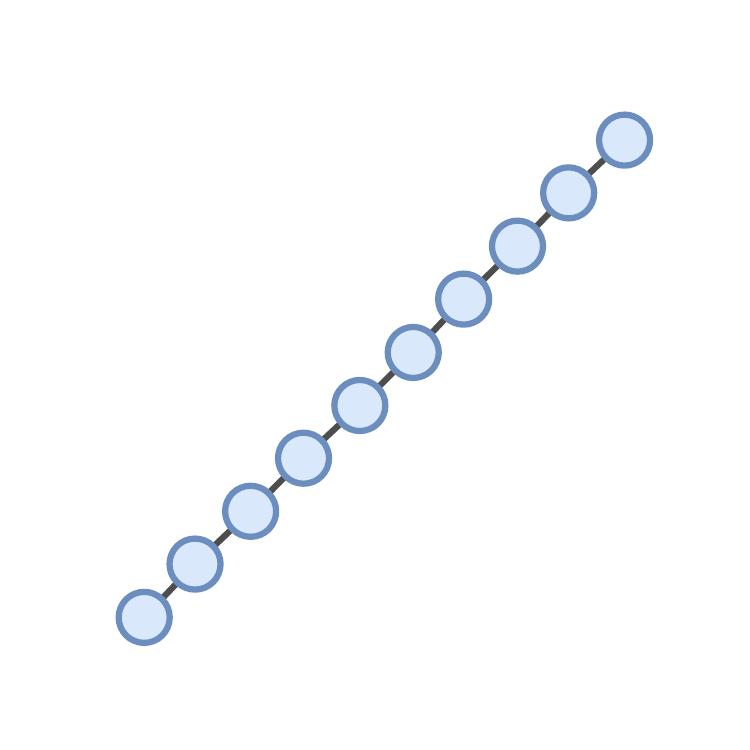}
         \caption{Path Graph}
     \end{subfigure}
     \hfill
     \begin{subfigure}[b]{0.32\textwidth}
         \centering
         \includegraphics[width=1.05\textwidth]{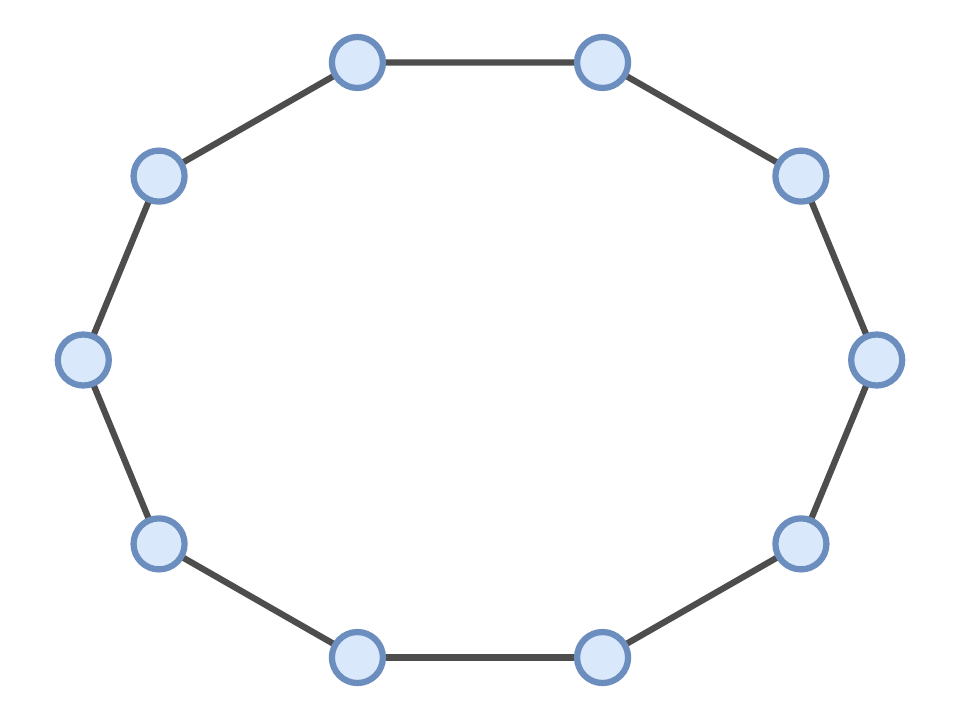}
         \caption{Ring Graph}
     \end{subfigure}
     \hfill
     \begin{subfigure}[b]{0.32\textwidth}
         \centering
         \includegraphics[width=1.05\textwidth]{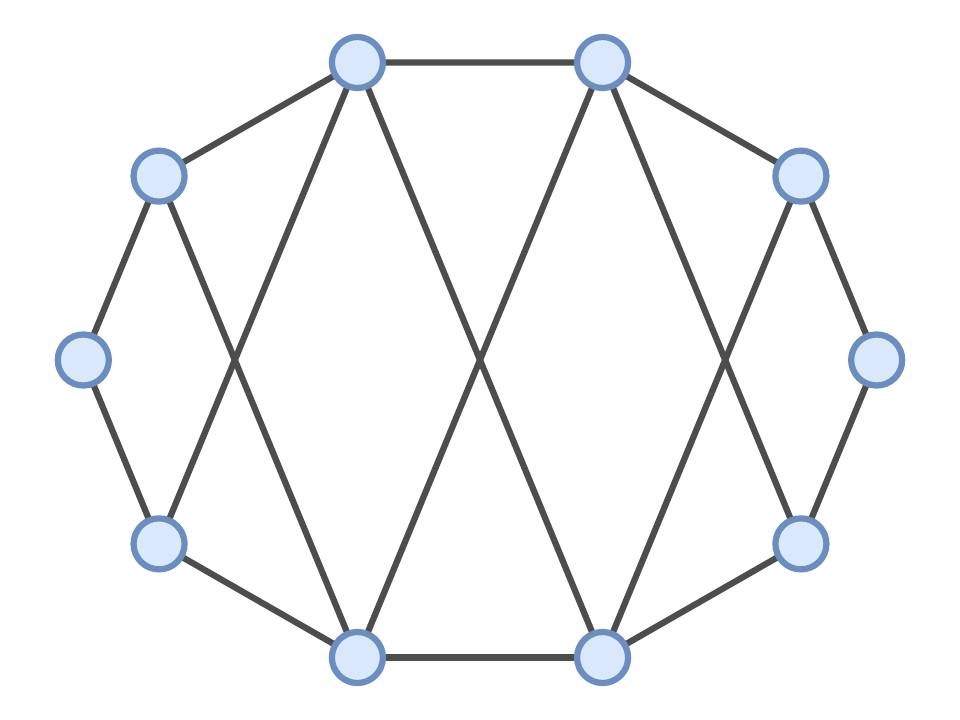}
         \caption{Crossed-ring Graph}
     \end{subfigure}
     \begin{subfigure}[b]{0.32\textwidth}
         \centering
         \includegraphics[width=1.05\textwidth]{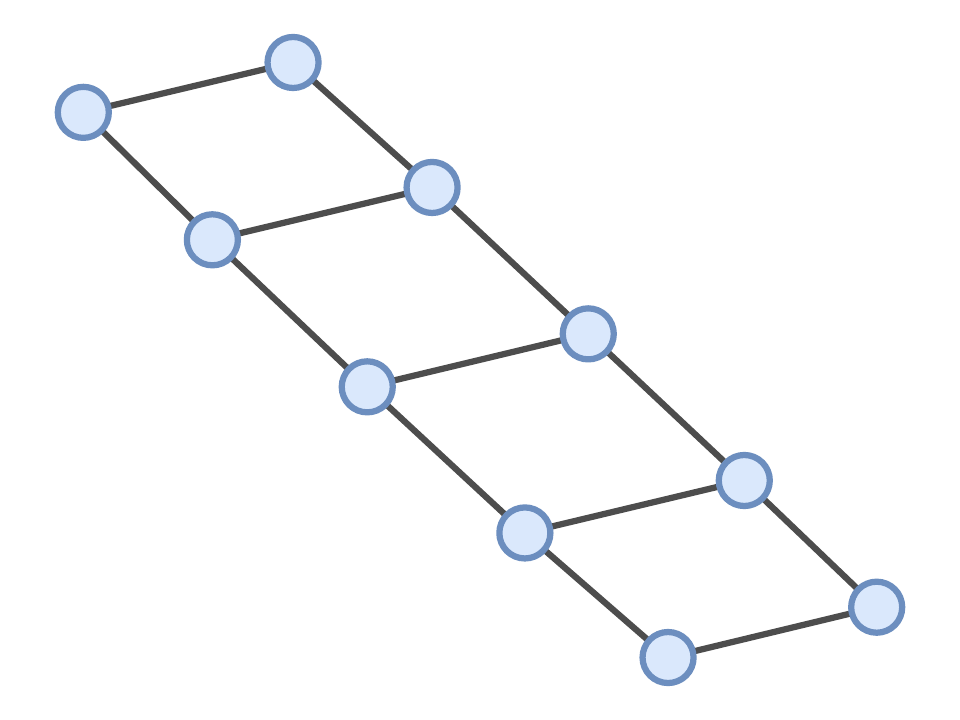}
         \caption{Grid  Graph}
     \end{subfigure}
     \begin{subfigure}[b]{0.32\textwidth}
         \centering
         \includegraphics[width=1.05\textwidth]{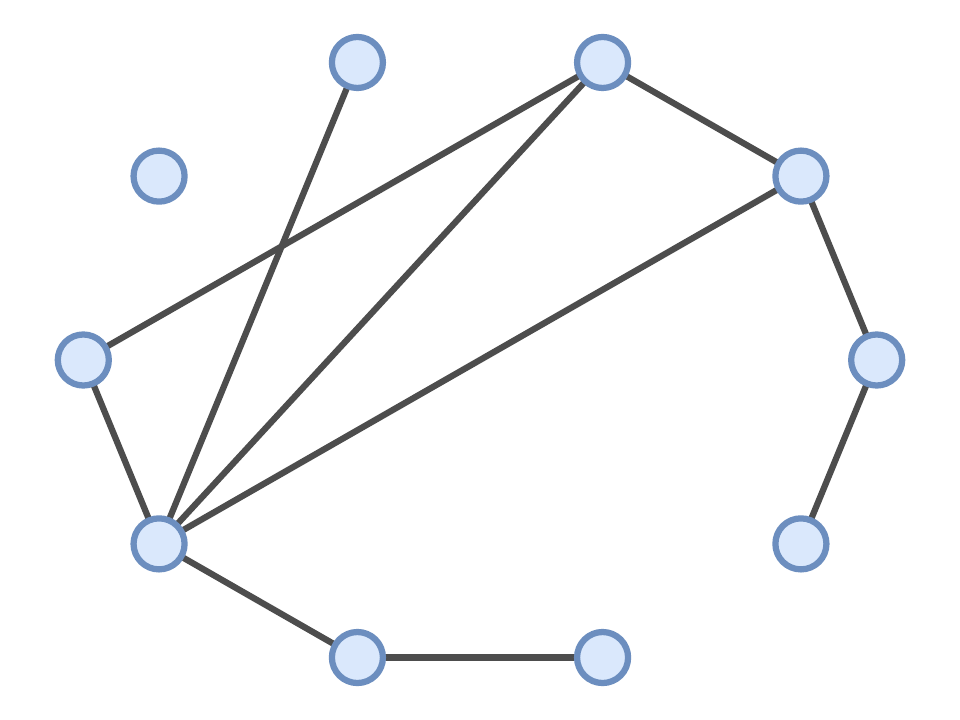}
         \caption{Erd\H{o}s–R\'{e}nyi}
     \end{subfigure}
     \begin{subfigure}[b]{0.32\textwidth}
         \centering
         \includegraphics[width=1.05\textwidth]{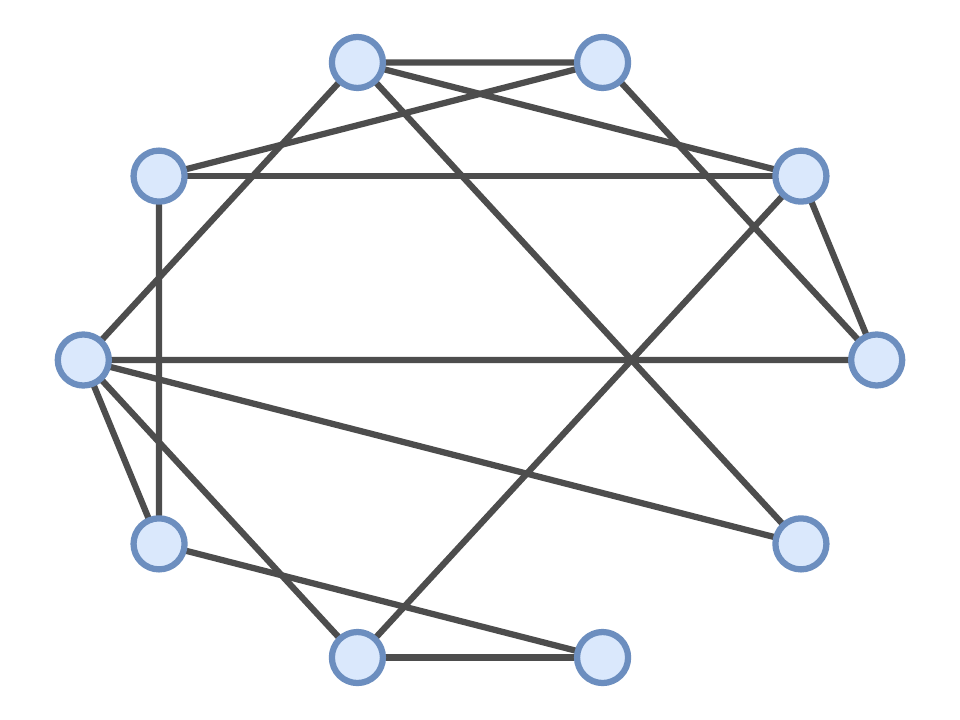}
         \caption{Barabasi-Albert}
     \end{subfigure}
        \caption{Instances of various graph families: (a) path graph, (b) ring graph, (c) crossed-ring graph, (d) grid graph, (e) Erd\H{o}s–R\'{e}nyi Graph with probability $p=0.2$, and (f) Barabasi-Albert Graph with $k=2$. Each graph consists of 10 nodes.}
        \label{fig:graph_distrib}
\end{figure}

Barabasi-Albert graphs are characterized by a node degree distribution that follows the power law, resulting with a few nodes with high degree and many nodes with low degree, since the probability that a new node $u$ will connect with a given node $v$ is proportional to the degree of $v$. Figure~\ref{fig:graph_distrib} illustrates an example for each type of a graph defined before.

\subsection{Dynamic Graphs}\label{sec:dynamic_graph_notation}
Static graphs may not fully capture the complexity of the entire realm of systems of relationships. This is because the static property can be limiting when considering real-world processes, both natural and synthetic. In these scenarios, the graph topology and features may evolve over time, making them dynamic in nature.
This highlights the necessity for a dynamic representation of a graph. 

\begin{definition}[Dynamic graph]
A \textbf{dynamic graph}\index{dynamic graph} (also referred to as \textbf{temporal} graph\index{temporal graph|see {dynamic graph}}) is a tuple $\gls*{G(t)}=(\gls*{V(t)}, \gls*{E(t)}, \gls*{Xf(t)}, \gls*{Ef(t)})$, defined for $\gls*{t}\geq0$.    
\end{definition}

 Differently from static graphs, all  elements in the tuple are functions of time $t$. Thus, $\mathcal{V}(t)$ provides the set of nodes which are present in the graph at time $t$, and $\mathcal{E}(t) \subseteq \{\{u,v\} \, | \, u,v \in \mathcal{V}(t)\}$ defines the links between them. Analogously, $\mathbf{X}(t)$ and $\mathbf{E}(t)$ define node states and edge attributes at time $t$. Although, $\mathcal{V}(t)$ can theoretically change over time, in practice it is often considered fixed for the ease of computation, which means that all the nodes that will appear in the dynamic graph are known in advance. Hence, $\mathcal{V}(t) = \mathcal{V}$ for $t\geq 0$.
\\

The way we observe a system of interacting entities plays a crucial role in the definition of the corresponding dynamic graph. We can distinguish between two distinct types: \textit{discrete-time} dynamic graphs and \textit{continuous-time} dynamic graphs.

Before providing such definitions, we define the concept of a graph snapshot.
\begin{definition}[Graph snapshot]
    A \textbf{graph snapshot}\index{snapshot of a, dynamic graph} is a tuple $\gls*{Gt} = (\gls*{Vt},\gls*{Et},\gls*{Xft},\\ \gls*{Eft})$ that refers to the (observed) static representation of a dynamic graph at a time $t$. 
\end{definition}
In other words, a graph snapshot is a static graph that provides a picture of the whole dynamic graph's state at a particular time.
Each snapshot maintains the notations and definitions outlined for static graphs (see Section~\ref{sec:static_graph_notation}).

\begin{definition}[Discrete-time dynamic graph]
A \textbf{discrete-time dynamic graph}\index{discrete-time dynamic graph|see {dynamic graph}}\index{dynamic graph!discrete-time} (\gls*{DTDG}) is a series of graph snapshots defined in the time interval $[t_0, t_n]$, $$\mathcal{G} = \{\gls*{Gt} \, | \, t\in[t_0, t_n]\}.$$ 
\end{definition}

\begin{figure}[ht]
\centering 
\includegraphics[width=0.8\textwidth]{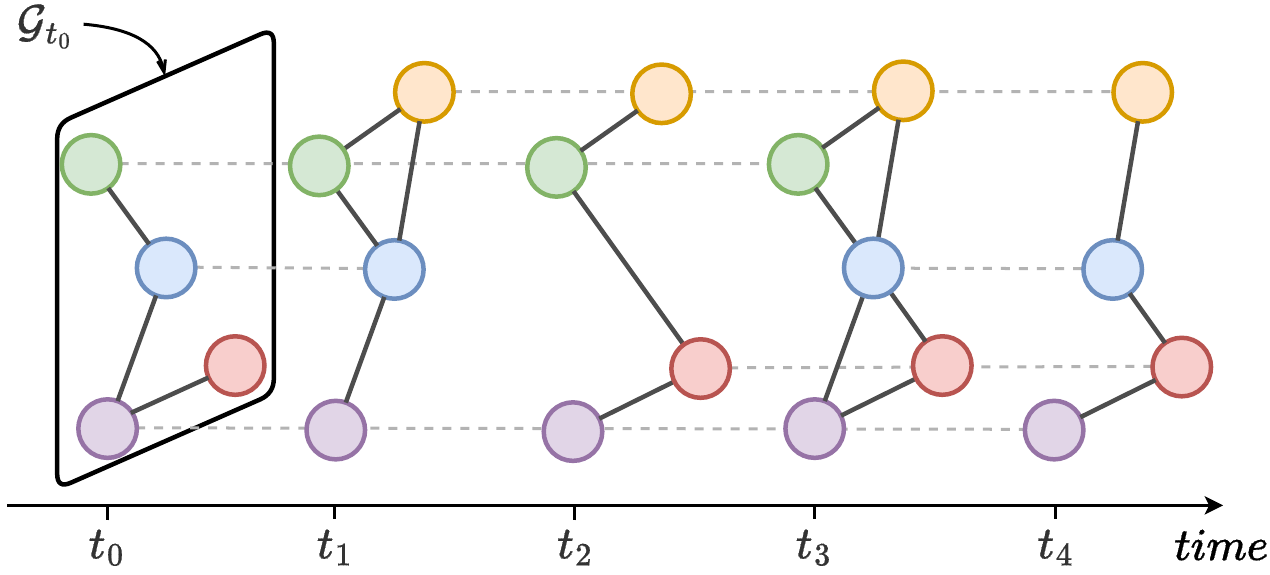}
\caption{A Discrete-Time Dynamic Graph defined over five timestamps and a set of five interacting entities.}
\label{fig:dtdg_graph}
\end{figure}

Therefore, a D-TDG models an evolving system that is fully observed at different timestamps. Commonly, D-TDG are captured at periodic intervals (\eg hours, days, etc.) Hence, considering $\Delta t > 0$ the interval between observations and $t_i$ the current timestamp, the next observation is captured at $t_{i+1} = t_i + \Delta t$. 
Since each snapshot at time $t$ provides an updated representation of the graph's topology, the temporal neighborhood of a node $u$ corresponds to the static neighborhood definition described in Section~\ref{sec:static_graph_notation}.
We present in Figure~\ref{fig:dtdg_graph} a visual exemplification of a D-TDG.

We note that if the set of nodes and edges are fixed over time (\ie $\mathcal{G}_t = (\mathcal{V}, \mathcal{E}, \overline{\bfX}_t, \bfE_t)$), then the dynamic graph is often referred to as \textbf{spatio-temporal} graph\index{spatio-temporal graph}.

A continuous-time dynamic graph is a more general formulation of a discrete-time dynamic graph. It models systems that are not fully observed over time. In facts, only new events in the system are observed. Therefore,

\begin{definition}[Continuous-time dynamic graph]
    A \textbf{continuous-time dynamic graph}\index{continuous-time dynamic graph|see {dynamic graph}}\index{dynamic graph!continuous-time} (\gls*{CTDG}) is a stream of \textbf{events}\index{event} (also referred to as \textbf{observations}\index{observation|see {event}}) defined in the time interval $[t_0, t_n]$ $$\mathcal{G} = \{o_t\, | \, t\in[t_0, t_n]\}.$$ An event, 
    $o_t = (t, \,EventType,\,u,\,v,\,\overline{\bfx}_u,\, \overline{\bfx}_v,\, \mathbf{e}_{uvt})$, is a tuple containing information regarding the timestamp, the event type, the involved nodes, and their (observed) states.  
\end{definition}
 
Such events have a clear interpretation as graph edits~\citep{Gao2010}, which have also been formulated for dynamic graphs~\citep{paassen2021graph}. We consider six edit types: node deletions, node insertions, node replacements (i.e. features change), edge deletions, edge insertions, and edge replacements (i.e., edge features change). Without loss of generality, we can group these events into three main categories: \textit{node-wise}\index{event!node-wise} events, when a node is created or its features are updated; \textit{interaction}\index{event!interaction} events, \ie a temporal edge is created; \textit{deletion}\index{event!deletion} events, \ie node/edge is deleted.

Despite the underlying discrete nature of events, the granularity of the observations is refined to the extent that the dynamic graph is considered as a continuous flow, allowing for events to happen at any moment (\ie characterized by irregular timestamps). This is in contrast to discrete-time dynamic graphs, where changes to the graph are typically aggregated.
\\

Generally, the temporal neighborhood of a node $u$ at time $t$, consists of all the historical neighbors of $u$, prior to current time $t$.
\begin{definition}[Temporal neighborhood in C-TDG]
Let the edge set $\mathcal{E}(t) \subseteq \{\{u,v, t^-\} \, | \, u,v \in \mathcal{V}(t),\,\, t^- < t\}$ be the set of edges that are present in a C-TDG at time $t$. The \textbf{temporal neighborhood}\index{temporal neighborhood|see {neighborhood}}\index{neighborhood!temporal neighborhood} of a node $u$ at time $t$ is the set 
$\gls*{Nut} = \{(v, t^-) \, |\,  \{u,v,t^-\} \in \mathcal{E}(t) \}$.
\end{definition}

\begin{figure}[ht]
\centering 
\includegraphics[width=0.74\textwidth]{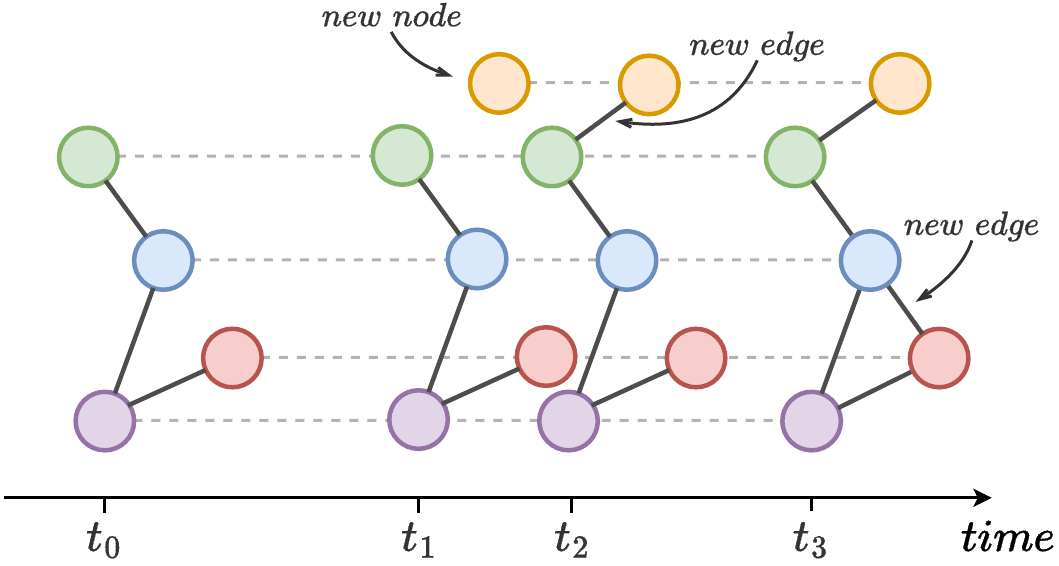}
\caption{The evolution of a Continuous-Time Dynamic Graph through the stream of events until the timestamp $t_3$.}
\label{fig:ctdg_graph}
\end{figure}

We observe that at any time point $t$, we can obtain a snapshot of the C-TDG by sequentially aggregating the events up to time $t$.  
Figure~\ref{fig:ctdg_graph} shows visually the temporal evolution of a C-TDG.


\section{Introduction to Differential Equations}\label{sec:intro_DE}

In this section, we survey the principal concepts about differential equations and dynamical systems, which will be used throughout the rest of this thesis. We base this section on classical theory of differential equations \citep{ross1984differential, amr, Ascher1998, EvansPDE, Mattheij2002, Ascher2008} and the more recent literature on neural differential equations \citep{HaberRuthotto2017, neuralODE, chang2018antisymmetricrnn, kidger2021a}.

\subsection{Differential Equations}\label{sec:diff_eq}
Differential equations serve as powerful mathematical tools for modeling and understanding various phenomena across diverse fields, ranging from physics and engineering to biology and economics. These equations describe the relationship between an unknown function and its derivatives, capturing the rate of change of a quantity with respect to one or more independent variables. 

To be more formal, a differential equation is defined as follows.
\begin{definition}[Differential equation]\label{def:diff_eq}
    A \textbf{differential equation}\index{differential equation} is any equation which contains derivatives, either ordinary or partial derivatives.
\end{definition}

To provide a practical intuition of the form of a differential equation, let's consider the example of an object of mass $m$ that is falling under the influence of constant gravity $g$. The differential equation written in terms of the position $x$ is
\begin{equation}\label{eq:newton_law}
    m\frac{d^2x}{dt^2}=-mg.
\end{equation}
Similarly, we can define the one-dimensional wave equation 
\begin{equation}\label{eq:1d-wave}
    \frac{\partial^2 u}{\partial t^2} = c^2\frac{\partial^2 u}{\partial x^2},
\end{equation}
which models the vibration of a string in one dimension $u = u(x,t)$, thus $u$ is the factor representing a displacement from rest situation. The constant $c$ gives the speed of propagation for the vibration, and $\partial^2 u/ \partial t^2$ describe how forcefully the displacement is being changed.

We observe that the order of the highest ordered derivative involved in the differential equation is called the \textbf{order}\index{order of a differential equation|see {differential equation}}\index{differential equation!order of} of the differential equation. Therefore, Equation~\ref{eq:newton_law} is a second-order differential equation, since the highest involved derivative is a second derivative.

Depending on the type of derivatives employed in the differential equation we can distinguish between \emph{ordinary differential equations} and \emph{partial differential equations}.

\begin{definition}[Ordinary differential equation]
    An \textbf{ordinary differential equation}
(\gls*{ODE})\index{ordinary differential equation} is a differential equation for a function of a single variable.
\end{definition}

\begin{definition}[Partial differential equation]
    A \textbf{partial differential equation}
(\gls*{PDE})\index{partial differential equation} is a differential equation for a function of two or more variables.
\end{definition}

Therefore, Equation~\ref{eq:newton_law} provides an example of an ordinary differential equation because it involves the position variable, $x$, and its derivatives with respect to time. In contrast, Equation~\ref{eq:1d-wave} exemplifies a partial differential equation as it relies on partial derivatives of $u$ with respect to both space and time.

In this thesis, we will focus only on ODEs, since they play a central role in the differential equation domain due to their simplicity and wide applicability in the description of dynamical systems. 

\begin{definition}[Dynamical system]
    A \textbf{dynamical system}\index{dynamical system} is a system whose state is uniquely specified by a set of variables and whose behavior is described by a predefined differential equation.
\end{definition}

\begin{figure}[h!]
    \centering
    \includegraphics[width=0.4\textwidth]{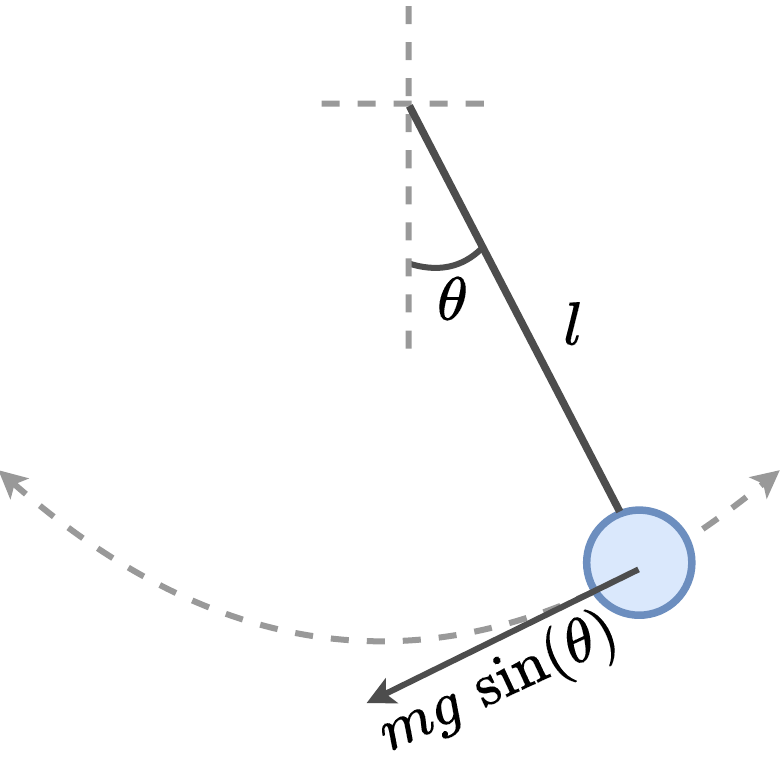}
    \caption{Motion of a pendulum.}
    \label{fig:pendulum}
\end{figure}

A classic example of a dynamical system is the pendulum. A pendulum is a body suspended from a fixed support that swings freely back and forth under the influence of gravity. When a pendulum is displaced from its resting position, it experiences a restoring force due to gravity that accelerates it towards equilibrium, causing the mass of the pendulum to oscillate about the equilibrium position. In the case of a simple pendulum (i.e., undampened pendulum with point mass), the differential equation that defines the behavior of the system is 
\begin{equation}\label{eq:pendulum}
    \frac{d^2\theta}{dt^2} + \frac{g}{l} \sin(\theta)=0,
\end{equation}
where $g$ is the gravitational constant, $l$ is the length of the rod, and $\theta$ is the angle from the vertical to the pendulum, \ie the rest position. Figure~\ref{fig:pendulum} visually represents the motion of a pendulum.

\begin{figure}
    \centering
    \includegraphics[width=0.8\textwidth]{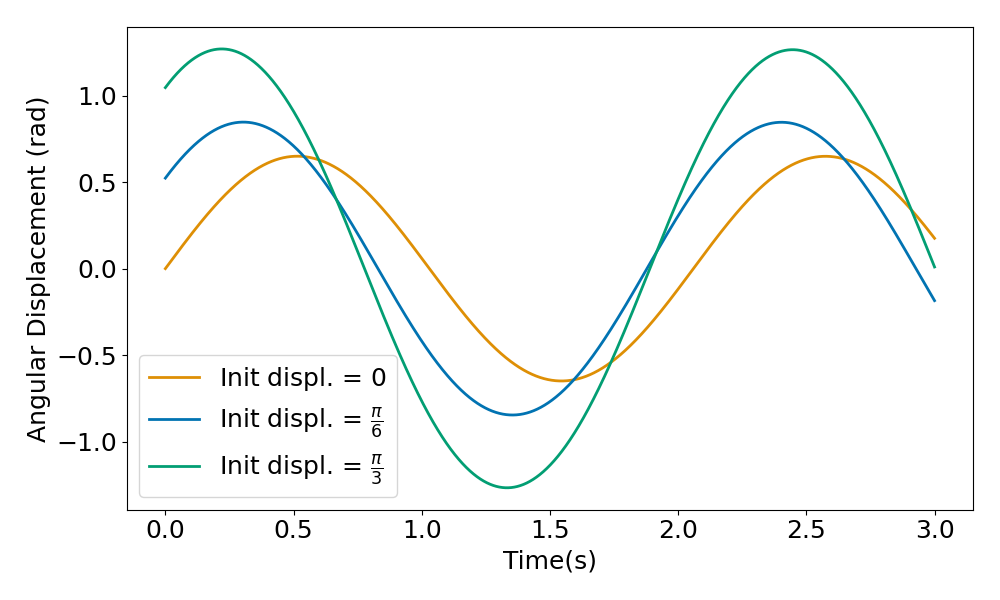}
    \caption{Simulation of a simple pendulum, with unitary rod length ($l=1$) and initial angular velocity of 2 rad/s, over a timespan of three seconds ($t\in[0,3]$) with different initial angular displacement, \ie $0$, $\pi/6$, and $\pi/3$.}
    \label{fig:pendulum_simulation}
\end{figure}

We note that, usually, a given differential equation has an infinite number of solutions, thus the state of the dynamical system is dependent on additional conditions associated with the differential equation. Generally, such condition corresponds to the initial value of the system's state. Thus, the dynamical system is described by an \emph{initial value problem}. Figure~\ref{fig:pendulum_simulation} shows how different initializations produce various system's dynamics, \ie  \textbf{trajectories}.

\begin{definition}[Initial value problem]
    An \textbf{initial value problem} (also known as \textbf{Cauchy problem})\index{initial value problem}\index{Cauchy problem|see {initial value problem}} is a differential equation together with a point in the domain of the function called initial condition
    \begin{equation}\label{eq:ivp}
        \begin{dcases}
            \frac{dx(t)}{dt} = f(t, x), \quad t\in[0,T]\\
            x(0) = c
        \end{dcases}
    \end{equation}
    where $f(t,x)$ is a general nonlinear function of $t$ and $x$, which is often referred to as \textbf{vector field}\index{vector field}; and  $x(t)$ represents a point in the space, initialized by some value $c$, that describes the state of the system. $x(t)$ is called \textbf{state vector}\index{state vector}. 
\end{definition}
A dynamical system defined by Equation~\ref{eq:ivp} is called \textbf{autonomous}\index{autonomous dynamical system|see {dynamical system}}\index{dynamical system!autonomous} when $f$ does not explicitly depend on $t$. The dynamical system is called \textbf{non-au\-ton\-o\-mous}\index{non-autonomous dynamical system|see {dynamical system}}\index{dynamical system!non-autonomous} otherwise. 

There are problems in physics and engineering that are modeled by \textbf{dissipative dynamical systems}\index{dissipative dynamical systems|see {dynamical system}}\index{dynamical system!dissipative}. These systems are characterized by the
property of possessing a bounded absorbing set in which all trajectories enter in a finite time, and thereafter remain inside. This can be seen as a gradual decrease in energy (\ie energy dissipation) as the system evolves. 
We now provide a definition of a dissipative system 
based on the one provided in \cite{dissipative}. 

\begin{definition}[Dissipative system]\label{def:dissipative}
Let define $E\subseteq \mathbb{R}^{d}$  a bounded set that contains any initial condition $\mathbf{x}_u(0)$ for the ODE in Equation~\ref{eq:simple_ode_dgn}. The system defined by the ODE in Equation~\ref{eq:simple_ode_dgn} is dissipative if there is a bounded set $B$ where, for any $E$, exists $t^*\geq 0$ such that $\left\{x(t) \mid x(0) \in E\right\} \subseteq B$ for $t> t^*$.
\end{definition}

\subsection{Hamiltonian Systems}\label{sec:hamiltonian_sys}
A specific case of dynamical system is that of \emph{Hamiltonian systems}, which describe the evolution equations of specific physical systems.

\begin{definition}[Hamiltonian system]
    An \textbf{Hamiltonian system}\index{Hamiltonian system|see {dynamical system}}\index{dynamical system!Hamiltonian} is a dynamical system whose state, $\bfx = (\bfp, \bfq)^\top\in \mathbb{R}^{2d}$, is described by the ODE
\begin{equation}\label{eq:hamiltonian_sys}
\frac{d\bfx(t)}{dt} = \mathcal{J}\nabla_\bfx H(\bfx(t)),
\end{equation}
where $\mathcal{J} = \begin{pmatrix}\mathbf{0} & \bfI_d\\-\bfI_d & \mathbf{0}\end{pmatrix}$ is an antisymmetric matrix~\footnote{A matrix $\bfM$ is antisymmetric (\ie skew-symmetric) if $\bfM^\top=-\bfM$.} with $\bfI_d$ the identity matrix of dimension $d$, $H:\mathbb{R}^{2d}\rightarrow\mathbb{R}$ a twice continuously differentiable function, and $\nabla_\bfx H(\bfx(t))$ denoting the gradient of $H$ with respect to $\bfx$.
\end{definition}

We observe that by decoupling $\bfx$ into its main components, \ie $\bfp\in \mathbb{R}^{d}$ and $\bfq\in \mathbb{R}^{d}$, the Hamiltonian system described by Equation~\ref{eq:hamiltonian_sys} can be rewritten as
\begin{equation}\label{eq:hamiltonian_sys_split}
\frac{d \bfp}{d t} = -\frac{\partial H}{\partial \bfq}(\bfp, \bfq), \quad    \frac{d \bfq}{d t} = +\frac{\partial H}{\partial \bfp}(\bfp, \bfq).    
\end{equation}
In this setting, $\bfp$ is usually referred to as the \emph{momentum}, $\bfq$ as the \emph{coordinates}, and $H$ is the \emph{Hamiltonian} which represents the total energy of the system.

To provide an example for Hamiltonian systems, let's consider again the example of the pendulum outlined in Equation~\ref{eq:pendulum}. The pendulum can be defined as a Hamiltonian system described by the equation
\begin{equation}
    H(\bfp, \bfq) = \frac{1}{2l}\bfp^2-gl\cos(\bfq)
\end{equation}
where $l$ is the length of the rod and $g$ the gravitational constant, as before. Following Equation~\ref{eq:hamiltonian_sys_split} we obtain
\begin{equation}
    \frac{d \bfp}{d t} = -gl\sin(\bfq), \quad    \frac{d \bfq}{d t} = \frac{\bfp}{l}.   
\end{equation}

The fundamental property of Hamiltonian systems is their conservative nature. Therefore, they are essential for characterizing systems with constant energy. If the Hamiltonian does not explicitly depend on time, \ie we are in the autonomous case, the Hamiltonian is a statement of the \textbf{conservation of energy}, therefore no energy can be created nor lost, because the Hamiltonian is constant over time. Indeed, by considering the derivative of $H(\bfx(t))$ we obtain
\begin{equation}
    \frac{dH(\bfx(t))}{dt} = \frac{\partial H(\bfx(t))}{\partial \bfx(t)}\frac{d\bfx(t)}{dt} = \frac{\partial H(\bfx(t))}{\partial \bfx(t)} \mathcal{J}\nabla_\bfx H(\bfx(t))=0,
\end{equation}
where the last equality holds since $\mathcal{J}$ is antisymmetric. Therefore, $H(\bfx(t)) = H(\bfx(0)) = \text{const}$ for all $t$. From a geometrical point of view, this implies that the Hamiltonian system operates as a symplectic map, \ie the area of the set of output trajectories remains constant over time, because the output trajectories result as a rotation of the set of initial conditions at a constant rate.

\begin{figure}[h]
    \centering
    \includegraphics[width=0.7\linewidth]{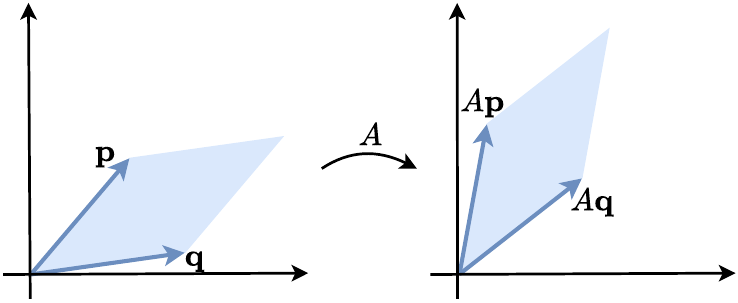}
    \caption{Symplecticity (\ie area preservation) of a linear mapping.}
    \label{fig:symplectic}
\end{figure}

\begin{definition}[Symplectic map]\label{def:symplectic}
    A linear mapping, $A: \mathbb{R}^{2d}\rightarrow \mathbb{R}^{2d}$, is called \textbf{symplectic}\index{symplectic map} if 
    \begin{equation}
        A^\top\mathcal{J}A = \mathcal{J}
    \end{equation}
    where $\mathcal{J} = \begin{pmatrix}\mathbf{0} & \bfI_d\\-\bfI_d & \mathbf{0}\end{pmatrix}$.
\end{definition}
Therefore, a symplectic map represents a transformation of state space (\ie the space in which all possible states of a dynamical system are represented) that preserves volume and orientation. In other words, as the system evolves over time the coordinates and the momentum change, but the total area is conserved. Figure~\ref{fig:symplectic} shows the case $d = 1$, where $\mathbf{p}$ and $\mathbf{q}$ identify a parallelogram.

\subsection{Discretization Methods\index{discretization method}}\label{sec:discretization_method_de}
Finding an analytical solution for a Cauchy problem is often impractical. Therefore, a common approach is to compute an approximate solution through a numerical discretization procedure. This section provides an introduction to common numerical methods for approximating solutions of Cauchy problems. Without loss of generality to higher-order systems, we restrict ourselves to first-order differential equations, because a higher-order ODE can be converted into a larger system of first-order equations by introducing extra variables\footnote{We refer an interested reader to \citet{Robinson_2004} for more details about the \emph{reduction of order} method.}. 

The fundamental concept behind numerical discretization methods is to divide time into discrete intervals, and then iteratively compute the solution over a discrete set of time points. Let $0=t_0, \dots, t_n=T$ be the set of time points in which we approximate the solution $x(t_i)$, then the set $\{x(t_i)\}_{i=0}^n$ is called a \emph{discretization} of the Cauchy problem in Equation~\ref{eq:ivp} 
.

Numerical methods often can be divided into two main categories: \emph{explicit} and \emph{implicit} methods.
\textbf{Explicit methods}\index{explicit methods|see {discretization method}}\index{discretization method!explicit} calculate the state of a system at a later time, $t_{i+1}$, from the state of the system at the current time, $t_{i}$, while \textbf{implicit methods}\index{implicit methods|see {discretization method}}\index{discretization method!implicit} find a solution by solving an equation involving both the current state of the system $x(t_{i})$ and the future state $x(t_{i+1})$.

The \textbf{forward Euler method}\index{forward Euler method} is the best known and simplest explicit numerical method for approximation. The idea is that it approximates the solution at the next time step based on the derivative at the current time step, thus moving along the tangent line of the approximate function at each point.
Formally, the forward Euler method we approximate the derivative on the left-hand side of Equation~\ref{eq:ivp} by a finite difference, and evaluate the right-hand side at $x(t_{i})$:
\begin{equation}
    \frac{x(t_i+\epsilon)- x(t_i)}{\epsilon} = f(t_i, x(t_i))
\end{equation}
where $\epsilon=t_{i+1}- t_i$ is the step size of the method, which defines the approximation accuracy of the solution. Therefore, the state at the next time step is
\begin{equation}\label{eq:forward_euler}
    x(t_{i+1})= x(t_i) + \epsilon f(t_i, x(t_i)).
\end{equation}

\begin{figure}[h!]
    \centering
     \begin{subfigure}[b]{0.49\textwidth}
         \centering
         \includegraphics[width=1.05\textwidth]{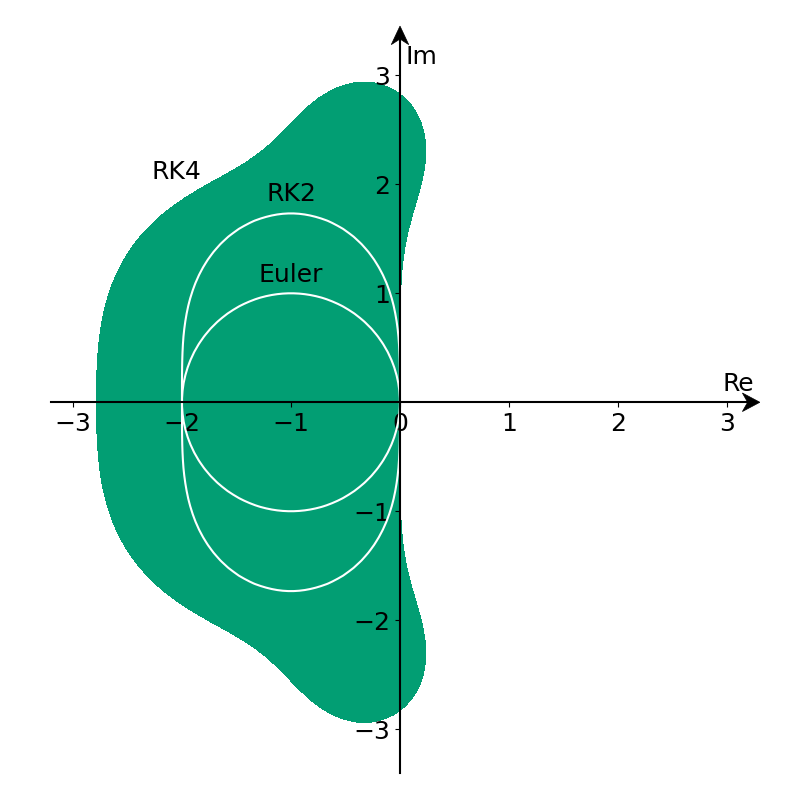}
         \caption{RK stability}
     \end{subfigure}
     \hfill
     \begin{subfigure}[b]{0.49\textwidth}
         \centering
         \includegraphics[width=1.05\textwidth]{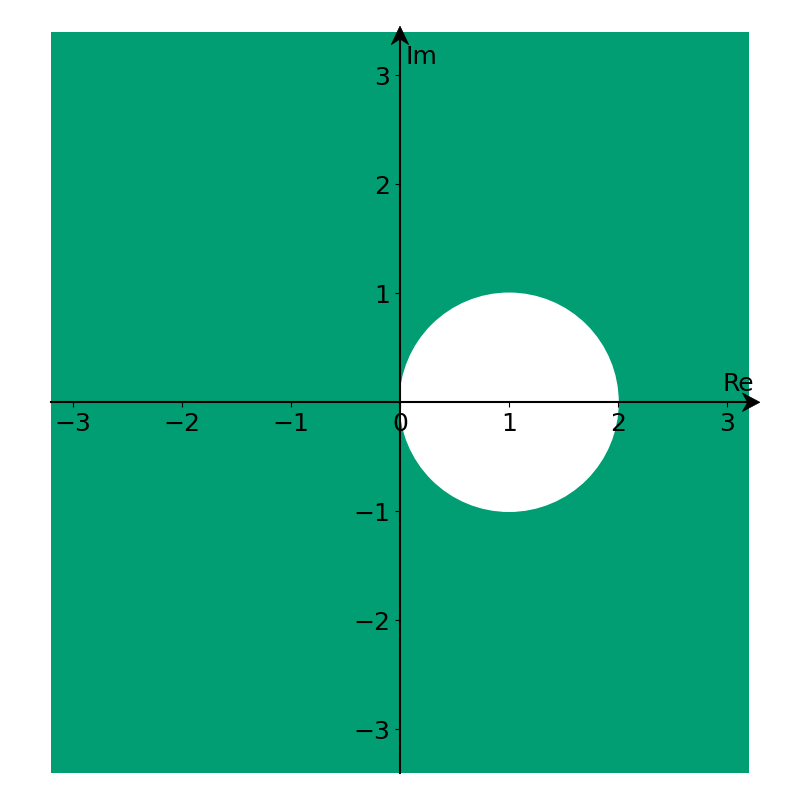}
         \caption{B-E stability}
     \end{subfigure}
    \caption{The stability regions, colored in green, of (a) various explicit Runge-Kutta methods and (b) the backward Euler method.}
    \label{fig:stability}
\end{figure}

To ensure that a small perturbation in the initial conditions does not cause the numerical approximation to diverge away from the true solution, \ie ensuring \emph{stability of the method}, we need that
\begin{equation}\label{eq:euler_stability}
   |x(t_{i+1})| \leq |x(t_i)|.
\end{equation}
Therefore, the forward Euler method is \textbf{(absolute) stable} if  
\begin{equation}
    |1+\epsilon \lambda| \leq 1,
\end{equation}
where $\lambda$ is the maximum eigenvalue of the system. In other words, the forward Euler method is considered stable when $(1+\epsilon\lambda)$ lies within the unit circle in the complex plane for all eigenvalues of the system (we refer the reader to the left-side of Figure~\ref{fig:stability} for a visual exemplification of the region). On the contrary, when the system is \emph{unstable}, the numerical solution diverges from the exact solution.
We observe that, while the choice of the step size should ideally be dictated only by approximation accuracy requirements, it also plays a crucial role in improving the stability of the method, since, for small enough step sizes, the numerical solution converges to the exact solution. 

To avoid the situation in which the stability demands a much smaller step size than what is needed to satisfy the approximation requirements, we can employ methods with less stringent stability constraints, such as the backward Euler method.
The \textbf{backward Euler method}\index{backward Euler method} (or \textbf{implicit} Euler method) is an implicit method that centers its computation at $t_{i+1}$, rather than $t_i$ like the explicit version of the method. Therefore, Equation~\ref{eq:forward_euler} is reformulated as the following
\begin{equation}\label{eq:backward_euler}
    x(t_{i+1})= x(t_i) + \epsilon f(t_{i+1}, x(t_{i+1})).
\end{equation}

Utilizing the same intuition as with the forward Euler method, here we employ the tangent at the \emph{future} point $(t_{i+1}, x(t_{i+1}))$ instead of the current point, thereby enhancing stability. Applying the condition in Equation~\ref{eq:euler_stability} to the backward euler method, we obtain that 
\begin{equation}
    \frac{1}{|1-\epsilon\lambda|} \leq 1
\end{equation}
which is always satisfied for all $\epsilon>0$ and $Re(\lambda) \leq 0$. Specifically, the stability region of the backward Euler method lies in the area in the complex plane outside the unit circle centered at (1, 0). The right-side of Figure~\ref{fig:stability} shows the stability region of the backward Euler method. As a result, we can choose the step size arbitrarily large, without compromising the stability of the method.
 
Since $x(t_{i+1})$ appears on both left and right sides of  Equation~\ref{eq:backward_euler}, for implicit methods like backward Euler, a nonlinear system of equations must be solved in each time step
. Therefore, each backward Euler step may be more expensive in terms of computing time than forward Euler. For backward Euler, the nonlinear system is 
\begin{equation}
    g(x(t_{i+1}))= x(t_{i+1}) - x(t_i) - \epsilon f(t_{i+1}, x(t_{i+1}))=0,
\end{equation}
which can be solved via a \emph{root-finding algorithm}, such as Netwton's method~\citep{Ascher1998}.

Another family of numerical methods is that of \textbf{Runge-Kutta methods}, which include both explicit and implicit approaches. 
The explicit Runge-Kutta computes the next state of the system as
\begin{equation}\label{eq:runge-kutta}
    x(t_{i+1}) = x(t_i)+\epsilon \sum_{j=0}^s \beta_jk_j
\end{equation}
where
\begin{align}
    k_1 &= f(t_i, x(t_i))\\
    k_2 &= f(t_i+\mu_2\epsilon, \; x(t_i)+\gamma_{21}k_1\epsilon)\\
    \vdots\nonumber\\
    k_s &= f\Bigl(t_i+\mu_s\epsilon, \; x(t_i)+(\gamma_{s1}k_1 + \gamma_{s2}k_2 + \dots + \gamma_{s,s-1}k_{s-1})\epsilon\Bigr).
\end{align}
Therefore, the next system state is computed as the sum of the current state and the weighted average of $s$ increments, where each increment is the product of the step size and an estimated slope specified by the function $f$ in various midpoints.

Providing specific values for the order of the method $s$, and the coefficients $\gamma_{nm}$ (for $1\leq m < n\leq s$), $\beta_j$ (for $j=1, \dots, s$) and $\mu_n$ (for $n=2, \dots, s$), we can derive different implementation of the method. The most widely known method in this family is the fourth order Runga-Kutta (RK4), thus $s=4$ and $\gamma_{nm}$, $\beta_j$, $\mu_n$ are:

\begin{equation}    
\renewcommand\arraystretch{1.3}
\begin{array}{c|cccc}
0\\
\mu_1 & \gamma_{21} \\
\mu_2 & \gamma_{31} & \gamma_{32} \\
\mu_3 & \gamma_{41} & \gamma_{42} & \gamma_{43}\\
\hline
& \beta_1 &\beta_2 &\beta_3 &\beta_4 
\end{array}
\;=\; 
\begin{array}{c|cccc}
0\\
\frac{1}{2} & \frac{1}{2}\\
\frac{1}{2} & 0 & \frac{1}{2} \\
1           & 0 & 0& 1\\
\hline
& \frac{1}{6} &\frac{1}{3} &\frac{1}{3} &\frac{1}{6} 
\end{array}
\end{equation}

We note that when $s=1$ then the explicit Runga-Kutta is equivalent to the forward Euler method, and that the stability of Runge-Kutta methods is strictly dependent on the order $s$. Larger values of $s$ correspond to a greater stability region. Figure~\ref{fig:stability} shows the stability regions for $s=1, 2, 4$. 

Since numerical methods compute approximate solutions, it's essential to assess the discrepancy between these approximations and the true solutions of differential equations, which are quantified by the concept of \textbf{local} and \textbf{global truncation errors}\index{truncation error}\index{truncation error!local}\index{truncation error!global}\index{local truncation error|see {truncation error}}\index{global truncation error|see {truncation error}}. In the first case, the error is computed over a single step of the method, under the assumptions that we start the step with the exact solution and that there is no round off error. On the other hand, the global truncation error is the accumulation of the local truncation error over all the iterations.

\begin{figure}
    \centering
    \includegraphics[width=0.67\textwidth]{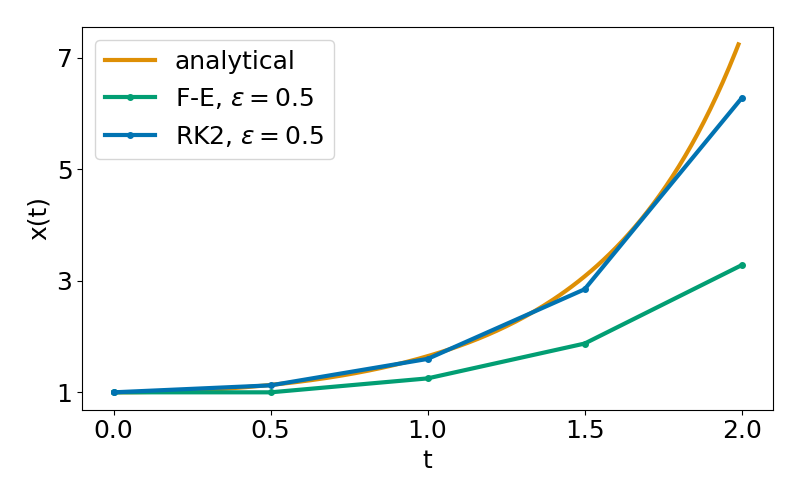}
    \caption{The numerical integration for the equation $\exp(x^2/2)$ leveraging forward Euler method with step size 0.5 (\ie F-E, $\epsilon=0.5$) and second order Runge-Kutta with step size 0.5 (\ie RK2, $\epsilon=0.5$).}
    \label{fig:euler_vs_rk2}
\end{figure}

Generally, higher-order methods produce more accurate solutions, since the errors are proportional to a power of the step size (\ie $\epsilon^s$, where $s$ is the order of the discretization method). This is visually exemplified in Figure~\ref{fig:euler_vs_rk2}, where the forward Euler method underperforms in approximating the function $\exp(x^2/2)$ with respect the second order Runge-Kutta method. We report in Table~\ref{tab:discretization_erros} the local and global truncation errors of various discretization methods \citep{Ascher1998, dormand1996numerical}.

While it is true that higher-order methods offer superior accuracy in generating solutions, their implementation requires an increased number of evaluations, which slows down the computational process. This results in a trade-off between discretization accuracy and speed.

\begin{table}[h]
\centering
\scriptsize
\caption{The local truncation error and global truncation error of various discretization schemes. \label{tab:discretization_erros}}
\begin{tabular}{lcc}
\toprule 
\textbf{Method}           & \textbf{Local error}             & \textbf{Global error}            \\\midrule
Forward Euler         & $\mathcal{O}(\epsilon^2)$ & $\mathcal{O}(\epsilon)$ \\
Backward Euler         & $\mathcal{O}(\epsilon^2)$ & $\mathcal{O}(\epsilon)$ \\
Runge-Kutta \nth{2} order & $\mathcal{O}(\epsilon^3)$ & $\mathcal{O}(\epsilon^2)$ \\
Runge-Kutta \nth{3} order & $\mathcal{O}(\epsilon^4)$ & $\mathcal{O}(\epsilon^3)$ \\
Runge-Kutta \nth{4} order & $\mathcal{O}(\epsilon^5)$ & $\mathcal{O}(\epsilon^4)$ \\
Runge-Kutta \nth{5} order & $\mathcal{O}(\epsilon^6)$ & $\mathcal{O}(\epsilon^5)$\\\bottomrule
\end{tabular}
\end{table}

Although the presented numerical methods effectively compute approximate solutions of Cauchy problems, they are not suitable for approximating Hamiltonian systems. These methods introduce numerical artifacts over time, \ie energy drift, that deviate from pure Hamiltonian dynamics. \textbf{Symplectic integrators}\index{symplectic methods|see {discretization method}}\index{discretization method!symplectic} are numerical methods designed to overcome this limitation, thus solving Hamiltonian systems while preserving their symplectic structure. Indeed, the key property of these integrators is that they reproduce the invariance of a symplectic map 
(see Definition~\ref{def:symplectic}). In other words, these integrators ensure that the energy of the system is preserved over long integration periods, making them  suitable for accurately modeling physical systems. 
The \textbf{symplectic Euler method}\index{symplectic Euler method} is a symplectic integrator that adapts the Euler's method for solving Hamiltonian's equations. It computes the state at the next time step as a combination of forward and backward Euler, \ie
\begin{align}
    \bfp(t_{i+1}) &= \bfp(t_{i}) - \epsilon \frac{\partial H}{\partial \bfq}(\bfp(t_{i}), \bfq(t_{i}))\\
    \bfq(t_{i+1}) &= \bfp(t_{i}) + \epsilon \frac{\partial H}{\partial \bfp}(\bfp(t_{i+1}), \bfq(t_{i})).
\end{align}
In other words, $\bfp$ is computed by approximating the solution at
the next time step based on the derivative at the current time step, while $\bfq$ leverages the state computed at the future point $t_{i+1}$.

\begin{figure}[h]
    \centering
    \includegraphics[width=0.4\linewidth]{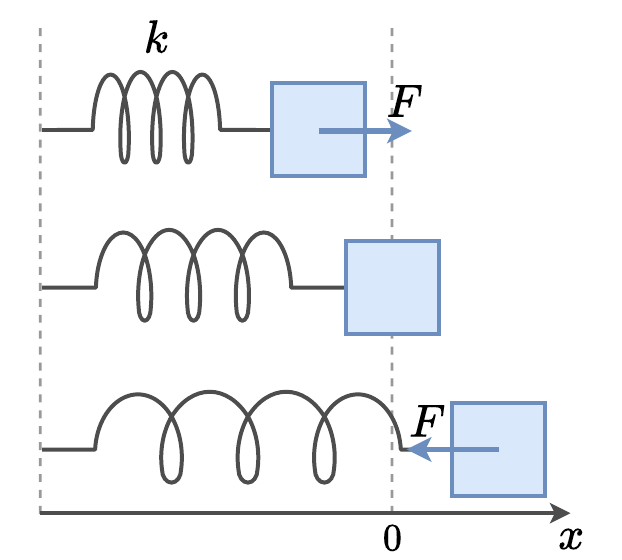}
    \caption{Motion of a simple harmonic oscillator.}
    \label{fig:harmonic_oscillator}
\end{figure}

To better illustrate the conservative properties of symplectic integrators, we report the example of a simple harmonic oscillator. A harmonic oscillator is a system that, when displaced from its equilibrium position, experiences a restoring force $F$ proportional to the displacement $x$, \ie $F=-kx$ with $k>0$ constant. Figure~\ref{fig:harmonic_oscillator} visually represents the motion of a simple harmonic oscillator. 

From a differential equation perspective, the system is described as 
\begin{equation}
\frac{d^2x}{dt^2} = -kx.    
\end{equation}

\begin{figure}[h!]
    \centering
     \begin{subfigure}[b]{0.49\textwidth}
         \centering
         \includegraphics[width=\textwidth]{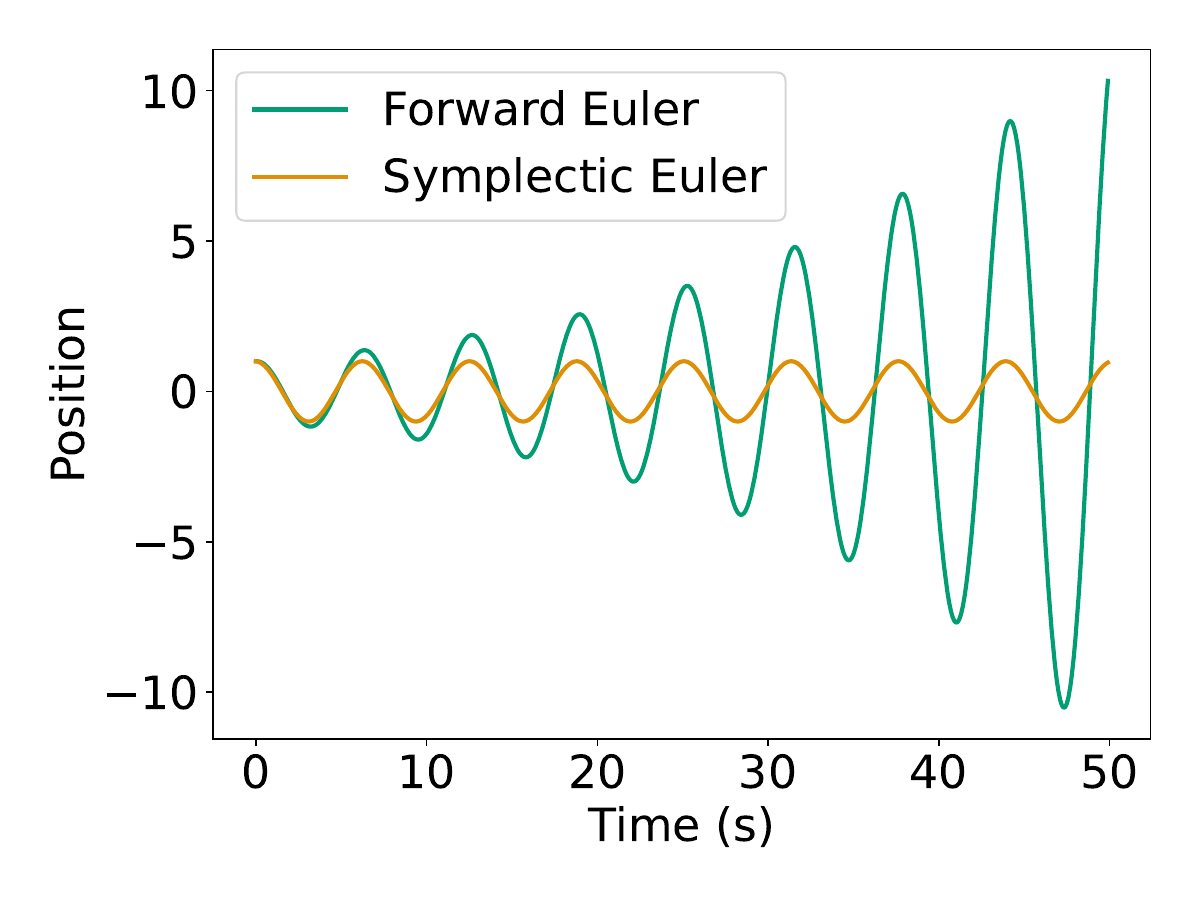}
         \caption{Position}
     \end{subfigure}
     \hfill
     \begin{subfigure}[b]{0.49\textwidth}
         \centering
         \includegraphics[width=\textwidth]{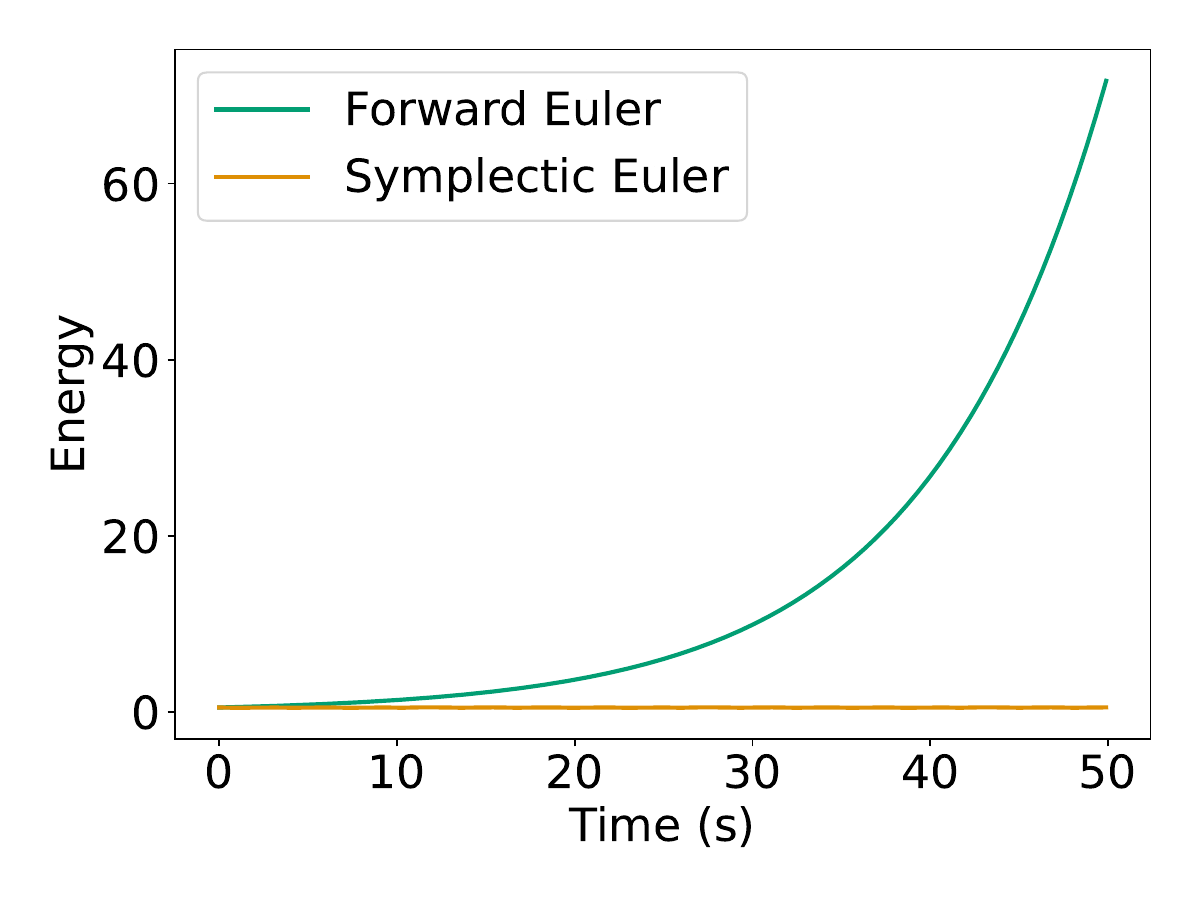}
         \caption{Energy}
     \end{subfigure}
    \caption{Simulation of a simple harmonic oscillator with unit mass (m=1) and spring constant $k=1$ over a timespan of 50 seconds ($t\in[0,50]$), comparing (a) the position and (b) the energy conservation over time using the Forward Euler and Symplectic Euler methods.}
    \label{fig:SE_vs_FE}
\end{figure}

Figure~\ref{fig:SE_vs_FE} illustrates the system's evolution using the Symplectic and Forward Euler methods. The Symplectic Euler method demonstrates a more stable oscillation, preserving the amplitude and total energy of the system. In contrast, the Forward Euler method exhibits a drift in amplitude and energy, indicating that such a method is less accurate in handling long-term dynamics and highlighting the importance of symplectic integration in ensuring the long-term stability of the system. 

Finally, we note that although in this section we mainly focus on the Euler and Runge-Kutta methods, for the sake of simplicity, the literature of discretization schemes allows for more complex strategies, such as those methods that approximate the solution using an adaptive step size.

\subsection{Neural Differential Equations}\label{sec:neural_de}
The conjoining of dynamical systems and deep learning has become a topic of great interest in recent years. In particular, neural differential equations (neural DEs) demonstrate that neural networks and differential equations are two sides of the same coin \citep{HaberRuthotto2017, neuralODE, chang2018antisymmetricrnn}. Indeed, many popular neural network architectures, such as residual networks (\gls*{ResNet}s)~\citep{resnet} and recurrent neural networks (\gls*{RNN}s)~\citep{RNN}, can be seen as a discretization of parameterized differential equations.

\begin{definition}[Neural differential equation]
A \textbf{neural differential equation}\index{neural differential equation} is a differential equation using a neural network to parameterize the vector field, \ie
    \begin{equation}\label{eq:neuralODE}
        \begin{dcases}
            \frac{d \bfx(t)}{dt} = f_\theta(\bfx(t)), \quad t\in[0,T]\\
            \bfx(0) = c
        \end{dcases}
    \end{equation}
where $f_\theta(\bfx(t)): \mathbb{R}^d \rightarrow \mathbb{R}^d$ is a neural architecture parametrized by $\theta$. 
\end{definition}

The most famous family of neural DEs is that of \textbf{neural ODEs}\index{neural ODE|see {neural differential equation}}\index{neural differential equation!neural ODE}, which consider ordinary differential equations in Equation~\ref{eq:neuralODE}.

\begin{figure}
    \centering
    \includegraphics[width=0.7\textwidth]{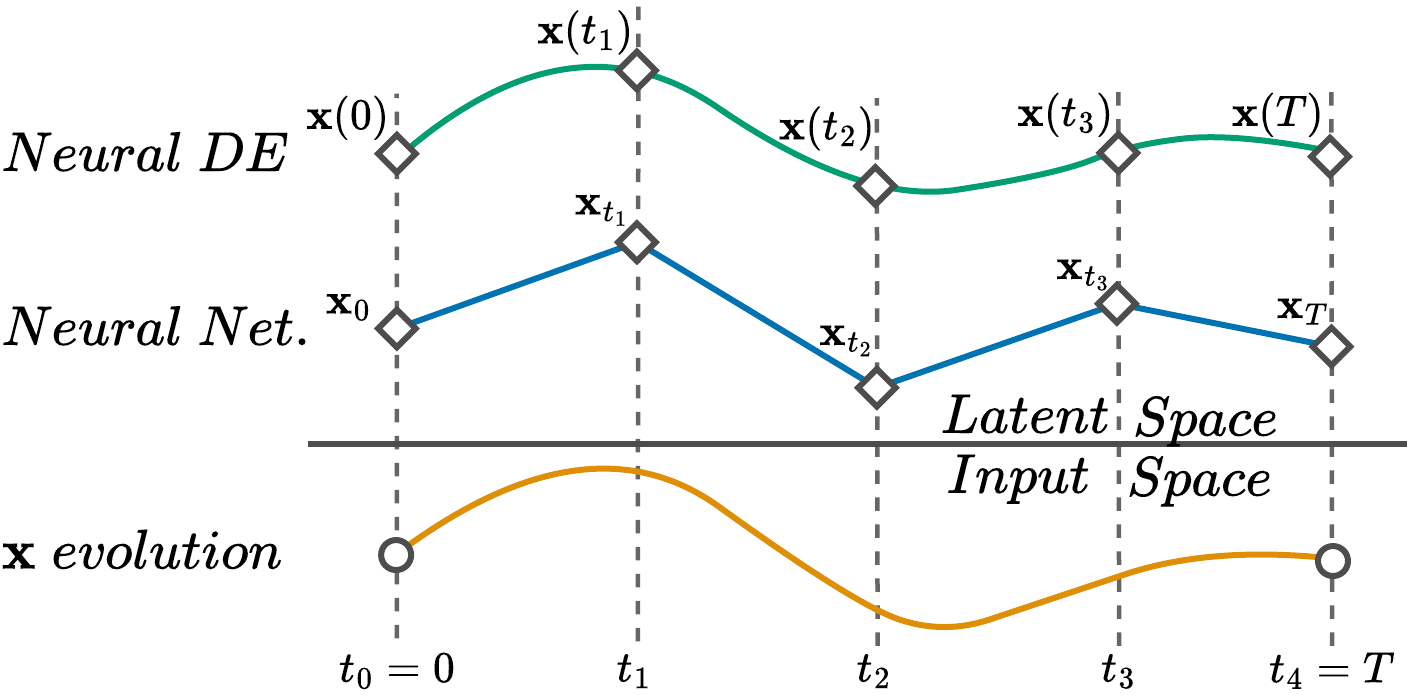}
    \caption{At the top the continuous neural DE hidden state, in the middle the time-discrete hidden state of neural networks (\eg ResNet), and at the bottom the real state evolution.}
    \label{fig:hidden_state}
\end{figure}

With the aim of drawing a link between neural networks and differential equations, we follow \cite{neuralODE} and note that ResNets, RNNs, and other similar architectures compute their outputs by composing a sequence of
transformations to a hidden state:
\begin{equation}
    \bfx_{t+1} = \bfx_t + f_\theta(\bfx_t). 
\end{equation}
Interestingly, these iterative updates can be seen as an Euler discretization of a continuous transformation.
As an example, we consider the ODE 
\begin{equation}\label{eq:exaple_ode}
\frac{d \bfx(t)}{dt} = \tanh(\theta\bfx(t))
\end{equation}
and its Euler discretization
\begin{equation}\label{eq:exaple_discretization}
    \bfx_{t+1} = \bfx_t + \epsilon\tanh(\theta\bfx(t)). 
\end{equation}
Equation~\ref{eq:exaple_discretization}~\footnote{We note that in Equation~\ref{eq:exaple_discretization} we slightly changed the notation with respect to Section~\ref{sec:discretization_method_de} to better highlight the similarity between discretized ODEs and neural architectures.} can be interpreted as one layer of a residual network or a recurrent network without input-drive data. Here, $\bfx_{t}$ represents the hidden state at the $t$-th step of the network (\ie $t$-th layer), $\theta$ is the model weight, and the step size $\epsilon$ is a hyperparameter.
Each discretization step 
can be equated as one layer of the network, thus providing a framework that maps neural architectures into discretized ODEs. Therefore, Equation~\ref{eq:exaple_ode} is the continuous interpretation of a neural architecture, which  computes the final hidden state $\bfx(T)$ starting from the initial condition $\bfx(0)$. This interpretation is visually represented in Figure~\ref{fig:hidden_state}. 

In summary, we can reinterpret neural architectures as neural DEs by establishing a correspondence between the layers of the neural architecture and the successive iterations of a discretization method. This reinterpretation involves representing the transformation applied by each layer of the neural network as the evolution of a continuous dynamical system governed by a differential equation.
Therefore, neural DEs are a powerful tool for deep learning. This family of models exploits the flexibility and generalization capabilities of neural networks while inheriting properties and theoretical understanding from the differential equation domain. Therefore, it provides a coherent theory for model design. In other words, thanks to neural DEs, we can design neural architectures based on differential equations to have desirable properties inherited by the theory of dynamical systems.

\section{Fundamentals of DGNs}\label{sec:static_dgn_fundamentals}
In this section, we introduce fundamental concepts about representation learning for graphs and deep graph networks, mainly taken from \citet{gravina_dynamic_survey} and \citet{gravina_swan}
, which will be used throughout the rest of this thesis. 
\\

In recent years, representation learning for graphs has emerged as a vibrant research area at the intersection of graph theory and machine learning. Graphs, which serve as powerful models for representing complex relationships and interactions among entities, are ubiquitous in various domains, as emerged in Section~\ref{sec:graphs}. The ability to effectively learn meaningful representations of nodes, edges, and entire graphs is fundamental for numerous downstream tasks, including node classification, link prediction, and graph classification.

Unlike traditional data structures such as vectors or sequences, graphs pose unique challenges for representation learning due to their irregular and heterogeneous nature. Nodes within a graph can exhibit diverse characteristics and may interact with their neighbors in complex ways, making it non-trivial to capture the underlying structural and semantic information. 
Therefore, the key challenge when learning from graph data is how to numerically represent the combinatorial structures for effective processing and prediction by the model. A classical predictive task of molecule solubility prediction, for instance, requires the model to encode both topological information and chemical properties of atoms and bonds. Graph representation learning solves the problem in a data-driven fashion, by \textit{learning} a mapping function that compresses the complex relational information of a graph into an information-rich feature vector that reflects both structural and label information in the original graph.


Representation learning for graphs has been pioneered by Graph Neural Network (GNN) \citep{GNN} and Neural Network for Graphs (NN4G) \citep{NN4G}, which were the first to provide learning models amenable also for cyclic and undirected graphs. The GNN leverages a \textit{recursive} approach, in which the state transition function updates the node representation through a diffusion mechanism that takes into consideration the current node and its neighborhood defined by the input graph. This procedure continues until it reaches a stable equilibrium. On the other hand, the NN4G leverages a \textit{feed-forward} approach where node representations are updated by composing representations from previous layers in the architecture.

\begin{figure}[ht]
\centering 
\includegraphics[width=0.95\textwidth]{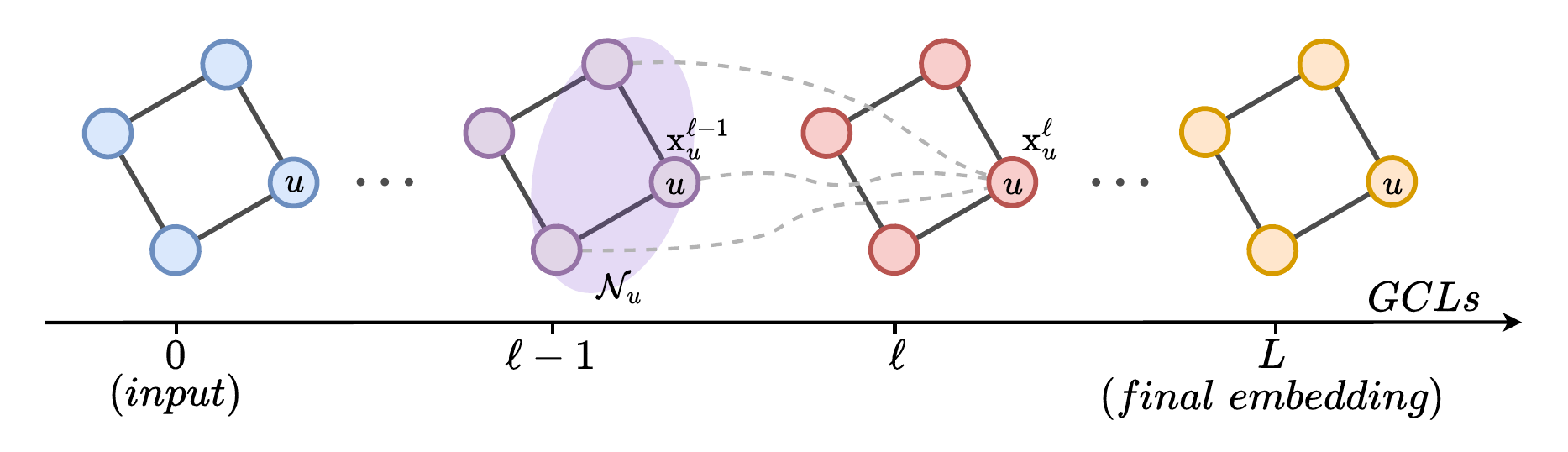}
\caption{Visual representation of a DGN wih $\gls*{Layers}$ layers. Given the input graph, the $(\ell + 1)$ GCL computes the new representation of a node $u$ as a transformation of $u$ and its neighbors representations at the previous layer, $\ell$.}\label{fig:GCLs}
\end{figure}

The original approaches by NN4G and GNN have been later extended and improved throughout a variety of approaches, which can be cast under the umbrella term of (static) \textbf{Deep Graph Networks}\index{deep graph network} (\gls*{DGN}s)~\citep{BACCIU2020203, GNNsurvey}. DGNs denote a family of approaches capable of learning the functional dependencies in a graph through a layered approach, where the single layers are often referred to as \textbf{graph convolutional layers}\index{graph convolutional layer} (\gls*{GCL}s). Each of these computes a transformation (often referred to as \textbf{graph convolution}\index{graph convolution}) of node representations by combining the previous node representations and their neighborhoods, following the message passing paradigm. We refer to the updated node representations resulting from the graph convolution as \textbf{node embedding}\index{node embedding} or \textbf{node (latent) state}\index{node state|see {node embedding}}. We visually represent this procedure in Figure~\ref{fig:GCLs}.

The number of GCLs plays a crucial role in determining the distance over which information propagates within the graph. 
As the number of layers increases, information is propagated further across the graph. 

\begin{figure}[ht]
\centering 
\includegraphics[width=0.8\textwidth]{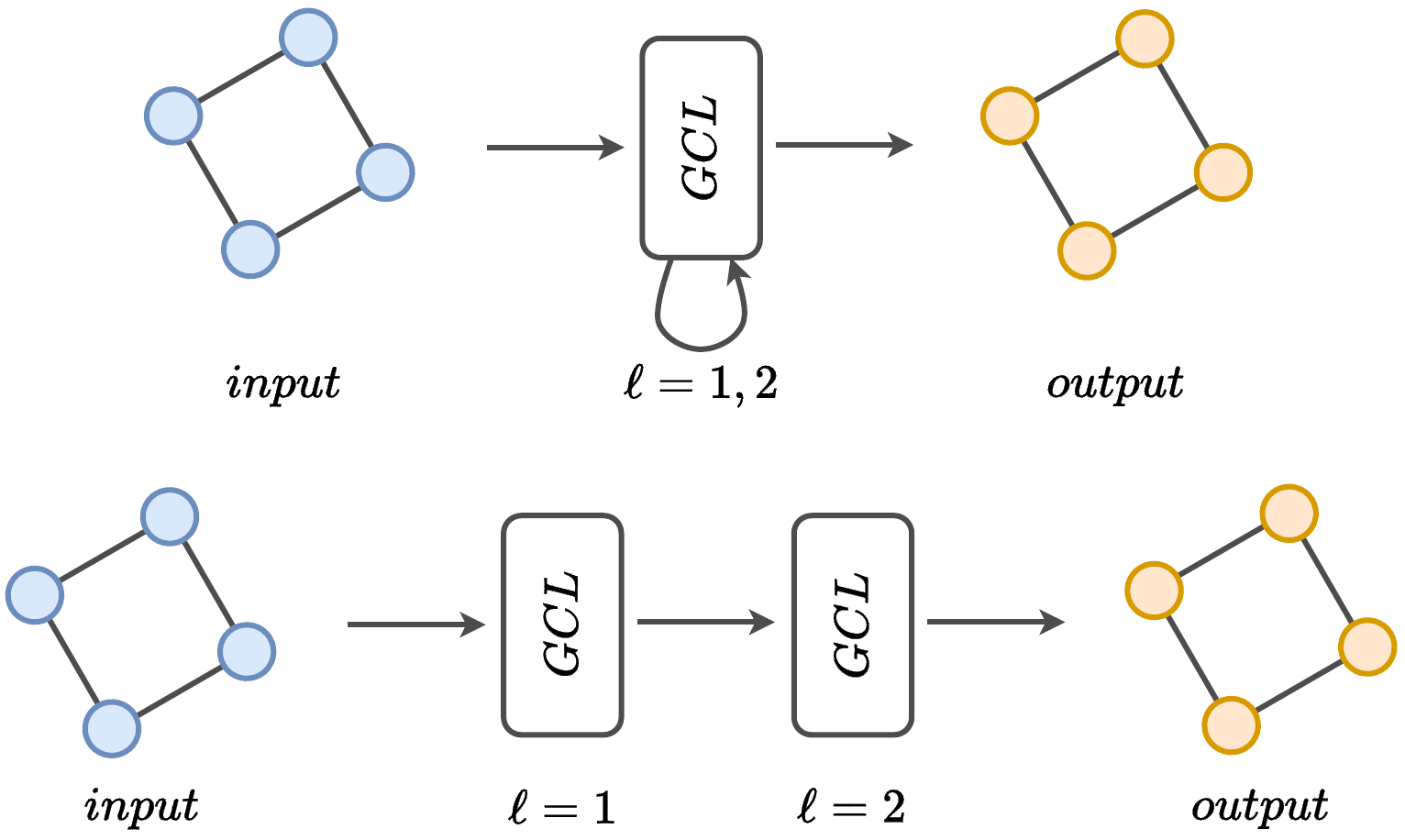}
\caption{A Deep Graph Network with 2 layers, $\ell$, showcasing recurrent and feed-forward architectures. In the recurrent architecture (at the top) a single shared layer is iteratively unfolded for 2 steps, as indicated by the recurrent connection. At the bottom, the feed-forward architecture leverages two Graph Convolutional layers to process the input graph.}\label{fig:rec_vs_feedfor}
\end{figure}

Depending on the use of GCLs we can distinguish between \textbf{recurrent} or \textbf{feed-forward} architectures. In the former case, a single parametrized GCL is unfolded iteratively for a number of steps equal to the desired number of layers. Thus, the layer is shared across the DGN, and the layer parametrization is the same across multiple steps. We refer to this architecture as implementing recurrence with \textbf{weight sharing}\index{weight sharing}. On the other hand, the feed-forward architecture employs multiple GCLs, each of which implements a different parametrization. We refer to this architecture as implementing \textbf{layer-dependent weights}\index{layer-dependent weights}. Following the line of reservoir computing and randomized networks~\citep{TANAKA2019100, randomizedNN}, we can also identify \textbf{randomized} DGNs in which the layers' parametrization is randomly initialized and kept fixed after initialization. We refer to this architecture as implementing \textbf{random weights}\index{random weights}.
Figure~\ref{fig:rec_vs_feedfor} visually represents the difference between recurrent and feed-forward architectures.

Finally, it is noteworthy that the implementation of each GCL, \ie the graph convolution, significantly influences the overall behavior of the model as it shapes the way node embeddings are learned. In fact, different graph convolutions determine distinct patterns in information propagation across a graph. 
The easiest implementation of a GCL is the one that updates the node state at layer $\gls*{l}+1$ as the sum of neighbors' representations at the previous layer
\begin{equation}\label{eq:simple_gcl}
    \gls*{xul1} = \sum_{v \in \mathcal{N}_u} \bfx^{\ell}_v.
\end{equation}
Form the graph point of view, Equation~\ref{eq:simple_gcl} can be rewritten as
\begin{equation}\label{eq:simple_gcl_graph}
    \bfX^{\ell+1} = \bfA\bfX^{\ell}=\bfA^\ell\bfX^0
\end{equation}
where $\bfA$ is the adjacency matrix, and $\gls*{Xl}$ is the node state matrix at layer $\ell$, $\bfA^\ell$ is the $\ell$-th power of the adjacency matrix, and $\bfX^0$ is the input feature matrix. We observe that in this case, the adjacency matrix assumes the role of \textbf{graph shift operator}\index{graph shift operator}, which is an operator that delineates the transformation of the signal as it traverses the graph, thus it captures the dynamic of the information flow. Additional instances of graph shift operators include the Laplacian matrix or its normalized variants.

In the following sections, we analyze graph convolutions that are realized either in the \textit{spectral} or \textit{spatial} domain. For the sake of completeness, we also review two alternative approaches for learning node embeddings, which are based on random walks and graph rewiring. The taxonomy behind our surveying methodology is depicted in Figure~\ref{fig:taxonomy_static}. We observe that graph rewiring is not displayed in the figure since it is a preprocessing technique for enhancing the downstream DGN, and it does not implement a new graph convolution method. 

Before analyzing such methods, we first define the common plights of graph learning models, such as oversmoothing, oversquashing and underreaching.

\tikzset{SG/.style={sibling distance = 2.5cm}}
\tikzset{DT/.style={sibling distance = 9cm}}
\tikzset{CT/.style={sibling distance = 3cm}}

\begin{figure}[h]
\centering
\scalebox{0.8}{
\begin{tikzpicture}
[
    level 1/.style = {very thick, sibling distance = 8cm, anchor=north,level distance=1cm},
    edge from parent path=
{(\tikzparentnode.south) .. controls +(0,-0.3) and +(0,0.5)
                           .. (\tikzchildnode.north)}]

\node {\textbf{\textit{Deep Graph Networks}}}
    child {node {\textbf{\textit{Static graphs}}}
        child[SG] {node[align=center] {\textbf{\textit{Spectral}}\\\small{(Sec \ref{sec:spectral_conv})}\\}}
        child[SG] {node[align=center] {\textbf{\textit{Spatial}}\\\small{(Sec \ref{sec:spatial_conv})}\\}}
        child[SG] {node[align=center] {\textbf{\textit{Random walks}}\\\small{(Sec \ref{sec:random_walks})}\\}}
        edge from parent []
    } 
    child[lightgray] {node[align=center] {\textbf{\textit{Dynamic graphs}}\\\small{(Ch \ref{ch:learning_dyn_graphs})}}};
\end{tikzpicture}
}
\caption{Taxonomy employed to structure our survey of DGN models for static graphs. Graph rewiring is not displayed in the figure since it is a preprocessing technique for enhancing the power of the downstream DGN.} \label{fig:taxonomy_static}
\end{figure}
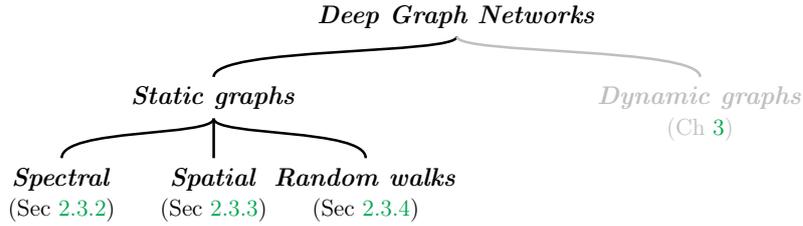

\subsection{Oversquashing, Oversmoothing, and Underreaching}\label{sec:dgn_plights}
Oversquashing~\citep{alon2021oversquashing,topping2022understanding, diGiovanniOversquashing},  oversmoothing~\citep{cai2020note, oono2020graph, rusch2023survey}, and underreaching~\citep{alon2021oversquashing} are common plights of DGNs. Although DGNs have achieved remarkable success in multiple domains \citep{bioinformatics,social_network,GNNsurvey, google_maps,  dwivedi2023benchmarking},  these challenges restrict their overall effectiveness. \textbf{Oversquashing}\index{oversquasing} refers to the shortcoming of a DGN when transferring information between distant nodes. This shortcoming typically increases in tandem with the distance between the nodes, which hampers the ability of DGNs to model complex behaviors that require long-range interactions, such as protein folding \citep{jumper2021}.
To allow a node to receive information from $L$-hops distant nodes, a DGN must employ at least $L$ layers, otherwise, it will suffer from \textbf{underreaching}\index{underreaching}, because the two nodes are too far to interact with each other. However, the stacking of multiple GCLs also causes each node to receive an exponentially growing amount of information, as multiple hops are considered, since each node state update incorporates neighborhood information. This exponential growth of information, combined with the finite and fixed number of channels (features), can lead to a potential loss of information. As a consequence, the long-range effectiveness of DGNs is reduced.

Lastly, \textbf{oversmoothing}\index{oversmoothing} is defined as the tendency for node embeddings to converge to an extremely small subspace (\ie becoming indistinguishable) as the number of layers increases. Thus, oversmoothing diminishes the ability of the DGN to distinguish between different nodes, as node embedding's similarity increases. A common metric to measure oversmoothing is the \emph{Dirichlet energy}\index{Dirichlet energy}~\citep{cai2020note} computed over the set of node embeddings at a specific layer $\ell$, $\bfX^\ell$, 
\begin{equation}
    \gls*{energy}(\bfX^\ell) = \frac{1}{|\mathcal{V}|}\sum_{u\in\mathcal{V}}\sum_{v\in\mathcal{N}_u} \|\bfx_u^\ell-\bfx_v^\ell\|^2.
\end{equation}
Therefore, oversmoothing not only diminishes the expressive power of node embeddings, but also prevents the long-range effectiveness of DGNs.

\subsection{Spectral Convolution}\label{sec:spectral_conv}
In the setting of \textit{spectral convolution}\index{graph convolution!spectral}\index{spectral convolution|see {graph convolution}}, graphs are processed and learned through a parameterization in the spectral domain of their Laplacian matrices. Specifically, given a filter $\bfg_\theta = diag(\theta)$ parametrized by $\gls*{theta} \in \mathbb{R}^{|\mathcal{V}|}$ and the graph signal $\bfx \in \mathbb{R}^{|\mathcal{V}|}$ for a graph $\mathcal{G}$, we can define the spectral graph convolution as a multiplication in the Fourier domain: 
\begin{equation}  
\label{eq:spectral_conv}
    \bfx * \bfg_\theta = \bfU(\bfU^\top \bfx \circ \bfg_\theta) =  \bfU \bfg_\theta \bfU^\top \bfx
\end{equation} 
where $\bfU^\top \bfx$ is the graph Fourier transform, and $\bfU$ is the matrix of eigenvectors of the normalized graph Laplacian $\bfL^{sym} = \bfI - \bfD^{-\frac{1}{2}}\bfA\bfD^{-\frac{1}{2}} = \bfU\Lambda \bfU^\top$ (see Definition~\ref{def:sym_laplacian}), with $\Lambda$ the diagonal matrix of the eigenvalue of $\bfL^{sym}$. 
The approach in Equation~\ref{eq:spectral_conv} is severely limited by the computational requirements associated to the Laplacian decomposition and by the spectral parameterization costs, which have motivated a whole body of followup works~\citep{chebnet, GCN}. Among these, the Graph Convolutional Network (\gls*{GCN})~\citep{GCN} is certainly the most successful one.  
GCN leverages the degree-normalized Laplacian introduced in \citep{chebnet}, hence, the output of the GCN's $(\ell+1)$-th layer for a node $u$ is computed as 
\begin{equation}
    \label{eq:GCN}
    \bfx^{\ell+1}_u = \sigma\left(\gls*{W} \bfx^{\ell}_u + \gls*{Vw} \sum_{v \in \mathcal{N}_u} \frac{\bfx^{\ell}_v}{\sqrt{\mathbf{deg}(v)\,\mathbf{deg}(u)}}\right),
\end{equation}
where $\gls*{sigma}$ is the activation function, while $\mathbf{deg}(v)$ and $\mathbf{deg}(u)$ are, respectively, the degrees of nodes $v$ and $u$. $\bfW\in\mathbb{R}^{\gls*{d}\times d}$ and $\bfV\in\mathbb{R}^{d\times d}$ are the learnable parameters, \ie weight matrices, of the method. With such formulation, GCN requires $\mathcal{O}(|\mathcal{E}|)$ time.

Although GCN has demonstrated effectiveness across several applications \citep{gcn_compvis,gnn_bioinf}, one limitation lies in its tendency for node embeddings to 
become indistinguishable (\ie oversmooth) as the number of layers increase, which can lead to weakened performance. To improve the resilience of GCN to this behavior, GCNII~\citep{gcnii} introduces \emph{initial residual}\index{initial residual} and \emph{identity mapping}\index{identity mapping}. The initial residual constructs a skip connection from the input layer, ensuring that the final embedding of each node retains at least a fraction of the input representation. On the other hand, identity mapping is employed to avoid information loss in the propagation, by preserving node identity as more layers are employed. Thus, Equation~\ref{eq:GCN} can be reformulated as
\begin{equation}
    \bfx^{\ell+1}_u = \sigma\Biggl(\Bigl((1-\beta) \sum_{v \in \mathcal{N}_u \cup \{u\}} \frac{\bfx^{\ell}_v}{\sqrt{\mathbf{deg}(v)\,\mathbf{deg}(u)}} + \beta \bfx^{0}_v\Bigr)\Bigl((1-\gamma)\bfI+\gamma\bfW\Bigr)\Biggr),
\end{equation}
with $\gls*{beta},\gls*{gamma}\in [0,1]$ two hyperparameters.

With a similar objective to GCNII, MixHop~\citep{abu2019mixhop} extends GCN with a \emph{neighbor mixing} strategy, enabling the learning of more informative embeddings that capture higher-order graph information, \ie coming from distant nodes. In this scenario, the updated node state is derived by concatenating several GCN layers (as described in Equation~\ref{eq:GCN}), each computed using a different power of the graph shift operator. This allows the integration of information from nodes at varying distances within the graph.

\subsection{Spatial Convolution}\label{sec:spatial_conv}
Spatial convolutions\index{graph convolution!spatial}\index{spatial convolution|see {graph convolution}} are typically framed in the Message Passing Neural Network (\gls*{MPNN})~\citep{MPNN} framework, where the representation for a node $u$ at a layer $\ell+1$ is computed as 
\begin{equation}
    \label{eq:MPNN}
    \bfx_u^{\ell+1} = \gls*{rhou}(\bfx_u^\ell,\, \gls*{bigoplus}\ \gls*{rhom}(\bfx_u^\ell, \bfx_v^\ell, \mathbf{e}_{uv}))
\end{equation}
where $\bigoplus$ is an aggregation invariant function, and $\rho_U$ and $\rho_M$ are respectively the \textit{update}\index{update function} and \textit{message}\index{message function} functions. The message function computes the message for each node, and then dispatches it among the neighbors. The update function collects incoming messages and updates the node state. A typical implementation of the MPNN use  sum  as $\bigoplus$ and $\rho_U$ functions, and $\rho_M(\bfx_u^\ell, \bfx_v^\ell, \mathbf{e}_{uv}) = \mathrm{MLP}(e_{uv})\bfx_v^\ell$, with \gls*{MLP} implementing a Multi-Layer Perceptron.

Depending on the definition of the update and message functions, it is possible to derive a variety of DGNs. The Graph Attention Network (\gls*{GAT}) \citep{GAT} introduces an \textit{attention mechanism} to learn neighbors' influences computing node representation as
\begin{equation}\label{eq:gat}
    \bfx_u^{\ell+1} = \sigma \left( \sum_{v \in \mathcal{N}_u} \alpha_{uv}\bfW \bfx_v^\ell\right)   
\end{equation}
where $\gls*{auv}$ is the classical softmax attention score between node $u$ and its neighbor $v$.
Similar to GAT, transformer-based approaches~\citep{graphtransformer, dwivedi2021generalization, ying2021transformers,wu2023difformer} utilize an attention mechanism to capture dependencies between nodes in a graph. However, while GAT applies localized attention across small neighborhoods, graph transformers enable the attention mechanism across the entire graph structure. 
In other words, such approaches enable message passing between all pairs of nodes, sidestepping oversquashing at the price of increased computational complexity
. SAN~\citep{san} improves the power of fully-connected graph transformers by separating the treatment of real and non-real graph edges and introducing a learned positional encoding module based on Laplacian eigenvectors and eigenvalues. \emph{Positional encodings}\index{positional encoding} (\gls*{PE}s) offer insights into the spatial location of individual nodes within the graph, ensuring that nodes in close proximity exhibit similar PE values. 
GraphGPS~\citep{graphgps} expands upon this idea by proposing not only to learn positional encodings but also to incorporate structural encodings. \emph{Structural encodings}\index{structural encoding} (\gls*{SE}s) capture the graph or subgraph structure to enhance the expressiveness and generalizability of DGNs. When nodes share similar subgraphs or graphs exhibit resemblance, their SE values should be closely aligned. Therefore, GraphGPS employs PE and SE schemes to enrich node features with local and global graph information.
\\

When graphs are large and dense, \ie $|\mathcal{E}|$ close to $|\mathcal{V}|^2$, it can be impractical to perform the convolution over the node's neighborhood. Neighborhood sampling has been proposed as a possible strategy to overcome this limitation, i.e. by using only a random subset of neighbors to update node representation. GraphSAGE~\citep{SAGE} exploits this strategy to improve efficiency and scale to large graphs.
GraphSAGE updates the representation of a node $u$ by fixing the subset of nodes treated as neighbors, and by leveraging aggregation and concatenation operations:
\begin{equation}
    \bfx_{u}^{\ell+1} = \sigma \Biggl(\bfW \cdot \Bigl[\bfx_u^\ell \,\gls*{concatenation}\, \bigoplus_{v\in \mathcal{N}_{S}(u)}\bfx_v^\ell\Bigr]\Biggr)
\end{equation}
where $\mathcal{N}_S: \mathcal{V} \rightarrow \mathcal{V}$ is the function that computes the fixed subset of neighbors for a node $u$
. Differently, ClusterGCN~\citep{clusterGCN} samples a block of nodes identified by a graph clustering algorithm to restrict the neighborhood dimension.

The way models aggregate neighbors representations to compute node embeddings affects the discriminative power of DGNs. \citet{GIN} showed that most DGNs are at most as powerful as 1-Weisfeiler-Lehman test~\citep{WL}. In particular, Graph Isomorphism Network (GIN) \citep{GIN} has been proven to be as powerful as 1-Weisfeiler-Lehman test by computing node representations as 
\begin{equation}
    \bfx_{u}^{\ell+1} = \mathrm{MLP} \left((1+\gamma)\bfx_u^\ell + \sum_{v \in \mathcal{N}_u} \bfx_v^\ell \right)
\end{equation}
with $\gamma$ as a learnable parameter or a fixed scalar. GINE~\citep{gine} extends GIN by enriching node embeddings with additional domain-specific information extracted from edge features.
\\

More recently, advancements in the field of representation learning for graphs have introduced new architectures that establish a connection between the domains of DGNs and ODEs, with the primary objective of optimizing various aspects of message passing. These new methods exploit the intrinsic properties of ODEs to enhance the efficiency and effectiveness of message passing within DGNs.
Specifically, works like GDE~\citep{GDE}, GRAND~\citep{GRAND}, and DGC~\citep{DGC} propose to view GCLs as time steps in the integration of the heat equation. This perspective allows controlling the diffusion (smoothing) in the network and understand the problem of oversmoothing. Other architectures like PDE-GCN \citep{pdegcn} and GraphCON \citep{graphcon} propose to mix diffusion and oscillatory processes as a feature energy preservation mechanism.

In conclusion, by formulating the propagation of information in graphs as an ODE system, these architectures effectively tackle multiple challenges, such as reducing the computational complexity of message passing and mitigating the oversmoothing phenomena.

In Chapter~\ref{ch:antisymmetry}, we will delve deeper into the analysis of DGNs formulated from the perspective of ODEs. Specifically, we will describe the mutual interpretability between DGNs and ODEs.

\subsection{Random Walks}\label{sec:random_walks}
A different strategy to learn node embeddings including local and global properties of the graph relies on \textit{random walks}\index{random walk|see {walk}}\index{walk!random walk}. Similarly to standard walks (see Section~\ref{sec:static_graph_notation}), a  random walk is a random sequence of edges which joins a sequence of nodes. \citet{DeepWalk} proposed DeepWalk, a method that learns continuous node embedding by modeling random walks as the equivalent of sentences. 
Specifically, the approach samples multiple walks of a specified length for each node in the graph, and then it leverages the SkipGram model~\citep{skipgram} to update node representations based on the walks, treating the walks as sentences and the node representations as words within them.

Node2Vec~\citep{node2vec} improves DeepWalk by exploiting \textit{biased} random walks, \ie we can control the likelihood of revisiting a node in the walk (allowing the walk to be more or less explorative) and bias the exploration of new nodes towards a breath first or a depth first strategy.

\subsection{Graph Rewiring}
\begin{figure}[h]
     \centering
     \begin{subfigure}[b]{0.43\textwidth}
         \centering
         \includegraphics[width=\textwidth]{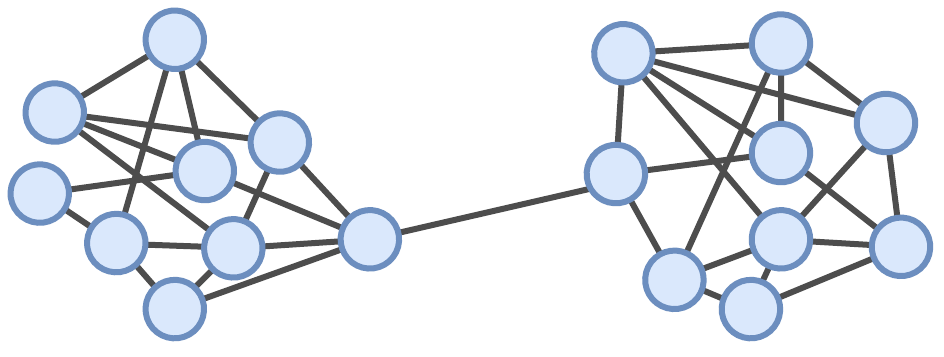}
         \caption{Original graph}
     \end{subfigure}
     \hspace{3mm}
     \begin{subfigure}[b]{0.43\textwidth}
         \centering
         \includegraphics[width=\textwidth]{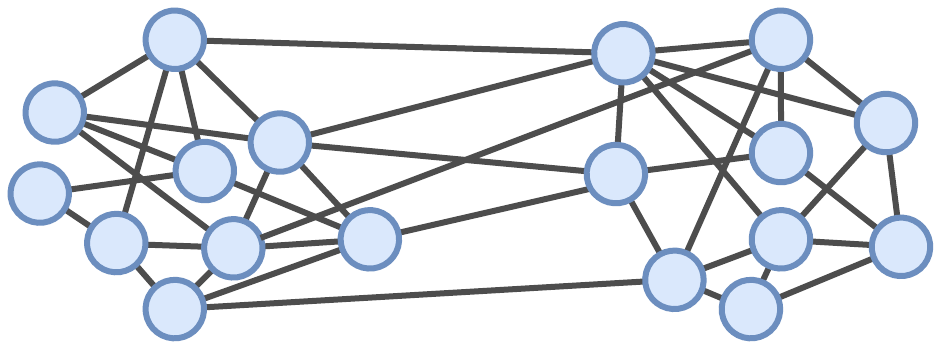}
         \caption{Original graph after rewiring}
     \end{subfigure}
    \caption{Comparison of original graph topology (left) and preprocessed graph after applying graph rewiring (right), illustrating densification of edges to facilitate information flow within the graph.}
    \label{fig:rewiring}
\end{figure}

The issue of oversquashing has been widely recognized due to its impact on the inability of DGNs to effectively transfer information across nodes that are far apart in the graph (refer to Section~\ref{sec:dgn_plights} for a comprehensive discussion on oversquashing). In recent years, the strategy of \emph{graph rewiring}\index{graph rewiring}~\citep{DIGL, shi2023exposition} has been studied as a potential solution to mitigate this issue. Graph rewiring is a technique used to modify the original edge set to densify the graph as a \emph{preprocessing step}, with the aim of enhancing the performance of DGNs. Consequently, graph rewiring allows 1-hop direct connection between nodes that would otherwise be $\ell$-hops away in the original graph, with $\ell\gg1$. 
This preprocessing step aims to facilitate the subsequent operation of the DGN by optimizing the flow of information within the graph. Figure~\ref{fig:rewiring} shows an example of graph rewiring application, comparing the original graph topology with the preprocessed version.

Initial methodologies applying graph rewiring include DIGL~\citep{DIGL} and SDRF~\citep{topping2022understanding}. DIGL leverages the graph heat kernel and personalize PageRank for rewiring, while SDRF employs a curvature-based graph rewiring strategy. In differential geometry, curvatures measure the degree to which the geometry determined by a given metric tensor deviates from a flat space. So, in the graph domain, curvatures identify bottlenecks in the information flow.
SDRF identifies edges with negative curvatures (indicators of potential oversquashing issues) and constructs additional supportive edges around 
them, effectively reinforcing their structural context within the graph.
Similarly, GRAND~\citep{GRAND}, BLEND~\citep{blend}, and DRew~\citep{drew} dynamically adjust graph connectivity based on updated node features.

Despite the success of these techniques in addressing oversquashing, a potential drawback is the increased complexity associated with propagating information at each update, often linked to denser graph shift operators.

\chapter{Learning Dynamic Graphs}\label{ch:learning_dyn_graphs}
In this chapter, building upon the work \citep{gravina_dynamic_survey}, we provide a survey of state-of-the-art approaches in the domain of representation learning for dynamic graphs under our unified formalism defined in Section~\ref{sec:graphs} (the taxonomy behind our surveying methodology is depicted in Figure~\ref{fig:taxonomy}). Finally, we provide the graph learning community with a fair performance comparison among the most popular DGNs for dynamic graphs, using a standardized and reproducible experimental environment
. Specifically, we performed experiments with a rigorous model selection and assessment framework, in which all models were compared using the same features, same datasets and the same data splits. As a by-product of our work, we also provide the community with a selection of datasets which we put forward as good candidates for the benchmarking of future works produced by the community.

\tikzset{L1/.style={sibling distance = 5cm,anchor=north,level distance=1.2cm}}
\tikzset{L2/.style={sibling distance = 10cm,anchor=north,level distance=1.3cm}}
\tikzset{DT/.style={sibling distance = 6.8cm,anchor=north,level distance=1.5cm}}
\tikzset{SG/.style={sibling distance = 2.3cm,text width=1.5cm, anchor=north,level distance=1.2cm}}
\tikzset{GDT/.style={sibling distance = 1.9cm,text width=1.5cm,level distance=1.2cm}}
\tikzset{CT/.style={sibling distance = 2cm,anchor=north,text width=1.5cm, level distance=1.2cm}}

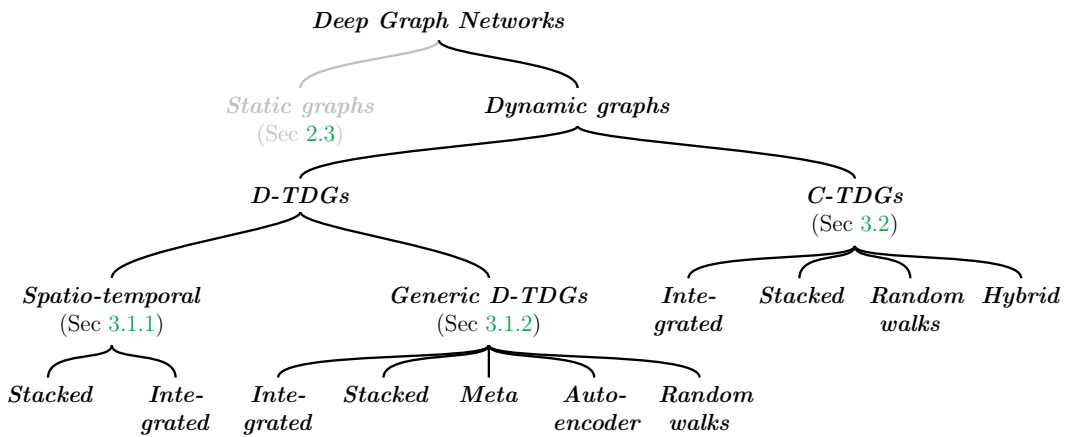
\begin{figure}[h]
\centering
\resizebox{\textwidth}{!}{
\begin{tikzpicture}
[
    level 1/.style = {very thick, sibling distance = 12cm,level distance=1cm},
    edge from parent path=
{(\tikzparentnode.south) .. controls +(0,-0.3) and +(0,0.5)
                           .. (\tikzchildnode.north)}]

\node {\textbf{\textit{Deep Graph Networks}}}
    child[L1,lightgray] {node[align=center] {\textbf{\textit{Static graphs}}\\(Sec \ref{sec:static_dgn_fundamentals})}} 
    child[L1] {node {\textbf{\textit{Dynamic graphs}}}
         child[L2] {node {\textbf{\textit{D-TDGs}}}
            child[DT]{node[align=center] {\textbf{\textit{Spatio-temporal}}\\(Sec~\ref{sec:survey_spatiotemporal})}
                 child[SG] {node[align=center] {\textbf{\textit{Stacked}}}}
                 child[SG] {node[align=center] {\textbf{\textit{Inte\-grated}}}}
            }
            child[DT]{node[align=center] {\textbf{\textit{Generic D-TDGs}}\\(Sec~\ref{sec:survey_generaldtdg})}
                 child [GDT] {node[align=center] {\textbf{\textit{Inte\-grated}}}}
                 child [GDT] {node[align=center] {\textbf{\textit{Stacked}}}}
                 child [GDT] {node[align=center] {\textbf{\textit{Meta}}}}
                 child[GDT] {node[align=center] {\textbf{\textit{Auto\-encoder}}}}
                 child[GDT] {node[align=center] {\textbf{\textit{Random walks}}}}
            }
        }  
        child[L2] {node[align=center] {\textbf{\textit{C-TDGs}}\\(Sec~\ref{sec:survey_ctdg})}
             child[CT] {node[align=center] {\textbf{\textit{Inte\-grated}}}}
             child[CT] {node[align=center] {\textbf{\textit{Stacked}}}}
             child[CT] {node[align=center] {\textbf{\textit{Random walks}}}}
             child[CT] {node[align=center] {\textbf{\textit{Hybrid}}}}
            }
        };
\end{tikzpicture}
}
\caption{Taxonomy employed to structure our survey of DGN models for dynamic graphs.\label{fig:taxonomy}}
\end{figure}

\section{Survey on Discrete-Time Dynamic Graphs}\label{sec:D_TDG}
Given the sequential structure of D-TDGs, a natural choice for many methods has been to extend 
recurrent neural networks (RNNs)~\citep{RNN} to graph data. Indeed, most of the models presented in the literature can be summarized as a combination of static DGNs and RNNs. In particular, some approaches adopt a stacked architecture, where DGNs and RNNs are used sequentially, enabling to separately model spatial and temporal dynamics. Other approaches integrate the DGN inside the RNN, allowing to jointly capture the temporal evolution and the spatial dependencies in the graph. Thus, the primary distinction between static and dynamic approaches lies in their architectures. Static approaches predominantly utilize feedforward or recurrent architectures. Both used for exploring and learning the inherent static graph structure. Differently, dynamic approaches are characterized by recurrent architectures for learning and capturing temporal and spatial dependencies within the evolving graph, which mirrors the increased complexity inherent in addressing dynamic graphs. In the following, we review state-of-the-art approaches for both spatio-temporal graphs and more general D-TDGs (for a detailed description of such graphs, please refer to
Section~\ref{sec:dynamic_graph_notation}).

\subsection{Spatio-Temporal Graphs}\label{sec:survey_spatiotemporal}
When dealing with spatio-temporal graphs, new methods are designed to solve the problem of predicting the node states at the next step, $\overline{\bfX}_{t+1}$, given the history of states, $\overline{\bfX}_{t}$. 
To do so, different types of architectures have been proposed to effectively solve this task.

\myparagraph{Stacked architectures}\index{stacked architectures}
\citet{GCRN} proposed Graph Convolutional Recurrent Network (GCRN), one of the earliest deep learning models able to learn spatio-temporal graphs. The authors proposed to stack a Chebyshev spectral convolution~\citep{chebnet} (Equation~\ref{eq:GCN} shows the first-order approximation of this convolution) for graph embedding computation and a Peephole-LSTM~\citep{peephole-LSTM1, peephole-LSTM2} for sequence learning:
\begin{equation}\label{eq:GCRN}
\begin{split}
    \overline{\bfX}'_t &= \text{Cheb}(\overline{\bfX}_t, \mathcal{E}, k, \bfW)\\
    \bfH_t &= \text{peephole-\textsc{lstm}}(\overline{\bfX}'_t)
\end{split}
\end{equation}
where $\text{Cheb}(\overline{\bfX}_t, \mathcal{E}, k, \bfW)$ represents Chebyshev spectral convolution (leveraging a polynomial of order $k$) computed on the snapshot $\mathcal{G}_t$ parametrized by $\bfW \in \mathbb{R}^{k\times d\times d_n}$. We recall that $d$ is the new latent dimension of node states. $\gls*{Ht}$ is the hidden state vector, which is equivalent to the node states at time $t+1$ (\ie $\overline{\bfX}_{t+1}$). To ease readability, in the following, we
drop from the equation the edge set, $\mathcal{E}$, and the polynomial degree, $k$, since they are fixed for the whole snapshot sequence.

Equation~\ref{eq:GCRN} can be reformulated to define a more abstract definition of a stacked architecture between a DGN and an RNN, \ie
\begin{equation}\label{eq:stacked_architecture}
\begin{split}
    \overline{\bfX}'_t &= \text{\textsc{dgn}}(\overline{\bfX}_t, \bfW)\\
    \bfH_t &= \text{\textsc{rnn}}(\overline{\bfX}'_t)
\end{split}
\end{equation}

\citet{DCRNN} implement Equation~\ref{eq:stacked_architecture} by leveraging the same spectral convolution as GCRN and a Gated Recurrent Unit (\gls*{GRU})~\citep{GRU} as RNN. Differently, \citet{T-GCN} employed the first-order approximation of the Chebyshev polynomials, which lead to the usage of a GCN to learn spatial features, and a GRU to extract temporal features. 

A3TGCN~\citep{a3tgcn} extends the implementation of \citet{T-GCN} with an attention mechanism to re-weight the influence of historical node states, with the aim of capturing more global information about the system.

\myparagraph{Integrated architectures}\index{integrated architectures}
In contrast to the aforementioned approaches, an alternative type of architecture is the one of an integrated architecture, where the DGN is incorporated into the RNN to simultaneously capture and integrate temporal evolution and spatial dependencies within the graph.
\citet{GCRN} proposed a second version of GCRN that exploit this type of architecture, by embedding the Chebyshev spectral convolution in the Peephole-LSTM. In this case, input, forget, and output gates can be reformulated as
\begin{equation}
    \hat{h} = \sigma(\text{Cheb}(\overline{\bfX}_t, \bfW_{x}) + \text{Cheb}(\bfH_{t-1}, \bfW_{h}) + \theta_{c} \odot c_{t-1}),
\end{equation}
the rest of the LSTM is defined as usual. We note that $\gls*{hadamard}$ denotes the Hadamard product, $\sigma$ is the activation function, and $\hat{h}$ if the output of a generic gate. The weights $\bfW_{h} \in \mathbb{R}^{k\times d\times d}$, $\bfW_{x} \in \mathbb{R}^{k\times d\times d_{n}}$, $\theta_{c} \in \mathbb{R}^{d}$
are the parameters of the model. We observe that (here and in the following) the bias term is omitted for ease of readability.

The Spatio-Temporal Graph Convolutional Network~\citep{STGCN} composes several spatio-temporal blocks to learn topological and dynamical features. Each block consists of two sequential convolution layers and one graph convolution in between. The temporal convolution layer contains a 1-D causal convolution followed by a Gated Linear Unit~\citep{GLU}, while the graph convolution is in the spectral domain. Let's consider $\mathrm{Conv}^\mathcal{G}$ the spectral graph convolution, and $\mathrm{Conv}^\mathcal{T}_1$ and $\mathrm{Conv}^\mathcal{T}_2$ the first and second temporal convolutions, respectively. Thus, each spatio-temporal block can be formulated as 
\begin{equation}
    \bfH_t = \mathrm{Conv}^\mathcal{T}_2\left(\, \mathrm{ReLU}(\,\mathrm{Conv}^\mathcal{G}(\,\mathrm{Conv}^\mathcal{T}_1(\overline{\bfX}_t)\,))\right).
\end{equation}

\citet{astgcn} extend the spatio-temporal blocks with an attention mechanism on both spatial and temporal dimensions to better capture the spatial-temporal dynamic of the graph.

\subsection{General D-TDGs}\label{sec:survey_generaldtdg}
Different from spatio-temporal graphs, the topology of a general D-TDG can evolve over time. In this case, the mere exploitation of the sole node states can lead to poor performance, since the new topology leads to different dynamics in the graph. In fact, the evolving topology is responsible for different information flows in the graph over time. Thus, excluding the evolution of the graph structure becomes a major limit of the method, leading to inaccurate predictions. 

Even in this case, we can categorize approaches for general D-TDGs depending on the architectural design.

\myparagraph{Integrated architectures}\index{integrated architectures}
\citet{GC-LSTM} proposed GC-LSTM an en\-coder-de\-coder model for link prediction, assuming fixed the node set $\mathcal{V}(t)$. The encoder consists of a GCN embedded in a standard LSTM. The GCN learns topological features of the cell state $c$ and of the hidden state $h$, which are used to save long-term relations and extract input information, respectively. The encoder takes as input the sequence of adjacency matrices and returns an embedding that encodes both temporal and spatial information. Thus, a generic gate in the LSTM can be expressed as:
\begin{equation}
\hat{h}_t = \sigma(\bfW_{h} \mathbf{A}_t + \text{\textsc{gcn}}_h(\bfH_{t-1}, \mathcal{E}_{t-1}))
\end{equation}
where $\bfW_h \in \mathbb{R}^{|\mathcal{V}| \times d}$ is the weight matrix, and $\gls*{At}$ is the adjacency matrix at time $t$, as usual. The decoder part of the model is an MLP that leverages the embedding generated from the encoder to predict the probability of each edge in future adjacency matrix $\mathbf{A}_{t+1}$. 

A similar strategy, which integrates topological changes into the computation, has also been employed by \citet{LRGCN}.  
Indeed, the authors proposed the so-called LRGCN that embed a Relational-GCN~\citep{RGCN} into a LSTM model.  
Different from GC-LSTM, LRGCN exploits the directionality of the edges in accordance with node features rather than the only stream of adjacency matrices, with the aim of effective modeling of the temporal dynamics. 
In LRGCN the input, forget, and output gates are computed as the result of the R-GCN model over the input node representations and the node embeddings computed at the previous step. The authors distinguish between four edge types to produce more informed latent representation: \textit{intra-incoming}, \textit{intra-outgoing}, \textit{inter-incoming}, \textit{inter-outgoing}. An inter-time relation corresponds to an arc $(u,v)$ present at the previous time $t-1$, while an intra-time relation is an arc $(u,v)$ present at current time $t$. To employ LRGCN in a \textit{path classification} task, the authors extend their model with a self-attentive path embedding (SAPE). Given the representations of $m$ nodes on a path, $P \in \mathbb{R}^{m \times d_o}$ with $d_o$ is the output dimension of LRGCN,
SAPE first applies LSTM to capture node dependency along the path sequence, \ie $\Gamma = \mathrm{LSTM}(P) \, \in \mathbb{R}^{m \times d_{new}}$. Then, SAPE uses the self-attentive mechanism to learn node importance and generate size-invariant representation of the path, \ie 
\begin{equation}
    S=\mathrm{softmax}(\text{\textsc{mlp}}(\mathrm{tanh}(\text{\textsc{mlp}}(\Gamma))))  \, \in \mathbb{R}^{r \times m}
\end{equation} 
with $r$ an hyper-parameter. Lastly, the final path representation is obtained by multiplying $S$ with $\Gamma$, \ie $e=S\Gamma  \, \in \mathbb{R}^{r \times d_{new}}$.

With the aim of speeding up the dynamic graph processing, \citet{dyngesn} propose DynGESN, an extension of the Graph Echo State Network~\citep{gesn} to the temporal domain. Specifically, DynGESN updates the embedding for a node $u$ at time $t$ as
\begin{equation}
    \gls*{ht} = (1-\gamma)\bfh^{t-1}_u + \gamma \text{tanh} \left( \bfW_{i} \bfh_u^{t} +  \sum_{v \in \mathcal{N}^t_u}  \bfW_{r} \gls*{xvt-1}\right),
\end{equation}
with $0<\gamma\leq 1$  being a leakage constant, $\bfW_{i}$ the input weights, and $\bfW_{r}$ the recurrent weights. Both input and recurrent weights are randomly initialized.

\myparagraph{Stacked architectures}\index{stacked architectures}
Instead of integrating the DGN into the RNN, \citet{mpnn_lsltm} stack an LSTM on top of a DGN (in this case an MPNN), as previously proposed for spatio-temporal graphs. Differently from those approaches, the authors leverage as input the new node features as well as the new topology. Thus, the MPNN updates node representations by exploiting the temporal neighborhoods in each snapshot. 
\citet{roland} extend the MPNN-LSTM method by proposing the exploitation of \textit{hierarchical node states}. Thus, the authors propose to stack multiple DGN's layers and interleave them with the sequence encoder, \eg the RNN, to better exploit the temporal dynamic at each degree of computation. Thus, the node state at each layer depends on both the node state from the previous layer and the historical node state. More formally, the  $\ell$-th layer of \citeauthor{roland}'s framework is 
\begin{equation}
\begin{split}
    \tilde{\bfH}^\ell_t &= \text{\textsc{dgn}}^\ell(\tilde{\bfH}_t^{\ell-1})\\
    \gls*{Htell} &= \text{\textsc{update}}(\tilde{\bfH}^\ell_t, \bfH_t^{\ell-1}).
\end{split}
\end{equation}
where \textsc{update} is the sequence encoder. Similarly \citet{10.1145/3292500.3330919} employed GCN to learn spatial dependencies and MLP as update function.

Contrarily from previous works, \citet{cini2023scalable} propose to first embed the history of the node time series into latent representations that encode the temporal dynamic of the system. Such representations are then processed leveraging multiple powers of a graph shift operator (\eg graph Laplacian or adjacency matrix) to encode the spatial dynamic of the system. Specifically, the authors propose to encode the temporal dynamics by means of an Echo State Networks (ESNs)~\citep{esn1, esn2}, a randomized recurrent neural networks\footnote{In randomized neural networks the hidden weights are randomly initialized and kept fixed after initialization. Only the weights in the final readout layer are learned, typically employing highly efficient methods like least-squared minimization~\citep{randomizedNN}.}, to efficiently compute node embedding and improve the scalability of DGNs for D-TDG.

\myparagraph{Meta architectures}\index{meta architectures}
We refer to \textit{meta architectures} as those methods that learn a function that maps the evolution of the graph into the evolution of the parameters of the employed DGN. This kind of architecture has been proposed by \citet{egcn} to deal with those scenarios where nodes may frequently appear and disappear. As observed by the authors, %
such dynamics can be challenging to model with RNN-based models, since they have difficulties in learning these irregular behaviors. 
In such situation, the authors proposed Evolving GCN (E-GCN) to capture 
the dynamism of such graphs by using an RNN to evolve the parameters of a GCN. Thus, only the RNN parameters are trained. The authors considered two versions of their model, depending on whether graph structure or node features play the more important role. The first treats the GCN weights as the hidden state of a GRU to assign more significance to node representations. 
The second computes the weights as the output of the LSTM model, and it is more effective when the graph structure is important for the task. Let's consider $\mathrm{GRU}(\overline{\bfX}_t, \bfW_{t-1})$ as an extended version of a standard GRU model that exploits both the weight matrix at time $t-1$, $\bfW_{t-1}$, and the previous node embedding, $\bfH_t$. The first E-GCN architecture can be formulated as
\begin{equation}
\begin{split}
    \bfW_{t} &= \mathrm{GRU}(\overline{\bfX}_t, \bfW_{t-1})\\
    \bfH_t &= \mathrm{GCN}(\overline{\bfX}_t, \mathcal{E}_t, \bfW_{t})
\end{split}
\end{equation}
while the second substitutes the GRU with an LSTM that takes as input only the weight matrix at time $t-1$.

\myparagraph{Autoencoder architectures}\index{autoencoder architectures}
\citet{dygrae} introduced DyGrAE, an autoencoder for D-TDGs. Specifically, DyGrAE leverages the Gated Graph Neural Network (GGNN)~\citep{gatedGNN} to capture spatial information, and LSTM encoder-decoder architecture to capture the dynamics of the network. GGNN is a DGN similar to the GNN introduced by \citeauthor{GNN}, but with a fixed number of iterations. DyGrAE consists of four components: a GGNN to learn the spatial dynamic; an RNN to propagate temporal information; an encoder to project the graph evolution into a fixed-size representation; and a decoder to reconstruct the structure of the dynamic graph. At each time step, 
at first, DynGrAE computes the snapshot embedding as the result of the average pooling on node embeddings at time $t$, \ie $emb(\mathcal{G}_t) = \mathrm{pool}_{avg}(\mathrm{GGNN}(\overline{\bfX}_{t}))$.
Then, the LSTM encoder-decoder uses the graph embeddings to encode and reconstruct the input graph sequence:
\begin{equation}
    \begin{split}
        \mathrm{encoder}:\, \bfh^{enc}_t &= \mathrm{LSTM}_{enc} (emb(\mathcal{G}_t), \bfh^{enc}_{t-1})\\
        \mathrm{decoder}:\, \bfh^{dec}_t &= \mathrm{LSTM}_{dec} (\Bar{\mathbf{A}}_{t-1}, \bfh^{dec}_{t-1})
    \end{split}
\end{equation}
where $\Bar{\mathbf{A}}_{t-1} = \mathrm{sigmoid}(\text{\textsc{mlp}}(\bfh^{dec}_{t-1}))$ is the reconstructed adjacency matrix at time $t-1$. The decoder uses $ h^{enc}_w$ to initialize its first hidden state, if $w$ is window size. To improve the performance, the authors introduced a temporal attention mechanism, which forces the model to focus on the time steps with significant impact. That mechanism causes the reformulation of the decoder as 
\begin{equation}
    \bfh^{dec}_t = \mathrm{LSTM}_{dec} ([\bfh_{t}^* ||\Bar{\mathbf{A}}_{t-1}], \bfh^{dec}_{t-1})
\end{equation}
where $\bfh_{t}^* = \sum_{i=t-w}^{t-1} \Bar{\alpha}_t^i \bfh_i^{enc}$ is the attention distribution, the attention weights $\Bar{\alpha}_t^i = \mathrm{softmax}(f(\bfh^{dec}_{t-1}, \bfh^{enc}_{i}))$, and $f$ is a function, \eg dot product or MLP.
Dyngraph2vec~\citep{dyngraph2vec} introduces an analogous encoder-decoder approach featuring a deep architecture comprising dense and recurrent layers. This design facilitates the utilization of a more extended temporal evolution for predictions. 

A different strategy has been proposed by \citet{dyngem} that developed DynGEM. Such method handles D-TDGs by varying the size of the autoencoder network depending on a heuristic, which determines the number of hidden units required for each snapshot. Such heuristic, named PropSize, ensures that each pair of consecutive layers, $\ell$ and $\ell+1$, satisfy the condition:
\begin{equation}
\label{eq:propsize}
    size(\ell+1) \geq \mu \cdot size(\ell)
\end{equation}
where $0<\mu<1$ is a hyper-parameter. This heuristic is applied to both encoder and decoder separately. If the condition in Equation~\ref{eq:propsize} is not satisfied for each pair of layers, then the number of $(\ell+1)$'s hidden units are increased. If PropSize is still unsatisfied between the penultimate and ultimate layers, a new layer is added in between. At each time step $t$ and before any application of PropSize, DynGEM initializes model parameters with those of the previous step $\bfW_t = \bfW_{t-1}$. This results in a direct transfer of knowledge between adjacent time steps, which guarantees a higher affinity between consecutive embeddings.

\myparagraph{Random walk based architectures}\index{random walk architectures}
Inspired by DeepWalk and Node2Vec, \citet{evolve2vec} propose a random walk approach designed for D-TDGs named Evolve2Vec. Given a sequence of graph snapshot, \citeauthor{evolve2vec} consider old interactions to contribute only in the propagation of topological information, while they use more recent interactions to encode the temporal dynamic. Thus, they proceed by aggregating old snapshots as a unique static graph.
Evolve2Vec starts random walks from all nodes with at least one outgoing edge in the static graph, as discussed in Section~\ref{sec:random_walks}. Then, in the temporal part, each walker move to a new neighbor if there is at least an outgoing edge in the current snapshot, otherwise it remains in the current node until an outgoing edge is added. Depending on how threshold between old and new is set, we can interpolate between a fully static or fully dynamic approach. After the computation of the random walks, node embeddings are computed by feeding the walks into a skip-gram model, as usual.

\section{Survey on Continuous-Time Dynamic Graphs}\label{sec:survey_ctdg}
In a scenario where the dynamic graph is observed only as new incoming events in the system, the methods defined in Section~\ref{sec:D_TDG} are unsuitable. In fact, approximating a C-TDG through a sequence of graph snapshots can introduce noise and loss of temporal information, since snapshots are captured at a more coarse level, with consequent performance deterioration. Moreover, the previously discussed methods usually do not allow including the time elapsed since the previous event. The majority of such methods update the embeddings only when new events occur. However, depending on how long it passed since the last event involving a node may result in the staleness\index{staleness problem} of the embedding. 
Intuitively, the embedding may change depending on the time elapsed since the previous event. For such reasons, new techniques have been introduced to handle C-TDGs. We classify literature approaches into four categories depending on the architectural choices.

\myparagraph{Integrated architectures}\index{integrated architectures}
\citet{jodie} proposed JODIE, a method that learns embedding trajectories to overcome the staleness problem. JODIE computes the projection of a node $u$ in a future timestamp $t$ as an element-wise Hadamard product of the temporal attention vector with the previous node embedding,
\begin{equation}
    \Hat{\bfx}_u(t) = (1+\mathbf{w}) \odot \gls*{xult} 
\end{equation}
where $(1+\mathbf{w})$ is the temporal attention vector, $\mathbf{w} = \bfW_p\Delta t$ is the context vector, and $\Delta t = t - t^-_u$ is the time since the last event involving $u$. Thanks to the projection, JODIE can predict more accurately future embeddings, thus new events. Similar to other models, when an interaction event occurs between nodes $u$ and $v$, JODIE computes the embeddings $\bfx_u$ and $\bfx_v$ by leveraging two RNNs. 

\citet{dyrep} proposed DyRep, a framework that  update the representation of a node as it appear in an event in the C-TDG. DyRep captures the continuous-time dynamics leveraging a temporal point process approach. A temporal point process is characterized by the conditional intensity function that models the likelihood of an event to happen given the previous events. 
 DyRep's conditional intensity function, computed for an event between nodes $u$ and $v$ at time $t$, is:
\begin{equation}
    \lambda_{uv}^k(t) = f_k(g^k_{uv}(t^-))
\end{equation}
where $k$ is the event type, $\gls*{t-}$ is the previous timestamp in which an event occur, and 
\begin{equation}
    f_k(z) = \theta_k \log\left(1 + \exp\left(\frac{z}{\theta_k}\right)\right)
\end{equation}
with $\theta_k$ a parameter to be learned. The inner function 
\begin{equation}
    g^k_{uv}(t^-) = \omega_k^T \cdot [\bfx_u(t^-) || \bfx_v(t^-)] 
\end{equation}
is a function of node representations learned through a DGN, with $\omega_k \in \mathbb{R}^{2|F|}$ the model parameters that learn time-scale specific compatibility. Node embeddings computed by the DGN are updated as
\begin{equation}\label{eq:dyrep_dgn}
    \bfh_u(t) = \sigma(\bfW_{i} \bfh_u^{loc}(t^-) + \bfW_{r} \bfh_u(t^-_u) + \bfW_{e}(t - t^-_u))
\end{equation}
where $h_u^{loc}(t^-) \in \mathbb{R}^{d}$ is the representation of the aggregation of $u$'s direct neighbors, $t^-_u$ 
is the timestamp of the previous event involving node $u$, and $\bfW_{i}, \bfW_{r}  \in \mathbb{R}^{d\times d}$ and $\bfW_{e} \in \mathbb{R}^{d}$ are learnable parameters. In Equation~\ref{eq:dyrep_dgn} the first addend propagates neighborhood information, the second self-information, while the third considers the exogenous force that may smoothly update node features during the interval time. To learn $\bfh_u^{loc}(t^-)$, DyRep uses an attention mechanism similar to the one proposed in the GAT model (see Section~\ref{sec:static_dgn_fundamentals} for more details). In this case, the attention coefficient is parametrized by $\mathcal{S} \in \mathbb{R}^{|\mathcal{V}| \times |\mathcal{V}|}$, which is a stochastic matrix denoting the likelihood of communication between each pair of nodes. $\mathcal{S}$ is updated according to the conditional intensity function. The aggregated neighborhood representation is 
\begin{equation}
    \bfh_u^{loc}(t^-) = \max(\{\sigma(\alpha_{uv}(t) \cdot \bfh_v(t^-)) \,|\, v \in \mathcal{N}_u^t) \}),
\end{equation}
with $\sigma$ the activation function and $\alpha_{uv}(t)$ the attention factor, as usual.

\myparagraph{Stacked architectures}\index{stacked architectures}
In the case of sequential encoding of spatial and temporal information, \citet{TGAT} introduce TGAT, a model that learns the parameters of a continuous function that characterize the continuous-time stream. Similar to GraphSAGE and GAT models, TGAT employs a local aggregator that takes as input the temporal neighborhood and the timestamp and computes a time-aware embedding of the target node by exploiting an attention mechanism. 
The $\ell$-th layer of TGAT computes the temporal embedding of node $u$ at time $t$ as
\begin{equation}
    \gls*{hult}= \text{\textsc{mlp}}^\ell_2(\mathrm{ReLU}(\text{\textsc{mlp}}^\ell_1([\hat{\bfh}(t) || \overline{\bfx}_u])))
\end{equation}
where $\hat{\bfh}(t)$ is the attentive hidden neighborhood representation obtained as
\begin{equation}
    \begin{split}
        \mathbf{q}(t) &= \left[\mathbf{Z}(t)\right]_0 \bfW_q\\
        \mathbf{K}(t) &= \left[\mathbf{Z}(t)\right]_{1:n} \bfW_K\\
        \mathbf{V}(t) &= \left[\mathbf{Z}(t)\right]_{1:n} \bfW_V\\
        \hat{\bfh}(t) &= \mathrm{attn}(\mathbf{q}(t), \mathbf{K}(t), \mathbf{V}(t)).
    \end{split}
\end{equation}
$\mathbf{Z}(t)= \left[\bfx^{\ell-1}_u(t) || \Omega_d(0), ..., \bfx^{\ell-1}_{v}(t) || \Omega_d(t-t_v)\right] \in \mathbb{R}^{(d+d_{t}) \times n}$ is the temporal feature matrix that concatenates the representation of each node in the neighborhood of $u$ with the time difference between the current time and the time of the previous event involving the neighbor. We consider $v\in \mathcal{N}_u^t$ and $n$ the size of $u$'s neighborhood.  $\mathbf{q}(t)$, $\mathbf{K}(t)$, and $\mathbf{V}(t)$ are the query, key and value projections of the matrix; and \emph{attn} is an attention mechanism similar to GAT. The dimensional functional mapping $\Omega_d: t \rightarrow \mathbb{R}^{d_t}$ is defined as
\begin{equation}\label{eq:time_enc}
    \Omega_d(t) = [\cos(\omega_1t)\sin(\omega_1t), ..., \cos(\omega_dt)\sin(\omega_dt)]
\end{equation}
where $\omega_i$ are learnable parameters.

Differently, \citet{streamgnn} proposed an approach, named StreamGNN, to learn the node embedding evolution as new edges appear in the dynamic graph. Thus, it is design to only deal with interaction events.  StreamGNN is composed of two main components: the update component, which is responsible for updating the node representations of the source and destination nodes of the new link; and the propagation component, which propagates the new event across the direct neighborhood of the involved nodes. When a new event is observed, the update component computes the representation of the event as the result of an MLP on the node representation of both source and destination. Then, such representation is updated by an LSTM to include historical information from previous interactions. The amount of the past node history used by the LSTM is inversely proportional to the time difference with the previous node interaction. Then, the lastly computed node embeddings of source and target nodes are merged with the output of the LSTM model.  
After these first steps, the propagation component diffuse the computed representations across the 1-hop neighborhood by leveraging an attention mechanism and by filtering out those neighbors which appear in an interaction before a predefined threshold.

\citet{tgn_rossi2020} extend previous concepts by proposing Temporal Graph Network (\gls*{TGN}), a general framework composed of five core modules: memory, message function, message aggregator, memory updater, and the embedding module. The memory at time $t$ is a matrix $\mathbf{s}(t)$ that has the objective of representing the node's history in a vectorial format. For this purpose, it is updated after every event. The message function has the role of encoding the event to update the memory module. Given an interaction event involving nodes $u$ and $v$ at time $t$, the message function computes two messages
\begin{equation}
    \begin{split}
    m_u(t) &= \mathrm{msg}_{src}(\mathbf{s}_u(t^-), \mathbf{s}_v(t^-), \Delta t,  \mathbf{e}_{uv}(t))\\
    m_v(t) &= \mathrm{msg}_{dst}(\mathbf{s}_v(t^-), \mathbf{s}_u(t^-), \Delta t,  \mathbf{e}_{uv}(t)),
    \end{split}
\end{equation}
where $\mathrm{msg}$ can be any learnable function, \eg a MLP. In case of a node event, it is sent a single message. The message aggregator is a mechanism to aggregate messages computed at different timestamps. It can be a learnable function, \eg RNN, or not, \eg message average or most recent message. After every event involving a node $u$, the memory of the node is updated by the memory updater as
\begin{equation}
    \mathbf{s}_u(t) = \mathrm{mem}(\Bar{m}_u(t), \mathbf{s}_u(t^-))
\end{equation}
where $\Bar{m}_u(t)$ represents the 
aggregation of computed messages in a batch related to node $u$, and $\mathrm{mem}$ is an RNN. Lastly, the embedding module generates the representation for a node $u$ at time $t$ by exploiting the information stored in the memory module of the node itself and its neighborhood up to time $t$
\begin{equation}
    \bfh_u(t) = \sum_{v \in \mathcal{N}_u^t} f(\mathbf{s}_u(t),  \mathbf{s}_v(t),  \overline{\bfx}_u(t),  \overline{\bfx}_v(t),  \mathbf{e}_{uv})
\end{equation}
with $f$ a learnable function and $\overline{\bfx}_u(t)$, $\overline{\bfx}_v(t)$ the input node representations of nodes $u$ and $v$.

\myparagraph{Random walk based architectures}\index{random walk architectures}
Even in the scenario of C-TDGs, it is possible to compute node embeddings relying on random walks. Differently from a standard random walk, in the continuous-time domain a valid walk is a sequence of interaction events with a non-decreasing timestamp. \citet{temporal_node2vec} extended the Node2Vec framework to exploit temporal random walks. Once decided the starting timestamp $t_0$, which is used to \textit{temporally bias} the walk, the framework samples new nodes for the walk by considering the temporal neighborhood. Differently from the general formulation of temporal neighborhood, \citeauthor{temporal_node2vec} apply a threshold to discriminate and filter old neighbors. 
The distribution to sample nodes in the walk can be either uniform, \ie $\mathbb{P}(v) = 1/|\mathcal{N}^t_u|$, or biased. Specifically, the authors proposed two ways to obtain a temporally weighted distribution. Let consider that the random walk is currently at the node $u$. In the first case, a node $v$ is sampled with the probability
\begin{equation}
    \mathbb{P}(v) = \frac{\exp(\mathcal{T}(v) - \mathcal{T}(u))}{\sum_{v' \in \mathcal{N}^t_u}\exp(\mathcal{T}(v') - \mathcal{T}(u))},
\end{equation}
where $\mathcal{T}: \mathcal{V} \rightarrow \mathbb{R}^+$ is the function that given a node return the corresponding timestamp of the event in which the node was involved; while in the second
\begin{equation}
    \mathbb{P}(v) = \frac{\delta(v, \mathcal{T}(v))}{\sum_{v' \in \mathcal{N}^t_u}\delta(v', \mathcal{T}(v'))},
\end{equation}
where $\delta : \mathcal{V}\times \mathbb{R}^+ \rightarrow \mathbb{Z}^+$ is a function that sorts temporal neighbors in descending order depending on time, thus returns a score that biases the distribution towards the selection of edges that are closer in time to the current node.

Instead of temporal random walks, \citet{CAW} exploited \textit{Causal Anonymous Walks} (CAW) to model C-TDGs. A CAW encodes the causality of network dynamics by starting from an edge of interest and backtracking adjacent edges over time. Moreover, a CAW is anonymous because it replaces node identities in a walk with relative identities based on the appearance order. The causality extraction helps the identification of temporal network motif, while node anonymization guarantees inductive learning. Given an edge $\{u,v\}$, the model extracts $M$ walks of length $m$ starting from both $u$ and $v$, and then performs the anonymization step. Afterward, an RNN encodes each walks leveraging two functions. The first consists of two MLPs ingested with the encoding of the correlation between the node $w$ and the sampled walks
\begin{equation}
    f_1(w) = \text{\textsc{mlp}}(g(w, S_u)) + \text{\textsc{mlp}}(g(w, S_v))
\end{equation}
where $S_u$ is the set of sampled walks started from $u$, and $g$ is the function that counts the times a node $w$ appears at certain positions in $S_u$. The second function encodes time as Equation~\ref{eq:time_enc}. All the encoded walks are aggregated through mean-pooling or the combination of self-attention and mean-pooling to obtain the final edge representation.

NeurTWs~\citep{neurtws} extends \citeauthor{CAW}'s work by employing a different sampling strategy for the temporal random walks, which integrates temporal constraints, topological properties, and tree traversals, allowing to sample \textit{spatiotemporal-biased random walks}. These walks prioritize neighbors with higher connectivity (promoting exploration of more diverse and potentially expressive motifs), while being aware of the importance of recent neighbors. Furthermore, NeurTWs replaces the RNN-based encoding approach for walks with a component based on neural ODEs to facilitate the explicit embedding of irregularly-sampled events. 

\myparagraph{Hybrid architectures}\index{hybrid architectures}
\citet{pint} propose to improve the expressive power of methods designed for C-TDGs by leveraging the strengths of both CAW and TGN-based architectures. Thus, by providing a hybrid architecture.
Specifically, the authors observe that for TGN-based architectures, most expressive power is achieved by employing injective embedding module, message aggregator and memory updater functions. On the other hand, the main advantage of CAW is its ability to leverage node identities to compute representative embeddings and capture correlation between walks. However, such approach imposes that walks have timestamps in decreasing order, which can limit its ability to distinguish events. Under such circumstances, the authors propose PINT, an architecture that leverages injective temporal message passing and relative positional features to improve the expressive power of the method. Specifically, the embedding module computes the representation of node $u$ at time $t$ and layer $\ell$ as
\begin{equation}
\begin{split}
    \hat{\bfh}^\ell_u(t) &= \sum_{v \in \mathcal{N}_u^t} \text{\textsc{mlp}}^\ell_{agg}(\bfh^{\ell-1}_v(t) || \mathbf{e}_{uv}) \alpha^{-\beta(t-t^-)}\\
    \bfh^\ell_u(t) &=\text{\textsc{mlp}}^\ell_{upd}(\bfh^{\ell-1}_u(t) || \hat{\bfh}^\ell_u(t))
\end{split}
\end{equation}
where $\alpha$ and $\beta$ are scalar hyper-parameters the node state is initialized with its memory representation, \ie $\bfh^{0}_u(t)=\mathbf{s}_u(t)$. To boost the power of PINT, the authors augment memory states with relative positional features, which include information about the number of existing temporal walks of a given length between two nodes.
\section{The Benchmarking Problem}\label{sec:benchmark}

In this section, we provide the graph learning community with a performance comparison among the most popular DGNs for dynamic graphs. The aim is to support the tracking of the progress of the state-of-the-art and to provide robust baseline performance for future works. To the best of our knowledge, in fact, there are no widely agreed standard benchmarks in the domain of dynamic graphs. For such a reason, nowadays, it is not easy to fairly compare models presented in different works, because they typically use different data and empirical settings. The latter plays a crucial role in the definition of a fair and rigorous comparison, \eg including multiple random weights initialization and hyperparameter search and similar data splits.

With this in mind, we designed three benchmarks to assess models that deal with spatio-temopral graphs, general D-TDGs, and C-TDGs.
Specifically, we evaluated methods for D-TDGs on both link and node prediction tasks while we constrained the evaluation of C-TDG methods to link prediction tasks due to the scarcity of suitable datasets.
To do so, we extended the library PyDGN~\citep{pydgn} to the D-TDG learning setting to foster reproducibility and robustness of results. With the same aim, we developed a Pytorch Geometric~\citep{Fey/Lenssen/2019} based framework to allow reproducible results in the continuous scenario. Lastly, in Appendix~\ref{app:benchmark_data_and_models} we provide the community with a selection of datasets useful for benchmarking future works. An interest reader is referred to SNAP~\citep{snapnets}, TSL~\citep{tsl}, and Network Repository~\citep{nr} for a broader data collections.
We release openly the code implementing our methodology and reproducing our
empirical analysis at \url{https://github.com/gravins/dynamic_graph_benchmark}.

\subsection{Spatio-Temporal Graph Benchmark}
In the spatio-temporal setting, we consider three graph datasets for traffic forecasting, \ie Metr-LA~\citep{DCRNN}, Montevideo~\citep{rozemberczki2021pytorch}, and PeMSBay~\citep{DCRNN}, and Traffic~\citep{LRGCN}. Specifically, 
\begin{itemize}
    \item \textbf{Metr-LA} consists of four months of traffic readings collected from 207 loop detectors in the highway of Los Angeles County every five minutes;
    \item \textbf{Montevideo} comprises one month of hourly passenger inflow at stop level for eleven bus lines from the city of Montevideo;
    \item \textbf{PeMSBay} contains six months of traffic readings collected by California Transportation Agencies (CalTrans) Performance Measurement System (PeMS) every five minutes by 325 traffic sensors in San Francisco Bay Area.
    \item \textbf{Traffic}: consists of traffic data collected over a period of three months, with hourly granularity, from 4,438 sensor stations located in the 7th District of California.
\end{itemize}
For all the three datasets, the objective is to perform \emph{temporal node regression}\index{temporal node regression}, thus, to predict the future node values, $\overline{\bfX}_{t+1}$, given the past graph history, $[\mathcal{G}_i]_{i=1}^t$.

The baseline performance for this type of predictive problems on graphs is based on five spatio-temporal DGNs (\ie A3TGCN~\citep{a3tgcn}, DCRNN \citep{DCRNN}, GCRN-GRU~\citep{GCRN}, GCRNN-LSTM~\citep{GCRN}, TGCN~\citep{T-GCN}), within the aim of assessing both stacked and integrated architectures, and the influence of an attention mechanism.

We designed each model as a combination of three main components. The first is the encoder which maps the node input features into a latent hidden space; the second is the DGN which computes the spatio-temporal convolution; and the third is a readout that maps the output of the convolution into the output space. The encoder and the readout are MLPs that share the same architecture among all models in the experiments.
We performed hyperparameter tuning via grid search, optimizing the Mean Absolute Error (\gls*{MAE}). We perform a time-based split of the dataset which reserves the first 70\% of the data as training set, 15\% of the following data as validation set, and the last 15\% as test set. We trained the models using Adam optimizer for a maximum of 1000 epochs and early stopping with patience of 50 epochs on the validation error. For each model configuration, we performed 5 training runs with different weight initialization and report the average of the results. We report in Table~\ref{tab:benchmark_st_grid} (Appendix~\ref{app:benchmark_hyperparam}) the grid of hyperparameters exploited for this experiment. 

\myparagraph{Results}
In Table~\ref{tab:benchmark_spatio-temp} we report the results on the spatio-temporal-based experiments, including Mean Squared Error (\gls*{MSE}) as an additional metric. 
Overall, DCRNN and GCRN-GRU achieve the better performance on the selected tasks. Interestingly, they both rely on Chebyshev spectral convolution and GRU, but with different architectural structure. Indeed, DCRNN employs a stacked architecture, while GCRN-GRU embeds the DGN into the RNN, enabling a combined modeling of the temporal and spatial information. This result shows that there is not a superior architectural design, in these tasks. However, it seems relevant to include a bigger neighborhood in the computation (\eg by exploiting a larger Chebishev polynomial filter size). Indeed, even if A3TGCN employs an attention mechanism to capture more global information, it is not enough to achieve comparable performance to DCRNN or GCRN-based approaches. Nevertheless, it is noteworthy that the superior performance of these approaches comes at the expense of computational speed, as it is shown in Table~\ref{tab:benchmark_time_st}.
\begin{table}[h]
\centering
\caption{Mean test scores of the spatio-temporal models and std averaged over 5 random weight initializations. MAE is the optimized metric. The lower, the better.}\label{tab:benchmark_spatio-temp}
\scriptsize
\begin{tabular}{lcc|cc}
\toprule
 & \multicolumn{2}{c}{\textbf{Montevideo}} & \multicolumn{2}{c}{\textbf{Metr-LA}} \\
\textbf{Model} & \textbf{MAE} & \textbf{MSE} & \textbf{MAE} & \textbf{MSE} \\\midrule
A3TGCN    & 0.3962$_{\pm0.0021}$       & \textbf{1.0416$_{\pm 0.0047}$} & 0.3401$_{\pm0.0008}$       & 0.3893$_{\pm0.0039}$          \\
DRCNN     & 0.3499$_{\pm0.0006}$       & 1.0686$_{\pm 0.0012}$          & \one{0.1218$_{\pm0.0013}$} & \textbf{0.0960$_{\pm0.0017}$} \\
GCRN-GRU  & \one{0.3481$_{\pm0.0008}$} & 1.0534$_{\pm 0.0062}$          & 0.1219$_{\pm0.0007}$       & 0.0973$_{\pm0.0009}$          \\
GCRN-LSTM &  0.3486$_{\pm 0.0026}$     & 1.0451$_{\pm 0.0099}$          & 0.1235$_{\pm0.0009}$       & 0.0985$_{\pm0.0004}$          \\
TGCN      & 0.4024$_{\pm 0.0022}$      & 1.0678$_{\pm 0.0049}$          & 0.3422$_{\pm 0.0046}$      & 0.3891$_{\pm0.0058}$          \\\midrule
& \multicolumn{2}{c}{\textbf{PeMSBay}}  & \multicolumn{2}{c}{\textbf{Traffic}}  \\
\textbf{Model} & \textbf{MAE} & \textbf{MSE} & \textbf{MAE} & \textbf{MSE}\\\midrule
A3TGCN    & 0.2203$_{\pm 0.0105}$       & 0.2540$_{\pm0.0039}$          & 0.1373$_{\pm 0.0300}$       & 0.0722$_{\pm 0.0053}$\\
DRCNN     & \one{0.0569$_{\pm 0.0004}$} & \textbf{0.0398$_{\pm0.0002}$} & \one{0.0153$_{\pm0.0002}$}  & \textbf{0.0018$_{\pm4\cdot 10^{-5}}$} \\
GCRN-GRU  & 0.0571$_{\pm 0.0005}$       & 0.0404$_{\pm0.0007}$          & \one{0.0153$_{\pm 0.0003}$} & \textbf{0.0018$_{\pm 0.0001}$} \\
GCRN-LSTM & 0.0593$_{\pm 0.0004}$       & 0.0436$_{\pm0.0001}$          & \one{0.0153$_{\pm 0.0006}$} & \textbf{0.0018$_{\pm 0.0001}$} \\
TGCN      & 0.2109$_{\pm 0.0039}$       & 0.2466$_{\pm0.0033}$          & 0.1375$_{\pm 0.0299}$       & 0.0724$_{\pm 0.0052}$\\\bottomrule
\end{tabular}
\end{table}

\begin{table}[h]
\centering
\caption{Average time to execute a forward pass on the whole dataset (measured in seconds) and std of the best configuration of each model in each task in the spatio-temporal setting, averaged over 5 repetitions. The evaluation was carried out on an Intel(R) Xeon(R) Gold 6240R CPU @ 2.40GHz.}\label{tab:benchmark_time_st}
\scriptsize
\begin{tabular}{lcccc}
\toprule
\textbf{Model} & \textbf{Montevideo} & \textbf{Metr-LA} & \textbf{PeMSBay}  & \textbf{Traffic}  \\\midrule
A3TGCN      & \one{1.59$_{\pm0.05}$}   & 95.12$_{\pm0.53}$  & 167.05$_{\pm1.96}$ & 18.81$_{\pm1.04}$\\
DCRNN       & 3.15$_{\pm0.1}$    & 291.44$_{\pm0.76}$ & 366.34$_{\pm2.73}$ & 112.49$_{\pm11.52}$\\
GCRN-GRU    & 4.89$_{\pm0.1}$    & 216.99$_{\pm3.77}$ & 289.38$_{\pm3.38}$ & 32.44$_{\pm1.03}$\\
GCRN-LSTM   & 6.92$_{\pm0.13}$   & 313.27$_{\pm4.09}$ & 534.65$_{\pm5.85}$ & 46.08$_{\pm1.6}$\\
TGCN        & 1.64$_{\pm0.07}$   & \one{95.08$_{\pm1.43}$}  & \one{165.19$_{\pm2.69}$} & \one{17.96$_{\pm0.52}$}\\
\bottomrule
\end{tabular}
\end{table}

\subsection{D-TDG Benchmark}
In the setting of general D-TDGs (\ie both nodes' state and topology may evolve over time), we consider the following datasets:
\begin{itemize}
    \item \textbf{Twitter Tennis}~\citep{twitter_tennis}: a mention graph in which nodes are Twitter accounts and their labels encode the number of mentions between them; 
    \item \textbf{Elliptic}~\citep{elliptic}: a network of bitcoin transactions, wherein a node represents a transaction and an edge indicate the payment flow. Node are also mapped to real entities belonging to licit categories (\eg exchanges, wallet providers, miners, licit services) versus illicit ones (\eg scams, malware, terrorist organizations, ransomware, Ponzi schemes);
    \item \textbf{AS-773}~\citep{as733}: the communication network of who-talks-to-whom defined in a timespan of almost 26 months from the BGP (Border Gateway Protocol) logs;
   \item \textbf{Bitcoin-$\alpha$}~\citep{bc-otc, bc-otc2}: a who-trusts-whom network of bitcoin users trading on the platform \url{http://www.bitcoin-alpha.com}. To convert this graph into a succession of snapshots, we adopted the same daily aggregation strategy as in \cite{egcn}.
\end{itemize}

We use the first two datasets to run node-level tasks. Specifically, similarly to the case of spatio-temporal setting, in Twitter tennis we perform temporal node regression, while in the Elliptic dataset \emph{temporal node classification}\index{temporal node classification}. Therefore, we predict the class associated to the nodes of the snapshot at time $t$ given the past graph history, $[\mathcal{G}_i]_{i=1}^t$.
We employ the last two datasets for \emph{temporal link prediction}\index{temporal link prediction} task, \ie to predict the future topology of the graph given its past history.
\\

In this benchmark we evaluate three different classes of architectures (\ie stacked, integrated and meta) and we show the potential of randomized networks in the tradeoff between performance and complexity. Thus, we consider five DGNs for our experiments: DynGESN \citep{dyngesn}, EvolveGCN-H \citep{egcn}, EvolveGCN-O \citep{egcn}, GC\-LSTM \citep{GC-LSTM}, LRGCN \citep{LRGCN}.

We performed hyperparameter tuning via grid search, optimizing the MAE in the case of node regression, Area Under the ROC curve (\gls*{AUC}) in the case of link prediction, and balanced accuracy (\gls*{B-Acc}) for node classification. We considered the same experimental setting, split strategy, and architectural choice as for the spatio-temporal graphs. In the case of link prediction, we perform negative sampling by randomly sampling non-occurring links from the next future snapshots. We note that in the case of DynGESN, the model employs fixed and randomized weights and only the final readout is trained.
We report in Table~\ref{tab:benchmark_dtdg_grid} (Appendix~\ref{app:benchmark_hyperparam}) the grid of hyperparameters exploited for this experiment. 

\myparagraph{Results}
Table~\ref{tab:benchmark_dtdg_complete_node} and Table~\ref{tab:benchmark_dtdg_complete_link} show the results on general D-TDGs on node-level and link-level tasks, respectively. 
Differently than the spatio-temporal setting, different tasks benefit from different architectures. Indeed, integrating topology's changes (such as in GCLSTM and LRGCN) is more effective in link prediction tasks, while evolving the parameters of the DGN is more beneficial for node-level tasks, since it is more difficult to change the parameters of a static DGN to predict the topological evolution of the system. Notably, DynGESN achieves comparable results by exploiting only few trainable parameters and reduced computational overhead (see Table~\ref{tab:benchmark_time_dtdg}), showing an advantageous tradeoff between performance and complexity. This makes it an ideal choice when the computational resources are limited. 
\begin{table}[h]
\centering
\caption{Mean test scores and std of DGNs for general D-TDGs averaged over 5 random weight initializations on \textbf{node-level} tasks. For MAE and MSE scores, lower values corresponds to better performances, while for B-Acc, AUC and F1 the higher values are better. The optimized metric is \one{colored}.}
\label{tab:benchmark_dtdg_complete_node}
\scriptsize
\begin{tabular}{l  c c | c c c }
\multicolumn{6}{c}{\textbf{Node-level tasks}}\\
\toprule
& \multicolumn{2}{c}{\textbf{Twitter tennis}}                                & \multicolumn{3}{c}{\textbf{Elliptic}}\\
 \textbf{Model} & \textbf{MAE}               & \textbf{MSE}                  & \textbf{AUC}              & \textbf{F1}                 & \textbf{B-Acc}  \\\midrule
DynGESN       & 0.1944$_{\pm0.0056}$         & 0.3708$_{\pm0.0411}$          & \textbf{51.12$_{\pm 1.30}$} & 79.2$_{\pm 19.62}$          & \one{50.56$_{\pm 1.10}$}     \\
EvolveGCN-H   & 0.1735$_{\pm0.0007}$         & 0.2858$_{\pm0.0074}$          & 48.43$_{\pm 2.71}$          & \textbf{92.54$_{\pm 8.39}$} & 49.52$_{\pm 1.55}$     \\
EvolveGCN-O   & 0.1749$_{\pm0.0007}$         & 0.3020$_{\pm0.0111}$          & 45.11$_{\pm 1.68}$          & 90.80$_{\pm 12.67}$         & 49.23$_{\pm 1.03}$     \\
GCLSTM        & \one{0.1686$_{\pm0.0015}$} & 0.2588$_{\pm0.0049}$          & 45.77$_{\pm 1.60}$          & 70.84$_{\pm 30.01}$         & 48.20$_{\pm 1.80}$     \\
LRGCN         & 0.1693$_{\pm0.0014}$         & \textbf{0.2507$_{\pm0.0057}$} & 45.82$_{\pm 3.81}$          & 65.69$_{\pm 20.21}$         & 47.84$_{\pm 3.37}$     \\
\bottomrule
\end{tabular}
\end{table}

\begin{table}[h]
\centering
\caption{Mean test scores and std of DGNs for general D-TDGs averaged over 5 random weight initializations on \textbf{link-level} tasks. The higher, the better. The optimized metric is \one{colored}.}
\label{tab:benchmark_dtdg_complete_link}
\scriptsize
\begin{tabular}{l  c c c | c c c }
\multicolumn{7}{c}{\textbf{Link-level tasks}}\\
\toprule
& \multicolumn{3}{c}{\textbf{AS-773}}                                                                     & \multicolumn{3}{c}{\textbf{Bitcoin $\alpha$}}\\
  \textbf{Model}           & \textbf{AUC}             & \textbf{F1}                  & \textbf{B-Acc}        & \textbf{AUC}               & \textbf{F1}                  & \textbf{B-Acc} \\\midrule
DynGESN      & 95.34$_{\pm 0.04}$         & 79.83$_{\pm 5.27}$           & 82.80$_{\pm 3.40}$           & 97.68$_{\pm 0.12}$           & 69.98$_{\pm 1.57}$           & 76.79$_{\pm 0.93}$ \\
EvolveGCN-H  & 59.52$_{\pm 17.53}$        & 39.85$_{\pm 34.24}$          & 53.72$_{\pm 16.79}$          & 51.35$_{\pm 2.88}$           & 29.55$_{\pm 30.58}$          & 50.69$_{\pm 1.69}$ \\
EvolveGCN-O  & 58.90$_{\pm 17.80}$        & 29.99$_{\pm 37.10}$          & 56.99$_{\pm 13.97}$          & 51.42$_{\pm 2.84}$           & 31.74$_{\pm 29.98}$          & 51.42$_{\pm 2.84}$ \\
GCLSTM       & \one{96.35$_{\pm 0.01}$} & \textbf{91.22$_{\pm 0.13}$}  & \textbf{91.11$_{\pm 0.06}$}  & 97.75$_{\pm 0.17}$           & 91.22$_{\pm 1.38}$           & 91.72$_{\pm 1.11}$ \\
LRGCN        & 94.77$_{\pm 0.23}$         & 89.59$_{\pm 0.33}$           & 89.07$_{\pm 0.34}$           & \one{98.05$_{\pm 0.03}$}   & \textbf{91.33$_{\pm 0.08}$}  & \textbf{91.89$_{\pm 0.07}$} \\
\bottomrule
\end{tabular}
\end{table}

\begin{table}[h]
\centering
\caption{Average time to execute a forward pass on the whole dataset (measured in seconds) and std of the best configuration of each model in each task in the generic D-TDG, averaged over 5 repetitions. The evaluation was carried out on an Intel(R) Xeon(R) Gold 6240R CPU @ 2.40GHz.}\label{tab:benchmark_time_dtdg}
\scriptsize
\begin{tabular}{lcccc}
\toprule
\textbf{Model} & \textbf{Twitter tennis} & \textbf{Elliptic} & \textbf{AS-773}  & \textbf{Bitcoin $\alpha$}  \\\midrule
DynGESN     & \one{0.04$_{\pm2\cdot10^{-3}}$}  & \one{0.03$_{\pm0.01}$}  & \one{0.53$_{\pm0.13}$} &  \one{0.15$_{\pm0.01}$}\\
EvolveGCN-H & 0.38$_{\pm0.01}$   &  21.14$_{\pm0.07}$   & 3.73$_{\pm0.18}$   & 2.33$_{\pm0.03}$\\
EvolveGCN-O & 0.11$_{\pm0.02}$  &  19.5$_{\pm0.25}$    & 2.35$_{\pm0.29}$   & 1.07$_{\pm0.03}$\\
GCLSTM      & 1.08$_{\pm0.42}$  &  1.2$_{\pm0.13}$     & 31.14$_{\pm0.83}$  & 28.37$_{\pm1.59}$\\
LRGCN       & 1.63$_{\pm0.08}$  &  3.28$_{\pm0.28}$    & 21.66$_{\pm2.65}$  & 24.77$_{\pm3.21}$\\
\bottomrule

\end{tabular}
\end{table}

\subsection{C-TDG Benchmark}
In the continuous scenario, we perform our experiment leveraging three datasets:
\begin{itemize}
    \item \textbf{Wikipedia}~\citep{jodie}: one month of interactions (\ie 157,474 interactions) between user and Wikipedia pages. Specifically, it corresponds to the edits made by 8,227 users on the 1,000 most edited Wikipedia pages;
    \item \textbf{Reddit}~\citep{jodie}: one month of posts (\ie interactions) made by 10,000 most active users on 1,000 most active subreddits, resulting in a total of 672,447 interactions;
    \item \textbf{LastFM}~\citep{jodie}: one month of who-listens-to-which song information. The dataset consists of 1000 users and the 1000 most listened songs, resulting in 1,293,103 interactions.
\end{itemize}

For all the datasets we considered the task of \emph{future link prediction}\index{future link prediction}, thus, predicting if a link between two nodes $u$ and $v$ exists at a future time $t$ given the history of past events.

For our experimental purposes, we consider the following DGNs: DyRep \citep{dyrep}, JODIE \citep{jodie}, TGAT \citep{TGAT}, and TGN \citep{tgn_rossi2020}. These methods allow us to evaluate the sequential encoding of spatial and temporal information as well as integrated architectures. Moreover, they allow assessing the contribution of attention mechanism, embedding trajectories, and memory components. We consider as additional baseline EdgeBank~\citep{edgebank} with the aim of showing the performance of a simple heuristic. EdgeBank is a method that merely stores previously observed interactions (without any learning), and then predicts stored links as positive.

We performed hyperparameter tuning via grid search, optimizing the AUC score. We considered the same experimental setting and split strategy as previous experiments. We perform negative sampling by randomly sampling non-occurring links in the graph, as follows: (1) during training we sample negative destinations only from nodes that appear in the training set, (2) during validation we sample them from nodes that appear in training set or validation set and (3) during testing we sample them from the entire node set.

We report in Table~\ref{tab:benchmark_ctdg_grid} (Appendix~\ref{app:benchmark_hyperparam}) the grid of hyperparameters exploited for this experiment. 

\myparagraph{Results}
We report the results of the C-TDG experiments in Table~\ref{tab:benchmark_ctdg_complete}. 
Overall, TGN generally outperforms all the other methods, showing consistent improvements over DyRep and JODIE. This result shows how the spatial information is fundamental for the effective resolution of the tasks. Indeed, an advantage of TGAT and TGN is that they can exploit bigger neighborhoods with respect to DyRep, which uses the information coming from one-hop distance, and JODIE, which only encode the source and destination information. Despite these results, we observe that the temporal information is still extremely relevant to achieve good performance. In fact, the EdgeBank baseline is able to exceed 91\% AUC score by only looking at the graph's history. 
This is even more evident in the LastFM task, which, as observed in \cite{edgebank}, contains more reoccurring edges with respect to Wikipedia and Reddit. Consequently, such a task is comparatively easier to solve by solely exploiting these temporal patterns. Considering that EdgeBank's performance is directly correlated to the number of memorized edges, in this task, it is able to outperform all the other methods.
Lastly, it is worth mentioning that the enhanced performance of TGN and TGAT is accompanied by a trade-off in computational speed, as illustrated in Table~\ref{tab:benchmark_time_ctdg}.
\begin{table}[h]
\centering
\caption{Mean test scores and std of DGNs for C-TDGs averaged over 5 random weight initializations. The higher, the better. The models are trained to maximize the AUC score.}\label{tab:benchmark_ctdg_complete}
\scriptsize
\begin{tabular}{lccc}
\toprule
& \multicolumn{3}{c}{\textbf{Wikipedia}}\\
\textbf{Model}                & \textbf{AUC}            & \textbf{F1}                 & \textbf{Acc}               \\\midrule
EdgeBank        & 91.82                     & \textbf{91.09}              & 91.82                    \\
DyRep           & 89.72$_{\pm0.59}$         & 79.02$_{\pm0.91}$           & 80.46$_{\pm0.63}$        \\
JODIE           & 94.94$_{\pm0.48}$         & 87.52$_{\pm0.39}$           & 87.85$_{\pm0.44}$        \\
TGAT            & 95.54$_{\pm0.22}$         & 88.11$_{\pm0.45}$           & 88.58$_{\pm0.31}$        \\
TGN             & \one{97.07$_{\pm0.15}$} & \textbf{90.49$_{\pm0.24}$}  & \textbf{90.66$_{\pm0.22}$} \\\midrule

& \multicolumn{3}{c}{\textbf{Reddit}}\\
\textbf{Model} & \textbf{AUC}               & \textbf{F1}                & \textbf{Acc}                        \\\midrule
EdgeBank & 96.42                        & \textbf{96.29}             & \textbf{96.42}                  \\
DyRep & 97.69$_{\pm0.04}$            & 92.12$_{\pm0.13}$          & 92.02$_{\pm0.19}$               \\
JODIE & 96.72$_{\pm0.21}$            & 89.97$_{\pm0.66}$          & 89.48$_{\pm0.86}$               \\
TGAT & 98.41$_{\pm0.01}$            & 93.58$_{\pm0.05}$          & 93.63$_{\pm0.04}$               \\
TGN    & \one{98.66$_{\pm0.04}$}    & \textbf{94.20$_{\pm0.15}$} & \textbf{94.19$_{\pm0.17}$} \\\midrule

& \multicolumn{3}{c}{\textbf{LastFM}}\\
\textbf{Model}         & \textbf{AUC}            & \textbf{F1}                & \textbf{Acc}      \\\midrule
EdgeBank & \one{94.72}             & \textbf{94.43}             & \textbf{94.72}                 \\
DyRep    & 78.41$_{\pm0.50}$         & \textbf{71.80$_{\pm0.92}$} & 68.63$_{\pm1.01}$ \\
JODIE    & 69.32$_{\pm1.33}$         & 63.95$_{\pm2.64}$          & 62.08$_{\pm2.78}$  \\
TGAT     & \one{81.97$_{\pm0.08}$} & 70.96$_{\pm0.24}$          & \textbf{72.64$_{\pm0.09}$}  \\
TGN & 79.84$_{\pm1.58}$         & 71.09$_{\pm2.36}$          & 63.13$_{\pm7.03}$ \\\bottomrule
\end{tabular}
\end{table}

\begin{table}[h]
\centering
\caption{Average time to execute a forward pass on the whole dataset (measured in seconds) and std of the best configuration of each model in each task in the C-TDG setting, averaged over 5 repetitions. The evaluation was carried out on an Intel(R) Xeon(R) Gold 6278C CPU @ 2.60GHz.}\label{tab:benchmark_time_ctdg}
\scriptsize
\begin{tabular}{lccc}
\toprule
\textbf{Model} & \textbf{Wikipedia} &  \textbf{Reddit}             &      \textbf{LastFM}\\\midrule
DyRep  & 13.95$_{\pm1.05}$  &   99.11$_{\pm9.04}$  & 143.15$_{\pm12.72}$\\
JODIE  & \one{12.66$_{\pm1.98}$}  &   \one{83.27$_{\pm6.47}$}  &  \one{117.17$_{\pm6.86}$}\\
TGAT   & 36.84$_{\pm2.09}$  &    303.70$_{\pm7.80}$  & 167.65$_{\pm13.33}$\\
TGN    & 28.35$_{\pm2.14}$  & 114.73$_{\pm14.77}$  &  178.15$_{\pm7.87}$\\
\bottomrule

\end{tabular}
\end{table}

\section{Summary}
Despite the field of representation learning for (static) graphs is now a consolidated and vibrant research area, there is still a strong demand for work in the domain of \textit{dynamic} graphs. 

In light of this, in this chapter we proposed, at first, a survey that focuses on recent representation learning techniques for dynamic graphs under the uniform formalism introduced in Section~\ref{sec:graphs}. Second, we provide the research community with a fair performance comparison among the most popular  methods of the three families of dynamic graph problems, by leveraging a reproducible experimental environment. We believe that this work will help to foster the research in the domain of dynamic graphs by providing a clear picture of the current development status and a good baseline to test new architectures and approaches.

We point out to the reader that to further improve the maturity of representation learning for dynamic graphs, we believe that certain aspects still represent open challenges and need further work from the community in the future.
A future interesting direction, in this sense, is to extend the work that has been done for \textit{heterophilic} (static) graphs~\citep{geom-gcn, heterophily_results2, cavallo2023gcnh} to the temporal domain. This will require addressing the problem of generating information-rich node representations when neighboring nodes tend to belong to different classes. A similar challenge is the one of \textit{heterogeneus} graphs~\citep{heterogeneous, heterogeneous2}, which contain different types of nodes and links. In this scenario, new architectures should learn the semantic-level information coming from node and edge types, in addition to topological and label information. While these are interesting future directions, we observe that there are compelling challenges that need addressing and that relate to studying, in the temporal domain, aspects such as robustness to adversarial attacks~\citep{Maddalena2022, adversarial_graph}, oversmoothing (see Section~\ref{sec:dgn_plights}), oversquashing (see Section~\ref{sec:dgn_plights}), and DGNs' expressive power~\citep{GIN, GNNBook-ch5-li}.

With respect to the challenge of mitigating the oversquashing phenomenon over space and time, in Chapter~\ref{ch:ctan} we propose a novel method for long-range propagation within dynamic graphs 
that addresses this challenge.

\part{Non-Dissipative Propagation for Static Graphs}
\chapter{The Antisymmetric Constraint}\label{ch:antisymmetry}
The primary challenge in the field of representation learning for graphs is how we capture and encode structural information in the learning model, as previously discussed in Section~\ref{sec:static_dgn_fundamentals}. 
However, in some problems, the exploitation of local interactions between nodes is not enough to learn representative embeddings. In this scenario, it is often the case that the DGN needs to capture information concerning interactions between nodes that are far away in the graph, \ie by stacking multiple layers. A specific predictive problem typically needs to consider a specific range of node interactions in order to be effectively solved, hence requiring a specific number (possibly large) of DGN layers.

Despite the progress made in recent years in the field, many of the proposed methods suffer from the \emph{oversquashing} problem when the number of layers increases (see Section~\ref{sec:dgn_plights}). Specifically, when increasing the number of layers to cater for longer-range interactions, one observes an excessive amplification or an annihilation of the information being routed to the node by the message passing process to update its fixed length encoding. 
As such, oversquashing prevents DGNs to learn long-range information. 

To overcome this limitation, we build on the observations and understandings of neural DEs in Section~\ref{sec:intro_DE} and use similar concepts to forge the field of differential-equations inspired DGNs (\gls*{DE-DGN}s). Through this view, we design DGNs with strong inductive biases. Specifically, we are interested in addressing the oversquashing problem in a principled manner, accompanied by theoretical understanding through the prism of DE-DGNs.

Inspired by \emph{stable} and \emph{non-dissipative} dynamical systems, in the following we provide the theoretical conditions for realizing DE-DGNs for long-range propagation, within static graphs, through the use of \emph{antisymmetric constraints}. In Section~\ref{sec:ADGN} we first explore the benefits of antisymmetric weight parametrization, building on \cite{gravina_adgn} and \cite{gravina_randomized_adgn}
. Afterward, in Section~\ref{sec:SWAN}, we improve the long range propagation (\ie the non-dissipative behavior) thanks to an additional antisymmetric constrain on the space domain, \ie in the neighborhood aggregation. We base Section~\ref{sec:SWAN} on  \cite{gravina_swan}
.

\section{Antisymmetric Weight Parametrization}\label{sec:ADGN}
In this section, we present \textit{Antisymmetric Deep Graph Network} (\gls*{A-DGN})\index{A-DGN}, a framework for effective long-term propagation of information in DGN architectures
designed through the lens of ordinary differential equations.
Leveraging the connections between ODEs and deep neural architectures, we provide theoretical conditions for realizing a \textit{stable} and \textit{non-dissipative} ODE system on graphs through the use of antisymmetric weight matrices.
The formulation of the A-DGN layer then results from the forward Euler discretization of the achieved graph ODE.
Thanks to the properties enforced on the ODE, our framework preserves the long-term dependencies between nodes as well as prevents from gradient explosion or vanishing. 
Interestingly, our analysis also paves the way for rethinking the formulation of standard DGNs as discrete versions of non-dissipative and stable ODEs on graphs.

The key contributions of this section can be summarized as follows:
\begin{itemize}
    \item We introduce A-DGN, a novel design scheme for deep graph networks stemming from an ODE formulation. Stability and non-dissipation are the main properties that characterize our method, allowing the preservation of long-term dependencies in the information flow.
    
    \item We theoretically prove that the employed ODE on graphs has stable and non-dissipative behavior. Such result leads to the absence of exploding and vanishing gradient problems during training, typical of unstable and lossy systems.
    
    \item We conduct extensive experiments to demonstrate the benefits of our method.  
    A-DGN can outperform classical DGNs over several datasets even when dozens of layers are used. Overall, A-DGN shows the ability to effectively explore long-range dependencies and leverage dozens of layers without any noticeable drop in performance. For such reasons, we believe it can be a step towards the mitigation of the over-squashing problem in DGNs.

\end{itemize}

\subsection{From graph-ODEs to DGNs}\label{sec:ADGN_mpnn_comparison}
Recent advancements in the field of representation learning propose to treat neural network architectures as an ensemble of continuous (rather than discrete) layers, thereby drawing connections between deep neural networks and ODEs, as described in Section~\ref{sec:neural_de}. 
This connection can be pushed up to neural processing of graphs as introduced in \cite{GDE}, by making a suitable ODE define the computation on a graph structure.

\begin{figure}[h!]
\begin{center}
    \includegraphics[width=0.95\linewidth]{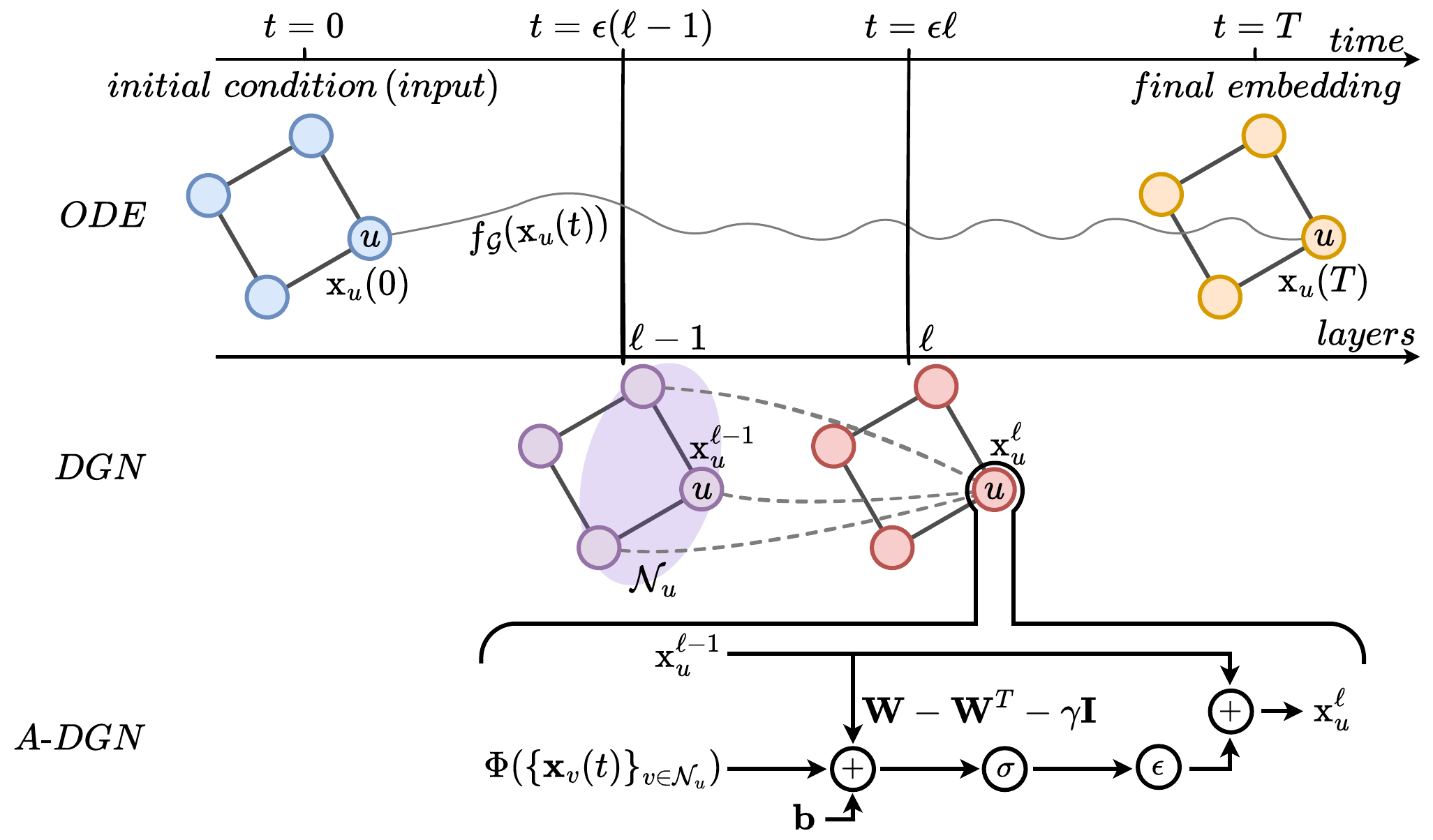}
\end{center}
\caption{A high level overview 
of our proposed framework, summarizing the involved concepts of ODE over a graph, its discretization as layers of a DGN, and the resulting node update of Antisymmetric DGN.
At the top, it is illustrated the continuous processing of nodes’ states in the time span between $0$ and $T>0$, as a Cauchy problem on graphs. The node-wise ODE $f_{\mathcal{G}}$ determines the evolution of the states $\bfx_u(t)$, while the initial conditions $\bfx_u(0)$ play the role of input information.
In the middle, the discretized solution of the graph ODE is interpreted  as a succession of DGN layers in a neural network architecture. The node state $\bfx_u^{\ell}$ computed at layer $\ell$ is updated iteratively by leveraging its neighborhood and self representations at the previous layer $\ell-1$. 
The bottom part sketches the computation performed by a layer in an Antisymmetric DGN (in Equation~\ref{eq:AGC_euler}), resulting from the forward Euler discretization of a stable and non-dissipative ODE on graphs. The more the layers, the more long-range dependencies are included in the final nodes' representations.}
\label{fig:framework}
\end{figure}

We focus on static graphs, \ie on structures with fixed sets of nodes and edges (see Section~\ref{sec:static_graph_notation}).
For each node $u \in \mathcal{V}$ we consider a state $\gls*{xut} \in \mathbb{R}^{d}$, which provides a representation of the node $u$ at time $t$. We can then define a Cauchy problem on graphs in terms of the following node-wise defined ODE:
\begin{equation}\label{eq:ode}
   \frac{d \bfx_u(t)}{d t} = f_{\mathcal{G}}(\bfx_u(t)),
\end{equation}
for time $t\in[0,T]$, and subject to the initial condition $\bfx_u(0)= \bfx_u^0\in\mathbb{R}^{d}$. 
The dynamics of node's representations is described by the function
$f_{\mathcal{G}}:\mathbb{R}^{d} \rightarrow \mathbb{R}^{d}$, while 
the initial condition $\bfx_u(0)$ can be interpreted as the initial configuration of the node's information, hence as the input for our computational model.
As a consequence, the ODE defined in Equation~\ref{eq:ode} can be seen as a continuous information processing system over the graph, which starting from the input configuration $\bfx_u(0)$ computes the final node's representation (i.e., embedding) $\bfx_u(T)$. Notice that this process shares similarities with standard DGNs, in what it computes nodes' states that can be used as an embedded representation of the graph and then used to feed a readout layer in a downstream task on graphs. The top of Figure~\ref{fig:framework} visually summarizes this concept, showing how nodes evolve following a specific graph ODE in the time span between $0$ and a terminal time $T>0$. 

Since for most ODEs it is impractical to compute an analytical solution (as discussed in Section~\ref{sec:discretization_method_de}) 
a common approach relies on finding an approximate one
through a numerical discretization procedure, such as the forward Euler method. 
In this way, the time variable is discretized and the ODE solution is computed by the successive application of an iterated map that operates on the discrete set of points between $0$ and $T$, with a step size $\epsilon>0$.
Crucially, as already observed for feed-forward and recurrent neural models in Section~\ref{sec:neural_de}, 
each step of the ODE discretization process can be equated to one layer of a DGN network. The whole neural architecture contains as many layers as the integration steps in the numerical method (i.e., $L = T/\epsilon$),
and each layer $\ell=1, ..., L$ computes nodes' states $\bfx_u^\ell$ which approximates 
$\bfx_u(\epsilon \, \ell)$.
This process is summarized visually in the middle of Figure~\ref{fig:framework}. 

\subsection{Antisymmetric Deep Graph Network}\label{sec:adgn}
Leveraging the concept of graph neural ODEs \citep{GDE}, we perform a further step by reformulating a DGN as a solution to a \emph{stable and non-dissipative}
Cauchy problem over a graph. The main goal of our work is therefore achieving preservation of long-range information between nodes, while laying down the conditions that prevent gradient vanishing or explosion.  
Inspired by the works on stable deep architectures that discretize ODE solutions \citep{HaberRuthotto2017,chang2018antisymmetricrnn}, 
we do so by first deriving conditions under which the graph ODE is constrained to the desired stable and non-dissipative behavior. 

Since we are dealing with static graphs, we instantiate Equation~\ref{eq:ode} for a node $u$ as follows:
\begin{equation}\label{eq:simple_ode_dgn}
    \frac{d \bfx_u(t)}{d t} = \sigma \left( \bfW_t \bfx_u(t) +  \gls*{Phi}(\{\bfx_v(t)\}_{v\in\mathcal{N}_u})+\bfb_t\right),
\end{equation}
where $\sigma$ is a monotonically non-decreasing activation function, 
$\bfW_t\in\mathbb{R}^{d\times d}$ and $\mathbf{b}_t\in\mathbb{R}^d$ are, respectively, a weight matrix and a bias vector that contain the trainable parameters of the system. 
We denote by $\Phi(\{\bfx_v(t)\}_{v\in\mathcal{N}_u})$ the aggregation function for the states of the nodes in the neighborhood of $u$\footnote{Following the notation in Section~\ref{sec:static_dgn_fundamentals}, the function $\Phi$ encompasses both the aggregation function $\bigoplus$ and the message function $\rho_M$, \ie $\Phi(\{\bfx_v(t)\}_{v\in\mathcal{N}_u}) = \bigoplus_{v\in\mathcal{N}_u}\rho_M(\bfx_v(t))$. We refer to the neighborhood aggregation function as $\Phi(\{\bfx_v(t)\}_{v\in\mathcal{N}_u})$ for brevity.}. 
For simplicity, in the following we keep $\bfW_t$ and $\bfb_t$ constant over time, hence dropping the $t$ subscript in the notation. Moreover, to ease readability, we drop the bias term $\bfb$.

Well-posedness and stability are essential concepts when designing DGNs as solutions to Cauchy problems, both relying on the continuous dependence of the solution from initial conditions. An ill-posed unstable system, even if potentially yielding a low training error, is likely to lead to a poor generalization error on perturbed data.
On the other hand, the solution of a Cauchy problem is stable if the long-term behavior of the system does not depend significantly on the initial conditions \citep{amr}. 
In our case, where the ODE defines a message passing diffusion over a graph, our intuition is that a stable encoding system will be robust to perturbations in the input nodes information. Hence, the state representations will change smoothly with the input, resulting in a 
non-exploding forward propagation and
better generalization. This intuition is formalized by the following definition.

\begin{definition}[Stability of a graph ODE]\label{def:stable}\index{stability of a graph ODE}
A solution $\bfx_u(t)$ of the ODE in Equation~\ref{eq:simple_ode_dgn}, with initial condition $\bfx_u(0)$, is stable if for any $\omega > 0$, there exists a $\delta > 0$ such that any other solution  $\mathbf{\tilde{x}}_u(t)$ of the ODE with initial condition $\mathbf{\tilde{x}}_u(0)$ satisfying $|\bfx_u(0) -\mathbf{\tilde{x}}_u(0)| \leq \delta$ also satisfies $|\bfx_u(t) -\mathbf{\tilde{x}}_u(t)| \leq \omega$, for all $t\ge0$.
\end{definition}

The idea is that a small perturbation of size $\delta$ of the initial state (\ie the node input features) results in a perturbation on the subsequent states that is at most $\omega$. 
As known from the stability theory of autonomous systems \citep{amr}, 
this condition is met when the maximum real part of the Jacobian's eigenvalues of $f_\mathcal{G}$ is smaller or equal than 0, \ie $\max_{i=1,...,d} Re(\lambda_i(\mathbf{J}(t))) \leq 0$, $\forall t\geq0$.

Although stability is a necessary condition for successful learning, it alone is not sufficient to capture long-term dependencies.
As it is discussed in \cite{HaberRuthotto2017}, if  $\max_{i=1,...,d} Re(\lambda_i(\mathbf{J}(t))) \ll 0$ the result is a lossy system subject to catastrophic forgetting during propagation. 
Thus, in the graph domain, this means that only local neighborhood information is preserved by the system, while long-range dependencies among nodes are forgotten.
If no long-range information is preserved, then it is likely that the DGN will underperform, since it will not be able to reach the minimum radius of inter-nodes interactions needed to effectively solve the task. 

Therefore, we can design an ODE for graphs which is stable and non-dis\-si\-pa\-tive (see Definition~\ref{def:dissipative} in Section~\ref{sec:intro_DE}) 
that leads to well-posed learning, when the criterion that guarantees stability is met and the Jacobian's eigenvalues of $f_\mathcal{G}$ are nearly zero. 
Under this condition, the forward propagation produces at most  moderate amplification or shrinking of the input, which enables to preserve long-term dependencies in the node states. During training, the backward propagation needed to compute the gradient of the loss $\partial \mathcal{L} / \partial \bfx_u(t)$ will have the same properties of the forward propagation. As such, no gradient vanish nor explosion is expected to occur. More formally:
\begin{proposition}\label{prep:stableCondition}
Assuming that $\mathbf{J}(t)$ does not change significantly over time, the forward and backward propagations of the ODE in Equation~\ref{eq:simple_ode_dgn} are stable and non-dissipative\index{non-dissipativity} if \begin{equation}\label{eq:eigenvalue_condition}
    Re(\lambda_i(\mathbf{J}(t))) 
    = 0, \quad \forall i=1, ..., d.
\end{equation}
\end{proposition}
\begin{boxedproof}
Let us consider the ODE defined in Equation~\ref{eq:simple_ode_dgn} and analyze the sensitivity of its solution to the initial conditions.
Following \citep{chang2018antisymmetricrnn}, we differentiate
both sides of Equation~\ref{eq:simple_ode_dgn}
with respect to $\mathbf{x}_u(0)$, obtaining:
\begin{equation}
    \label{eq:appendix1}
   \frac{d}{dt}\left(\frac{\partial \mathbf{x}_u(t)}{\partial \mathbf{x}_u(0)}\right) = \mathbf{J}(t) \frac{\partial \mathbf{x}_u(t)}{\partial\mathbf{x}_u(0)}.
\end{equation}
Assuming the Jacobian does not change significantly over time, 
we can apply results from autonomous differential equations \citep{glendinning_1994} and solve Equation~\ref{eq:appendix1} analytically as follows:
\begin{equation}
   \frac{\partial \mathbf{x}_u(t)}{\partial \mathbf{x}_u(0)} = e^{t \mathbf{J}} = \mathbf{T} e^{t \mathbf{\Lambda}}\mathbf{T}^{-1}
   = \mathbf{T} 
   \big(\sum_{k=0}^\infty \frac{(t \mathbf{\Lambda})^k}{k!}\big)
   \mathbf{T}^{-1},
\end{equation}
where $\mathbf{\Lambda}$ is the diagonal matrix whose non-zero entries contain the eigenvalues of $\mathbf{J}$, and $\mathbf{T}$ has the eigenvectors of $\mathbf{J}$ as columns. 
The qualitative behavior of $\partial \mathbf{x}_u(t)/\partial \mathbf{x}_u(0)$ is then determined by the real parts of the eigenvalues of $\mathbf{J}$. 
When $\max_{i=1,...,d} Re(\lambda_i(\mathbf{J}(t))) > 0$, a small perturbation of the initial condition (i.e., a perturbation on the input graph) would cause an exponentially exploding difference in the nodes representations, and the system would be unstable.
On the contrary, for 
$\max_{i=1,...,d} Re(\lambda_i(\mathbf{J}(t))) < 0$, 
the term $\partial \mathbf{x}_u(t)/\partial \mathbf{x}_u(0)$ would vanish exponentially fast over time, thereby making the nodes' representation insensitive to differences in the input graph. Accordingly, the system states $\mathbf{x}_u(t)$ would asymptotically approach the same embeddings for all the possible initial conditions $\mathbf{x}_u(0)$, and the system would be dissipative.
Notice that the effects of explosion and dissipation are progressively more evident for larger absolute values of $\max_{i=1,...,d} Re(\lambda_i(\mathbf{J}(t)))$.
If $Re(\lambda_i(\mathbf{J}(t))) 
= 0$ for $i=1,...,d$ then the magnitude of $\partial \mathbf{x}(t)/\partial \mathbf{x}(0)$ is constant over time, and the input graph information is effectively propagated through the successive transformations into the final nodes' representations. 
In this last case, the system is hence both stable and non-dissipative.

Let us now consider a loss function $\mathcal{L}$, and observe that its sensitivity to the initial condition (i.e., the input graph) $\partial \mathcal{L}/\partial \mathbf{x}_u(0)$ is proportional to $\partial \mathbf{x}_u(t)/\partial \mathbf{x}_u(0)$. 
Hence, in light of the previous considerations, if $Re(\lambda_i(\mathbf{J}(t))) =0$ for $i=1,...,d$, then the magnitude of $\partial \mathcal{L}/\partial \mathbf{x}_u(0)$, which is the 
longest gradient chain 
that we can obtain during
back-propagation, stays constant over time. The backward propagation is then stable and non-dissipative, and no gradient vanishing or explosion can occur during training.
\end{boxedproof}

A simple way to impose the condition in Equation~\ref{eq:eigenvalue_condition} is to use an antisymmetric 
weight matrix in Equation~\ref{eq:simple_ode_dgn}. 
Under this assumption, we can rewrite Equation~\ref{eq:simple_ode_dgn} as follows:
\begin{equation}\label{eq:stableode}
    \frac{d \bfx_u(t)}{d t} = \sigma \left( (\bfW-\bfW^T) \bfx_u(t) +  \Phi(\{\bfx_v(t)\}_{v\in\mathcal{N}_u})\right) 
\end{equation}
where $(\bfW-\bfW^T)\in\mathbb{R}^{d\times d}$ is the antisymmetric weight matrix. 
The next Proposition~\ref{prep:jacobian} ensures that when the aggregation function $\Phi(\{\bfx_v(t)\}_{v\in\mathcal{N}_u})$ is independent of $\bfx_u(t)$ (see for example Equation~\ref{eq:simple_aggregation}), the  Jacobian of the 
ODE has imaginary eigenvalues, hence it is stable and non-dissipative according to Proposition~\ref{prep:stableCondition}.
\begin{proposition}\label{prep:jacobian}
Provided that $\Phi(\{\bfx_v(t)\}_{v\in\mathcal{N}_u})$ is independent of $\bfx_u(t)$, the Jacobian matrix of the ODE in Equation~\ref{eq:stableode} 
has purely imaginary eigenvalues, i.e.
 $$Re(\lambda_i(\mathbf{J}(t))) = 0, \forall i=1, ..., d.$$
Therefore the ODE in Equation~\ref{eq:stableode} is stable and non-dissipative.
\end{proposition}
\begin{boxedproof}
    Under the assumption that the aggregation function $\Phi(\{\bfx_v(t)\}_{v\in\mathcal{N}_u})$ does not include a term that depends on $\mathbf{x}_u(t)$ itself (see Equation~\ref{eq:simple_aggregation}
for an example), the Jacobian matrix of Equation~\ref{eq:stableode} is given by:
\begin{equation}
    \label{eq:jacobian_appendix}
    \mathbf{J}(t) = \mathrm{diag}\left[\sigma' \left((\mathbf{W}-\mathbf{W}^T) \mathbf{x}_u(t) +\Phi(\{\bfx_v(t)\}_{v\in\mathcal{N}_u}) \right)\right](\mathbf{W}-\mathbf{W}^T). 
\end{equation}
Following \citep{chang2018reversible,chang2018antisymmetricrnn},
we can see the right-hand side of Equation~\ref{eq:jacobian_appendix} as the result of a matrix multiplication between an invertible diagonal matrix and an antisymmetric matrix. 
Specifically,
defining 
\begin{align}
    \mathbf{A} &= \mathrm{diag}\left[\sigma' \left((\mathbf{W}-\mathbf{W}^T) \mathbf{x}_u(t) +\Phi(\{\bfx_v(t)\}_{v\in\mathcal{N}_u}) \right)\right]\\ 
    \mathbf{B} &= \mathbf{W}-\mathbf{W}^T,
\end{align} 
we have $\mathbf{J}(t) = \mathbf{A}\mathbf{B}$.

Let us now consider an eigenpair of $\mathbf{A} \mathbf{B}$, where the eigenvector is denoted by $\mathbf{v}$ and the eigenvalue by $\lambda$. Then:
\begin{align}
\label{eq:eigen}
    \mathbf{A}\mathbf{B}\mathbf{v} &= \lambda \mathbf{v},\notag \\
    \mathbf{B}\mathbf{v} &= \lambda \mathbf{A}^{-1}\mathbf{v},\notag \\
    \mathbf{v}^*\mathbf{B}\mathbf{v} &= \lambda (\mathbf{v}^*\mathbf{A}^{-1}\mathbf{v})
\end{align}
where $*$ represents the conjugate transpose.
On the right-hand side of Equation~\ref{eq:eigen}, we can notice that the $(\mathbf{v}^*\mathbf{A}^{-1}\mathbf{v})$ term  is a real number. 
Recalling that $\mathbf{B}^* = \mathbf{B}^T=-\mathbf{B}$ for a real antisymmetric matrix, we can notice that 
$(\mathbf{v}^*\mathbf{B}\mathbf{v})^* = \mathbf{v}^*\mathbf{B}^*\mathbf{v} = -\mathbf{v}^*\mathbf{B}\mathbf{v}$. Hence, 
the $\mathbf{v}^*\mathbf{B}\mathbf{v}$ term on the left-hand side 
of Equation~\ref{eq:eigen}
is an imaginary number.
Thereby, $\lambda$ needs to be purely imaginary, and, as a result, all eigenvalues of $\mathbf{J}(t)$ are purely imaginary.
\end{boxedproof}

Whenever $\Phi(\{\bfx_v(t)\}_{v\in\mathcal{N}_u})$ includes $\bfx_u(t)$ in its definition (see for example Equation~\ref{eq:gcn_aggregation}), the eigenvalues of the resulting Jacobian are still bounded in a small neighborhood around the imaginary axis. Let us consider here the case in which $\Phi(\{\bfx_v(t)\}_{v\in\mathcal{N}_u})$ is defined such that it depends on $\mathbf{x}_u(t)$ (see for example 
Equation~\ref{eq:gcn_aggregation}).
In this case, the Jacobian matrix of Equation~\ref{eq:stableode} can be written (in a more general form than Equation~\ref{eq:jacobian_appendix}), as follows:
\begin{equation}
    \label{eq:jacobian2_appendix}
    \mathbf{J}(t) = \mathrm{diag}\left[\sigma' \left((\mathbf{W}-\mathbf{W}^T) \mathbf{x}_u(t) +\Phi(\{\bfx_v(t)\}_{v\in\mathcal{N}_u}) \right)\right]
    \left(
    (\mathbf{W}-\mathbf{W}^T)+\mathbf{C}
    \right),
\end{equation}
where the term $\mathbf{C}$ represents the derivative of $\Phi(\{\bfx_v(t)\}_{v\in\mathcal{N}_u})$ with respect to $\mathbf{x}_u(t)$.
Similarly to the proof of Proposition~\ref{prep:jacobian}
, we can see the right-hand side of Equation~\ref{eq:jacobian2_appendix} as $\mathbf{J}(t) = \mathbf{A}(\mathbf{B} + \mathbf{C}) = \mathbf{A} \mathbf{B} + \mathbf{A} \mathbf{C}$. 
Thereby, we can bound the eigenvalues of $\mathbf{J}(t)$ around those of $\mathbf{A} \mathbf{B}$ by applying the results of the Bauer-Fike's theorem \citep{bauer1960norms}. Recalling that the eigenvalues of $\mathbf{A} \mathbf{B}$ are all imaginary (as proved in 
Proposition~\ref{prep:jacobian}), we can conclude that the eigenvalues of $\mathbf{J}(t)$ are contained in a neighborhood of the imaginary axis with radius $r = \|\mathbf{A} \mathbf{C}\| \leq \|\mathbf{C}\|$. 
Although this result does not guarantee that the eigenvalues of the Jacobian are imaginary, in practice it crucially limits their position around the imaginary axis, limiting the dynamics of the system on the graph to show at most moderate amplification or loss of signals over the structure.
\\

We now proceed to discretize the ODE in Equation~\ref{eq:stableode} by means of the \textit{forward  Euler's method}. To preserve stability of the discretized system 
(see Section~\ref{sec:discretization_method_de}), we add a diffusion term to Equation~\ref{eq:stableode}, yielding the following node state update equation:
\begin{equation}\label{eq:AGC_euler}
    \bfx^{\ell}_u = \bfx^{\ell-1}_u + \epsilon \sigma \left( (\bfW-\bfW^T-\gamma \mathbf{I}) \bfx_u^{\ell-1} +  \Phi(\{\bfx^{\ell-1}_v\}_{v\in\mathcal{N}_u}) \right) 
\end{equation}
where $\mathbf{I}$ is the identity matrix,  $\gamma$ is a hyperparameter that regulates the strength of the diffusion, and $\epsilon$ is the discretization step. In particular, subtracting a small positive constant $\gamma>0$ from the diagonal elements of the weight matrix $\mathbf{W}$ allows positioning $(1+\epsilon\lambda(\mathbf{J}(t)))$ inside the unit circle, thus improving the stability of the numerical discretization method. By building on the relationship between the discretization and the DGN layers, we have introduced  $\bfx_u^\ell$ as the state of node $u$ at layer $\ell$, i.e. the discretization of state at time $t=\epsilon \ell$. 

Now, both ODE and its Euler discretization are stable and non-dissipative. We refer to the framework defined by Equation~\ref{eq:AGC_euler} as \emph{Antisymmetric Deep Graph Network} (A-DGN), whose 
state update process is schematically illustrated in the bottom of Figure~\ref{fig:framework}.
Notice that having assumed the parameters of the ODE constant in time, A-DGN can also be interpreted as a recursive DGN with weight sharing between layers.

We recall that $\Phi(\{\bfx^{\ell-1}_v\}_{v\in\mathcal{N}_u})$ can be any function that aggregates nodes (and edges) information. 
Therefore, the general formulation of $\Phi(\{\bfx^{\ell-1}_v\}_{v\in\mathcal{N}_u})$ in A-DGN allows casting all standard DGNs through in their non-dissipative, stable and well-posed version. As a result, A-DGN can be implemented leveraging the aggregation function that is more adequate for the specific task, while allowing to preserve long-range relationships in the graph. As a demonstration of this, in Section \ref{sec:adgn_experiments} we explore two neighborhood aggregation functions, that are 
\begin{equation}\label{eq:simple_aggregation}
    \Phi(\{\bfx^{\ell-1}_v\}_{v\in\mathcal{N}_u}) = \sum_{v\in\mathcal{N}_u} \mathbf{V} \bfx^{\ell-1}_v,
\end{equation}
(which is also employed in \cite{graphconv}) and the classical GCN aggregation
\begin{equation}\label{eq:gcn_aggregation}
    \Phi(\{\bfx^{\ell-1}_v\}_{v\in\mathcal{N}_u}) = \mathbf{V} \sum_{v \in \mathcal{N}_u \cup \{ u \}} \frac{1}{\sqrt{\hat{d}_v\hat{d}_u}} \bfx^{\ell-1}_v,
\end{equation} 
where $\mathbf{V}$ is the weight matrix, $\hat{d}_v$ and $\hat{d}_u$ are, respectively, the degrees of nodes $v$ and $u$.

Finally, although we designed A-DGN with weight sharing in mind (for ease of presentation), 
a more general version of the framework, with 
layer-dependent weights
$\bfW^\ell-(\bfW^\ell)^T$, is possible\footnote{ 
The dynamical properties discussed in this section are
in fact still true even in the case of time varying $\bfW_t$ in Equation~\ref{eq:simple_ode_dgn}, 
provided that  $\max_{i=1,...,d} Re(\lambda_i(\mathbf{J}(t))) \leq 0$ and $\mathbf{J}(t)$ changes sufficiently slow over time (see \cite{amr, HaberRuthotto2017}). We refer the reader to Appendix~\ref{app:ldw} for the analysis on the layer-dependent weights continuity.}.

\subsection{Experiments}
\label{sec:adgn_experiments}

In this section, we discuss the empirical assessment of our method. Specifically, we show the efficacy of preserving long-range information between nodes and mitigating the over-squashing by evaluating our framework on graph property prediction tasks where we predict single source shortest path, node eccentricity, and graph diameter (see Section~\ref{sec:adgn_graph_prop_pred}). 
Moreover, we assess the performance of the proposed A-DGN approach on classical graph homophilic (see Section~\ref{sec:adgn_graph_benchmark}) and heterophilic (see Section~\ref{sec:adgn_exp_hetero}) benchmarks. The performance of A-DGN is assessed against DGN variants from the literature. 

To show that A-DGN allows \emph{by design} for effective propagation of long-term propagation of information in DGN architectures, we tested a variant of A-DGN (on graph property prediction and homophilic tasks) in which we limited the training algorithm to act only on a minor set of weights, leaving the internal connections untrained after random initialization. Therefore, the final performance of the model almost solely relies on the employed architectural bias.

We refer the reader to Appendix~\ref{app:adgn_data_stats} for  more details about the employed datasets.
We report in Table~\ref{tab:adgn_configs} (Appendix~\ref{app:adgn_hyperparam}) the grid of hyperparameters employed in our experiments. We observe that even if we do not directly explore in the hyperparameter space the terminal time $T$ in which the node evolution produces the best embeddings, that is done indirectly by fixing the values of the step size $\epsilon$ and the maximum number of layers $L$, since $T=L\epsilon$.

We carried the experiments on a Dell server with 4 Nvidia GPUs A100. We release openly the code implementing our methodology and reproducing our empirical analysis at \url{https://github.com/gravins/Anti-SymmetricDGN}. 

\subsubsection{Graph Property Prediction}\label{sec:adgn_graph_prop_pred}
\myparagraph{Setup}
For the graph property prediction task, we considered three datasets extracted from the work of \citet{PNA}. The analysis consists of classical graph theory tasks on undirected unweighted randomly generated graphs sampled from a wide variety of distributions. Specifically, we considered two node level tasks and one graph level task, which are single source shortest path (SSSP), node eccentricity, and graph diameter. Such tasks require capturing long-term dependencies in order to be solved, thus mitigating the over-squashing phenomenon. Indeed, in the SSSP task, we are computing the shortest paths between a given node $u$ and all other nodes in the graph. Thus, it is fundamental to propagate not only the information of the direct neighborhood of $u$, but also the information of nodes which are extremely far from it. Similarly, for diameter and eccentricity.

We employed the same seed and generator as \citet{PNA} to generate the datasets, but we considered graphs with 25 to 35 nodes, instead of 15-25 nodes as in the original work
, to increase the task complexity and lengthen long-range dependencies required to solve the task. As in the original work, we used 5120 graphs as training set, 640 as validation set, and 1280 as test set.

We explored the performance of three versions of A-DGN, \ie weight sharing, layer-dependent weights, and weight sharing with random fixed weights. Moreover, we employed two instances of our method leveraging the two aggregation functions in Equation \ref{eq:simple_aggregation} and \ref{eq:gcn_aggregation}. We will refer to the former as simple aggregation and to the latter as GCN-based aggregation. We compared our method to three DE-DGN models, \ie DGC~\citep{DGC}, GRAND~\citep{GRAND}, and GraphCON~\citep{graphcon}; and the five most popular MPNN-based DGNs, \ie GCN~\citep{GCN}, GraphSAGE~\citep{SAGE}, GAT~\citep{GAT}, GIN~\citep{GIN}, and GCNII~\citep{gcnii}. We refer the reader to Section~\ref{sec:static_dgn_fundamentals} for a more in depth description of such methods.

We designed each model as a combination of three main components. The first is the encoder which maps the node input features into a latent hidden space; the second is the graph convolution (\ie A-DGN or the DGN baseline); and the third is a readout that maps the output of the convolution into the output space. The encoder and the readout are MLPs that share the same architecture among all models in the experiments.

We performed hyperparameter tuning via grid search, optimizing the Mean Square Error (MSE). We trained the models using Adam optimizer for a maximum of 1500 epochs and early stopping with patience of 100 epochs on the validation error. For each model configuration, we performed 4 training runs with different weight initialization and report the average of the results. 


\myparagraph{Results}
We present the results on the graph property prediction in Table~\ref{tab:results_GraphProp}. Specifically, we report $log_{10}(\mathrm{MSE})$ as the evaluation metric. We observe that our method, A-DGN, outperforms all the DGNs employed in this experiment. Indeed, by employing GCN-based aggregation, we achieve an error score that is on average 
0.70 points better than the selected baselines. 
A-DGN with simple aggregation 
shows a decisive improvement with respect to baselines, with an improvement of 0.53 points (on average) when randomized weights are employed. Specifically, A-DGN achieves a performance that is 
up to 2.02 points better than the best baseline in each task. Moreover, it is on average 
$3.3\times$ faster than the baselines (see Table~\ref{tab:results_GraphProp_time}). If the model is left untrained after random initialization, this speedup increases to 
$4.7\times$.

We observe that the main challenge when predicting diameter, eccentricity, or SSSP is to leverage not only local information but also global graph information. Such knowledge can only be learned by exploring long-range dependencies. Indeed, the three tasks are extremely correlated. All of them require computing the shortest paths in the graph. Thus, as for standard algorithmic solutions (\eg Bellman–Ford~\citep{Bellman}, Dijkstra's algorithm~\citep{Dijkstra}), more messages between nodes need to be exchanged in order to achieve accurate solutions. This suggests that A-DGN can better capture and exploit such information. Moreover, this indicates also that the simple aggregator is more effective than the GCN-based because the tasks are mainly based on counting distances. Thus, exploiting the information derived from the Laplacian operator is not helpful for solving these kinds of algorithmic tasks. 

\begin{table}[ht]
\centering
\caption{Mean test set $log_{10}(\mathrm{MSE})$ and std averaged over 4 random weight initializations for each configuration. The subscript \textit{\small ws} stands for weight sharing, while \textit{\small ldw} for layer-dependent weights. The lower the better. \one{First}, \two{second}, and \three{third} best results for each task are color-coded. \label{tab:results_GraphProp}}

\scriptsize
\begin{tabular}{lccc}
\toprule

\textbf{Model} &\textbf{Diameter} & \textbf{SSSP} & \textbf{Eccentricity} \\\midrule
\textbf{MPNNs} \\
$\,$ GCN            & 0.7424$_{\pm0.0466}$ & 0.9499$_{\pm9.18\cdot10^{-5}}$ & 0.8468$_{\pm0.0028}$ \\
$\,$ GAT            & 0.8221$_{\pm0.0752}$ & 0.6951$_{\pm0.1499}$           & 0.7909$_{\pm0.0222}$  \\
$\,$ GraphSAGE      & 0.8645$_{\pm0.0401}$ & 0.2863$_{\pm0.1843}$           &  0.7863$_{\pm0.0207}$\\
$\,$ GIN            & 0.6131$_{\pm0.0990}$ & -0.5408$_{\pm0.4193}$          & 0.9504$_{\pm0.0007}$\\
$\,$  GCNII          & 0.5287$_{\pm0.0570}$ & -1.1329$_{\pm0.0135}$          & 0.7640$_{\pm0.0355}$\\
\midrule
\textbf{DE-GNNs} \\
$\,$ DGC            & 0.6028$_{\pm0.0050}$ & -0.1483$_{\pm0.0231}$          & 0.8261$_{\pm0.0032}$\\
$\,$ GRAND          & 0.6715$_{\pm0.0490}$ & -0.0942$_{\pm0.3897}$          & \three{0.6602$_{\pm0.1393}$} \\
$\,$ GraphCON       & \four{0.0964$_{\pm0.0620}$} & -1.3836$_{\pm0.0092}$ & 0.6833$_{\pm0.0074}$\\

\midrule
\multicolumn{4}{l}{\textbf{Ours - random weights}}\\
$\,$ A-DGN\textsubscript{fix} & 0.1243$_{\pm0.0626}$ & -1.2731$_{\pm0.1622}$ & 0.8061$_{\pm0.0188}$ \\
$\,$ A-DGN\textsubscript{fix}(GCN) & 0.5549$_{\pm0.2156}$ & -0.9510$_{\pm0.0511}$ & 0.7649$_{\pm0.0108}$\\

\midrule
\multicolumn{4}{l}{\textbf{Ours - weight sharing}}\\
$\,$ A-DGN\textsubscript{ws}  & \two{-0.5188$_{\pm0.1812}$} & \two{-3.2417$_{\pm0.0751}$} & \two{0.4296$_{\pm0.1003}$} \\
$\,$ A-DGN\textsubscript{ws}(GCN) & 0.2646$_{\pm0.0402}$ & -1.3659$_{\pm0.0702}$ & 0.7177$_{\pm0.0345}$ \\

\midrule
\multicolumn{4}{l}{\textbf{Ours - layer-dependent weights}}\\
$\,$ A-DGN\textsubscript{ldw} &\one{-0.5455$_{\pm0.0328}$} & \one{-3.4020$_{\pm0.1372}$} & \one{0.3046$_{\pm0.1181}$} \\
$\,$ A-DGN\textsubscript{ldw}(GCN) & 0.2271$_{\pm0.0804}$ & \three{-1.8288$_{\pm0.0607}$} & 0.7235$_{\pm0.0211}$\\

\bottomrule
\end{tabular}
\end{table}
\begin{table}[ht]
\centering
\caption{Average time per epoch (measured in seconds) and std, averaged over 4 random weight initializations. Each time is obtained by employing 20 layers and an embedding dimension equal to 30. The subscript \textit{\small ws} stands for weight sharing, \textit{\small ldw} for layer-dependent weights. The evaluation was carried out on an AMD EPYC 7543 CPU @ 2.80GHz. \one{First}, \two{second}, and \three{third} best results.\label{tab:results_GraphProp_time}}
\scriptsize
\begin{tabular}{lccc}
\toprule
\textbf{Model} &\textbf{Diameter} & \textbf{SSSP} & \textbf{Eccentricity} \\\midrule
\textbf{MPNNs} \\
$\,$ GCN &   32.45$_{\pm2.54}$ &  17.44$_{\pm3.85}$ & 11.78$_{\pm2.43}$\\
$\,$ GAT &  20.20$_{\pm5.18}$ & 26.41$_{\pm8.34}$ & 17.28$_{\pm1.92}$\\
$\,$ GraphSAGE & 13.12$_{\pm2.99}$ & 13.12$_{\pm2.99}$ & \three{8.20$_{\pm0.75}$}\\
$\,$ GIN & \one{6.63$_{\pm0.28}$}  &21.16$_{\pm2.33}$ & 14.22$_{\pm3.17}$\\
$\,$ GCNII & 13.13$_{\pm6.85}$ & 14.96$_{\pm7.17}$ & 15.70$_{\pm3.92}$\\

\midrule
\textbf{DE-GNNs} \\
$\,$ DGC & 8.97$_{\pm9.07}$ & 12.54$_{\pm1.62}$ & \two{7.21$_{\pm11.10}$}\\
$\,$ GRAND & 133.84$_{\pm42.57}$ & 109.15$_{\pm27.49}$ & 202.46$_{\pm85.01}$\\
$\,$ GraphCON & 9.26$_{\pm0.47}$ & 7.76$_{\pm0.05}$ & 7.80$_{\pm0.05}$ \\

\midrule
\multicolumn{4}{l}{\textbf{Ours - random weights}}\\
$\,$ A-DGN\textsubscript{fix} & \two{7.77$_{\pm6.98}$} & \one{7.69$_{\pm3.56}$} & \one{5.33$_{\pm3.82}$}\\
$\,$ A-DGN\textsubscript{fix}(GCN) & 10.73$_{\pm1.50}$ & 14.80$_{\pm2.72}$ & 10.92$_{\pm1.77}$\\

\midrule
\multicolumn{4}{l}{\textbf{Ours - weight sharing}}\\
$\,$ A-DGN\textsubscript{ws} & \three{8.42$_{\pm2.71}$} & \two{7.86$_{\pm2.11}$} & 13.18$_{\pm9.07}$\\
$\,$ A-DGN\textsubscript{ws}(GCN) & 13.08$_{\pm5.49}$ & 28.74$_{\pm10.92}$ & 16.26$_{\pm4.58}$\\

\midrule
\multicolumn{4}{l}{\textbf{Ours - layer-dependent weights}}\\
$\,$ A-DGN\textsubscript{ldw} & 14.59$_{\pm8.67}$ & \three{10.47$_{\pm6.95}$} & 14.04$_{\pm11.60}$\\
$\,$ A-DGN\textsubscript{ldw}(GCN) & 40.50$_{\pm16.45}$ & 26.72$_{\pm17.98}$ & 24.43$_{\pm19.10}$\\
\bottomrule

\end{tabular}
\end{table}

\subsubsection{Graph Benchmarks}\label{sec:adgn_graph_benchmark}
\myparagraph{Setup}
In the graph benchmark setting we consider five well-known graph datasets for node classification, \ie PubMed~\citep{pubmed}; coauthor graphs CS and Physics; and the Amazon co-purchasing graphs Computer and Photo from \citet{pitfalls}. Also for this class of experiments, we considered the same baselines and architectural choices as for the graph property prediction task. However, in this experiment we study only the versions of A-DGN with weight sharing and randomized weights, since it achieve good performances with low training costs.

Within the aim to accurately assess the generalization performance of the models, we randomly split the datasets into multiple train/validation/test sets. Similarly to \citet{pitfalls}, we use 20 labeled nodes per class as the training set, 30 nodes per class as the validation set, and the rest as the test set. We generate 5 random splits per dataset and 5 random weight initialization for each configuration in each split.

We perform hyperparameter tuning via grid search, optimizing the accuracy score. We train for a maximum of 10000 epochs to minimize the Cross-Entropy loss. We use an early stopping criterion that stops the training if the validation score does not improve for 50 epochs. 

\myparagraph{Results}
We present the results on the graph benchmark in Table~\ref{tab:results}. Specifically, we report the accuracy as the evaluation metric and ratio between the accuracy score in percentage points and the employed total number of trainable hidden units, \ie $acc / N_{tot}$, for a more fair evaluation. Even in this scenario, A-DGN outperforms the selected baselines, except in PubMed and Amazon Computers where GCNII is slightly better than our method.
In this benchmark, results that the GCN-based aggregation produces higher scores with respect to the simple aggregation. Thus, additional local neighborhood features extracted from the graph Laplacian seem to strengthen the final predictions. 
It appears also that, in the weight sharing version, there is less benefit from including global information with respect to the graph property prediction scenario. As a result, exploiting extremely long-range dependencies do not strongly improve the performance as the number of layers increases. Differently, A-DGN with randomized weights benefits from propagating information across distant nodes in the input graph, demonstrating the effectiveness of A-DGN's architectural bias.

\begin{figure}[!h]
\begin{adjustbox}{center}
    \includegraphics[width=0.45\textwidth]{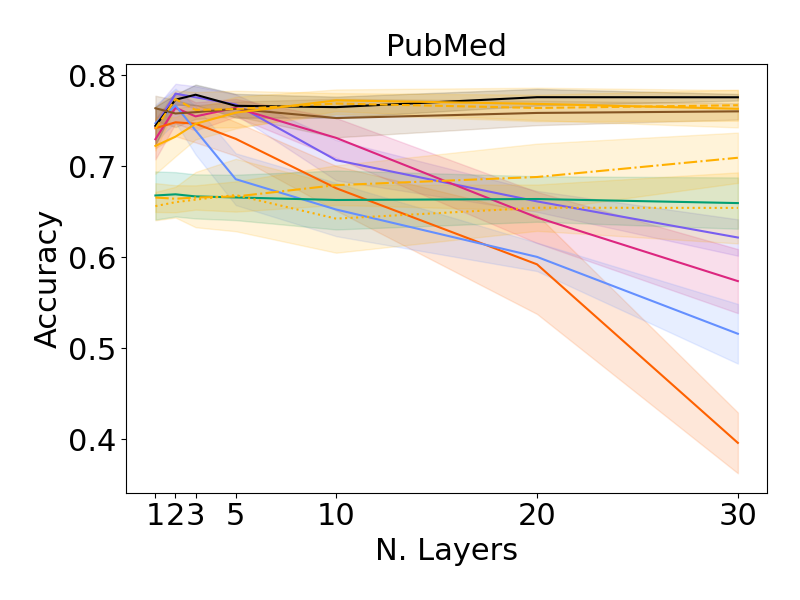}
    \includegraphics[width=0.45\textwidth]{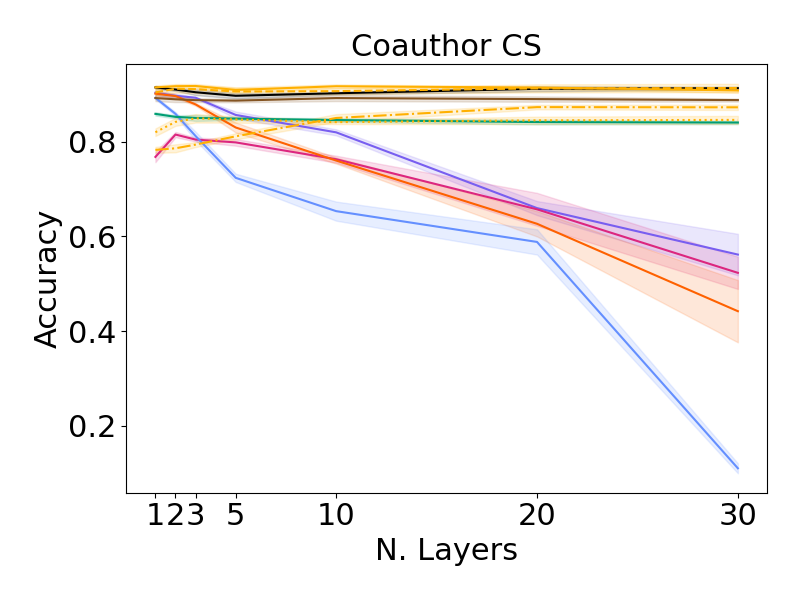}
\end{adjustbox}
\begin{adjustbox}{center}
    \includegraphics[width=0.45\textwidth]{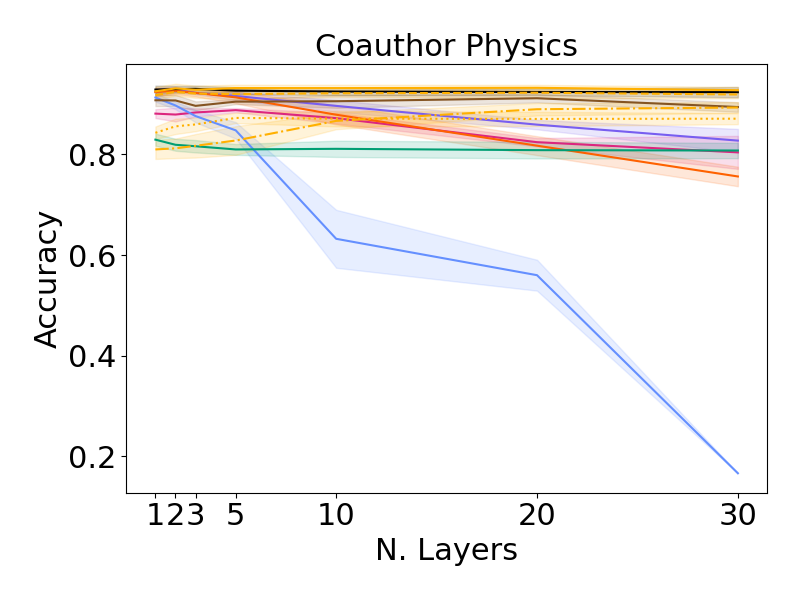}
\end{adjustbox}
\begin{adjustbox}{center}
    \includegraphics[width=0.45\textwidth]{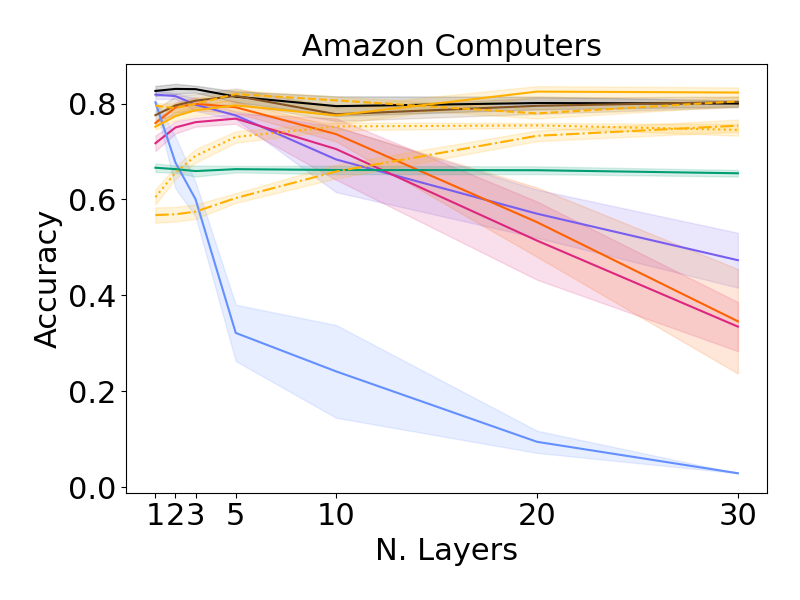}
    \includegraphics[width=0.45\textwidth]{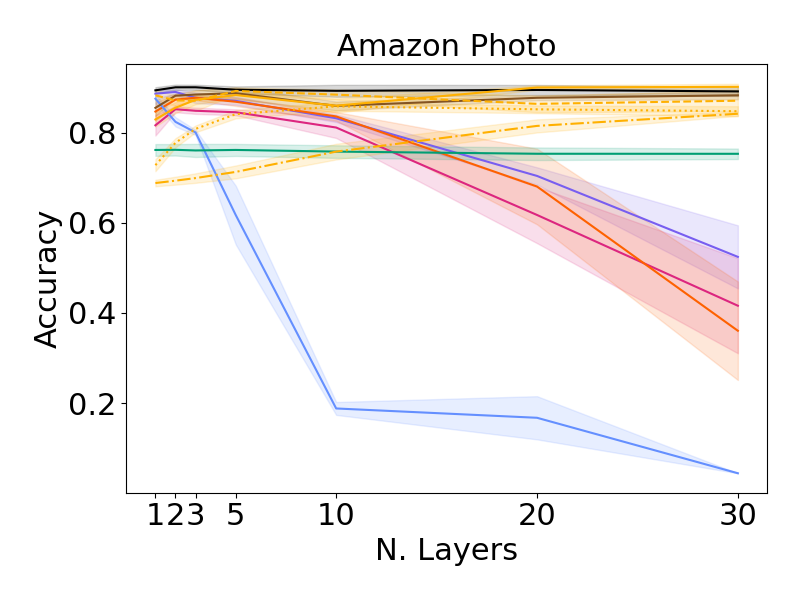}
\end{adjustbox}

\begin{adjustbox}{center}
    \includegraphics[width=0.95\textwidth]{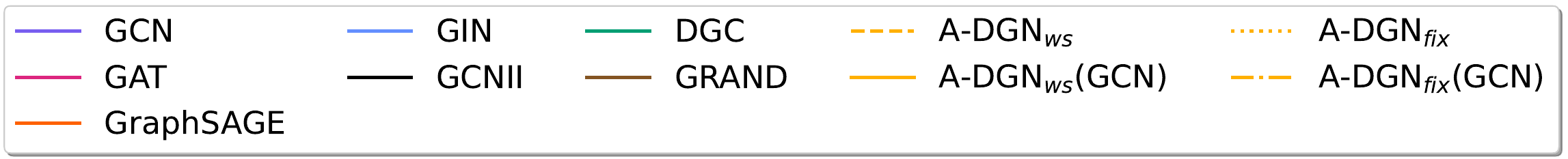}
\end{adjustbox}
\caption{The 
test accuracy with respect to the number of layers on all the graph benchmark datasets. From the top left to the bottom, we show: PubMed, Coauthor CS, Coauthor Physics, Amazon Computers, and Amazon Photo. The accuracy is averaged over 5 random train/validation/test splits and 5 random weight initialization of the best configuration per split.}
\label{fig:all_comparison}
\end{figure}

To demonstrate that our approach performs well with many layers, we show in Figure~\ref{fig:all_comparison} how the number of layers affects the accuracy score. Our model maintains or improves the performance as the number of layers increases. On the other hand, all the baselines obtain good scores only with one to two layers, and most of them exhibit a strong performance degradation as the number of layers increases. Indeed, in the Coauthor CS dataset we obtain that GraphSAGE, GAT, GCN and GIN lose 24.5\% to 78.2\% of accuracy. We observe that DGC does not degrade its performance since the convolution does not contain parameters.

Although extreme long-range dependencies do not produce the same boost as in the graph property prediction scenario, including more than 5-hop neighborhoods is fundamental to improve state-of-the-art performances. As clear from Figure~\ref{fig:all_comparison}, this is not practical when standard DGNs are employed. On the other hand, A-DGN demonstrates that can capture and exploit such information without any performance drop.

\begin{table}[h!]
\centering
\caption{Mean test set accuracy and std in percent averaged over 5 random train/validation/test splits and 5 random weight initializations for each configuration in each split. The 
normalized test accuracy (\ie $acc/N_{tot}$)
is reported in between parenthesis. The higher, the better. \one{First}, \two{second}, and \three{third} best results.\label{tab:results}}

\scriptsize
\begin{adjustbox}{center} 
\begin{tabular}{lccccc}
\toprule

\multirow{2}{*}{\textbf{Model}} &  \multirow{2}{*}{\textbf{PubMed}} & \textbf{Coauthor} & \textbf{Coauthor} & \textbf{Amazon} & \textbf{Amazon}\\
 & & \textbf{CS} & \textbf{Physics} &\textbf{Computers} & \textbf{Photo}\\\midrule
\textbf{MPNNs} \\
{$\,$ GCN}     & \three{76.75$_{\pm1.29}$}
        & 90.34$_{\pm0.31}$ 
        & \three{92.80$_{\pm0.44}$}
        & 81.63$_{\pm0.93}$
        & \three{89.14$_{\pm0.59}$}
        \\
        & \tiny${(0.0006)}$ 
        & \tiny${(0.0001)}$
        & \tiny${(0.0001)}$
        & \tiny${(0.0009)}$ 
        & \tiny${(0.0011)}$ \vspace{1mm}\\
{$\,$ GAT}     & 75.64$_{\pm1.27}$
& 81.57$_{\pm1.02}$
& 89.25$_{\pm0.82}$
& 76.36$_{\pm0.89}$
& 85.58$_{\pm0.91}$
\\
& \tiny${(0.0010)}$ 
& \tiny${(0.0001)}$ 
& \tiny${(0.0001)}$  
& \tiny${(0.0004)}$ 
& \tiny${(0.0006)}$ \vspace{1mm}\\

{$\,$ GraphSAGE} & 74.96$_{\pm1.69}$
& 89.93$_{\pm0.79}$
& 92.47$_{\pm0.94}$
& 79.37$_{\pm1.38}$
& 88.04$_{\pm0.85}$
\\
& \tiny${(0.0006)}$ 
& \tiny${(0.0001)}$  
& \tiny${(0.0001)}$ 
& \tiny${(0.0003)}$ 
& \tiny${(0.0005)}$ \vspace{1mm}\\

{$\,$ GIN}       & 76.24$_{\pm1.86}$
& 89.26$_{\pm0.31}$
& 91.40$_{\pm0.70}$
& 79.64$_{\pm0.72}$
& 87.69$_{\pm1.16}$
\\
&\tiny${(0.0013)}$ 
&\tiny${(0.0001)}$ 
&\tiny${(0.0001)}$  
&\tiny${(0.0009)}$ 
&\tiny${(0.0009)}$ \vspace{1mm}\\

{$\,$ GCNII}     & \one{77.39$_{\pm1.36}$}
& \three{91.16$_{\pm0.28}$}
& \two{92.97$_{\pm0.60}$}
& \one{82.72$_{\pm0.98}$}
& \two{89.98$_{\pm0.86}$}
\\ 
& \tiny${(0.0005)}$ 
& \tiny${(0.0001)}$ 
& \tiny${(0.0001)}$
& \tiny${(0.0009)}$ 
& \tiny${(0.0009)}$\\

\midrule
\textbf{DE-GNNs} \\
{$\,$ DGC}    & 66.71$_{\pm2.55}$
& 85.84$_{\pm0.01}$
& 82.95$_{\pm1.20}$
& 66.44$_{\pm0.63}$
& 76.13$_{\pm0.01}$
\\ 
& \tiny${(0.0015)}$ 
& \tiny${(0.0001)}$ 
& \tiny${(0.0001)}$ 
& \tiny${(0.0013)}$ 
& \tiny${(0.0014)}$ \vspace{1mm}\\

{$\,$ GRAND}  & 76.18$_{\pm1.56}$
& 89.20$_{\pm0.62}$
& 90.72$_{\pm0.87}$
&  81.09$_{\pm0.70}$
& 89.05$_{\pm0.73}$
\\ 
& \tiny${(0.0012)}$ 
& \tiny${(0.0001)}$ 
& \tiny${(0.0002)}$ 
& \tiny${(0.0009)}$ 
& \tiny${(0.0014)}$\\

\midrule
\multicolumn{4}{l}{\textbf{Ours - random weights}}\\
{$\,$ A-DGN\textsubscript{fix}}    & 67.47$_{\pm2.63}$
& 84.84$_{\pm0.66}$
& 87.37$_{\pm1.18}$
& 75.65$_{\pm0.45}$
& 85.30$_{\pm0.86}$
\\
& \tiny${(0.0438)}$ 
& \tiny${(0.0110)}$ 
& \tiny${(0.0341)}$ 
& \tiny${(\one{0.0147})}$ 
& \tiny${(\one{0.0208})}$
\vspace{1mm}\\
{$\,$ A-DGN\textsubscript{fix}(GCN)} & 69.95$_{\pm3.27}$
& 87.25$_{\pm0.57}$
& 89.21$_{\pm1.02}$
& 75.39$_{\pm1.27}$
& 84.25$_{\pm0.47}$
\\
& \tiny${(\one{0.0455})}$ 
& \tiny${(\one{0.0113})}$ 
& \tiny${(\one{0.0348})}$ 
& \tiny${(\one{0.0147})}$ 
& \tiny${(0.0206)}$
\\
\midrule
\multicolumn{4}{l}{\textbf{Ours - weight sharing}}\\
{$\,$ A-DGN\textsubscript{ws}} & 76.57$_{\pm1.00}$
& \two{91.35$_{\pm0.88}$}
& 92.45$_{\pm0.53}$
& \three{81.83$_{\pm0.75}$}
& 88.83$_{\pm1.12}$
\\
& \tiny${(0.0013)}$ 
& \tiny${(0.0001)}$
& \tiny${(0.0001)}$ 
& \tiny${(0.0008)}$
& \tiny${(0.0007)}$ 
\vspace{1mm}\\
{$\,$ A-DGN\textsubscript{ws}(GCN)}  & \two{76.82$_{\pm0.86}$}
& \one{91.71$_{\pm0.43}$}
& \one{93.27$_{\pm0.62}$}
& \two{82.35$_{\pm0.89}$}
& \one{90.52$_{\pm0.40}$}
\\
& \tiny${(0.0011)}$
& \tiny${(0.0001)}$
& \tiny${(0.0001)}$
& \tiny${(0.0007)}$
& \tiny${(0.0010)}$
\\

\bottomrule
\end{tabular}
\end{adjustbox}
\end{table}

\subsubsection{Heterophilic Graph Benchmarks}\label{sec:adgn_exp_hetero}
\myparagraph{Setup} In the graph heterophilic benchmarks we consider six well-known graph datasets for node classification, \ie Chameleon, Squirrel, Actor, Cornell, Texas, and Wisconsin. We employed the same experimental setting as \cite{geom-gcn}. Given the good performance observed in Section~\ref{sec:adgn_graph_prop_pred} and Section~\ref{sec:adgn_graph_benchmark}, here we study only the version of A-DGN with weight sharing. We perform hyperparameter tuning via grid search, optimizing the accuracy score. 

\myparagraph{Results} We present the results on the graph heterophilic benchmarks in Table~\ref{tab:results_hetero}, reporting the achieved accuracy.
We observe that our method obtains comparable results to state-of-the-art methods on four out of six datasets (\ie Actor, Cornell, Texas, Wisconsin). As stated in the work of \cite{heterophily_results2}, the main cause of performance degradation in heterophilic benchmarks is strongly related to over-smoothing. Therefore, since A-DGN is not designed to tackle the over-smoothing problem, the achieved level of accuracy on these datasets is a remarkable performance. In fact, our method outperforms most of the DGNs specifically designed to mitigate this phenomenon, ranking third and fourth among all the models when considering the average rank of each model across all benchmarks.

\begin{figure}[!h]
\begin{adjustbox}{center}
    \includegraphics[width=0.45\textwidth,trim={0.8cm 0 0 0},clip]{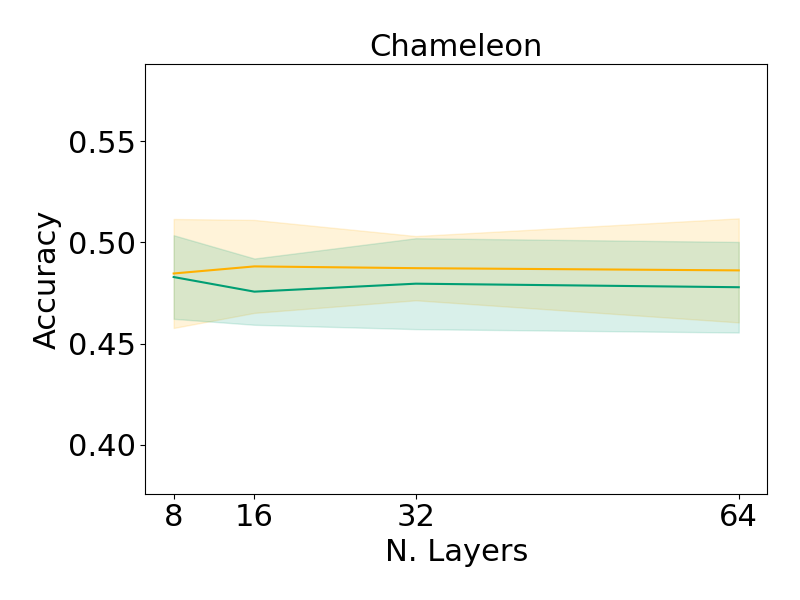}
    \includegraphics[width=0.45\textwidth]{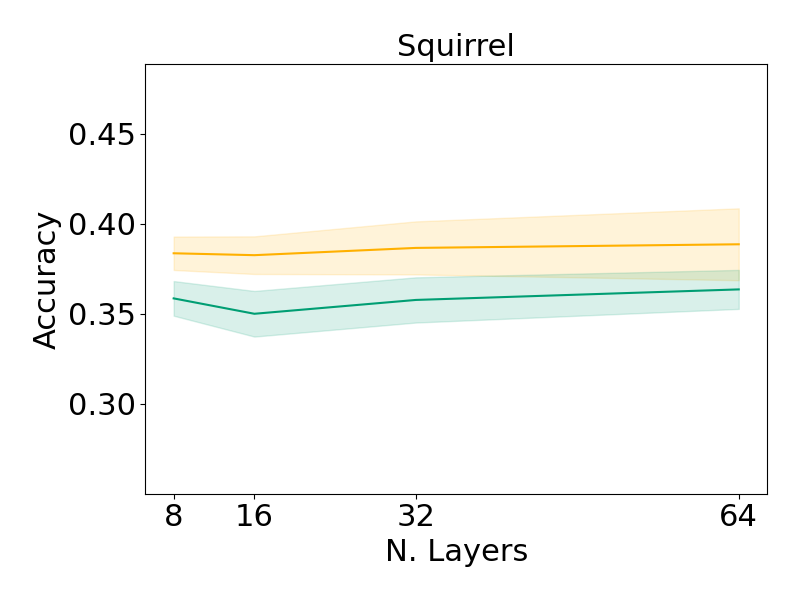}
\end{adjustbox}
\begin{adjustbox}{center}
\includegraphics[width=0.45\textwidth,trim={0.8cm 0 0 0},clip]{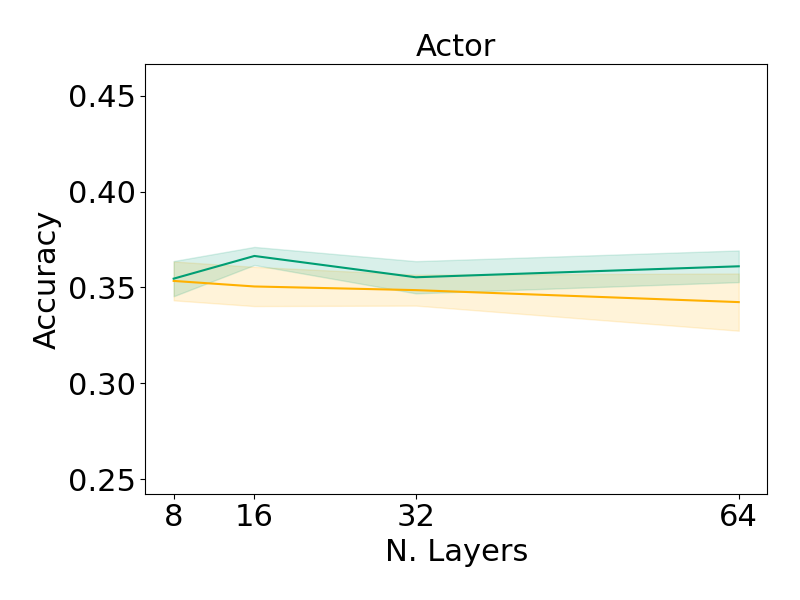}
\includegraphics[width=0.45\textwidth]{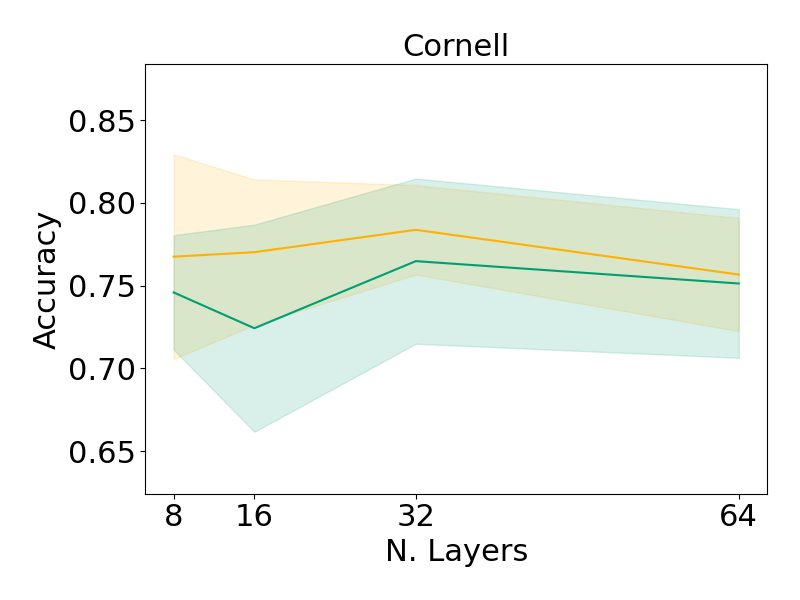}
\end{adjustbox}
\begin{adjustbox}{center}
    \includegraphics[width=0.45\textwidth,trim={0.8cm 0 0 0},clip]{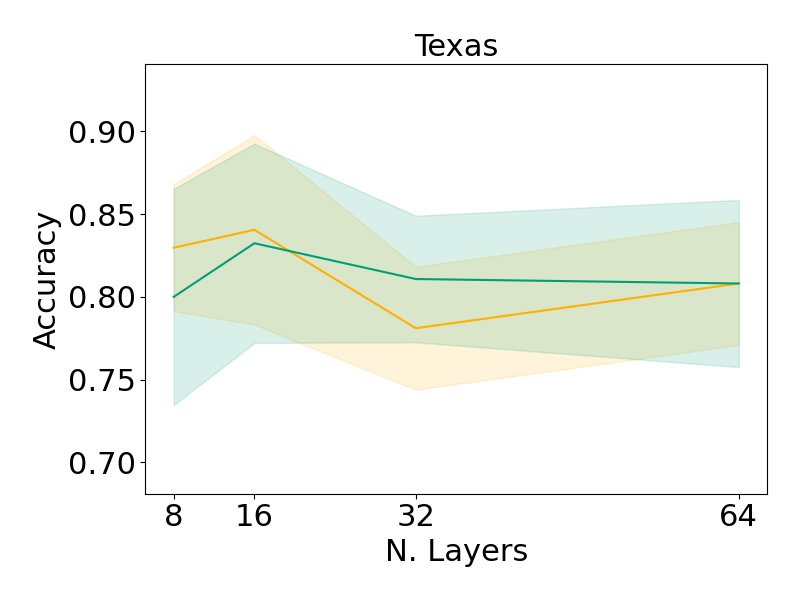}
    \includegraphics[width=0.45\textwidth]{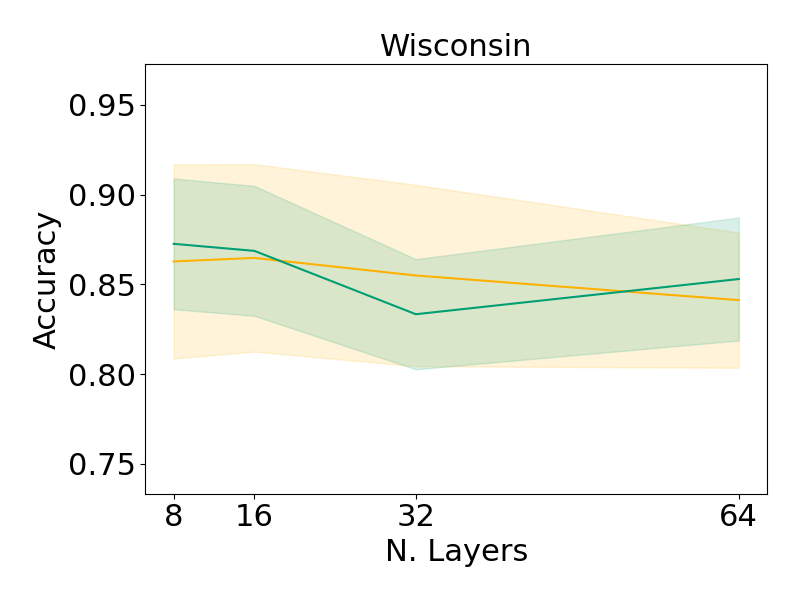}
\end{adjustbox}
\begin{adjustbox}{center}
    \includegraphics[width=0.17\textwidth]{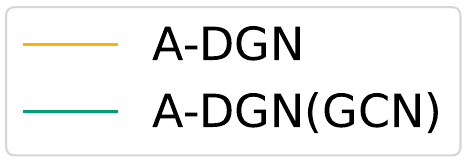}
\end{adjustbox}
\caption{The test accuracy of A-DGN with respect to the number of layers on all the graph heterophilic datasets. From the top left to the bottom, we show: Chameleon, Squirrel, Actor, Cornell, Texas, and Wisconsin. The accuracy is averaged over 10 train/validation/test splits.}
\label{fig:comparison_hetero}
\end{figure}

Similarly to the graph benchmarks in Section~\ref{sec:adgn_graph_benchmark}, our approach maintains or improves the performance as the number of layers increases, as Figure~\ref{fig:comparison_hetero} shows. Moreover, in this experiment, we show that A-DGN has outstanding performance even with 64 layers. Thus, A-DGN is able to effectively propagate the long-range information between nodes even in the scenario of graphs with high heterophily levels. Such result suggests that the presented approach can be a starting point to mitigate the over-smoothing problem as well.

\setlength{\tabcolsep}{3pt}
\begin{table}[ht]
\centering
\caption{Mean test set accuracy and std in percent averaged over different train/validation/test splits. The higher the better. 
The ``$*$'' results are obtained from \citet{heterophily_results2}, while the ``$\diamond$'' results are obtained from \citet{topping2022understanding}.  
We also report the average ranking of each model across all benchmarks, showing that the proposed method ranks third and fourth (in the GCN variant) among all the models considered.
\label{tab:results_hetero}}
\scriptsize
\begin{adjustbox}{center}
\begin{tabular}{lcccccc|c}
\toprule
\multirow{2}{*}{\textbf{Model}}& \multirow{2}{*}{\textbf{Chameleon}} & \multirow{2}{*}{\textbf{Squirrel}} & \multirow{2}{*}{\textbf{Actor}} & \multirow{2}{*}{\textbf{Cornell}} & \multirow{2}{*}{\textbf{Texas}} & \multirow{2}{*}{\textbf{Wisconsin}} & \textbf{avg}\\
&&&&&&&\textbf{rank}\\
\midrule
\textbf{Baseline}\\
$\,$ MLP$^*$                       & 46.21$_{\pm2.99}$       & 28.77$_{\pm1.56}$        & \three{36.53$_{\pm0.70}$} & \three{81.89$_{\pm6.40}$} & 80.81$_{\pm4.75}$ & 85.29$_{\pm3.31}$ & 7.67  \\
\midrule
\textbf{MPNNs}\\
$\,$ GGCN$^*$                      & \one{71.14$_{\pm1.84}$}  & \one{55.17$_{\pm1.58}$}   & \one{37.54$_{\pm1.56}$} & \one{85.68$_{\pm6.63}$} & \one{84.86$_{\pm4.55}$} & \three{86.86$_{\pm3.29}$} & \one{1.33}  \\
$\,$ GPRGNN$^*$                    & 46.58$_{\pm1.71}$        & 31.61$_{\pm1.24}$         & 34.63$_{\pm1.22}$       & 80.27$_{\pm8.11}$       & 78.38$_{\pm4.36}$       & 82.94$_{\pm4.21}$ & 8.50  \\
$\,$ H2GCN$^*$                     & 60.11$_{\pm2.15}$        & 36.48$_{\pm1.86}$         & 35.70$_{\pm1.00}$       & \two{82.70$_{\pm5.28}$} & \one{84.86$_{\pm7.23}$}       & \one{87.65$_{\pm4.98}$} & \two{4.67}  \\
$\,$ GCNII$^*$                     & 63.86$_{\pm3.04}$        & 38.47$_{\pm1.58}$         & \two{37.44$_{\pm1.30}$} & 77.86$_{\pm3.79}$       & 77.57$_{\pm3.83}$       & 80.39$_{\pm3.40}$ & \three{5.83}  \\
$\,$ Geom-GCN$^*$                  & 60.00$_{\pm2.81}$        & 38.15$_{\pm0.92}$         & 31.59$_{\pm1.15}$       & 60.54$_{\pm3.67}$       & 66.76$_{\pm2.72}$       & 64.51$_{\pm3.66}$ & 9.17  \\
$\,$ PairNorm$^*$                  & \three{62.74$_{\pm2.82}$}& \three{50.44$_{\pm2.04}$} & 27.40$_{\pm1.24}$       & 58.92$_{\pm3.15}$       & 60.27$_{\pm4.34}$       & 48.43$_{\pm6.14}$ & 11.00 \\
$\,$ GraphSAGE$^*$                 & 58.73$_{\pm1.68}$        & 41.61$_{\pm0.74}$         & 34.23$_{\pm0.99}$       & 75.95$_{\pm5.01}$       & 82.43$_{\pm6.14}$       & 81.18$_{\pm5.56}$ & 6.67  \\
$\,$ GCN$^*$                       & \two{64.82$_{\pm2.24}$}  & \two{53.43$_{\pm2.01}$}   & 27.32$_{\pm1.10}$       & 60.54$_{\pm5.30}$       & 55.14$_{\pm5.16}$       & 51.76$_{\pm3.06}$ & 9.83  \\
$\,$ GAT$^*$                       & 60.26$_{\pm2.50}$        & 40.72$_{\pm1.55}$         & 27.44$_{\pm0.89}$       & 61.89$_{\pm5.05}$       & 52.16$_{\pm6.63}$       & 49.41$_{\pm4.09}$ & 11.00 \\
\midrule
\multicolumn{8}{l}{\textbf{Multi-hop DGNs}}\\
$\,$ FA$^\diamond$                 & 42.33$_{\pm0.17}$        & 40.74$_{\pm0.13}$         & 28.68$_{\pm0.16}$       & 58.29$_{\pm0.49}$       & 64.82$_{\pm0.29}$       & 55.48$_{\pm0.62}$ & 11.33 \\
$\,$ DIGL$^\diamond$               & 42.02$_{\pm0.13}$       & 33.22$_{\pm0.14}$       & 24.77$_{\pm0.32}$       & 58.26$_{\pm0.50}$       & 62.03$_{\pm0.43}$       & 49.53$_{\pm0.27}$ & 15.33 \\
$\,$ DIGL+Undir$^\diamond$  &
42.68$_{\pm0.12}$       & 32.48$_{\pm0.23}$       & 25.45$_{\pm0.30}$       & 59.54$_{\pm0.64}$       & 63.54$_{\pm0.38}$       & 52.23$_{\pm0.54}$ & 14.00 \\
$\,$ SDRF$^\diamond$               & 42.73$_{\pm0.15}$       & 37.05$_{\pm0.17}$       & 28.42$_{\pm0.75}$       & 54.60$_{\pm0.39}$       & 64.46$_{\pm0.38}$       & 55.51$_{\pm0.27}$ & 12.67 \\ 
$\,$ SDRF+Undir$^\diamond$  & 
44.46$_{\pm0.17}$       & 37.67$_{\pm0.23}$       & 28.35$_{\pm0.06}$       & 57.54$_{\pm0.34}$       & 70.35$_{\pm0.60}$       & 61.55$_{\pm0.86}$ & 11.67 \\

\midrule
\multicolumn{8}{l}{\textbf{Ours - weight sharing}}\\
$\,$ A-DGN                           & 49.69$_{\pm2.59}$     & 38.70$_{\pm1.26}$       & 35.34$_{\pm1.01}$       & 78.38$_{\pm2.70}$        & \three{82.97$_{\pm2.72}$}      & 86.67$_{\pm3.70}$ & \three{5.83}  \\
$\,$ A-DGN(GCN)                      & 48.71$_{\pm3.07}$     & 36.36$_{\pm1.08}$       & 36.11$_{\pm0.83}$       & 76.49$_{\pm4.99}$        & \two{83.24$_{\pm6.02}$}      & \two{87.25$_{\pm3.64}$} & 6.50  \\
\bottomrule
\end{tabular}
\end{adjustbox}
\end{table}

\clearpage
\newpage
\section{Space and Weight Antisymmetry}\label{sec:SWAN}
\begin{figure}[h]
    \centering
    \hspace{-0.6cm}
  \begin{subfigure}{0.21\textwidth}
    \centering \includegraphics[width=0.8\linewidth]{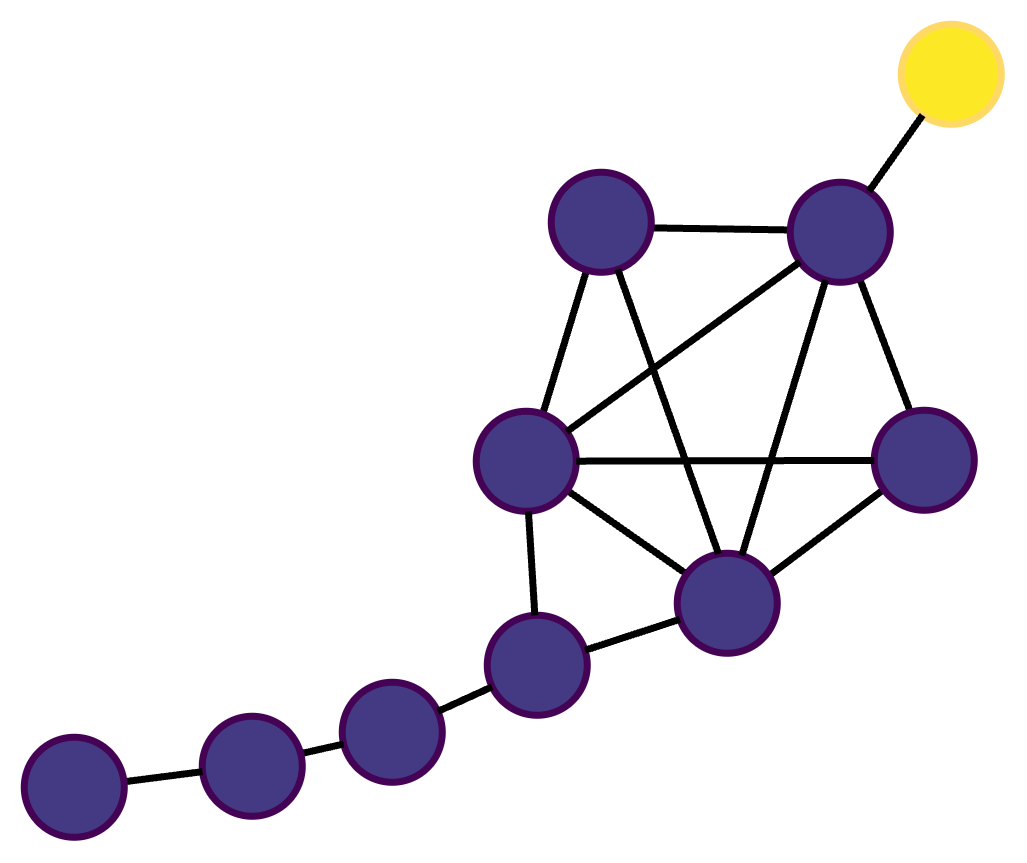}
    \caption{\scriptsize Source\vspace{0.32cm}}
    \label{fig:sub1}
  \end{subfigure}\hspace{-5mm}
   \begin{subfigure}{0.21\textwidth}
    \centering \includegraphics[width=0.8\linewidth]{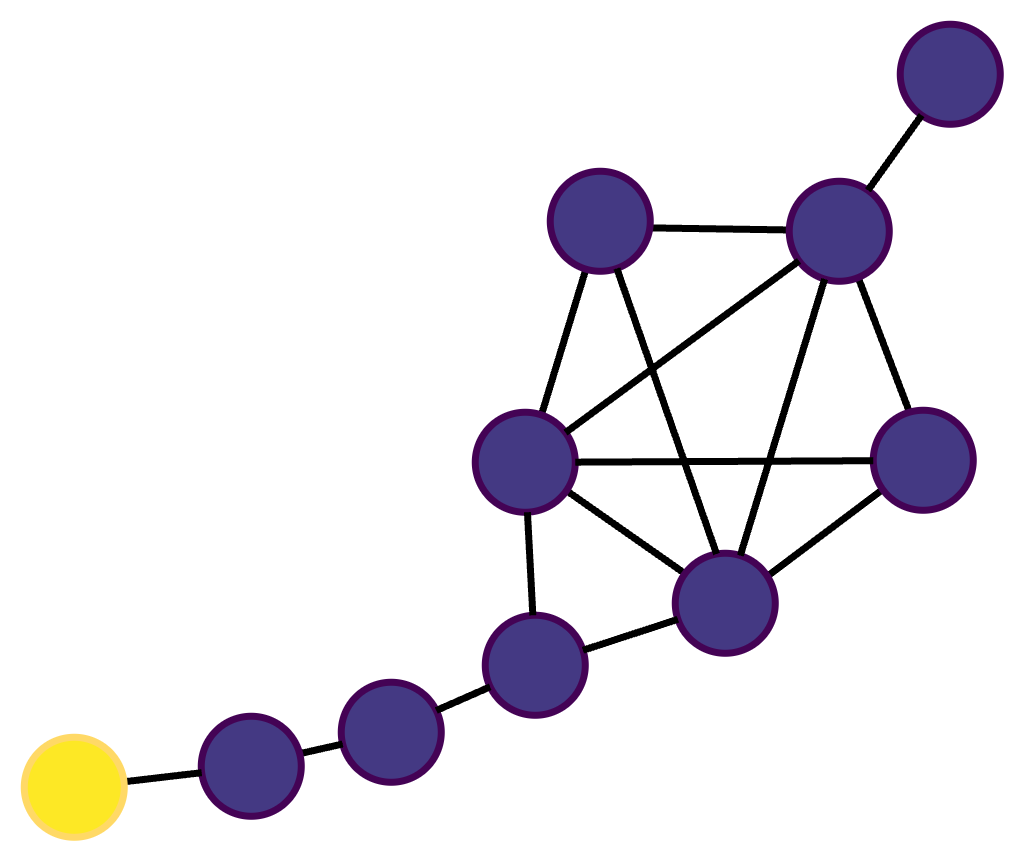}
    \caption{\scriptsize Target\vspace{0.32cm}}
    \label{fig:sub2}
  \end{subfigure}\hspace{-5mm}
   \begin{subfigure}{0.21\textwidth}
    \centering \includegraphics[width=0.8\linewidth]{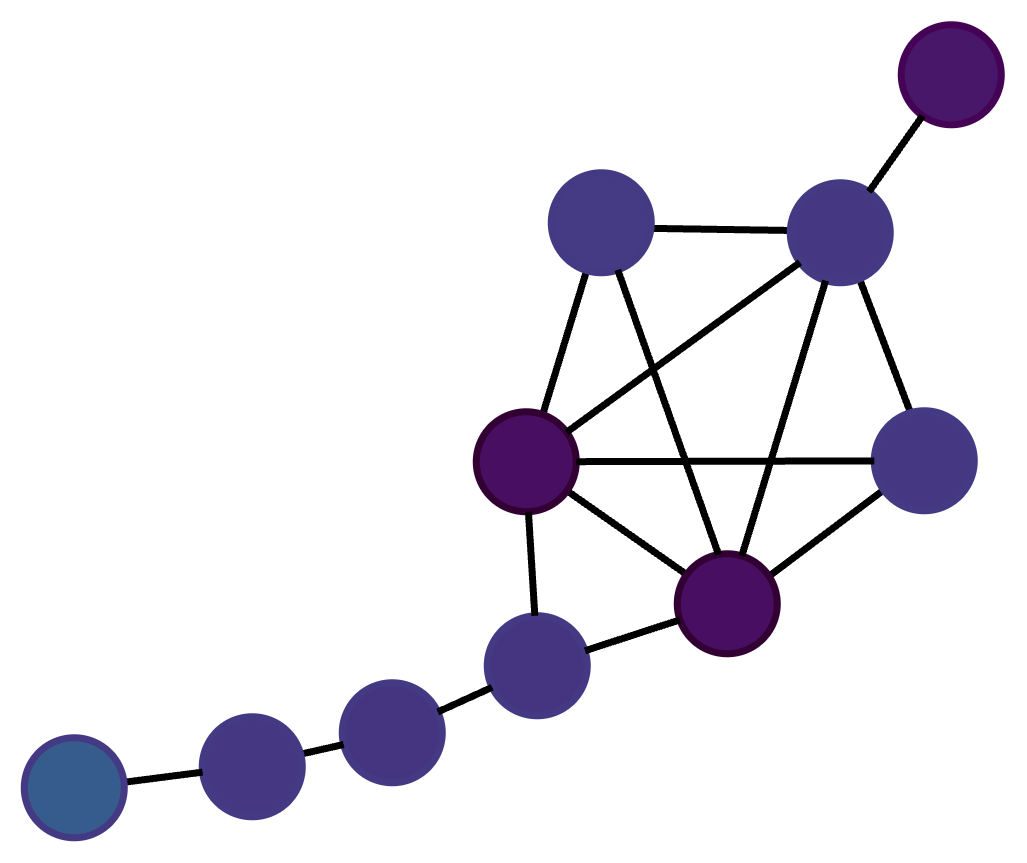}
    \caption{\scriptsize Diffusion\vspace{0.32cm}}
    \label{fig:sub3}
  \end{subfigure}\hspace{-2mm}
   \begin{subfigure}{0.21\textwidth}
    \centering \includegraphics[width=0.8\linewidth]{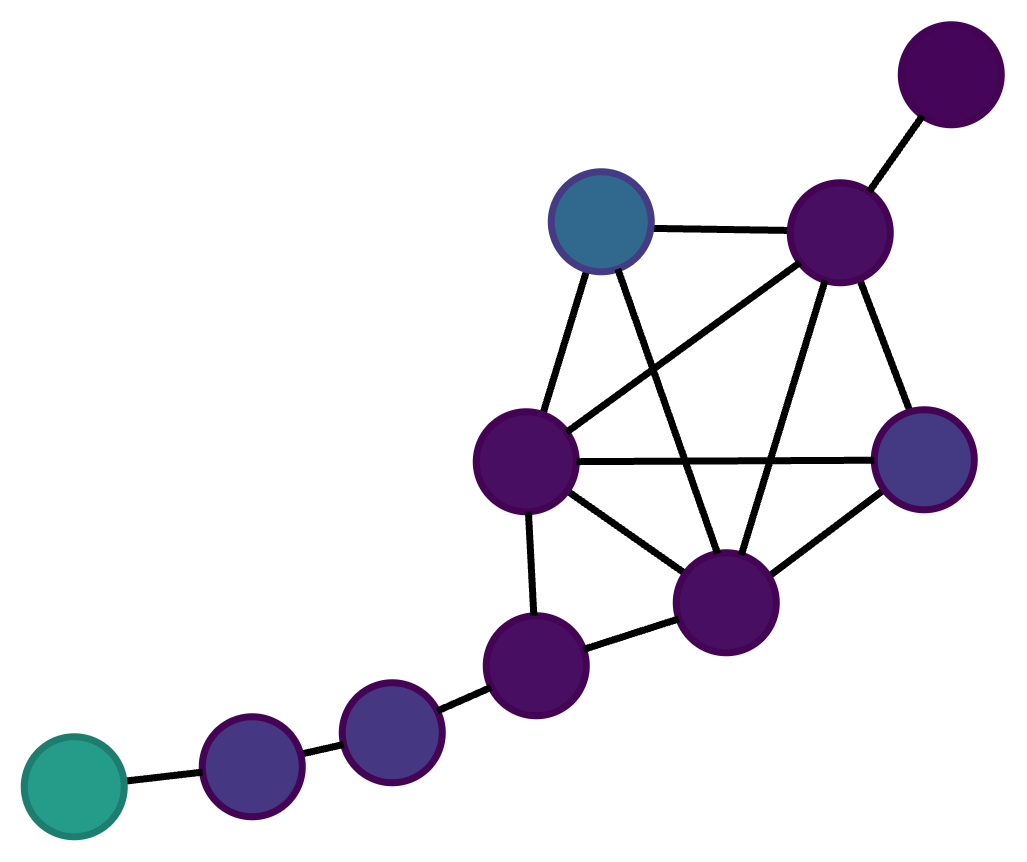}
    \caption{\scriptsize Local \\Non-Dissipativity }
    \label{fig:sub4}
  \end{subfigure}\hspace{-2mm}
   \begin{subfigure}{0.21\textwidth}
    \centering \includegraphics[width=0.8\linewidth]{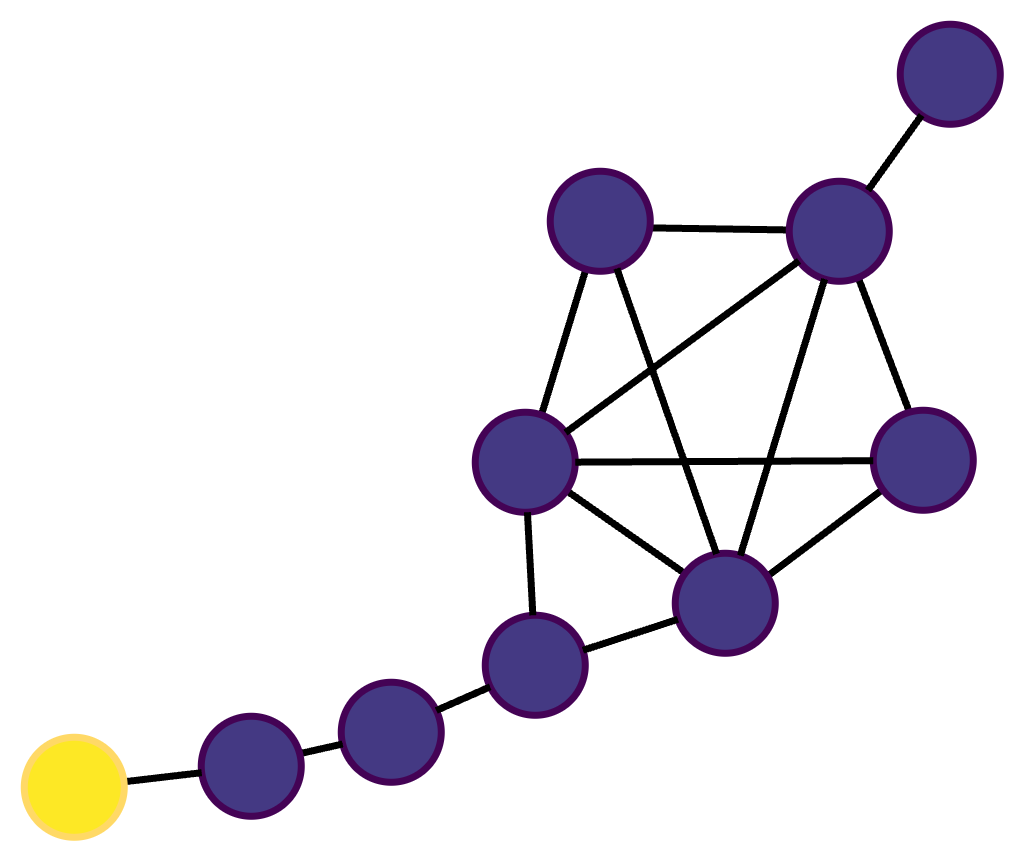}
    \caption{\scriptsize Local and Global Non-Dissipativity}
    \label{fig:sub5}
   \end{subfigure}
  \caption{An illustration of the ability of Global and Local Non-Dissipativity (e) to propagate information to distant nodes, from the source (a) to the target (b). Other dynamics, such as diffusion (c) cannot achieve the desired behavior, while Local Non-Dissipativity (d) has only limited effect. A-DGN implements (d) and SWAN implements (e). The color yellow represents the signal at its maximum strength, while the color blue indicates the absence of signal. Intermediate signal strengths are depicted through a gradient of colors between yellow and blue.}
    \label{fig:propagation}
\end{figure}

In Section~\ref{sec:ADGN}, we showed that by incorporating an antisymmetric transformation to the learned channel-mixing weights, it is possible to obtain a \emph{locally}, \ie node-wise, non-dissipative behavior. Here, we extend the model introduced in the previous section by proposing \textbf{\gls*{SWAN}} (\underline{S}pace-\underline{W}eight \underline{An}tisymmetric Deep Graph Network)\index{SWAN}, a novel DGN model that is both \emph{globally} (\ie graph-wise) and \emph{locally} (\ie node-wise) non-dissipative, achieved by space and weight antisymmetric parameterization. To understand the behavior of SWAN and its effectiveness in mitigating oversquashing, we propose a \emph{global}, \ie graph-wise, analysis, and show that compared to A-DGN (Section~\ref{sec:ADGN}), our SWAN is both globally and locally non-dissipative. The immediate implication of this property is that SWAN is guaranteed to have a constant information flow rate, thereby mitigating oversquashing. 
This property is illustrated in Figure~\ref{fig:propagation}, which depicts the signal strength received by a target node from a distant node in the graph. In this scenario, SWAN demonstrates improved capabilities in propagating information across the graph, as the target node receives the entire signal. In contrast, diffusion and local non-dissipative approaches can only transmit a portion of the signal to the target node.
To complement our theoretical analysis, we experiment with several long-range benchmarks, from synthetic to real-world datasets and tasks. 

In this section, we present the following contributions:
\begin{itemize}
    \item A novel graph perspective theoretical analysis of the stability and non-dissipativity of antisymmetric DE-GNNs, providing a general design principle for introducing non-dissipativity as an inductive bias in any DE-GNN model. 
    \item We propose SWAN, a space and weight antisymmetric DE-GNN with a constant information flow rate, and an increased distant node interaction sensitivity.
    \item We experimentally verify our theoretical understanding of SWAN, on both synthetic and real-world datasets. 
\end{itemize} 

\subsection{SWAN: Space-Weight Antisymmetric DGN}\label{sec:swan_method}

In this section, we are interested in achieving both local and global non-dis\-si\-pa\-tive behavior in our system. 
\begin{definition}[Local non-dissipativity]
    A system is \textbf{locally non-dis\-si\-pa\-tive}\index{local non-dissipativity|see {non-dissipativity}}\index{non-dissipativity!local} when the final representation of each node retains its historical information.
\end{definition}
\begin{definition}[Global non-dissipativity]
    A system is \textbf{globally non-dis\-si\-pa\-tive}\index{global non-dissipativity|see {non-dissipativity}}\index{non-dissipativity!global} when the information between nodes continues to propagate at a constant rate.
\end{definition}
Therefore, local non-dissipativity ensures the long-term memory capacity of individual nodes, and global non-dissipativity ensures that the system retains its ability to share information between nodes with the same effectiveness regardless of node distance.

To achieve such properties, we keep being interested in studying the \emph{stability} and \emph{non-dis\-si\-pa\-tiv\-ity} propagation of information. 
Therefore, we follow the analysis techniques presented in Section~\ref{sec:ADGN} and focus on analyzing the sensitivity of an ODE solution with respect to its initial condition (Equation~\ref{eq:appendix1}).

Again, we consider the Jacobian, $\mathbf{J}(t)$, to not change significantly over time. An interested reader is referred to Appendix~\ref{appendix:jacobian_stable} where we provide a discussion of the justification as well as numerical verification of our the assumption. 

We recall from Section~\ref{sec:adgn} that, by solving Equation~\ref{eq:appendix1} analytically, the qualitative behavior of $\partial \mathbf{x}(t)/\partial \mathbf{x}(0)$ is determined by the real parts of the eigenvalues of $\mathbf{J}$, leading to three different behaviors: (i) instability, (ii) dissipativity (i.e., information loss), (iii) non-dissipativity (i.e., information preservation). 

\begin{figure}[h]
\centering
\begin{subfigure}{0.32\textwidth}
    \includegraphics[width=\textwidth]{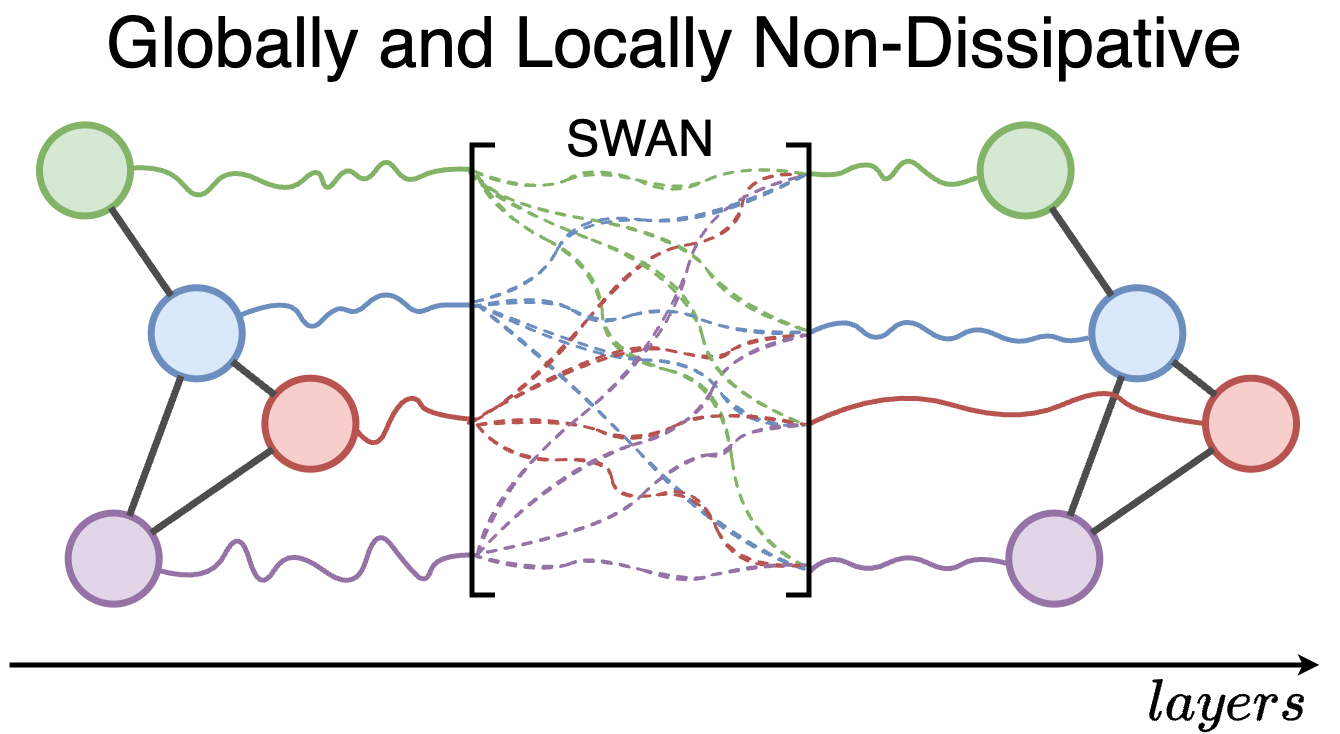}
    \caption{}
        \label{fig:globalNonDis}
\end{subfigure}
\begin{subfigure}{0.32\textwidth}
    \includegraphics[width=\textwidth]{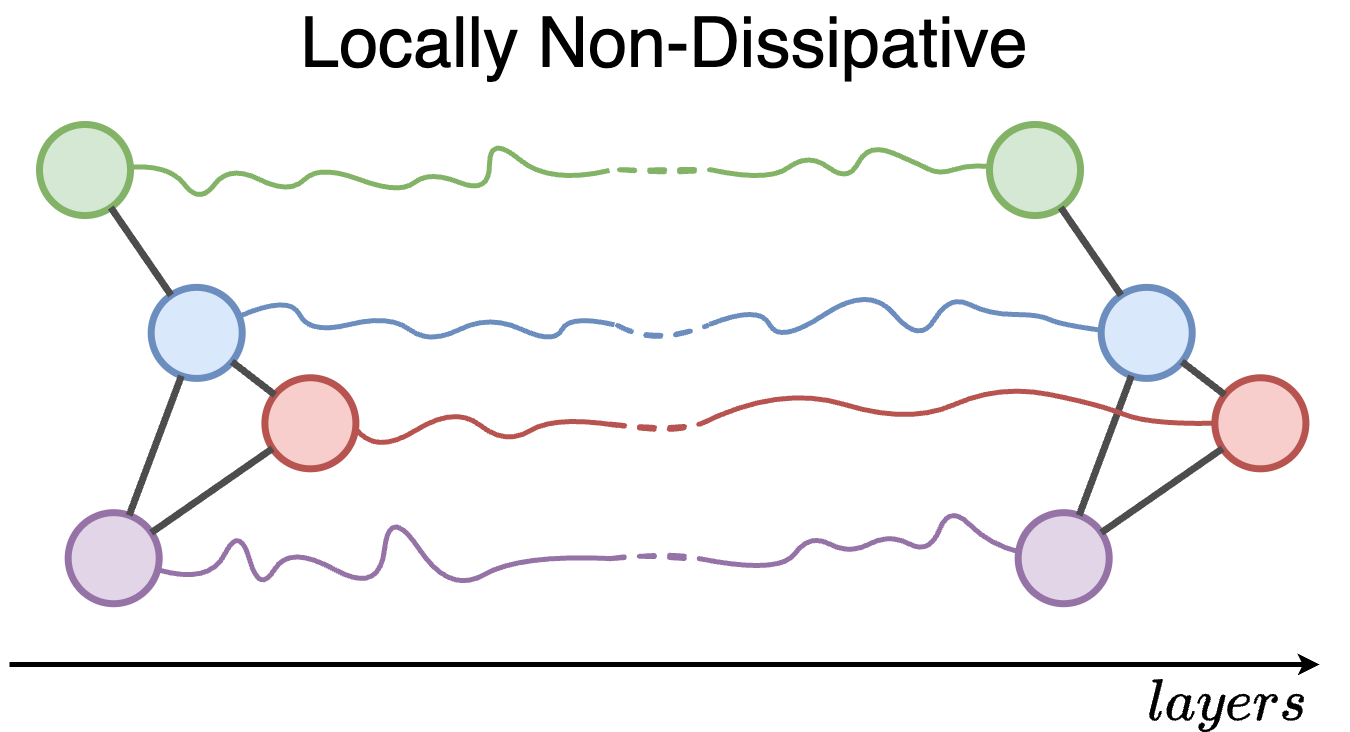}
    \caption{}
    \label{fig:localNonDis}
\end{subfigure}
\begin{subfigure}{0.32\textwidth}
    \includegraphics[width=\textwidth]{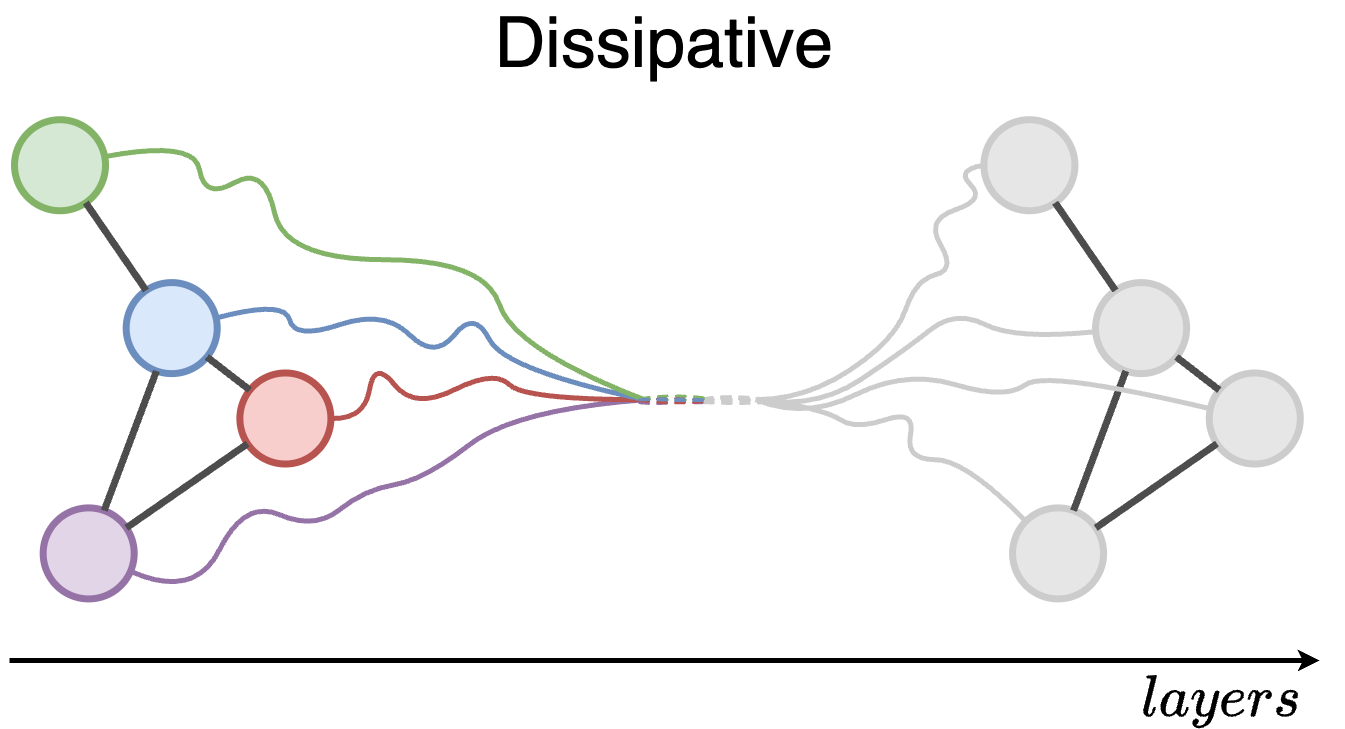}
    \caption{}
        \label{fig:dissipative}

\end{subfigure}
    \caption{The difference among non-dissipative and dissipative behaviors. With global (\ie graph-wise) and local (\ie node-wise) 
    non-dissipative behavior (a), information is propagated between any pair of nodes with a viable path in the graph. Therefore, such a behavior increases the long-range effectiveness of the model, with propagation reaching distant nodes. A model exhibiting local 
    non-dissipative behavior (b) enhances only the long-term memory capacity of individual nodes. A model demonstrating dissipative behavior (c) exhibits a convergence of node features toward non-informative values.}
    \label{fig:multiple_behaviors}
\end{figure}

We now turn to present \textbf{SWAN}, \underline{s}pace-\underline{w}eight  \underline{an}tisymmetric DGN. We analyze its theoretical behavior and show that it is both \emph{global} (\ie graph-wise) and \emph{local} (\ie node-wise) \emph{non-dissipative}. As a consequence, one of the key features of SWAN is that it has a constant \emph{global} information flow rate, unlike common diffusion DGNs. Therefore, SWAN should theoretically be able to propagate information between any nodes with a viable path in the graph, allowing to mitigate oversquashing.  
Figure~\ref{fig:multiple_behaviors}  exemplifies the differences between dissipative, local non-dissipative, and global and local non-dissipative systems. 

\myparagraph{Space and Weight Antisymmetry} We define SWAN by including a new term that introduces antisymmetry in the aggregation function to A-DGN, i.e.,
\begin{equation} \label{eq:new_adgn_nodewise}
    \frac{d\mathbf{x}_u(t)}{d t} =  \sigma\Bigl((\mathbf{W}-\mathbf{W}^\top)\mathbf{x}_u(t) + \Phi(\{\mathbf{x}_v\}_{v\in\mathcal{N}_u}) + \beta\gls*{Psi}(\{\mathbf{x}_v\}_{v\in\mathcal{N}_u})\Bigr),
\end{equation}
where $\mathbf{W}$ is a learnable weigh matrix (as usual). 
$\Phi$ and $\Psi$ are permutation invariant neighborhood aggregation functions, where $\Psi$ performs antisymmetric aggregation. 
While $\Psi$ can assume various forms of  antisymmetric aggregation functions with imaginary eigenvalues, and $\Phi$ can be any aggregation function, in this section we explore the following family of parametrizations:
\begin{equation}\label{eq:aggFuncs}
\begin{split}
\Phi = \sum_{v\in\mathcal{N}_u}(\hat{\bfA}_{uv}+\hat{\bfA}_{vu})(\mathbf{V}-\bfV^\top)\mathbf{x}_v(t),\\
\Psi = \sum_{v\in\mathcal{N}_u}(\tilde{\bfA}_{uv}-\tilde{\bfA}_{vu})(\mathbf{Z}+\mathbf{Z}^\top)\mathbf{x}_v(t),
\end{split}
\end{equation}    
where $\mathbf{V}$, and $\mathbf{Z}$ are learnable weight matrices, and $\tilde{\mathbf{A}},\hat{\mathbf{A}} \in \mathbb{R}^{|\mathcal{V}|\times |\mathcal{V}|}$ 
are neighborhood aggregation matrices that can be either pre-defined or learned.  In our experiments, we consider two instances of Equation~\ref{eq:aggFuncs}. The first, where $\tilde{\mathbf{A}},\hat{\mathbf{A}}$ are pre-defined by the random walk and symmetrically normalized adjacency matrices, respectively. The second, allows $\tilde{\mathbf{A}},\hat{\mathbf{A}}$ to be learned (see Section~\ref{sec:swan_learn_A}). More importantly, as we show now, the blueprint of $\Psi$ and $\Phi$, which is described in Equation~\ref{eq:aggFuncs} stems from a theoretical analysis of SWAN, that shows its ability to be non-dissipative both locally and globally, therefore leading to a \emph{globally} constant information flow, regardless of time, that is, the model's depth. Lastly, we note that the general formulation of $\Phi$ and $\Psi$ provide a general design principle for introducing non-dissipativity as an inductive bias in any DE-DGNs.

\subsubsection{Node-wise Analysis of SWAN} \label{sec:nodewise}
We reformulate Equation~\ref{eq:new_adgn_nodewise} to consider the formulation of $\Phi$ and $\Psi$ as in Equation~\ref{eq:aggFuncs}, reading:
\begin{align}\label{eq:new_adgn_node} 
    \frac{d\mathbf{x}_u(t)}{d t} =  \sigma\Bigl((\mathbf{W}-\mathbf{W}^\top)\mathbf{x}_u(t) &+ \sum_{v\in\mathcal{N}_u}(\hat{\bfA}_{uv}+\hat{\bfA}_{vu})(\mathbf{V}-\bfV^\top)\mathbf{x}_v(t) \nonumber\\  
    &+ \beta\sum_{v\in\mathcal{N}_u}(\tilde{\bfA}_{uv}-\tilde{\bfA}_{vu})(\mathbf{Z}+\mathbf{Z}^\top)\mathbf{x}_v(t)\Bigr).
\end{align}

\myparagraph{SWAN is locally non-dissipative} Following the sensitivity analysis introduced in Section~\ref{sec:ADGN}, we show that SWAN is stable and non-dissipative from the node perspective, \ie it is \emph{locally} non-dissipative
\index{non-dissipativity!local}. Therefore, we study the Jacobian matrix of our node-wise reformulation of SWAN.

In this case, the Jacobian $\mathbf{J}(t) = \mathbf{M}_1\mathbf{M}_2$ of Equation~\ref{eq:new_adgn_node} is composed of :
\begin{align}
\nonumber \mathbf{M}_1 &= \mathrm{diag}\Bigl[\sigma'\Bigl((\mathbf{W}-\mathbf{W}^\top)\mathbf{x}_u(t) + \sum_{v\in\mathcal{N}_u}(\hat{\bfA}_{uv}+\hat{\bfA}_{vu})(\mathbf{V}-\bfV^\top)\mathbf{x}_v(t) \\  &\hspace{4.07cm}+ \beta\sum_{v\in\mathcal{N}_u}(\tilde{\bfA}_{uv}-\tilde{\bfA}_{vu})(\mathbf{Z}+\mathbf{Z}^\top)\mathbf{x}_v(t)\Bigl)\Bigr],\\
\mathbf{M}_2 &= (\mathbf{W}-\mathbf{W}^\top) + (\hat{\bfA}_{uu}+\hat{\bfA}_{uu})(\mathbf{V}-\bfV^\top).
\end{align}
Following results from \cite{chang2018antisymmetricrnn} and Section~\ref{sec:ADGN}, only the eigenvalues of $\mathbf{M}_2$ determine the \emph{local} stability and non-dissipativity of the system in Equation~\ref{eq:new_adgn_nodewise} for the final behavior of the model. Specifically, if the real part of all the eigenvalues of $\bfM_2$ is zero, then stability and non-dissipativity are achieved. We note that this is indeed the case in our system, since the real part of the eigenvalues of antisymmetric matrices is zero, and $\bfM_2$ is composed of a summation of two antisymmetric matrices.

\subsubsection{Graph-wise Analysis of SWAN}\label{sec:graphwise}
While the node perspective analysis is important because it shows the long-term memory capacity of individual nodes, as illustrated in Figure~\ref{fig:localNonDis}, it overlooks the \emph{pairwise} interactions between nodes, which is equivalently described by considering the properties of Equation~\ref{eq:new_adgn_nodewise} with respect to the graph, and is illustrated in Figure \ref{fig:globalNonDis}.

As we show below, our SWAN is a globally non-dissipative architecture, and therefore it allows a constant rate of information flow and interactions between all nodes, independently of time $t$, equivalent to the network's depth.  Therefore, we deem that SWAN's non-dissipativity behavior is beneficial to address oversquashing in MPNNs.

\myparagraph{SWAN is globally non-dissipative}\index{non-dissipativity!global} We start by reformulating Equation \ref{eq:new_adgn_node} from a node-wise formulation to a graph-perspective formulation, as follows:
\begin{align}
\label{eq:new_adgn_graphwise} \nonumber
    \frac{d\mathbf{X}(t)}{d t} = \sigma \Bigl(\mathbf{X}(t){(\mathbf{W}-\mathbf{W}^\top)} &+{(\hat{\mathbf{A}}+\hat{\mathbf{A}}^\top)\mathbf{X}(t)(\mathbf{V}-\bfV^\top)} \nonumber\\ 
    &+ \beta{(\tilde{\mathbf{{A}}}-\tilde{\mathbf{{A}}}^\top)\mathbf{X}(t)(\mathbf{Z}+\mathbf{Z}^\top)}\Bigr).
\end{align}
We therefore turn to analyze the ODE in Equation~\ref{eq:new_adgn_graphwise} from a graph perspective, again, using the sensitivity analysis introduced in Section~\ref{sec:ADGN}. 
Applying the vectorization operator, 
the Jacobian $\mathbf{J}(t) = \mathbf{M}_1\mathbf{M}_2$ writes as the multiplication of:
\begin{align}
\label{eq:J_SWAN} \nonumber
\nonumber \mathbf{M}_1 &= \mathrm{diag}\Bigl[\mathrm{vec}\Bigl(\sigma'\Bigl(\mathbf{I}\mathbf{X} (\mathbf{W}-\mathbf{W}^\top)+ \\ \nonumber&\hspace{1cm}+ (\hat{\mathbf{A}}+\hat{\mathbf{A}}^\top)\mathbf{X}(t)(\mathbf{V}-\bfV^\top) \\ &\hspace{1cm}+\beta(\tilde{\mathbf{{A}}}-\tilde{\mathbf{{A}}}^\top)\mathbf{X}(t)(\mathbf{Z}+\mathbf{Z}^\top) \Bigr)\Bigr) \Bigr] \\
\label{eq:graphwiseM2}
\nonumber \mathbf{M}_2 &= (\mathbf{W}-\mathbf{W}^\top)^\top \otimes \bfI+ \\ \nonumber &\hspace{1cm} + (\mathbf{V}-\bfV^\top)^\top\otimes (\hat{\mathbf{A}}+\hat{\mathbf{A}}^\top) \\  &\hspace{1cm} + \beta(\mathbf{Z}+\mathbf{Z}^\top)^\top \otimes (\tilde{\mathbf{{A}}}-\tilde{\mathbf{{A}}}^\top),
\end{align}
where $\bfI \in \mathbb{R}^{|\mathcal{V}|\times|\mathcal{V}|}$ 
is the identity matrix, $\rm{vec}$ is the vectorization operator, and $\otimes$ is the Kronecker product. An interested reader is referred to Appendix~\ref{app:jacobian} for the explicit derivations of the Jacobian.

Similar to our node-wise analysis in Section~\ref{sec:nodewise}, $\mathbf{M}_1$ in Equation~\ref{eq:J_SWAN} is a diagonal matrix, and therefore to obtain stability and non-dissipativity, we need to demand that $\bfM_2$ from Equation~\ref{eq:graphwiseM2} has eigenvalues with real part equal to zero. Indeed, we see that $\bfM_2$ satisfies this condition because it is composed of a summation of three antisymmetric matrices, whose eigenvalues are all purely imaginary, i.e., have a real part of zero. Thus, we conclude that SWAN (Equation~\ref{eq:new_adgn_graphwise}) 
is both stable and globally and locally non-dissipative. 

We now show that the properties of global 
stability and non-dissipativity allow us, theoretically, to design DGNs that can mitigate oversquashing.

\begin{theorem}[SWAN has a constant global  information propagation rate]
\label{thm:swanConstantRate} The information propagation rate among the graph nodes $\mathcal{V}$ is constant, $c$, independently of time $t$:
\begin{equation}\label{eq:swanConstant}   
\left\Vert\frac{\partial \rm{vec}(\mathbf{X}(t))}{\partial \rm{vec}(\mathbf{X}(0))}\right\Vert = c.
\end{equation}
\end{theorem}

\begin{boxedproof}
Let us consider the following equation:
\begin{equation}    \label{eq:proof_graph_wise}   
\frac{d}{dt}\left(\frac{\partial \mathbf{X}(t)}{\partial \mathbf{X}(0)}\right) = \frac{d}{dt}\left(\frac{\partial \rm{vec}(\mathbf{X}(t))}{\partial \rm{vec}(\mathbf{X}(0))}\right) = \mathbf{J}(t) \frac{\partial \rm{vec}(\mathbf{X}(t))}{\partial\rm{vec}(\mathbf{X}(0))}. 
\end{equation}
We follow the assumption in Section~\ref{sec:ADGN} that the Jacobian, $\mathbf{J}(t)$, does not change significantly over time, 
then we can apply results from autonomous differential equations and solve Equation~\ref{eq:proof_graph_wise}: 
\begin{equation}
    \frac{\partial \rm{vec}(\mathbf{X}(t))}{\partial\rm{vec}(\mathbf{X}(0))} = e^{t \mathbf{J}} = \mathbf{T} e^{t \mathbf{\Lambda}}\mathbf{T}^{-1} = \mathbf{T} 
    \big(\sum_{k=0}^\infty \frac{(t \mathbf{\Lambda})^k}{k!}\big)
    \mathbf{T}^{-1},
\end{equation}
where $\mathbf{\Lambda}$ is the diagonal matrix whose non-zero entries contain the eigenvalues of $\mathbf{J}$, and $\mathbf{T}$ has the eigenvectors of $\mathbf{J}$ as columns. 
As we have previously shown, 
it holds that 
 $Re(\lambda_i(\mathbf{J}(t))) 
= 0$ for $i=1,...,d$, since the Jacobian is the result of the multiplication between a diagonal matrix and an antisymmetric matrix. Therefore, the magnitude of $\partial \mathbf{X}(t)/\partial \mathbf{X}(0)$ is constant over time, and the input graph information is effectively propagated through the successive layers into the final nodes' features. 
\end{boxedproof}

The outcome of Theorem 3.1 is that regardless of the integration time $t$ (equivalent to $\ell = t/\epsilon$ layers of SWAN, where $\epsilon$ is the step size), the information between nodes will continue to propagate at a constant rate, unlike diffusion-based DGNs that exhibit an exponential decay in the propagation rate with respect to time, as shown below.

\begin{theorem}[Time Decaying Propagation in Diffusion DGNs] \label{thm:diffusionExpDecay} A diffusion-based network with Jacobian eigenvalues with magnitude $\bfK_{ii}= |\mathbf{\Lambda}_{ii}| \ , \ i\in\{0,\ldots,|\mathcal{V}|-1\}$ 
has an exponentially decaying information propagation rate, as follows:
\begin{equation}    
    \label{eq:diffusionDecay}   
    \left\Vert\frac{\partial \rm{vec}(\mathbf{X}(t))}{\partial \rm{vec}(\mathbf{X}(0))}\right\Vert = \|e^{-t\mathbf{K}}\|, \end{equation}
\end{theorem}

\begin{boxedproof}
Let us assume a diffusion-based network whose Jacobian's eigenvalues are represented by the diagonal matrix $\mathbf{\Lambda}$, and let denote the eigenvalues magnitude by a diagonal matrix $\bfK\in\mathbb{R}_+^{|\mathcal{V}|\times|\mathcal{V}|}$ 
such that $\bfK_{ii} = | \mathbf{\Lambda}_{ii} |$ for $i \in \{0,\ldots, |\mathcal{V}|-1\}$. 
Applying the vectorization operator, it is true that :
\begin{equation}\label{eq:proofdiffusionExpDecay}
   \frac{\partial \rm{vec}(\mathbf{X}(t))}{\partial\rm{vec}(\mathbf{X}(0))} = e^{t \mathbf{J}} 
\end{equation}
As it is known from \citet{EvansPDE}, diffusion-based network are characterized by Jacobian's eigenvalues with negative real part. Indeed, diffusion DE-DGNs are based on the heat equation. Therefore, the right-hand side of the ODE is the graph Laplacian, reading $\frac{\partial\bfX(t)}{\partial t} = -\bfL\bfX(t)$. Therefore, 
we analyze $-\bfL$. It is known that the graph Laplacian has non-negative eigenvalues, and therefore $-\bfL$ has non-positive eigenvalues. Thus, the Jacobian has non-positive eigenvalues. Assuming a connected graph (\ie it exists at least one edge), then there is at least one value that is strictly negative in $-\bfL$, and any entry that is not non-negative will be equal to zero. Therefore, we can write $\bfJ= - \bfK$, leading to the equation
\begin{equation}       
    \left\Vert\frac{\partial \rm{vec}(\mathbf{X}(t))}{\partial \rm{vec}(\mathbf{X}(0))}\right\Vert = \|e^{-t\bfK}\|. 
\end{equation}
Therefore, the information propagation rate among the graph nodes exponentially decays over time. Note, that at $t \rightarrow \infty$, it will converge to 1, which is exponentially lower than the propagation rate at early time $t$.
\end{boxedproof}

This result means that with diffusion methods, after sufficiently large integration time $t$, i.e., layers, they will not be able to effectively share new information between nodes, as in early layers of the network. On the contrary, our SWAN maintains the same capacity, independent of time, meaning it retains its ability to share information across nodes with the same effectiveness in each layer of the network, regardless of its depth. We illustrate this discussion in Figure~\ref{fig:propagation} and Figure~\ref{fig:multiple_behaviors}.

\subsubsection{The Benefit of Spatial Antisymmetry}
\label{sec:benefitSpatial}
While oversquashig was not mathematically defined in \cite{alon2021oversquashing}, it was recently proposed in \cite{topping2022understanding, diGiovanniOversquashing} to quantify the level, or lack of oversquashing, by measuring the sensitivity of node embedding after $\ell$ layers with respect to the input of another node 
$\left\|\partial\mathbf{x}_v(\ell)/\partial\mathbf{x}_u(0)\right\|$. Furthermore, they bounded this sensitivity score on a MPNN
as:
\begin{equation}\label{eq:mpnn_oversquashing_1}
    \left\|\frac{\partial\mathbf{x}_v(\ell)}{\partial\mathbf{x}_u(0)}\right\| \leq \underbrace{(c_\sigma w p)^\ell}_{model}\underbrace{(\mathbf{O}^\ell)_{vu}}_{topology}
\end{equation}    
where $c_\sigma$ is the Lipschitz constant of non linearity $\sigma$, $w$ is the maximal entry-value over all weight matrices, and $p$ is the embedding dimension. The term $\mathbf{O}=c_r\mathbf{I}+c_a\mathbf{A}\in\mathbb{R}^{|\mathcal{V}|\times|\mathcal{V}|}$ 
is the message passing matrix adopted by the MPNN, with $c_r$ and $c_a$ being the weighted contributions of the residual and aggregation term, respectively. Oversquashing occurs if the right-hand side of Equation~\ref{eq:mpnn_oversquashing_1} is too small.
 
\myparagraph{The sensitivity of SWAN}
In light of the discussion above, we now provide a theoretical bound of our SWAN. 
\begin{theorem}[SWAN sensitivity upper bound]\label{thm:swan_sensitivity}
    Consider SWAN (Equation \ref{eq:new_adgn_nodewise}), with $\ell$ layers, and $u,v\in\mathcal{V}$ two connected nodes of the graph. The sensitivity of $v$'s embedding after $\ell$ layers with respect to the input of node $u$ is 
    \begin{equation}\label{eq:swan_oversquashing}
    \left\|\frac{\partial\mathbf{x}_v(\ell)}{\partial\mathbf{x}_u(0)}\right\| \leq \underbrace{(c_\sigma w p)^\ell}_{model}\underbrace{((c_r \mathbf{I} + c_a \mathbf{A} + \beta c_b \mathbf{S})^\ell)_{vu}}_{topology}
    \end{equation}
     with $c_\sigma$ the Lipschitz constant of non-linearity $\sigma$, $w$ is the maximal entry-value of all weight matrices, $p$ the embedding dimension, $\mathbf{A}$ the graph shift operator, {$\mathbf{S}=(\tilde{\mathbf{{A}}}-\tilde{\mathbf{{A}}}^\top)$} the antisymmetric graph operator, and $c_r$ and $c_a$ the weighted contribution of the residual term and aggregation term.
\end{theorem}
\begin{boxedproof}
This proof follows the one proposed in \citet{diGiovanniOversquashing} (Appendix B). 
We proceed by induction and we show only the inductive step (\ie $\ell > 1$), since the case $\ell=1$ is straightforward (we refer to \citet{diGiovanniOversquashing} for more details). Assuming the Einstein summation convention and given that $\mathbf{\hat{W}} = \mathbf{W}- \mathbf{W}^\top$, $\mathbf{\hat{Z}} = \mathbf{Z}+ \mathbf{Z}^\top$, and $\alpha,\beta\in [p]$, we have:
\begin{alignat*}{3}
\left|\frac{\partial \mathbf{x}_v^\alpha(\ell+1)}{\partial \mathbf{x}_u^\gamma(0)}\right| &= \left|\sigma^\prime\Bigl(c_r \mathbf{\hat{W}}^{\ell}_{\alpha, k} \frac{\partial \mathbf{x}_v^k(\ell)}{\partial \mathbf{x}_u^\gamma(0)} + c_a \mathbf{V}^{\ell}_{\alpha, k} \mathbf{A}_{vz} \frac{\partial \mathbf{x}_z^k(\ell)}{\partial \mathbf{x}_u^\gamma(0)} + \beta c_b \mathbf{\hat{Z}}^{\ell}_{\alpha, k} \mathbf{S}_{vz} \frac{\partial \mathbf{x}_z^k(\ell)}{\partial \mathbf{x}_u^\gamma(0)}\Bigr)\right|\\
 &\leq |\sigma^\prime|\Bigl(c_r |\mathbf{\hat{W}}^{\ell}_{\alpha, k}| \left|\frac{\partial \mathbf{x}_v^k(\ell)}{\partial \mathbf{x}_u^\gamma(0)}\right| + c_a |\mathbf{V}^{\ell}_{\alpha, k}| \mathbf{A}_{vz} \left|\frac{\partial \mathbf{x}_z^k(\ell)}{\partial \mathbf{x}_u^\gamma(0)}\right|  
 + \\ &\hspace{6,8cm}+\beta c_b |\mathbf{\hat{Z}}^{\ell}_{\alpha, k}| \mathbf{S}_{vz} \left|\frac{\partial \mathbf{x}_z^k(\ell)}{\partial \mathbf{x}_u^\gamma(0)}\right|\Bigr)\\
 &\leq c_\sigma w\Bigl(c_r \left\Vert\frac{\partial \mathbf{x}_v(\ell)}{\partial \mathbf{x}_u(0)}\right\Vert + c_a \mathbf{A}_{vz} \left\Vert\frac{\partial \mathbf{x}_z(\ell)}{\partial \mathbf{x}_u(0)}\right\Vert + \beta c_b \mathbf{S}_{vz} \left\Vert\frac{\partial \mathbf{x}_z(\ell)}{\partial \mathbf{x}_u(0)}\right\Vert\Bigr)\\
    &\leq c_\sigma w (c_\sigma wp)^\ell\Bigl(c_r ((c_r \mathbf{I} + c_a \mathbf{A} + \beta c_b \mathbf{S})^{\ell})_{vu} 
    + \\ &\hspace{5,3cm}+c_a \mathbf{A}_{vz} ((c_r \mathbf{I} + c_a \mathbf{A} + \beta c_b \mathbf{S})^{\ell})_{vz} + \\ &\hspace{5,3cm}+\beta c_b \mathbf{S}_{vz}((c_r \mathbf{I} + c_a \mathbf{A} + \beta c_b \mathbf{S})^{\ell})_{vz}\Bigr)\\
    &\leq c_\sigma w (c_\sigma wp)^\ell\Bigl((c_r \mathbf{I} + c_a \mathbf{A} + \beta c_b \mathbf{S})^{\ell+1}\Bigr)_{vu}
\end{alignat*}
where $|\cdot|$ denotes an absolute value of a real number, $w$ is the maximal entry-value over all weight matrices, and $c_\sigma$, $c_r$, $c_a$ , and $c_b$ are the Lipschitz maps of the components in the computation of SWAN.
We can now sum over $\alpha$ on the left, generating an extra $p$ on the right side.
\end{boxedproof}

The result of Theorem~\ref{thm:swan_sensitivity} indicates that the added antisymmetric term $\Psi$ contributes to an increase in the measured upper bound. This result, together with the constant rate of information flow obtained from Theorem~\ref{thm:swanConstantRate}, holds the potential to theoretically mitigate oversquashing using SWAN.

\subsection{Architectural Details of SWAN}
\subsubsection{Integration of SWAN}
The ODE that defines SWAN follows the general DE-DGNs form, presented and discussed in Section~\ref{sec:ADGN}. While there are various ways to integrate these equations (see for example various integration techniques in \citet{Ascher1998}), as for A-DGN, we follow the common forward Euler discretization approach. Formally, using the forward Euler method to discretize SWAN (Equation~\ref{eq:new_adgn_graphwise}) yields the following graph neural layer:
\begin{equation}
    \bfX^{\ell+1} = \bfX^{\ell} + \epsilon \sigma \Bigl(\mathbf{X}^{\ell}{(\mathbf{W}-\mathbf{W}^\top)} +{\Phi(\mathbf{A},\mathbf{X}^{\ell},\mathbf{V})} + \beta{\Psi(\mathbf{A}, \mathbf{X}^{\ell}, \mathbf{Z})}\Bigr).
\end{equation}
Note, that in this procedure we replace the notion of \emph{time} with \emph{layers}, and therefore instead of using $t$ to denote time, we use $\ell$ to denote the step or layer number. Here, $\epsilon$ is the discretization time steps which replaces the infinitesimal $dt$ from Equation~\ref{eq:new_adgn_graphwise}.

To strengthen the stability of the Euler's numerical discretization method, similar to A-DGN, we introduce a small positive constant $\gamma > 0$, which is subtracted from the diagonal elements of the weight matrix $\mathbf{W}$, with the aim of placing back $(1+\epsilon\lambda(\mathbf{J}(t)))$ within the unit circle.

\subsubsection{Spatial Aggregation Terms}
\label{sec:swan_learn_A}
In Equation~\ref{eq:aggFuncs} we utilize two aggregation terms, denoted by $\hat{\bfA}$ and $\tilde{\bfA}$. The first, $\tilde{\bfA}$, is used to populate the \emph{weight} antisymmetric term $\Phi$, while the second, $\hat{\bfA}$, is used to populate the \emph{space} antisymmetric $\Psi$. As discussed, in our experiments we consider two possible parameterizations of these terms, on which we now elaborate.

\myparagraph{Pre-defined $\hat{\bfA}, \ \tilde{\bfA}$} In this case, we denote our architecture as SWAN and utilize the symmetric normalized adjacency matrix and the random walk normalized adjacency matrix for $\hat{\bfA}, \ \tilde{\bfA}$, respectively. Formally:
\begin{equation}
    \label{eq:fixedOperators}
    \hat{\bfA} = \bfD^{-1/2}\bfA \bfD^{-1/2}, \quad \tilde{\bfA} = \bfD^{-1}\bfA,
\end{equation}
where $\bfA$ is the standard binary adjacency matrix that is induced by the graph connectivity $\mathcal{E}$, and $\bfD$ is the degree matrix of the graph.
We observe that the implementation of $\hat{\bfA}$ and $\tilde{\bfA}$ can be treated as a hyperparameter. To show this, in our experiments, we consider both the symmetric normalized adjacency matrix and the original adjacency matrix as implementations of $\hat{\bfA}$.

\myparagraph{Learnable $\hat{\bfA}, \ \tilde{\bfA}$} Here, we denote our architecture by \emph{SWAN}-\textsc{learn}, and we use a multilayer perceptron (MLP) to learn edge-weights according to the original graph connectivity, to implement learnable $\hat{\bfA}, \ \tilde{\bfA}$. Specifically, we first define edge features as the concatenation of the initial embedding of input node features $\bfX^{0}$ of neighboring edges. Formally, the edge features of the $(u,v) \in \mathcal{E}$ edge, read: 
\begin{equation}
    \label{eq:edgeFeat}
    f_{(u,v)\in \mathcal{E}}^{in} = \bfx_u^{0} \oplus \bfx_v^{0},
\end{equation}
where $f_{(u,v)\in \mathcal{E}}^{in} \in \mathbb{R}^{2d}$ and $\oplus$ denotes the channel-wise concatenation operator.
Then, we embed those features using a 2 layer MLP:
\begin{equation}
    \label{eq:embedEdges}
    f_{(u,v)\in \mathcal{E}}^{emb} = {\rm{ReLU}}(\bfK_2 \sigma(\bfK_1(f_{((u,v)\in \mathcal{E})}^{in}))),
\end{equation}
where $\bfK_1 \in \mathbb{R}^{d \times 2d}$ and $\bfK_2 \in \mathbb{R}^{|\mathcal{V}|\times|\mathcal{V}|}$ 
are learnable linear layer weights, and $\sigma$ is an activation function which is a hyperparameter of our method. By averaging the feature dimension and gathering the averaged edge features $f_{(u,v)\in \mathcal{E}}^{emb}$ into a sparse matrix $\bfF \in \mathbb{R}^{|\mathcal{V}|\times|\mathcal{V}|}$, such that $\bfF_{u,v} = \frac{1}{d} \sum f_{(u,v)\in \mathcal{E}}^{emb}$ we define the learned spatial aggregation terms as follows:
\begin{equation}
    \label{eq:learnOperators}
    \hat{\bfA}_{\bfF} = \bfD_{\bfF}^{-1/2}\bfF \bfD_{\bfF}^{-1/2}, \quad \tilde{\bfA}_{\bfF} = \bfD_{\bfF}^{-1}\bfF,
\end{equation}
where $\bfD_{\bfF}$ is the degree matrix of $\bfF$, i.e., a matrix with the column sum of $\bfF$ on its diagonal and zeros elsewhere.

\subsubsection{SWAN versions}\label{sec:swan_variants}
In the following, we leverage the general formulation of our method and explore two main variants of SWAN, each distinguished by the implementation of the aggregation terms $\hat{\bfA}, \ \tilde{\bfA}$ in the functions $\Psi$ and $\Phi$, as shown in Equation~\ref{eq:aggFuncs}. Specifically, we consider 
\begin{enumerate}[label=(\arabic*)]
\item  SWAN which implements the aggregation terms using pre-defined operators, which are the symmetric normalized and random walk adjacency matrices, as described in Section~\ref{sec:swan_learn_A}. 
\item SWAN-\textsc{learn} which utilizes the learned aggregation terms described in Section~\ref{sec:swan_learn_A}.
\end{enumerate}
Both variants follow our parameterization from Equation~\ref{eq:aggFuncs}, and therefore are in line with the theoretical analysis provided in Section~\ref{sec:SWAN}, and in particular Theorem~\ref{thm:swanConstantRate}.
In addition to these variants, we explore other two versions of our SWAN method, depending on the implementation of $\Phi$ and $\Psi$, as reported in Table~\ref{tab:swan_versions}. All these four versions can be grouped into two main groups, which differ in the resulting non-dissipative behavior. In the first group, which contains SWAN and SWAN-\textsc{learn} both $\Phi$ and $\Psi$ lead to purely imaginary eigenvalues, since we use symmetric graph shift operators and antisymmetric weight matrices, thus allowing for a node- and graph-wise non-dissipative behavior. The last group, which includes SWAN-\textsc{ne} and SWAN-\textsc{learn-ne}, can deviate from being globally non-dissipative. Indeed, if the weight matrix, $\bfV$, is an arbitrary matrix, then the eigenvalues of the Jacobian matrix of the system, $\mathbf{J}(t)$, are contained in a neighborhood of the imaginary axis with radius $r \leq ||\mathbf{V}||$ (Bauer-Fike’s theorem~\cite{bauer1960norms}). Although this result does not guarantee that the eigenvalues of the Jacobian are imaginary, in practice it crucially limits their position, limiting the dynamics of the system on the graph to show at most moderate amplification or loss of signals through the graph.

\begin{table}[h]
\centering
\caption{The grid of the evaluated SWAN versions. We consider $\hat{\bfA}$ to be either the original adjacency matrix $\bfA$ or the symmetric normalized adjacency matrix, $\bfD^{-1/2}\bfA \bfD^{-1/2}$, while $\tilde{\bfA}$ is the random walk normalized adjacency matrix, $\bfD^{-1}\bfA$. The learned versions (-\textsc{learn}) employ learnable $\hat{\bfA}$ and $\tilde{\bfA}$ as described in Section~\ref{sec:swan_learn_A}, here referred as $\hat{\bfA}_\bfF$ and $\tilde{\bfA}_\bfF$.\label{tab:swan_versions}}

\scriptsize
\begin{tabular}{lll}
\toprule
\textbf{Name} & $\Phi$ & $\Psi$\\\midrule

\textbf{Weight Antisymmetry Only} \\
$\,$ SWAN$_{\beta=0}$ & $(\hat{\bfA}+\hat{\bfA}^\top)\mathbf{X}(\mathbf{V}-\mathbf{V}^\top)$ & -- \\
\midrule
\textbf{Bounded Non-Dissipative}  \\
$\,$ SWAN-\textsc{ne} & $\hat{\bfA}\bfX\bfV$ & $(\tilde{\bfA}-\tilde{\bfA}^\top)\mathbf{X}(\mathbf{Z}+\mathbf{Z}^\top)$ \\
$\,$ SWAN-\textsc{learn-ne} & $\hat{\bfA}_\bfF\mathbf{X}\mathbf{V}$ & $(\mathbf{\tilde{A}}_\bfF-\mathbf{\tilde{A}}_\bfF^\top)\mathbf{X}(\mathbf{Z}+\mathbf{Z}^\top)$ \\

\midrule
\textbf{Global and Local Non-Dissipative}  \\ 

$\,$ SWAN & $(\hat{\bfA}+\hat{\bfA}^\top)\mathbf{X}(\mathbf{V}-\mathbf{V}^\top)$ & $(\tilde{\bfA}-\tilde{\bfA}^\top)\mathbf{X}(\mathbf{Z}+\mathbf{Z}^\top)$ \\
$\,$ SWAN-\textsc{learn} & $(\hat{\bfA}_\bfF+\hat{\bfA}_\bfF^\top)\mathbf{X}(\mathbf{V}-\mathbf{V}^\top)$ & $(\mathbf{\tilde{A}}_\bfF-\mathbf{\tilde{A}}_\bfF^\top)\mathbf{X}(\mathbf{Z}+\mathbf{Z}^\top)$ \\

\bottomrule
\end{tabular}
\end{table}

\subsubsection{Applicability of SWAN to general MPNNs}\label{sec:applicability_of_swan}
Nowadays, most DGNs rely on the concepts introduced by the MPNN as introduced in Section~\ref{sec:static_dgn_fundamentals}, which is a general framework based on the message passing paradigm. 
A general MPNN updates the representation for a node $u$ by using message and update functions. The first (message function) is responsible for defining the messages between nodes and their neighbors. On the other hand, the update function has the role of collecting (aggregating) messages and updating the node representation. As shown for A-DGN in Section~\ref{sec:ADGN_mpnn_comparison}, our SWAN, in Equation~\ref{eq:new_adgn_nodewise}, operates according to the MPNN paradigm, with the functions $\Phi$ and $\Psi$ that operate as the message function, while the sum operator among the node and neighborhood representations is the update function. Therefore, SWAN can be interpreted as a special case of an MPNN with the aformentioned parameterization, and can potentially be applied to different types of MPNNs.

\subsection{Experiments}

In this section, we empirically evaluate our SWAN and compare it with literature methods.
Specifically, we seek to address the following questions:
\begin{enumerate}[label=(\roman*)]
    \item Can SWAN effectively propagate information to distant nodes?
    \item Can SWAN accurately predict graph properties that are related to long-range interactions?
    \item How does SWAN perform on real-world long-range benchmarks?
\end{enumerate}

In our experiments, the performance of SWAN is compared with various DGN baselines, namely: 
(i) MPNNs with linear complexity with respect to the size of nodes and edges, similarly to the complexity of our SWAN, \ie GCN~\citep{GCN}, GraphSAGE~\citep{SAGE}, GAT~\citep{GAT}, GatedGCN~\citep{gatedgcn}, GIN~\citep{GIN}, GINE~\citep{gine}, and GCNII~\citep{gcnii}.
(ii) DE-DGNs, as a direct comparison to the family of DGNs considered in this chapter, such as DGC~\citep{DGC}, GRAND~\citep{GRAND}, GraphCON~\citep{graphcon}, and A-DGN (Section~\ref{sec:ADGN}).
(iii) Oversquashing designated techniques such as Graph Transformers and Higher-Order DGNs, \ie Transformer~\citep{vaswani2017attention, dwivedi2021generalization}, DIGL~\citep{DIGL}, MixHop~\citep{abu2019mixhop}, SAN~\citep{san}, GraphGPS \citep{graphgps}, and DRew~\citep{drew}.

In the following experiments, we evaluate the versions of SWAN discussed in Section~\ref{sec:swan_variants}.
We refer the reader to Appendix~\ref{app:swan_data} for more details about the employed
datasets. We report in Table~\ref{tab:hyperparams} (Appendix~\ref{app:swan_hyperparams}) the grid of hyperparameters employed in our experiments. 

\subsubsection{Graph Transfer}\label{sec:exp_transfer}
\myparagraph{Setup}
We consider three graph transfer tasks, whose goal is to transfer a label from a source node to a target node, placed at distance $k$ in the graph. Due to oversquashing and dissipative behavior of existing DGN methods that are based on diffusion, the performance is expected to degrade as $k$ increases. We initialize nodes with a random valued feature 
and we assign values ``1'' and ``0'' to source and target nodes, respectively. We consider three graph distributions, \ie line, ring, crossed-ring, as illustrated in Figure~\ref{fig:graph_distrib}. We consider four different distances $k=\{3,5,10,50\}$. By increasing $k$, we increase the complexity of the task and require long-range information. 

We design each model as a combination of three main components. The first is the encoder which maps the node input features into a latent hidden space; the second is the graph convolution (\ie SWAN or the other baselines); and the third is a readout that maps the output of the convolution into the output space. The encoder and the readout share the same architecture among all models in the experiments. SWAN and A-DGN are implemented using \textbf{weight sharing}.

We perform hyperparameter tuning via grid search, optimizing the Mean Squared Error (MSE) computed on the node features of the whole graph. We train the models using Adam optimizer for a maximum of 2000 epochs and early stopping with patience of 100 epochs on the validation loss. For each model configuration, we perform 4 training runs with different weight initialization and report the average of the results.

\begin{figure}[h]
    \begin{adjustbox}{center}
        \begin{subfigure}{0.45\textwidth}
            \centering \includegraphics[width=\linewidth]{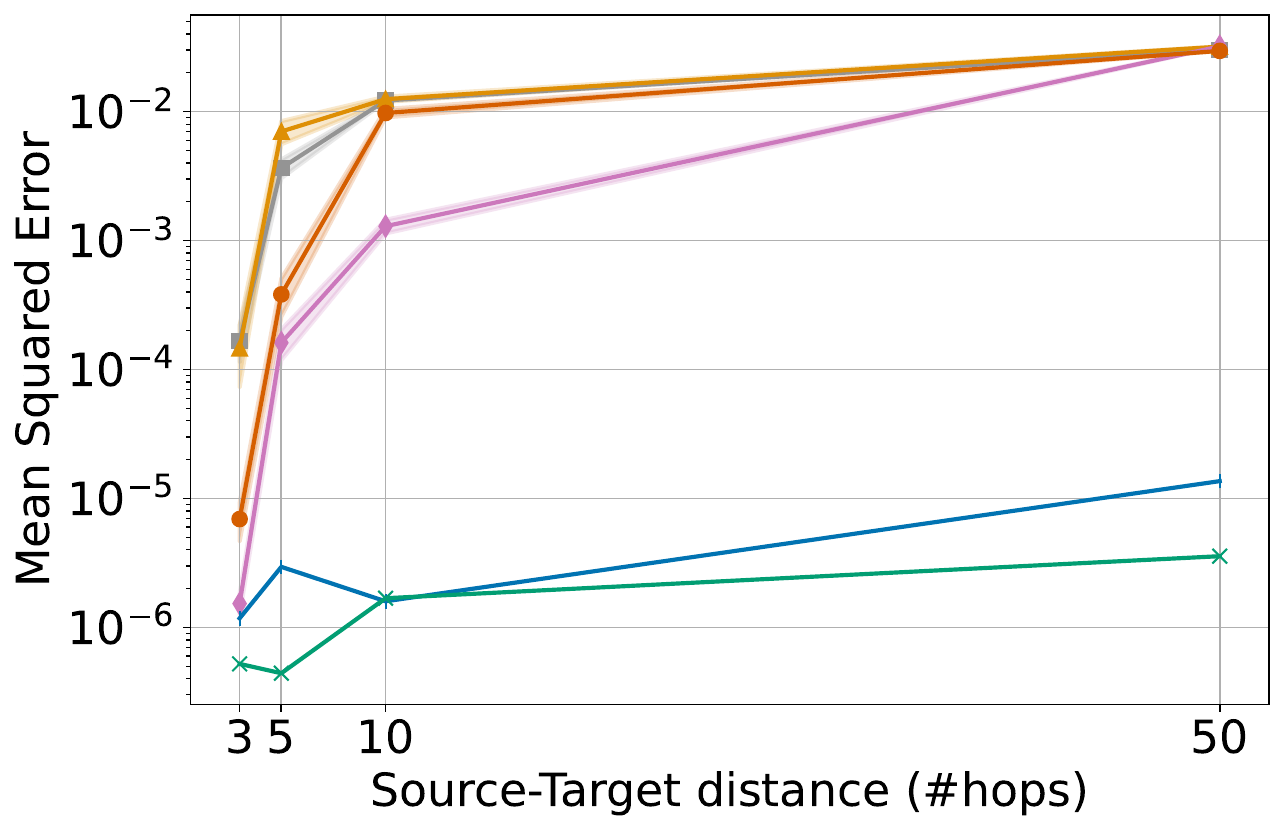}
            \caption{\small Line}
        \end{subfigure}\hspace{5mm}
        \begin{subfigure}{0.45\textwidth}
            \centering \includegraphics[width=\linewidth]{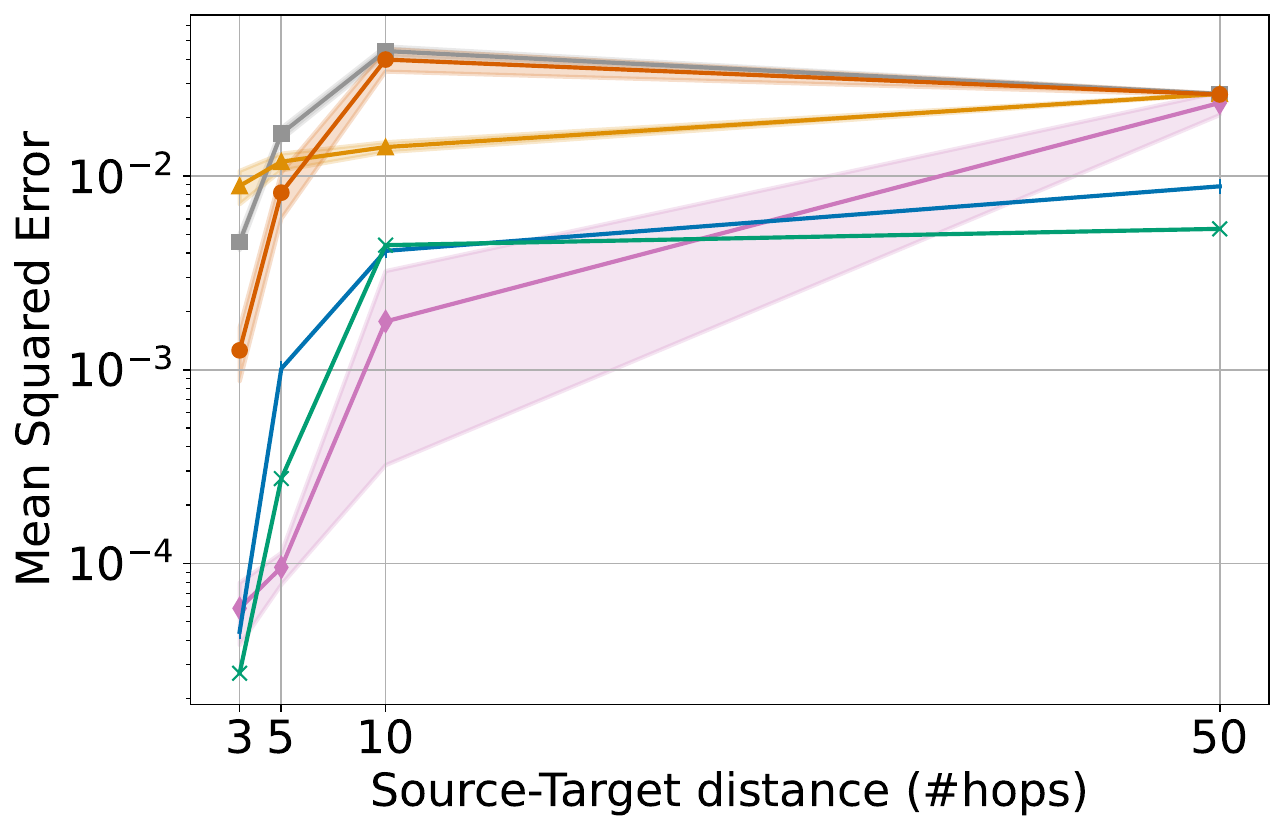}
            \caption{\small Ring}
        \end{subfigure}
    \end{adjustbox}\vspace{2mm}
    \begin{adjustbox}{center}
        \begin{subfigure}{\textwidth}
            \centering \includegraphics[width=0.45\linewidth]{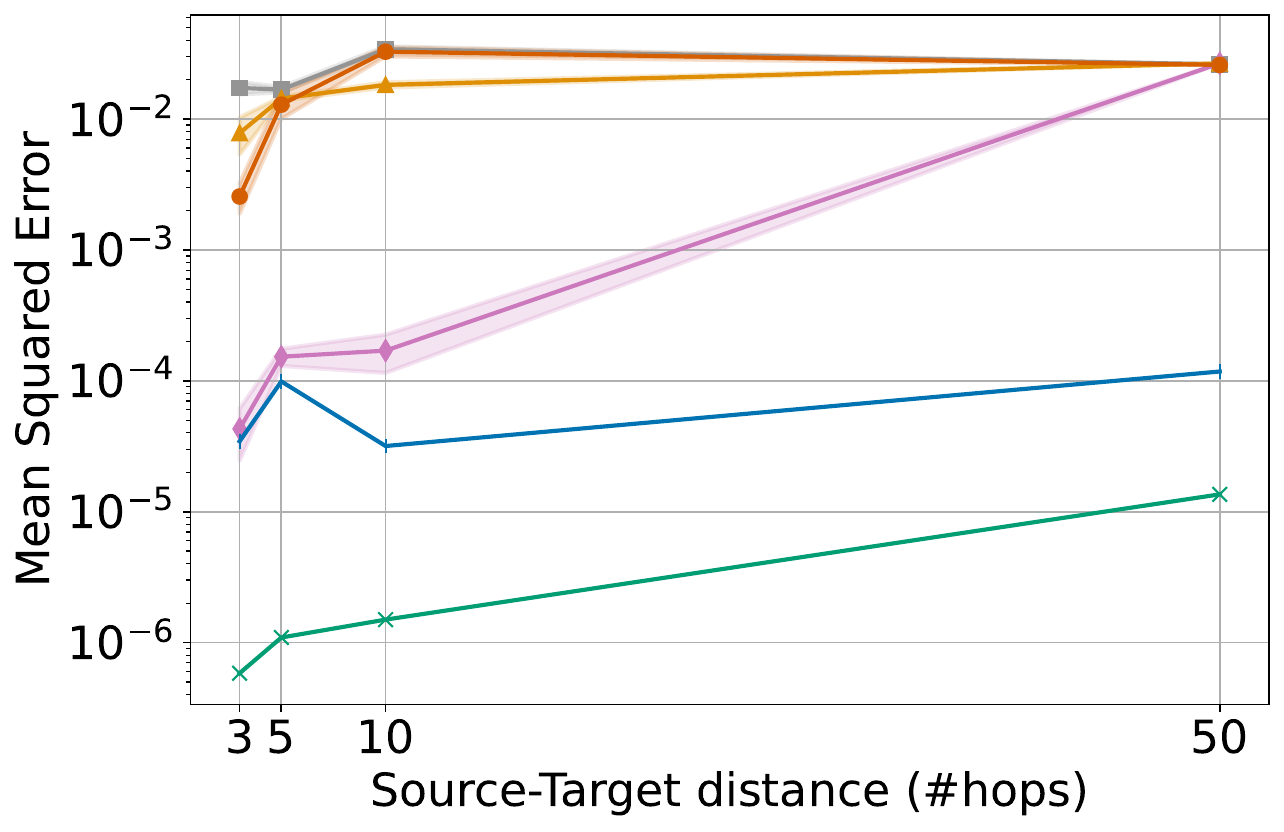}
            \caption{\small Crossed-Ring }
        \end{subfigure}
    \end{adjustbox}\vspace{2mm}
    \begin{subfigure}{1\textwidth}
        \begin{adjustbox}{center}
            \centering \includegraphics[width=1.3\linewidth]{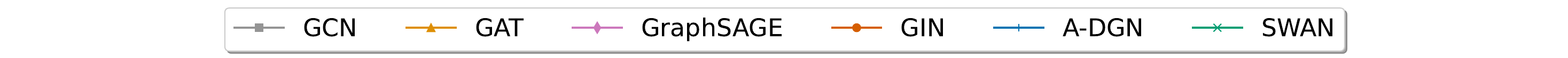}
        \end{adjustbox}
    \end{subfigure}
\caption{Information transfer on (a) Line, (b) Ring, and (c) Crossed-Ring graphs. While 
baseline approaches struggle to transfer the information accurately, non-dissipative methods like A-DGN and our SWAN achieve low errors.}
\label{fig:graph_transfer_res}
\end{figure}

\myparagraph{Results}
Figure~\ref{fig:graph_transfer_res} reports the results of the transfer graph tasks. The ring-graph proved to be the hardest, leading to higher errors when compared with other graphs. Overall, baselines struggle to accurately transfer the information through the graph, especially when the distance is high, \ie $\#hops\geq10$. Differently, non-dissipative methods, such as A-DGN and SWAN, achieve low errors across all distances. Moreover, SWAN consistently outperforms A-DGN, empirically supporting our theoretical findings that SWAN can better propagate information among distant nodes.


\subsubsection{Graph Property Prediction}\label{sec:exp_graph_prop_pred}
\myparagraph{Setup}
We address tasks involving the prediction of three graph properties -  Diameter, Single-Source Shortest Paths (SSSP), and node Eccentricity on synthetic graphs as introduced in Section~\ref{sec:adgn_graph_prop_pred}. 

\myparagraph{Results}
In Table~\ref{tab:results_GraphProp_complete} and \ref{tab:results_GraphProp_complete_ldw}  we compare SWAN and SWAN-\textsc{learn} with other MPNNs and DE-DGNs. Our results indicate that SWAN improves performance across all baselines. Specifically, SWAN-\textsc{learn} yields the lowest (best) results across all models, showing an improvement of up to 117\% with respect to the runner up model, in the weight sharing scenario. Such improvements are further increased when layer-dependent weights are employed.
We note that similarly to the transfer task in Section~\ref{sec:exp_transfer}, solving  Graph Property Prediction tasks necessitates capturing long-term dependencies. Hence, successful prediction  requires the mitigation of oversquashing. For instance, in the eccentricity task, the goal is to calculate the maximal shortest path between a node $u$ and all other nodes. Thus, it is crucial to propagate information not only from the direct neighborhood of $u$ but also from nodes that are considerably distant. This requirement holds true for diameter and SSSP tasks as well.

\begin{table}[h]
\centering
\caption{Mean test set {\small$log_{10}(\mathrm{MSE})$} and std averaged over 4 random weight initializations on the Graph Property Prediction tasks. SWAN and DE-DGN baselins employ \textbf{weight sharing}. The lower, the better. 
\one{First}, \three{third} and \four{fourth} best results for each task are color-coded.
\label{tab:results_GraphProp_complete}}

\scriptsize
\begin{tabular}{lccc}
\toprule
\textbf{Model} &\textbf{Diameter} & \textbf{SSSP} & \textbf{Eccentricity} \\\midrule
\textbf{MPNNs} \\
$\,$ GCN            & 0.7424$_{\pm0.0466}$ & 0.9499$_{\pm9.18\cdot10^{-5}}$ & 0.8468$_{\pm0.0028}$ \\
$\,$ GAT            & 0.8221$_{\pm0.0752}$ & 0.6951$_{\pm0.1499}$           & 0.7909$_{\pm0.0222}$  \\
$\,$ GraphSAGE      & 0.8645$_{\pm0.0401}$ & 0.2863$_{\pm0.1843}$           &  0.7863$_{\pm0.0207}$\\
$\,$ GIN            & 0.6131$_{\pm0.0990}$ & -0.5408$_{\pm0.4193}$          & 0.9504$_{\pm0.0007}$\\
$\,$  GCNII          & 0.5287$_{\pm0.0570}$ & -1.1329$_{\pm0.0135}$          & 0.7640$_{\pm0.0355}$\\
\midrule
\textbf{DE-DGNs} \\
$\,$ DGC            & 0.6028$_{\pm0.0050}$ & -0.1483$_{\pm0.0231}$          & 0.8261$_{\pm0.0032}$\\
$\,$ GRAND          & 0.6715$_{\pm0.0490}$ & -0.0942$_{\pm0.3897}$          & \four{0.6602$_{\pm0.1393}$} \\
$\,$ GraphCON       & \four{0.0964$_{\pm0.0620}$} & \four{-1.3836$_{\pm0.0092}$} & 0.6833$_{\pm0.0074}$\\

\midrule
\textbf{Ours} \\
$\,$ A-DGN & \three{-0.5188$_{\pm0.1812}$} & \two{-3.2417$_{\pm0.0751}$} & \three{0.4296$_{\pm0.1003}$}  \\
$\,$ SWAN & \two{-0.5249$_{\pm0.0155}$} &  \three{-3.2370$_{\pm0.0834}$} & \two{0.4094$_{\pm0.0764}$} \\

$\,$ SWAN-\textsc{learn} & \one{-0.5981$_{\pm0.1145}$}  & \one{-3.5425$_{\pm0.0830}$}  & \one{-0.0739$_{\pm0.2190}$} \\
\bottomrule 
\end{tabular}
\end{table}

\begin{table}[h]
\centering
\caption{Mean test set {\small$log_{10}(\mathrm{MSE})$} and std averaged over 4 random weight initializations on the Graph Property Prediction tasks. SWAN and A-DGN emply \textbf{layer-dependent weights}. The lower, the better. 
\one{First} and \two{second} best results for each task are color-coded.
\label{tab:results_GraphProp_complete_ldw}}

\scriptsize
\begin{tabular}{lccc}
\toprule
\textbf{Model} &\textbf{Diameter} & \textbf{SSSP} & \textbf{Eccentricity} \\\midrule
A-DGN &-0.5455$_{\pm0.0328}$ & -3.4020$_{\pm0.1372}$ & 0.3046$_{\pm0.1181}$ \\
SWAN & \one{-0.6381$_{\pm0.0358}$} &  \one{-3.9342$_{\pm0.1993}$} & \one{-0.2706$_{\pm0.0948}$} \\
SWAN-\textsc{learn} & \two{-0.5905$_{\pm0.0372}$}  & \two{-3.8258$_{\pm0.0950}$}  & \two{-0.2245$_{\pm0.0840}$} \\
\bottomrule 
\end{tabular}
\end{table}

\subsubsection{Long-Range Graph Benchmark}\label{sec:exp_lrb}
\myparagraph{Setup} 
We employ the same datasets and experimental setting presented in \citet{LRGB}. Therefore, we perform hyperparameter tuning via grid search, optimizing the Average Precision (AP) in the Peptide-func task, the Mean Absolute Error (MAE) in the Peptide-struct task, and F1 in Pascal-VOC, training the models using AdamW optimizer for a maximum of 300 epochs. For each model configuration, we perform 3 training runs with different weight initialization and report the average of the results. Also, we follow the guidelines in \cite{LRGB, drew} and stay within the 500K parameter budget. SWAN and other DE-DGN baselines are implemented using \textbf{weight sharing}.

\myparagraph{Results}
In Table \ref{tab:lrgb_results} we are interested in directly comparing the performance of SWAN and its non-dissipative properties with other MPNNs and DE-DGNs, as well as the popular approach of using graph transformer to address long-range interaction modeling. We provide additional comparisons with other methods that utilize multi-hop information, and are therefore more computationally expensive than our SWAN, while also utilizing additional features such as the Laplacian positional encoding. In our evaluation of SWAN, we chose not to use additional feature enhancements, in order to provide a clear exposition of the contribution and importance of the local and global non-dissipativity offered by SWAN.  

In the Peptide-struct task, SWAN demonstrates superior performance compared to all the compared methods. On the Peptide-func task,  SWAN outperforms MPNNs, Graph Transformers, DE-DGNs, and some multi-hop DGNs, while being second to multi-hop methods such as DRew. The PascalVOC-SP results show the effectiveness of SWAN with other DE-DGNs (GRAND, GraphCON, A-DGN), which are the focus of this chapter, while offering competitive results to multi-hop and transformer methods. However, it is essential to note that multi-hop DGNs incur higher complexity, while SWAN maintains a linear complexity. Therefore, we conclude that SWAN offers a highly effective approach for tasks that require long-range interactions, as in the LRGB benchmark.

To summarize, our results in Table \ref{tab:lrgb_results} suggest the following:
(i) SWAN achieves significantly better results than standard MPNNs such as GCN, GINE, or GCNII (\eg  SWAN-\textsc{learn} improves GCN's performance by 7.2\% Peptides-func).
(ii) Compared with Transformers, which are of complexity ${\mathcal{O}}(|\mathcal{V}|^2)$, our SWAN achieves better performance while remaining with a linear complexity of $\mathcal{O}(|\mathcal{V}|+|\mathcal{E}|)$.
(iii) Among its class of DE-DGNs, our SWAN offers overall better performance.

\begin{table}[h!]
\centering
\caption{Performance of various classical, multi-hop and static rewiring MPNN, graph Transformer benchmarks, DE-GNNs, and our SWAN across two LRGB tasks. Results are averaged over 3 weight initializations. The \one{first}, \two{second}, and \three{third} best results for each task are color-coded. Beseline results are reported from \cite{drew}.\label{tab:lrgb_results}
}
\scriptsize
\begin{tabular}{@{}lccc@{}}
\toprule
\multirow{3}{*}{\textbf{Model}} & \textbf{Peptides-}  & \textbf{Peptides-}  & \textbf{Pascal}              \\
& \textbf{func} & \textbf{struct} & \textbf{voc-sp}
                               \\
                                & \scriptsize{AP $\uparrow$}                             & \scriptsize{MAE $\downarrow$} & \scriptsize{F1 $\uparrow$}                             \\ \midrule  
\textbf{MPNNs} \\
$\,$ GCN           & 0.5930$_{\pm0.0023}$ & 0.3496$_{\pm0.0013}$ & 0.1268$_{\pm0.0060}$\\
$\,$ GINE          & 0.5498$_{\pm0.0079}$ & 0.3547$_{\pm0.0045}$ & 0.1265$_{\pm0.0076}$\\
$\,$ GCNII         & 0.5543$_{\pm0.0078}$ & 0.3471$_{\pm0.0010}$ & 0.1698$_{\pm0.0080}$\\
$\,$ GatedGCN      & 0.5864$_{\pm0.0077}$ & 0.3420$_{\pm0.0013}$ & 0.2873$_{\pm0.0219}$\\
$\,$ GatedGCN+PE   & 0.6069$_{\pm0.0035}$ & 0.3357$_{\pm0.0006}$ & 0.2860$_{\pm0.0085}$\\ 
\midrule
\textbf{Multi-hop GNNs}\\
$\,$ DIGL+MPNN           & 0.6469$_{\pm0.0019}$         & 0.3173$_{\pm0.0007}$ & 0.2824$_{\pm0.0039}$\\
$\,$ DIGL+MPNN+LapPE     & 0.6830$_{\pm0.0026}$         & 0.2616$_{\pm0.0018}$ & 0.2921$_{\pm0.0038}$\\
$\,$ MixHop-GCN          & 0.6592$_{\pm0.0036}$         & 0.2921$_{\pm0.0023}$ & 0.2506$_{\pm0.0133}$\\
$\,$ MixHop-GCN+LapPE    & 0.6843$_{\pm0.0049}$         & 0.2614$_{\pm0.0023}$ & 0.2218$_{\pm0.0174}$\\ 
$\,$ DRew-GCN            & \three{0.6996$_{\pm0.0076}$} & 0.2781$_{\pm0.0028}$ & 0.1848$_{\pm0.0107}$\\
$\,$ DRew-GCN+LapPE      & \one{0.7150$_{\pm0.0044}$}   & 0.2536$_{\pm0.0015}$ & 0.1851$_{\pm0.0092}$\\
$\,$ DRew-GIN            & 0.6940$_{\pm0.0074}$         & 0.2799$_{\pm0.0016}$ & 0.2719$_{\pm0.0043}$\\
$\,$ DRew-GIN+LapPE      & \two{0.7126$_{\pm0.0045}$}   & 0.2606$_{\pm0.0014}$ & 0.2692$_{\pm0.0059}$\\
$\,$ DRew-GatedGCN       & 0.6733$_{\pm0.0094}$         & 0.2699$_{\pm0.0018}$ & 0.3214$_{\pm0.0021}$\\
$\,$ DRew-GatedGCN+LapPE & 0.6977$_{\pm0.0026}$         & 0.2539$_{\pm0.0007}$ & \two{0.3314$_{\pm0.0024}$}\\


\midrule
\textbf{Transformers} \\
$\,$ Transformer+LapPE & 0.6326$_{\pm0.0126}$ & \three{0.2529$_{\pm0.0016}$} & 0.2694$_{\pm0.0098}$\\
$\,$ SAN+LapPE         & 0.6384$_{\pm0.0121}$ & 0.2683$_{\pm0.0043}$         & \three{0.3230$_{\pm0.0039}$}\\
$\,$ GraphGPS+LapPE    & 0.6535$_{\pm0.0041}$ & \two{0.2500$_{\pm0.0005}$}   & \one{0.3748$_{\pm0.0109}$}\\ 
\midrule
\textbf{DE-GNNs} \\
$\,$ GRAND     & 0.5789$_{\pm0.0062}$ & 0.3418$_{\pm0.0015}$ &  0.1918$_{\pm0.0097}$ \\
$\,$  GraphCON & 0.6022$_{\pm0.0068}$ & 0.2778$_{\pm0.0018}$ &  0.2108$_{\pm0.0091}$ \\
\midrule
\textbf{Ours} \\    
$\,$ A-DGN      & 0.5975$_{\pm0.0044}$ & 0.2874$_{\pm0.0021}$ &  0.2349$_{\pm0.0054}$\\ 
$\,$ SWAN                & 0.6313$_{\pm0.0046}$ & 0.2571$_{\pm0.0018}$       & 0.2796$_{\pm0.0048}$\\
$\,$ SWAN-\textsc{learn} & 0.6751$_{\pm0.0039}$ & \one{0.2485$_{\pm0.0009}$} & 0.3192$_{\pm0.0250}$\\
\bottomrule
\end{tabular}
\end{table}

\subsubsection{Ablation Study}\label{sec:ablation}
In this section, we study the different components of SWAN, namely, the contribution of guaranteed global (\ie graph-wise) and local (\ie node-wise) non-dissipativity, and the benefit the spatial antisymmetry.

\begin{table}[h]
\centering
\caption{Performance of different versions of SWAN on Graph Property Prediction and LRGB tasks. Results are averaged over 4 random weight initializations on the Graph Property Prediction, while over 3 on the LRGB. The \one{first}, \two{second}, and \three{third} best results for each task are color-coded.\label{tab:importance}}

\scriptsize
\begin{tabular}{lccccc}
\toprule

\multirow{3}{*}{\textbf{Model}} &\multirow{2}{*}{\textbf{Diam.}} & \multirow{2}{*}{\textbf{SSSP}} & \multirow{2}{*}{\textbf{Ecc.}} & \textbf{Peptides-}  & \textbf{Peptides-}                \\
& & & & \textbf{func} & \textbf{struct} 
                               \\
& \scriptsize{$log_{10}$(MSE)$\downarrow$} & \scriptsize{$log_{10}$(MSE)$\downarrow$} & \scriptsize{$log_{10}$(MSE)$\downarrow$} & \scriptsize{AP $\uparrow$} & \scriptsize{MAE $\downarrow$}      \\\midrule
\multicolumn{4}{l}{\textbf{Weight Antisymmetry Only}} \\
$\,$ SWAN$_{\beta=0}$       & -0.3882$_{\pm0.0610}$         & -3.2061$_{\pm0.0416}$         & 0.5573$_{\pm0.0247}$         & 0.6195$_{\pm0.0067}$         & 0.2703$_{\pm0.0023}$\\\midrule
\multicolumn{4}{l}{\textbf{Bounded Non-Dissipative}}  \\  
$\,$ SWAN-\textsc{ne}       & \three{-0.5497$_{\pm0.0766}$} & -3.1913$_{\pm0.0762}$         & \three{0.3792$_{\pm0.1514}$} & 0.6119$_{\pm0.0037}$         & 0.2672$_{\pm0.0012}$ \\
$\,$ SWAN-\textsc{ne-learn} & \two{-0.5631$_{\pm0.0694}$}   & \two{-3.5296$_{\pm0.0831}$}   & \two{0.1317$_{\pm0.1253}$}   & \three{0.6249$_{\pm0.0051}$} & \three{0.2606$_{\pm0.0007}$}\\
\midrule
\multicolumn{4}{l}{\textbf{Global and Local Non-Dissipative}}  \\ 
$\,$ SWAN                   & -0.5249$_{\pm0.0155}$         & \three{-3.2370$_{\pm0.0834}$} & 0.4094$_{\pm0.0764}$         & \two{0.6313$_{\pm0.0046}$}   & \two{0.2571$_{\pm0.0018}$} \\
$\,$ SWAN-\textsc{learn}    & \one{-0.5981$_{\pm0.1145}$}   & \one{-3.5425$_{\pm0.0830}$}   & \one{-0.0739$_{\pm0.219}$}   & \one{0.6751$_{\pm0.0039}$}   & 
\one{0.2485$_{\pm0.0009}$}\\

\bottomrule      
\end{tabular}
\end{table}

\myparagraph{The importance of global and local non-dissipativity}
To verify the contribution of the global and local non-dissipativity offered by SWAN, we also evaluate the performance of the following variants, which can deviate from being globally and locally non-dissipative, although in a bounded manner, as we discuss in Section~\ref{sec:swan_variants}. Specifically, we consider the non-enforced (NE) variants of SWAN and SWAN-\textsc{learn}, which differ from the original version by the use of unconstrained weight matrix $\mathbf{V}$, rather than forcing it to be antisymmetric. These two additional variants are called SWAN-\textsc{ne} and SWAN-\textsc{learn-ne}, respectively. 

Table~\ref{tab:importance} presents the performance of several SWAN variants on the Graph Property Prediction and LRGB tasks. The best performance on both synthetic and real-world problems is achieved through a global and local non-dissipative behavior by SWAN-\textsc{learn}, showcasing the importance of global and local non-dissipativity.  

\myparagraph{The benefit of spatial antisymmetry} 
Table~\ref{tab:importance} highlights the advantages of the spatial antisymmetry in Equation~\ref{eq:new_adgn_nodewise}.  By setting $\beta=0$, we obtain a model with antisymmetry solely in the weight space, exhibiting only a local non-dissipative behavior. Our results showcase a noteworthy performance improvement when spatial antisymmetry is employed. This is further supported by the improved performance of -\textsc{ne} versions of SWAN, which do not guarantee both global and local non-dissipative behavior, compared to SWAN$_{\beta=0}$. Thus, spatial antisymmetry emerges as a beneficial factor in effective information propagation.

\myparagraph{Complexity Analysis}

Our SWAN architecture remains within the computational complexity of MPNNs (e.g., \citet{morris2019weisfeiler, GIN}) and other DE-GNNs such as GRAND. Specifically, each SWAN layer is linear in the number of nodes $|\mathcal{V}|$ and edges $|\mathcal{E}|$, therefore it has a time complexity of $\mathcal{O}(|\mathcal{V}|+|\mathcal{E}|)$. Assuming we compute $L$ steps of the ODE defined by SWAN, the overall complexity is $\mathcal{O} \left(L\cdot(|\mathcal{V}|+|\mathcal{E}|) \right)$.

\myparagraph{Runtimes} We measure the training and inference (i.e., the epoch-time and test-set time) runtime of SWAN on the Peptides-struct dataset and compare it with other baselines. Our runtimes show that SWAN obtains high performance, while retaining a linear complexity 
aligned to other MPNNs and DE-GNNs. Our measurements 
in Table \ref{tab:runtimes} present the training and inference runtimes in seconds. The runtimes were measured on an NVIDIA RTX-3090 GPU with 24GB of memory. In all measurements, we use a batch size of 64, 128 feature channels, and 5 layers. For reference, we also provide the reported downstream task performance of each method.

\begin{table}[h]
    \centering
    \caption{Measured training and inference runtimes in seconds, and the obtain MAE of SWAN and other baselines on the Peptides-func dataset.}
    \scriptsize
    \begin{tabular}{lcccc}
    \toprule
         \textbf{Method} & \textbf{Training} & \textbf{Inference} & \textbf{MAE} {\scriptsize$\downarrow$}  \\
         \midrule
         GCN &  2.90 & 0.32 &  0.3496$_{\pm0.0013}$  \\
         \midrule
         GraphGPS+LapPE & 23.04 & 2.39 &   0.2500$_{\pm0.0005}$ \\
         \midrule
         GraphCON & 3.03 & 0.27 & 0.2778$_{\pm0.0018}$\\
         A-DGN & 2.83 & 0.25 & 0.2874$_{\pm0.0021}$\\
         \midrule
         SWAN & 2.88 & 0.24 & 0.2571$_{\pm0.0018}$\\
         SWAN-\textsc{learn} & 2.93 & 0.26  & 0.2485$_{\pm0.0009}$  \\
         \bottomrule
    \end{tabular}
    \label{tab:runtimes}
\end{table}

\section{Related Work}\label{sec:rel_work_adgn}
As introduced in Section~\ref{sec:static_dgn_fundamentals}, most of the DGNs typically relies on the concepts introduced by the Message Passing Neural Network (MPNN)~\citep{MPNN}, which is a general framework based on the message passing paradigm. As observed in Sections~\ref{sec:ADGN_mpnn_comparison} and \ref{sec:applicability_of_swan}, by relaxing the concepts of stability and local and global non-dissipation from our frameworks, MPNN becomes a specific discretization instance of A-DGN and SWAN.

Depending on the definition of the update and message functions, it is possible to derive a variety of DGNs that mainly differ on the neighbor aggregation scheme \citep{GCN,GAT,SAGE,GIN, chebnet, gine}. However, all these methods focus on presenting new and effective functions without questioning the stability and non-dissipative behavior of the final network. As a result, most of these DGNs are usually not able to capture long-term interactions. Indeed, only few layers can be employed without falling into the oversquashing phenomenon, as it is discussed in Section~\ref{sec:dgn_plights}.  

Since the previous methods are all specific cases of MPNN, they are all instances of the discretized and unconstrained version of A-DGN and SWAN. Moreover, a proper design of the functions $\Phi$ and $\Psi$ in A-DGN and SWAN allows rethinking the discussed DGNs through the lens of non-dissipative and stable ODEs. 
\cite{gcnii}, \cite{egnn}, and \cite{pathGCN} proposed three methods to alleviate oversmoothing (see Section~\ref{sec:dgn_plights}). Similarly to the forward Euler discretization, the first method employs identity mapping. It also exploits initial residual connections to ensure that the final representation of each node retains at least a fraction of input. The second method proposes a DGN that constrains the Dirichlet energy at each layer and leverages initial residual connections, while the latter tackles oversmoothing by aggregating random paths over the graph nodes. Thus, the novelty of our methods is still preserved since both A-DGN and SWAN define a map between DGNs and stable and non-dissipative graph ODEs to preserve long-range dependencies between nodes.


Inspired by the NeuralODE approach~\citep{neuralODE}, \citet{GDE} develops a DGN defined as a continuum of layers. In such a work, the authors focus on building the connection between ODEs and DGNs. We extend their work to include stability and non-dissipation, which are fundamental properties to preserve long-term dependencies between nodes and prevent gradient explosion or vanishing during training. Thus, by relaxing these two properties from our frameworks, the work by \citet{GDE} becomes a specific instance of A-DGN and SWAN. \citet{GRAND} propose GRAND an architecture to learn graph diffusion as a PDE. Differently from GRAND, our frameworks design an architecture that is theoretically non-dissipative and free from gradient vanishing or explosion. 
DGC~\citep{DGC} and SGC~\citep{SGC} propose linear models that propagate node information as the discretization of the graph heat equation, $d \mathbf{X}(t)/d t = -\mathbf{L}\mathbf{X}(t)$, without learning. Specifically, DGC mainly focus on exploring the influence of the step size $\epsilon$ in the Euler discretization method.
\citet{pdegcn} and \cite{graphcon} present two methods to preserve the energy of the system, \ie they mitigate over-smoothing, instead of preserving long-range information between nodes. Differently from our methods, which employs a first-order ODE, the former leverages the conservative mapping defined by hyperbolic PDEs, while the latter is defined as second-order ODEs that preserve the Dirichlet energy. In general, this testifies that  non-dissipation in graph ODEs is an important property to pursue, not only when preserving long-range dependencies. However, to the best of our knowledge, we are the first to propose a non-dissipative graph ODEs to effectively propagate the information on the graph structure.

The graph representation learning community explored various strategies in recent years to effectively transfer information across distant nodes, such as graph rewiring and Transformer-based architectures. In the first setting, methods like SDRF~\citep{topping2022understanding}, GRAND~\citep{GRAND}, BLEND~\citep{blend}, and DRew~\citep{drew} (dynamically) alter the original edge set to densify the graph during preprocessing to facilitate node communication.  Differently, Transformer-based methods~\citep{graphtransformer, dwivedi2021generalization, ying2021transformers,wu2023difformer} enable message passing between all node pairs. FLODE~\citep{maskey2024fractional} incorporates non-local dynamics by using a fractional power of the graph shift operator. Although these techniques are effective in addressing the problem of long-range communication, they can also increase the complexity of information propagation due to denser graph shift operators, whereas our approaches avoid this issue.
\section{Summary}
In this chapter, we have presented \emph{Antisymmetric Deep Graph Network} (A-DGN) and \emph{\underline{S}pace-\underline{W}eight \underline{An}tisymmetric Deep Graph Network} (SWAN), two novel differential-equation inspired DGNs (DE-DGNs) designed to address the oversquashing problem. 

Unlike previous approaches, by imposing stability and conservative constraints on the differential equation through the use of antisymmetric constraints, the proposed frameworks can learn and preserve long-range dependencies between nodes. 
We prove theoretically that the differential equation corresponding to A-DGN is both stable and non-dissipative. Consequently, typical problems of systems with unstable and lossy dynamics, \eg gradient explosion or vanishing, do not occur. On the other hand, SWAN extends the concepts introduced by A-DGN to include both global (\ie graph-wise) and local (\ie node-wise) non-dissipative properties through space and weight antisymmetric parameterization.
Thanks to their formulations, both A-DGN and SWAN provide a general design principle for introducing non-dissipativity as an inductive bias in any DE-DGN.

Our theoretical and experimental results emphasize the importance of global and local non-dissipativity achieved by SWAN and A-DGN. Moreover, our experimental analysis shows that our frameworks largely outperform standard DGNs when capturing long-range dependencies on several graph benchmarks. For these reasons, we believe that our methods represent a significant step forward in addressing oversquashing in DGNs.

\chapter{A Physics-Inspired DGN}\label{ch:phdgn}
Chapter~\ref{ch:antisymmetry} introduces antisymmetric constraints to achieve stable and non-dissi\-pa\-ti\-ve dynamical systems enabling long-range propagation in DGNs for static graphs. In this chapter, we are interested in designing the information flow within a (static) graph as a solution of a port-Hamiltonian system~\citep{van2000l2}, which is a generalization of Hamiltonian systems introduced in Section~\ref{sec:hamiltonian_sys}, that provides a general formalism for physical systems that allows for both conservative and non-conservative dynamics with the aim of allowing flexible long-range propagation in DGNs. 

Thus, we provide a theoretically grounded framework through the prism of Hamiltonian-inspired DE-DGNs, named \emph{(port-)Hamiltonian Deep Graph Network} (PH-DGN), which defines a new message-passing scheme inspired by port-Hamil\-ton\-ian dynamics. By design, our method introduces the flexibility to balance non-dissipative long-range propagation and non-conservative behaviors as required by the specific task at hand. Therefore, when using purely Hamiltonian dynamics, our method allows the preservation and propagation of long-range information by obeying the conservation laws. In contrast, when our method is used to its full extent, internal damping and additional forces can deviate from this purely conservative behavior, potentially increasing effectiveness in the downstream task. Leveraging the connection with Hamiltonian systems, we provide theoretical guarantees that information is conserved over time, \ie spatial hops. Lastly, the general formulation of our approach can seamlessly incorporate any neighborhood aggregation function (\ie DGN), thereby endowing these methods with the distinctive properties of our PH-DGN.

Our main contributions of this chapter are:
\begin{itemize}
    \item We introduce PH-DGN, a novel general DE-DGN inspired by (port-)Hamil\-ton\-ian dynamics, which enables the balance and integration of non-dissi\-pa\-ti\-ve long-range propagation and non-conservative behavior while seamlessly incorporating the most suitable aggregation function.
    \item We theoretically prove that, when pure Hamiltonian dynamic is employed, both the continuous and discretized versions of our framework allow for long-range propagation in the message passing flow, since node states retain their past.
    \item We introduce tools inspired by mechanical systems that deviate the Hamiltonian dynamic from its conservative behavior, thus facilitating a clear interpretation from the physics perspective.
    \item We conduct extensive experiments to demonstrate the benefits of our method and the ability to stack thousands of layers. Our PH-DGN outperforms existing state-of-the-art methods on both synthetic and real-world tasks.
\end{itemize} 

We base this chapter on \cite{gravina_phdgn}.
\section{(Port-)Hamiltonian Deep Graph Network}\label{sec:phdgn_method}
We introduce a new DE-DGN framework that designs the information flow within a graph as the solution of a (port-)Hamiltonian system~\citep{van2000l2}. 
Differently from the pure Hamiltonian systems introduced in Section \ref{sec:hamiltonian_sys}, port-Hamiltonian systems let us introduce non-conservative phenomena, such as internal dampening and external forcing, thereby relaxing the guarantee of energy preservation.

\begin{figure}[h]
    \centering
    \includegraphics[width=\textwidth]{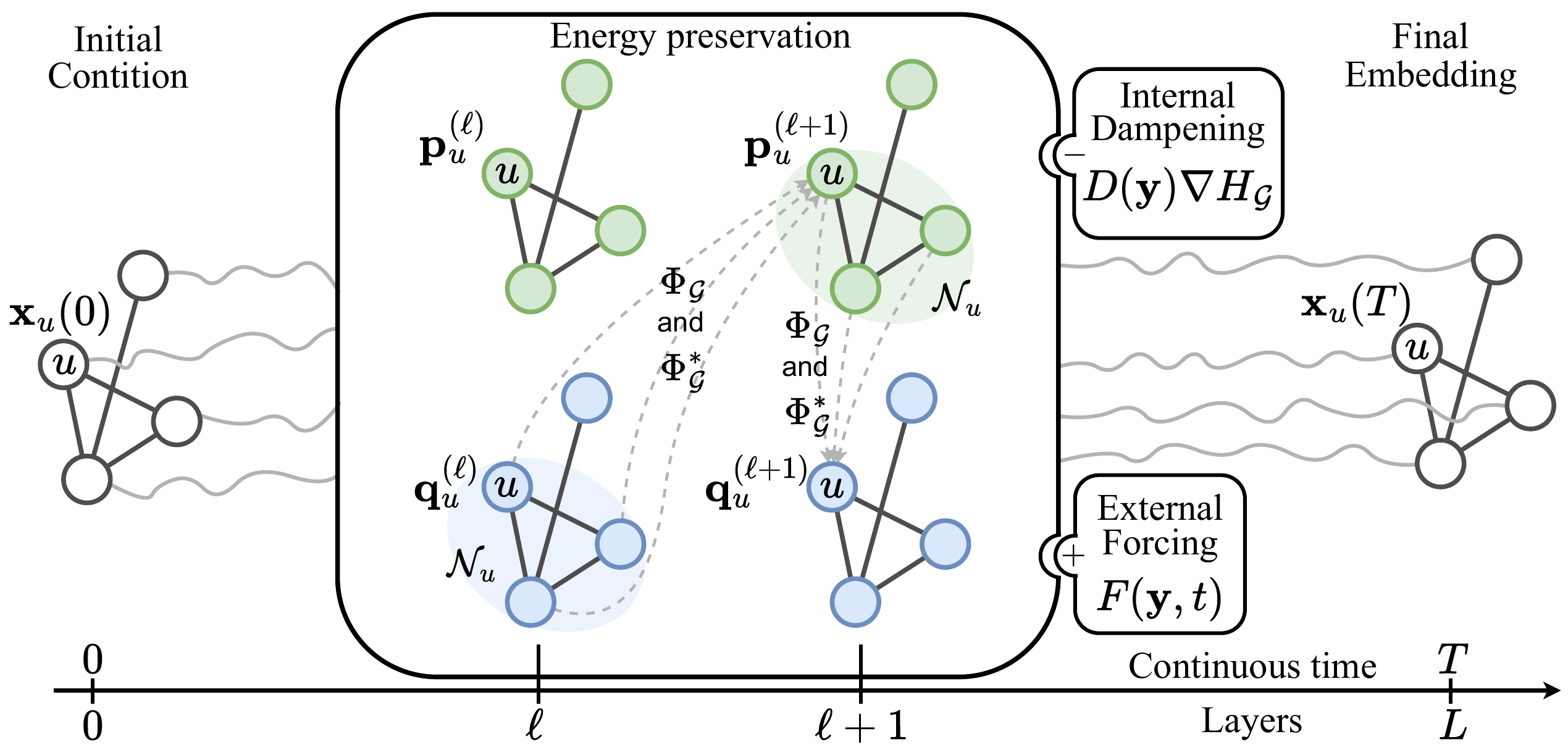}
    \caption{A high-level overview of the proposed (port-)Hamiltonian Deep Graph Network. It summarizes how the initial node state $\bfx_u(0)$ is propagated by means of energy preservation up until the terminal time $T$ (\ie layer $L$), $\bfx_u(T)$. While the global system's state $\mathbf{y}$ evolves preserving energy, external forces (\ie dampening $D(\mathbf{y})$ and external control $F(\mathbf{y},t)$) can intervene to alter its conservative trajectory. The gray trajectories between the initial and final states represent the continuous evolution of the system. The discrete message passing step from layer $\ell$ to $\ell +1$, which is shown in middle of the figure, is given by the coupling of coordinates $\bfq$ and momenta $\bfp$ in terms of neighborhood aggregation $\Phi_{\mathcal{G}}$ and influence to adjacent neighbors $\Phi^*_{\mathcal{G}}$. Self-influence on both $\bfq$ and $\bfp$ from the previous step $\ell$ are omitted for simplicity.
    }
    \label{fig:phdgn_model}
\end{figure}

Here, we show how the Hamiltonian formulation provides the backing to preserve and propagate long-range information between nodes, in adherence to the laws of conservation. The casting of the system in the more general port-Hamiltonian setting, then, introduces the possibility of 
trading non-dissipation with non-conservative behaviors when needed by the task at hand.
Our approach is general, as it can be applied to any message-passing DGN, and frames in a theoretically sound way the integration of non-dissipative propagation and non-conservative behaviors.
In the following, we refer to a \emph{Hamiltonian Deep Graph Network}\index{H-DGN} (\gls*{H-DGN}) when the framework is instantiated to a purely conservative DGN, and \emph{port-Hamiltonian Deep Graph Network}\index{PH-DGN} (\gls*{PH-DGN}) otherwise. Figure~\ref{fig:phdgn_model} shows our high-level architecture hinting at how the initial state of the system is propagated up to the terminal time $T$. While the state evolves preserving energy, internal dampening and additional forces can intervene to alter its conservative trajectory.

In the following, we first derive our method from a purely Hamiltonian system and prove its conservative behavior theoretically. We then complete our method by incorporating non-conservative behaviors through port-Hamiltonian dynamics.

\subsection{Hamiltonian-Inspired Message Passing}
To inject the Hamiltonian dynamics into a DE-DGN, we start by considering the graph Hamiltonian system described by the following ODE
\begin{equation}\label{eq:global_gde}
    \frac{d\bfy(t)}{dt} = \Jskew \nabla H_{\mathcal{G}}(\bfy(t)),
\end{equation}
for time $t\in[0, T]$ and subject to an initial condition $\bfy(0) = \bfy^0$. 
The term $\bfy(t)\in \mathbb{R}^{n d}$ is the vectorized view of $\bfX(t)$ that represents the global state of the graph at time $t$, with an even dimension $d$, following the notation of Hamiltonian systems \citep{Hairer2006-ks}. 
$H_{\mathcal{G}}: \mathbb{R}^{nd} \to \mathbb{R}$ is a neural-parameterized Hamiltonian function capturing the energy of the system. 
The skew-symmetric matrix $\Jskew = \begin{pmatrix}\bf0 & -\bfI_{nd/2}\\\bfI_{nd/2} & \bf0\end{pmatrix}$,  with $\bfI_{nd/2}$ being the identity matrix of dimension $nd/2$, reflects a rotation of the gradient $\nabla H_{\mathcal{G}}$ and couples the position and momentum of the system.

Since we are dealing with a Hamiltonian system, the global state $\bfy(t)$ is composed by two components which are the momenta, $\bfp(t) = (\bfp_1(t), \dots, \bfp_n(t))$, and the position, $\bfq(t) = (\bfq_1(t), \dots, \bfq_n(t))$, of the system, thus $\bfy(t) = (\bfp(t), \bfq(t))$. Therefore, from the node (local) perspective, each node state is expressed as $\bfx_u(t) = (\bfp_u(t), \bfq_u(t))$. 

Under this local node-wise perspective, Equation~\ref{eq:global_gde} can be equivalently written as 
\begin{equation}\label{eq:local_gde}
    \myodv{\bfx_u(t)}{t} = \begin{pmatrix}
        \dot{\bfp}_u(t)\\\dot{\bfq}_u(t)
    \end{pmatrix} = \begin{pmatrix}
        -\nabla_{\bfq_u}H_{\mathcal{G}}(\bfp(t),\bfq(t))\\\phantom{-}\nabla_{\bfp_u}H_{\mathcal{G}}(\bfp(t),\bfq(t))
    \end{pmatrix},\,\, \forall u \in \mathcal{V}.
\end{equation}

With the aim of designing a Hamiltonian system based on message passing, we instantiate the Hamiltonian function $H_\mathcal{G}$ as
\begin{equation}\label{eq:hamiltonian}
    H_{\mathcal{G}}(\bfy(t)) = \sum_{u \in \mathcal{V}}\Tilde{\sigma}(\bfW\xk(t) + \Phi_{\mathcal{G}}(\{\bfx_v(t)\}_{v\in\mathcal{N}_u}) + \bfb)^{\top}\mathbf{1}_d,
\end{equation}
where $\Tilde{\sigma}(\cdot)$ is the anti-derivative of a monotonically non-decreasing activation function $\sigma$, $\mathcal{N}_u$ is the neighborhood of node $u$, and $\Phi_{\mathcal{G}}$ is a neighborhood aggregation permutation-invariant function. Terms $\bfW\in\mathbb{R}^{d\times d}$ and $\bfb\in\mathbb{R}^d$ are the weight matrix and the bias vector, respectively, containing the trainable parameters of the system; $\mathbf{1}_d$ denotes a vector of ones of length $d$. 

By computing the gradient $\nabla_{\bfx_u}H_{\mathcal{G}}(\bfy(t))$ we obtain an explicit version of Equation~\ref{eq:local_gde}, which can be rewritten from the node-wise perspective
of the information flow as the sum of the self-node evolution influence and its neighbor's evolution influence (referred to as $\Phi_\mathcal{G}^*$). More formally, for each node $u\in\mathcal{V}$
\begin{multline}\label{eq:local_dyn}
\myodv{\xk(t)}{t} = \Jkskew\Biggl[\bfW^{\top}\sigma(\bfW\xk(t) + \Phi_{\mathcal{G}}(\{\bfx_v(t)\}_{v\in\mathcal{N}_u}) + \bfb)
 \\+\underbrace{\sum_{v \in \mathcal{N}_u\cup \{u\}}{\pdv{\Phi_{\mathcal{G}}(\{\bfx_v(t)\}_{v\in\mathcal{N}_u})}{\xk{(t)}}^\top\sigma(\bfW\xk(t) + \Phi_{\mathcal{G}}(\{\bfx_v(t)\}_{v\in\mathcal{N}_u}) + \bfb)}}_{\Phi^*_\mathcal{G}}\Biggr].
\end{multline}
Here, $\Jkskew$ has the same structure as $\Jskew$, but the identity blocks have dimension $d/2$ as it is applied to the single node $u$. We refer to the system in Equation~\ref{eq:local_dyn} as \emph{Hamiltonian Deep Graph Network} (H-DGN) as it adheres solely to conservation laws.

Now, given an initial condition $\bfx_u(0)$ for a node $u$, and the other nodes in the graph, the ODE defined in Equation~\ref{eq:local_dyn} (\ie H-DGN) is a continuous information processing system over a graph governed by conservation laws that computes the final node representation $\bfx_u(T)$. This is visually summarized in Figure~\ref{fig:phdgn_model} when dampening and external forcing are excluded.

Moreover, we observe that the general formulation of the neighborhood aggregation function $\Phi_\mathcal{G}(\{\bfx_v(t)\}_{v\in\mathcal{N}_u})$ allows implementing any function that aggregates nodes (and edges) information. Therefore, $\Phi_\mathcal{G}(\{\bfx_v(t)\}_{v\in\mathcal{N}_u})$ allows enhancing a standard DGNs with our Hamiltonian conservation. As a demonstration of this, in Section~\ref{sec:phdgn_experiments}, we experiment with two neighborhood aggregation functions, which are the classical GCN aggregation \citep{GCN} and the simple aggregation function presented in Equation~\ref{eq:simple_aggregation}.

\subsection{H-DGN allows Long-Range Propagation}
We show that our H-DGN in Equation~\ref{eq:local_dyn} adheres to the laws of conservation, allowing long-range propagation in the message-passing flow. 

As discussed in \citep{HaberRuthotto2017, gravina_adgn}, non-dissipative (\ie long-range) propagation is directly linked to the sensitivity of the solution of the ODE to its initial condition, thus to the stability of the system. Such sensitivity is controlled by the Jacobian's eigenvalues of Equation~\ref{eq:local_dyn}. Under the assumption that the Jacobian varies sufficiently slow over time and its eigenvalues are purely imaginary, then the initial condition is effectively propagated into the final node representation, making the system both stable and non-dissipative, thus allowing for long-range propagation.
\begin{theorem}\label{th:im_eig}
    The Jacobian matrix of the system defined by the ODE in Equation~\ref{eq:local_dyn} possesses eigenvalues purely on the imaginary axis, \ie $$\operatorname{Re}\left(\lambda_i\left(\pdv{}{\bfx_u}\Jkskew\nabla_{\bfx_u}H_{\mathcal{G}}(\bfy(t))\right)\right) = 0,\quad \forall i,$$
where $\lambda_i$ represents the $i$-th eigenvalue of the Jacobian.
\end{theorem}
\begin{boxedproof}
First, we note that 
\begin{equation}
    \pdv{}{\bfx_u}\Jkskew\nabla_{\bfx_u}H_{\mathcal{G}}(\mathbf{y}(t)) = \nabla^2_{\bfx_u}H_{\mathcal{G}}(\mathbf{y}(t))\Jkskew^{\top},
\end{equation} 
 where $\nabla^2_{\bfx_u}H_{\mathcal{G}}$ is the symmetric Hessian matrix.
 Hence, the Jacobian is shortly written as $\mathbf{AB}$, where $\mathbf{A}$ is symmetric and $\mathbf{B}$ is antisymmetric. Consider an eigenpair of $\mathbf{A B}$, where the eigenvector is denoted by $\mathbf{v}$ and the eigenvalue by $\lambda \ne 0$. Then:
\begin{align*}
\mathbf{v}^*\mathbf{AB} & = \lambda\mathbf{v}^*  \\
\mathbf{v}^*\mathbf{A} & =\lambda\mathbf{v}^*\mathbf{B}^{-1} \\
\mathbf{v}^* \mathbf{A} \mathbf{v} & =\lambda\left(\mathbf{v}^* \mathbf{B}^{-1} \mathbf{v}\right)
\end{align*}
where $*$ represents the conjugate transpose. On the left-hand side, it is noticed that the $\left(\mathbf{v}^* \mathbf{A} \mathbf{v}\right)$ term is a real number. Recalling that $\mathbf{B}^{-1}$ remains antisymmetric and for any real antisymmetric matrix $\mathbf{C}$ it holds that $\mathbf{C}^*=\mathbf{C}^{\top}=-\mathbf{C}$, it follows that $\left(\mathbf{v}^* \mathbf{C} \mathbf{v}\right)^*=\mathbf{v}^* \mathbf{C}^* \mathbf{v}=-\mathbf{v}^* \mathbf{C} \mathbf{v}$. Hence, the $\mathbf{v}^* \mathbf{B}^{-1} \mathbf{v}$ term on the right-hand side is an imaginary number. Thereby, $\lambda$ needs to be purely imaginary, and, as a result, all eigenvalues of $\mathbf{A B}$ are purely imaginary.
\end{boxedproof}
Then, we take a further step and strengthen such result by proving that the nonlinear vector field defined by H-DGN is divergence-free, thus preserving information within the graph during the propagation process and helping to maintain informative node representations.
In other words, the H-DGN's dynamics possess a non-dissipative behavior independently of both the assumption regarding the slow variation of the Jacobian and the position of the Jacobian eigenvalues on the complex plane. 
\begin{theorem}\label{th:divergence}
    The autonomous Hamiltonian $H_{\mathcal{G}}$ of the system in Equation~\ref{eq:local_dyn} with learnable weights shared across time stays constant at the energy level specified by the initial value $H_{\mathcal{G}}(\bfy(0))$, \ie
    \begin{equation}
        \myodv{H_{\mathcal{G}}}{t} = 0.
    \end{equation}
    H-DGN also possesses a divergence-free nonlinear vector field
    \begin{equation}
    \nabla \cdot \Jkskew \nabla_{\xk}H_{\mathcal{G}}(\mathbf{y}(t)) = 0,\quad t\in[0,T].
    \end{equation}
\end{theorem}
\begin{boxedproof}
When the system in Equation~\ref{eq:local_dyn} employs shared weights across time, then the resulting Hamiltonian is autonomous and does not depend explicitly on time, \ie $H_\mathcal{G}(\mathbf{y}(t), t) = H_\mathcal{G}(\mathbf{y}(t))$. 
In such case, the energy is naturally conserved in the system it represents.

The time derivative of $H(\mathbf{y}(t))$ is given by means of the chain-rule:
    \begin{equation}\label{eq:H_constant}
    \myodv{H(\mathbf{y}(t))}{t} = {\pdv{H(\mathbf{y}(t))}{\mathbf{y}(t)}}\cdot\myodv{\mathbf{y}(t)}{t} = {\pdv{H(\mathbf{y}(t))}{\mathbf{y}(t)}}\cdot\Jskew\pdv{H(\mathbf{y}(t))}{\mathbf{y}(t)} = 0,
\end{equation}
where the last equality holds since $\Jskew$ is antisymmetric. Having no change over time implies that $H(\mathbf{y}(t)) = H(\mathbf{y}(0)) = \text{const}$ for all $t$.

Since the Hessian $\nabla^2H_{\mathcal{G}}(\mathbf{y}(t))$ is symmetric, it follows directly
\begin{align*}
    \nabla \cdot \Jkskew\nabla_{\bfx_v}H_{\mathcal{G}}(\mathbf{y}(t)) 
    &=\sum_{i=1}^d{-\frac{\partial^2H_{\mathcal{G}}(\mathbf{y}(t))}{\partial q_{v}^{i} \partial p_{v}^{i}}+\frac{\partial^2H_{\mathcal{G}}(\mathbf{y}(t))}{\partial p_{v}^{i} \partial q_{v}^{i}}} = 0
\end{align*}
\end{boxedproof}
This allows us to interpret the system dynamics as purely rotational, without energy loss, and demonstrates that H-DGN is governed by conservation laws. 

We now provide a sensitivity analysis, following \citet{chang2018antisymmetricrnn,galimberti2023hamiltonian} and Chapter~\ref{ch:antisymmetry},
to prove that H-DGN effectively allows for long-range information propagation. Specifically, we measure the sensitivity of a node state after an arbitrary time $T$ of the information propagation with respect to its previous state, $\|\partial\bfx_u(T)/\partial\bfx_u(T-t)\|$. In other words, we compute the backward sensitivity matrix (BSM). We now provide a theoretical bound of our H-DGN.
\begin{theorem}\label{th:lowerb}
Considering the continuous system defined by Equation~\ref{eq:local_dyn}, the backward sensitivity matrix (BSM) is bounded from below:
$$\left\|\pdv{\xk(T)}{\xk(T-t)}\right\| \ge 1,\quad \forall t \in [0,T].$$
\end{theorem}
\begin{boxedproof}
In order to prove the lower bound on the BSM, we need a technical lemma that describes the time evolution of the BSM itself, which extends the result from \cite{galimberti2023hamiltonian}.
\begin{lemma}
\label{lemma:time_bsm_self_loop}
    Given the system dynamics of the ODE in Equation~\ref{eq:global_gde} governing the H-DGN, we have that
    \begin{equation}\label{eq:BSM_global}
        \myodv{}{t}\pdv{\mathbf{y}(T)}{\mathbf{y}(T-t)} = \Jskew\left.\pdv{H}{\mathbf{y}}\right|_{\mathbf{y}(T-t)}\pdv{\mathbf{y}(T)}{\mathbf{y}(T-t)}
    \end{equation}
    as in \citep{galimberti2023hamiltonian}. The same applies, with a slightly different formula, for each node $u$, that is the BSM satisfies
    \begin{equation}\label{eq:BSM_local}
        \myodv{}{t}\pdv{\xk(T)}{\xk(T-t)} = \left.\pdv{\mathbf{y}}{\xk}\right\rvert_{(T-t)} \left.\pdv{f_u}{\mathbf{y}}\right\rvert_{\mathbf{y}(T-t)}\pdv{\xk(T)}{\xk(T-t)} = F_u \pdv{\xk(T)}{\xk(T-t)}
    \end{equation}
    where $f_u$ is the restriction of $f = \Jskew\pdv{H}{\mathbf{y}}$ to the components corresponding to $\xk$, that is the dynamics of node $u$, which can be written as
    \begin{equation}
        f_u = L_u f = L_u \Jskew\pdv{H}{\mathbf{y}}
    \end{equation}
    where $L_u$ is the readout matrix, of the form
    \begin{equation}
         L_u = \begin{bmatrix}
            0_{\frac{d}{2}\times \frac{d}{2}(u-1)} & I_{\frac{d}{2}\times \frac{d}{2}} & 0_{\frac{d}{2}\times \frac{d}{2}(n-u)} & 0_{\frac{d}{2}\times \frac{d}{2}(u-1)} & 0_{\frac{d}{2}\times \frac{d}{2}} & 0_{\frac{d}{2}\times \frac{d}{2}(n-u)}\\
            0_{\frac{d}{2}\times \frac{d}{2}(u-1)} & 0_{\frac{d}{2}\times \frac{d}{2}} & 0_{\frac{d}{2}\times \frac{d}{2}(n-u)} & 0_{\frac{d}{2}\times \frac{d}{2}(u-1)} & I_{\frac{d}{2}\times \frac{d}{2}} & 0_{\frac{d}{2}\times \frac{d}{2}(n-u)}
        \end{bmatrix}
    \end{equation}
    which is a projection on the coordinates of a single node $u$.
    Notice as well that, in denominator notation
    \begin{equation}
        \pdv{\mathbf{y}}{\xk} = L_u
    \end{equation}
\end{lemma}
We only show Equation~\ref{eq:BSM_local}, as it is related to graph networks and is actually a harder version of Equation~\ref{eq:BSM_global}, with the latter being already proven in \cite{galimberti2023hamiltonian}. We also show the last part of the proof in a general sense, without using the specific matrices of the Hamiltonian used.
\begin{proof}
    Following \cite{galimberti2023hamiltonian}, the solution to the ODE $\myodv{\xk}{t} = f_u(\mathbf{y}(t))$ can be written in integral form as
    \begin{equation}
        \xk(T) = \xk(T-t) + \int_{T-t}^{T}f_u(\mathbf{y}(\tau))\dsmth \tau = \xk(T-t) + \int_{0}^{t}f_u(\mathbf{y}(T-t + s))\dsmth s
    \end{equation}
    Differentiating by the solution at a previous time $\xk(T-t)$ we obtain
    \begin{align}
        \pdv{\xk(T)}{\xk(T-t)} & = I_{u} + \pdv{\int_0^{\top} f_u(\mathbf{y}(T-t + s))\dsmth s}{\xk(T-t)} \nonumber\\
        & = I_u + \int_{0}^{t}{\pdv{f_u(\mathbf{y}(T-t+s))}{\xk(T-t)}} \nonumber\\ 
        & = I_u + \int_{0}^{t}{\pdv{\mathbf{y}(T-t+s)}{\xk(T-t)}\left.\pdv{f_u}{\mathbf{y}}\right\rvert_{\mathbf{y}(T-t+s)}\dsmth s}
    \end{align}
    where in the second equality we brought the derivative term under the integral sign and in the third we used the chain rule of the derivative (recall we are using denominator notation). Considering a slight perturbation in time $\delta$, we consider $\pdv{\xk(T)}{\xk(T-t-\delta)}$ as this will be used to calculate the time derivative of the BSM. Using again the chain rule for the derivative and the formula above with $T-t-\delta$ instead of $T-t$, we have that
    \begin{multline}\label{eqn:time_jacobian}
            \pdv{\xk(T)}{\xk(T-t-\delta)} = \pdv{\xk(T-t)}{\xk(T-t-\delta)}\pdv{\xk(T)}{\xk(T-t)} \\
            = \left( I_u + \int_{0}^{\delta}{\pdv{\mathbf{y}(T-t-\delta+s)}{\xk(T-t-\delta)}\left.\pdv{f_u}{\mathbf{y}}\right\rvert_{\mathbf{y}(T-t-\delta+s)}\dsmth s}\right)\pdv{\xk(T)}{\xk(T-t)}
    \end{multline}
    This way, we have expressed $\pdv{\xk(T)}{\xk(T-t-\delta)}$ in terms of $\pdv{\xk(T)}{\xk(T-t)}$. To calculate our objective, we want to differentiate with respect to $\delta$. We first calculate the difference:
    \begin{multline}
        \pdv{\xk(T)}{\xk(T-t-\delta)} - \pdv{\xk(T)}{\xk(T-t)} = \\
        =\left(\int_{0}^{\delta}{\pdv{\mathbf{y}(T-t-\delta+s)}{\xk(T-t-\delta)}\left.\pdv{f_u}{\mathbf{y}}\right\rvert_{\mathbf{y}(T-t-\delta+s)}\dsmth s}\right)\pdv{\xk(T)}{\xk(T-t)}
    \end{multline}
    We can now divide by $\delta$ and take the limit $\delta\rightarrow 0$
    \begin{align}
        & \lim_{\delta\rightarrow 0}\frac{1}{\delta}\left(\pdv{\xk(T)}{\xk(T-t-\delta)}  - \pdv{\xk(T)}{\xk(T-t)}\right) = \nonumber\\
        & = \lim_{\delta\rightarrow 0}\left(\frac{1}{\delta}\int_{0}^{\delta}{\pdv{\mathbf{y}(T-t-\delta+s)}{\xk(T-t-\delta)}\left.\pdv{f_u}{\mathbf{y}}\right\rvert_{\mathbf{y}(T-t-\delta+s)}\dsmth s}\right)\pdv{\xk(T)}{\xk(T-t)}\nonumber\\ 
        & =  \left.\pdv{y}{\xk}\right\rvert_{(T-t)}\left.\pdv{f_u}{y}\right\rvert_{\mathbf{y}(T-t)}\pdv{\xk(T)}{\xk(T-t)}
    \end{align}
    Where in the final equality we used the fundamental theorem of calculus. Finally
    \begin{equation}\label{eq:bsm_time_ode}
        \myodv{}{t}\pdv{\xk(T)}{\xk(T-t)} = \left.\pdv{\mathbf{y}}{\xk}\right\rvert_{(T-t)}\left.\pdv{f_u}{\mathbf{y}}\right\rvert_{\mathbf{y}(T-t)}\pdv{\xk(T)}{\xk(T-t)} = F_u \pdv{\xk(T)}{\xk(T-t)}
    \end{equation}
    giving us the final result.
\end{proof}

We are now ready to prove Theorem~\ref{th:lowerb}. First, we calculate that
\begin{equation}
    \pdv{f_u}{\mathbf{y}} = \pdv{}{y}\left(L_u \Jskew \pdv{H}{\mathbf{y}}\right) = \pdv[order={2}]{H}{\mathbf{y}}\Jskew^{\top} L_u^{\top} = S\Jskew^{\top} L_u^{\top}
\end{equation}
so that $F_u = L_u S\Jskew^{\top} L_u^{\top}$. This will be helpful in the following matrix calculations.
\begin{proof}
    For brevity, we call $\left[\pdv{\xk(T)}{\xk(T-t)}\right] = \Psi_u(T,T-t)$, which will be indicated simply as $\Psi_u$. When $t=0$, $\Psi_u$ is just the Jacobian of the identity map $\Psi_u(T,T) = I_u$ and the result $\Psi_u^{\top}\Jkskew \Psi_u = \Jkskew$ is true for $t=0$. Calculating the time derivative on $\Psi_u^{\top}\Jkskew \Psi_u$ we have that
    \begin{align}
            \myodv{}{t}[\Psi_u^{\top}\Jkskew \Psi_u] & = \dot{\Psi}_u^{\top} \Jkskew \Psi_u + \Psi_u^{\top} \Jkskew \dot{\Psi}_u \nonumber\\
             & = (F_u\Psi_u)^{\top} \Jkskew \Psi_u + \Psi_u^{\top} \Jkskew F_u \Psi_u  \nonumber\\
             & = \Psi_u^{\top} L_u \Jskew S^{\top}  L_u^{\top} \Jskew_u \Psi_u + \Psi_u^{\top} \Jskew_u L_u S\Jskew^{\top} L_u^{\top} \Psi \nonumber\\
             & = \Psi_u^{\top} \left(L_u \Jskew S L_u^{\top} \Jskew_u + \Jskew_u L_u S \Jskew^{\top} L_u^{\top}\right)\Psi_u
    \end{align}
    where in the second equality we used the result from Lemma~\ref{lemma:time_bsm_self_loop}.
    We just need to show that the term in parentheses is zero so that the time derivative is zero. Using the relations $\Jskew^{\top} L_u^{\top}= -L_u^{\top}\Jkskew$ and $J_u L_u= L_u\Jskew$ we easily see that, finally
    \begin{align}
        \myodv{}{t}\Bigl(\left[\pdv{\xk(T)}{\xk(T-t)}\right]^{\top}& \Jskew_u \left[\pdv{\xk(T)}{\xk(T-t)}\right]\Bigr) = \nonumber\\
        &=\Psi_u^{\top}\left(L_u\Jskew S L_u^{\top}\Jkskew +L_u\Jskew S (-L_u^{\top}\Jkskew) \right)\Psi_u \nonumber\\
        &= 0
    \end{align}
    which means that $\left[\pdv{\xk(T)}{\xk(T-t)}\right]^{\top} \Jskew_u \left[\pdv{\xk(T)}{\xk(T-t)}\right]$ is constant and equal to $\Jskew_u$ for all $t$, that is our thesis.
Now, the bound on the gradient follows by considering any sub-multiplicative norm $\|\cdot\|$:
\begin{equation*}
     \left\|\Jkskew \right\| = \left\| \left[\pdv{\xk(T)}{\xk(T-t)}\right]^{\top} \Jskew_k \left[\pdv{\xk(T)}{\xk(T-t)}\right] \right\| \le \left\| \pdv{\xk(T)}{\xk(T-t)}\right\|^2 \left\|\Jkskew\right\| 
\end{equation*}
and simplifying by $\left\|\Jkskew\right\| = 1$.
\end{proof}
This concludes the proof of Theorem~\ref{th:lowerb}.
\end{boxedproof}

The result of Theorem~\ref{th:lowerb} indicates that the gradients in the backward pass do not vanish, enabling the effective propagation of previous node states through successive transformations to the final nodes' representations. Therefore, H-DGN has a conservative message passing, where the final representation of each node retains its complete past. 
We observe that Theorem~\ref{th:lowerb} holds even during discretization when the Symplectic Euler method is employed (see Section~\ref{sec:phdgn_discretization}).

Although the sensitivity of a node state after a time $t$ with respect to its previous state can be bounded from below, allowing effective conservative message passing in H-DGN, we observe that it is possible to compute an upper bound on such a measure, which we provide in the following theorem, giving the full picture of the time dynamics of the gradients. While the theorem shows that, theoretically, the sensitivity measure may grow (\ie potentially causing gradient explosion), we emphasize that during our experiments we did not encounter such a problem.

\begin{theorem}\label{th:upperb}
Consider the continuous system defined by Equation~\ref{eq:local_dyn}, if $\sigma$ is a non-linear function with bounded derivative, i.e. $\exists M > 0, |\sigma'(x)| \le M$, and the neighborhood aggregation function is of the form $\Phi_{\mathcal{G}} = \sum_{v \in \mathcal{N}_u}\mathbf{V}\bfx_v$, the backward sensitivity matrix (BSM) is bounded from above:
$$\left\|\pdv{\xk(T)}{\xk(T-t)}\right\| \leq \sqrt{d}\,\text{exp}(QT),\quad \forall t \in [0,T],$$
where $Q = \sqrt{d}\,M\|\mathbf{W}\|_2^2 + \sqrt{d}\,M\text{max}_{i\in[n]}|\mathcal{N}_i|\|\mathbf{V}\|_2^2$.
\end{theorem}
\begin{boxedproof}
    To prove the upper bound, we use the following technical lemma:
\begin{lemma}[\cite{galimberti2023hamiltonian}]\label{lemma:columnbound}
Consider a matrix $\mathbf{A} \in \mathbb{R}^{n \times n}$ with columns $\mathbf{a}_i \in \mathbb{R}^n$, i.e., $\mathbf{A}=\left[\begin{array}{llll}\mathbf{a}_1 & \mathbf{a}_2 & \cdots & \mathbf{a}_n\end{array}\right]$, and assume that $\left\|\mathbf{a}_i\right\|_2 \leq$ $\gamma^{+}$ for all $i=1, \ldots, n$. Then, $\|\mathbf{A}\|_2 \leq \gamma^{+} \sqrt{n}$.
\end{lemma}
This lemma gives a bound on the spectral norm of a matrix when its columns are uniformly bounded in norm.
Therefore, our proof strategy for Theorem~\ref{th:upperb} lies in bounding each column of the BSM matrix.
\begin{proof}
Consider the ODE in Equation~\ref{eq:bsm_time_ode} from Lemma~\ref{lemma:time_bsm_self_loop} and split $\pdv{\bfx_u(T)}{\bfx_u(T-t)}$ into columns $\pdv{\bfx_u(T)}{\bfx_u(T-t)}=\left[\begin{array}{llll}\mathbf{z}_1(\mathbf{t}) & \mathbf{z}_2(t) & \ldots & \mathbf{z}_d(t)\end{array}\right]$. Then, Equation~\ref{eq:bsm_time_ode} is equivalent to
\begin{equation}\label{eq:columnwise_dyn}
    \dot{\mathbf{z}}_i(t)=\mathbf{A}_u(T-t) \mathbf{z}_i(t), \quad t \in[0, T], i=1,2 \ldots, d,
\end{equation}
subject to $\mathbf{z}_i(0)=e_i$, where $e_i$ is the unit vector with a single nonzero entry in position $i$. The solution of the linear system of ODEs in Equation~\ref{eq:columnwise_dyn} is given by the integral equation
\begin{equation}\label{eq:columnwise_solu}
    \mathbf{z}_i(t)=\mathbf{z}_i(0)+\int_0^t \mathbf{A}_u(T-s) \mathbf{z}_i(s) d s, \quad t \in[0, T].
\end{equation}
By assuming that $\|\mathbf{A}_u(\tau)\|_2 \leq Q$ for all $\tau \in[0, T]$, and applying the triangular inequality in Equation~\ref{eq:columnwise_solu}, it is obtained that:
$$
\left\|\mathbf{z}_i(t)\right\|_2 \leq\left\|\mathbf{z}_i(0)\right\|_2+Q \int_0^t\left\|\mathbf{z}_i(s)\right\|_2 d s=1+Q \int_0^t\left\|\mathbf{z}_i(s)\right\|_2 d s,
$$
where the last equality follows from $\left\|\mathbf{z}_i(0)\right\|_2=\left\|e_i\right\|_2=1$ for all $i=1,2, \ldots, d$. Then, applying the Gronwall inequality, it holds for all $t \in[0, T]$
\begin{equation}\label{eq:general_column_bound}
    \left\|\mathbf{z}_i(t)\right\|_2 \leq \exp (Q T) .
\end{equation}
By applying Lemma \ref{lemma:columnbound}, the general bound follows.

Lastly, we characterize $Q$ by bounding the norm $\|\mathbf{A}_u(\tau)\|_2 \; \forall \tau \in [0,T]$.
From Lemma \ref{lemma:time_bsm_self_loop} $\mathbf{A}_v$ can be expressed as $\mathbf{A}_u = \mathbf{L}_u  \mathbf{S} \Jskew^{\top} \mathbf{L}_u^{\top}$, which is equivalently $ \mathbf{A}_u =  \nabla^2_{\bfx_u}{H}_{\mathcal{G}}(\mathbf{y})\Jkskew^{\top}$, since $\Jskew^{\top} \mathbf{L}_u^{\top}= \mathbf{L}_v^{\top}\Jkskew^{\top}$.
The Hessian $\nabla^2_{\bfx_u}{H}_{\mathcal{G}}(\mathbf{y})$ is of the form:
\begin{multline*}
\nabla^2_{\bfx_u}H_{\mathcal{G}}(\mathbf{y}) = \mathbf{W}^{\top}\operatorname{diag}(\sigma'(\mathbf{W}\bfx_u + \Phi_u + b))\mathbf{W} \\
+\sum_{v \in \mathcal{N}_u} \mathbf{V}^{\top}\operatorname{diag}(\sigma'(\mathbf{W}\bfx_v +\Phi_v +b))\mathbf{V}.
\end{multline*}
After noting that $\|\nabla^2_{\bfx_u}{H}_{\mathcal{G}}(\mathbf{y})\Jkskew^{\top}\|_2 \le \|\nabla^2_{\bfx_u}{H}_{\mathcal{G}}(\mathbf{y})\|_2\|\Jkskew^{\top}\|_2$, the only varying part is the Hessian $\nabla^2_{\bfx_u}{H}_{\mathcal{G}}(\mathbf{y})$ since $\|\Jkskew^{\top}\|_2=1$.
By Lemma~\ref{lemma:columnbound} $\operatorname{diag}(\sigma'(x)) \le \sqrt{d}\,M$ and noting that $\|\mathbf{X}^{\top}\| = \|\mathbf{X}\|$ for any square matrix $\mathbf{X}$, then
\begin{equation*}
    \left\|\nabla^2_{\bfx_u}H_{\mathcal{G}}(\mathbf{y})\right\|_2 \le \sqrt{d}\,M \,\|\mathbf{W}\|_2^2 + \sqrt{d}\,M\, \operatorname{max}_{i \in [n]}|\mathcal{N}_i| \; \|\mathbf{V}\|_2^2 \eqqcolon Q.
\end{equation*}
This also justifies our previous assumption that $\|\mathbf{A}_u(\tau)\|_2$ is bounded.
\end{proof}
\end{boxedproof}

\subsection{Introducing Dissipative Components} 
A purely conservative Hamiltonian inductive bias forces the node states to follow trajectories that maintain constant energy, potentially limiting the effectiveness of the DGN on downstream tasks by restricting the system's ability to model all complex nonlinear dynamics.
To this end, we complete the formalization of our port-Hamiltonian framework by introducing tools from mechanical systems, such as friction and external control, to learn how much the dynamic should deviate from this purely conservative behavior. Therefore, we extend the dynamics in Equation~\ref{eq:local_dyn} to a \emph{port}-Hamiltonian by including two new terms $D(\bfq)\in\mathbb{R}^{d/2 \times d/2}$ and $F(\bfq,t)\in\mathbb{R}^{d/2}$, \ie
\begin{equation}\label{eq:port-hamiltonian}
    \myodv{\xk(t)}{t} = \left[\Jkskew-\begin{pmatrix}
    D(\bfq(t)) & \bf0\\\bf0 & \bf0 
\end{pmatrix} \right]\nabla_{\xk} H_{\mathcal{G}}(\bfy(t)) + \begin{pmatrix}
    F(\bfq(t),t)\\\bf0
\end{pmatrix},\; \forall u \in \mathcal{V}.
\end{equation}
Depending on the definition of $D(\bfq(t))$ we can implement different forces. Specifically, if $D(\bfq(t))$ is positive semi-definite then it implements internal dampening, while a negative semi-definite implementation leads to internal acceleration. A mixture of dampening and acceleration is obtained otherwise. In the case of dampening, the energy is decreased along the flow of the system \citep{van2000l2}. 
To further enhance the modeling capabilities, we integrate the learnable state- and time-dependent external force $F(\bfq(t),t)$, which further drives node representation trajectories. Figure~\ref{fig:phdgn_model} visually summarizes how such tools can be plugged in our framework during node update. 

Although $D(\bfq(t))$ and $F(\bfq(t),t)$ can be implemented as static (fixed) functions, in our experiments in Section~\ref{sec:phdgn_experiments} we employ neural networks to learn such terms. We provide additional details on the specific architectures in Appendix~\ref{app:phdgn_hyperparams}.
In the following, we refer to a DGN following Equation~\ref{eq:port-hamiltonian} as \emph{port-Hamiltonian Deep Graph Network} (PH-DGN) to distinguish it from the purely conservative H-DGN in Equation~\ref{eq:local_dyn}. We provide further details about the discretization of PH-DGN in Section~\ref{sec:phdgn_discretization}.

\subsection{Discretization of (port-)Hamiltonian DGNs} \label{sec:phdgn_discretization}
As for standard DE-DGNs a numerical discretization method is needed to solve Equation~\ref{eq:local_dyn}. However, as observed in \cite{HaberRuthotto2017, galimberti2023hamiltonian}, not all standard techniques can be employed for solving Hamiltonian systems. Indeed, symplectic integration methods need to be used to preserve the conservative properties in the discretized system (see Section~\ref{sec:discretization_method_de}). 

For the ease of simplicity, in the following we focus on the Symplectic Euler method (see Section~\ref{sec:discretization_method_de}), however, we observe that more complex methods such as Str\"omer-Verlet can be employed \citep{Hairer2006-ks}.

The Symplectic Euler scheme, applied to our H-DGN in Equation~\ref{eq:local_dyn}, updates the node representation at the $(\ell+1)$-th step as 
\begin{equation}\label{eq:implicit_symplectic_euler_local}
        \bfx_u^{\ell+1} = \begin{pmatrix}
            \bfp^{\ell+1}_{u}\\[0.2cm]
            \bfq^{\ell+1}_{u} 
        \end{pmatrix}
        = \begin{pmatrix}
            \bfp^{\ell}_{u}\\[0.2cm]
            \bfq^{\ell}_{u}
        \end{pmatrix} + \epsilon \Jkskew
        \begin{pmatrix}
    \nabla_{\bfp_u} H_{\mathcal{G}}(\bfp^{\ell
    },\bfq^{\ell})\\[0.2cm]\nabla_{\bfq_u} H_{\mathcal{G}}(\bfp^{\ell+1},\bfq^{\ell})
\end{pmatrix},\quad  \forall u \in \mathcal{V}.
\end{equation}
with $\epsilon$ the step size of the numerical discretization.
We note that Equation~\ref{eq:implicit_symplectic_euler_local} relies on both the current and future state of the system, hence marking an implicit scheme that would require solving a linear system of equations in each step. To obtain an explicit version of Equation~\ref{eq:implicit_symplectic_euler_local}, we consider the neighborhood aggregation function in Equation~\ref{eq:simple_aggregation} and impose a structure assumption on $\bfW$ and $\bfV$, namely $\bfW = \begin{pmatrix}\bfW_p & \bf0\\\bf0 & \bfW_q\end{pmatrix}$ and $\bfV = \begin{pmatrix}\bfV_p & \bf0\\\bf0 & \bfV_q
\end{pmatrix}$. We note that a comparable assumption can be made for other neighborhood aggregation functions, such as GCN aggregation.

Therefore, the gradients in Equation~\ref{eq:implicit_symplectic_euler_local} can be rewritten in the explicit form as
\begin{align}
\bfp^{\ell+1}_u &= \bfp^{\ell}_u - \epsilon \Biggl[\bfW_q^{\top}\sigma(\bfW_q\bfq^{\ell}_u + \Phi_{\mathcal{G}}(\{\bfq^{\ell}_v\}_{v\in\mathcal{N}_u} ) + \bfb_q) \nonumber\\ 
&\hspace{2cm} + \sum_{v \in \mathcal{N}_u\setminus \{u\}}{\bfV_q^{\top}}\sigma(\bfW_q\bfq^{\ell}_v + \Phi_{\mathcal{G}}(\{\bfq^{\ell}_j\}_{j\in\mathcal{N}_v} ) + \bfb_q)\Biggr] \label{eq:discretized_local_hamil_p}\\
\bfq^{\ell+1}_u &= \bfq^{\ell}_u + \epsilon\Biggl[ \bfW_p^{\top}\sigma(\bfW_p\bfp^{\ell+1}_u + \Phi_{\mathcal{G}}(\{\bfp^{\ell+1}_v\}_{v\in\mathcal{N}_u} ) + \bfb_p) \nonumber\\ 
&\hspace{2cm}+ \sum_{v \in \mathcal{N}_u\setminus \{u\}}{\bfV_p^{\top}}\sigma(\bfW_p\bfp^{\ell+1}_v + \Phi_{\mathcal{G}}(\{\bfp^{\ell+1}_j\}_{j\in\mathcal{N}_v} ) + \bfb_p)\Biggr]. \label{eq:discretized_local_hamil_q}
\end{align}

We observe that Equations~\ref{eq:discretized_local_hamil_p} and \ref{eq:discretized_local_hamil_q} can be understood as coupling
two DGN layers. 
This discretization mechanism is visually summarized in the middle of Figure~\ref{fig:phdgn_model} where a message-passing step from layer $\ell$ to layer $\ell+1$ is performed.

In the case of PH-DGN in Equation~\ref{eq:port-hamiltonian} the discretization employs the same step for $\mathbf{q}^{\ell+1}$ in Equation~\ref{eq:discretized_local_hamil_q} while Equation~\ref{eq:discretized_local_hamil_p} includes the dissipative components, thus it can be rewritten as
\begin{equation}\label{eq:discretized_port_ham}
\mathbf{p}_u^{\ell+1} = \mathbf{p}_u^{\ell} + \epsilon\Biggl[-\nabla_{\mathbf{q}_u}{{H}_{\mathcal{G}}}(\mathbf{p}^{\ell},\mathbf{q}^{\ell})
- D_u(\mathbf{q}^{\ell
}) \nabla_{\mathbf{p}_u}{{H}_{\mathcal{G}}}(\mathbf{p}^{\ell},\mathbf{q}^{\ell}) + F_u(\mathbf{q}^{\ell},t)\Biggr].
\end{equation}

To provide a clear understanding of our PH-DGN, Algorithm~\ref{alg:phdgn} presents how node embeddings are computed by 
the discretized version of our model. Additional insights on practical implementations of the dampening and external force components are presented in Appendix~\ref{app:phdgn_hyperparams}.

Lastly, it is important to acknowledge that properties observed in the continuous domain may not necessarily hold in the discrete setting due to the limitations of the discretization method. In the following theorem, we show that when the Symplectic Euler method is employed, then Theorem~\ref{th:lowerb} holds.
\begin{theorem}\label{th:lowerb_discrete}
Considering the discretized system in Equation~\ref{eq:implicit_symplectic_euler_local} obtained by Symplectic Euler discretization, the backward sensitivity matrix (BSM) is bounded from below:
$$\left\|\pdv{\xk^{L}}{\xk^{L-\ell}}\right\| \ge 1,\quad \forall \ell \in [0,L].$$
\end{theorem}
\begin{boxedproof}
    In the discrete case, since the semi-implicit Euler integration scheme is a symplectic method, it holds that:
 \begin{equation}\label{eq:bsm_symp}
     \left[\pdv{\bfx_u^{\ell}}{\bfx_u^{\ell-1}}\right]^{\top} \Jskew_u \left[\pdv{\bfx_u^{\ell}}{\bfx_u^{\ell-1}}\right] = \Jskew_u
 \end{equation}
Further, by using the chain rule and applying Equation~\ref{eq:bsm_symp} iteratively we get:
 \begin{equation*}
     \left[\pdv{\bfx_u^{L}}{\bfx_u^{L-\ell}}\right]^{\top} \Jkskew \left[\pdv{\bfx_u^{L}}{\bfx_u^{L-\ell}}\right] = \left[\prod_{i=L-\ell}^{L-1}\pdv{\bfx_u^{i+1}}{\bfx_u^{i}}  \right]^{\top} \Jkskew \left[\prod_{i=L-\ell}^{L-1}\pdv{\bfx_u^{i+1}}{\bfx_u^{i}}  \right] = \Jkskew
 \end{equation*}
Hence, the BSM is symplectic at arbitrary depth and we can conclude the proof with:
\begin{equation}
     \left\|\Jkskew \right\| = \left\| \left[\pdv{\bfx_u^{L}}{\bfx_u^{L-\ell}}\right]^{\top} \Jkskew \left[\pdv{\bfx_u^{L}}{\bfx_u^{L-\ell}}\right] \right\| \le \left\| \pdv{\bfx_u^{L}}{\bfx_u^{L-\ell}}\right\|^2 \left\|\Jkskew\right\|.
\end{equation}
\end{boxedproof}
Again, this indicates that even the discretized version of H-DGN enables for effective propagation and conservative message passing.

\begin{algorithm}[h]
  \caption{PH-DGN node embeddings computation}%
  \label{alg:phdgn}
  \SetAlgoLined
  \DontPrintSemicolon
  \KwIn{%
    A static graph $\mathcal{G}=(\mathcal{V}, \mathcal{E}, \bfX, \bfE)$, Symplectic Euler step size $\epsilon$, the number of discretization steps $L$.
  }
  \KwResult{%
    Final nodes' embeddings $\bfX^L$.
  }
  \SetCommentSty{small}
  \SetKwComment{Comment}{$\triangleright$ }{}
  \BlankLine
  \For(\Comment*[f]{Iterate the Symplectic Euler's method}){$\ell\in\{1, \dots,L\}$}{
    \For(\Comment*[f]{Iterate over the graph}){$u\in\mathcal{V}$}{
        $(\bfp_u^{\ell-1}, \bfq_u^{\ell-1}) \gets \bfx_u^{\ell-1}$\\
        $z_p \gets \nabla_{\bfq_u} H_{\mathcal{G}}(\bfp^{\ell-1},\bfq^{\ell-1})$\\
        \If(){apply dampening}{
        $z_p \gets D_u(\bfq^{\ell-1})z_p$
        \Comment*[f]{PH-DGN used to its maximum potential}
        }
        \If(){apply external force}{
        $z_p \gets z_p + F(\bfq_u^{\ell-1}, t)$
        \Comment*[f]{PH-DGN used to its maximum potential}
        }
        $\bfp_u^\ell \gets \bfp_u^{\ell-1} - \epsilon(z_p)$
        \Comment*[f]{Update $\bfp_u$ as in Eq.~\ref{eq:discretized_local_hamil_p} or Eq.~\ref{eq:discretized_port_ham}}\\
        $z_q \gets \nabla_{\bfp_u} H_{\mathcal{G}}(\bfp^{\ell},\bfq^{\ell-1})$\\
        $\bfq_u^\ell \gets \bfq_u^{\ell-1} + \epsilon(z_q)$
        \Comment*[f]{Update $\bfq_u$ as in Eq.~\ref{eq:discretized_local_hamil_q}}\\
        $\bfx_u^{\ell} \gets (\bfp_u^{\ell}, \bfq_u^{\ell})$\\
    }
  }
%
\end{algorithm}

\section{Experiments}\label{sec:phdgn_experiments}
We empirically verify both theoretical claims and practical benefits of our framework on popular graph benchmarks for long-range propagation. First (Section \ref{sec:phdgn_numerical}), we conduct a controlled synthetic test showing non-vanishing gradients even 
when thousands of layers are used. Afterward (Section~\ref{sec:phdgn_graph_transfer_exp}), we run a graph transfer task inspired by \cite{diGiovanniOversquashing} to assess the efficacy in preserving long-range information between nodes. Then, we assess our framework in popular benchmark tasks requiring the exchange of messages at large distances over the graph, including graph property prediction (Section~\ref{sec:phdgn_gpp_exp}) and the long-range graph benchmark \citep{LRGB} (Section~\ref{sec:phdgn_lrgb_exp}). We compare our performance to state-of-the-art methods, such as MPNN-based models, DE-DGNs (which represent a direct competitor to our method), higher-order DGNs, and graph transformers, as in Section~\ref{sec:SWAN}. We investigate two neighborhood aggregation functions for our H-DGN and PH-DGN, which are the classical GCN aggregation and that proposed in Equation~\ref{eq:simple_aggregation}. 
We report in Table~\ref{tab:phdgn_hyperparams} (Appendix~\ref{app:phdgn_hyperparams}) the grid of hyperparameters employed in our experiments.
Our experimental results were obtained using NVIDIA A100 GPUs and Intel Xeon Gold 5120 CPUs.

\subsection{Numerical Simulations}\label{sec:phdgn_numerical}
\myparagraph{Setup}
We empirically verify that our theoretical considerations on H-DGN hold true by an experiment requiring to propagate information within a Carbon-60 molecule graph without training on any specific task, \ie we perform no gradient update step. While doing so, we measure the energy level captured in $H_{\mathcal{G}}(\bfy(\ell\epsilon))$ in the forward pass and the sensitivity, $\|\partial{\bfx_u^{L}}/\partial{\bfx_u^{\ell}}\|$, from each intermediate layer $\ell=1,\dots,L$ in the backward pass. 
We consider the 2-d position of the atom in the molecule as the input node features, fixed terminal propagation time $T=10$ with various integration step sizes $\epsilon \in \{0.1, 0.01, 0.001\}$ and $T=300$ with $\epsilon=0.3$. Note that the corresponding number of layers is computed as $L=T/\epsilon$, \ie we use tens to thousands of layers. For the ease of the simulation, we use $\tanh$-nonlinearity, fixed learnable weights that are randomly initialized, and the aggregation function in Equation~\ref{eq:simple_aggregation}. 

\begin{figure}[h]
    \centering
    \begin{subfigure}[b]{0.3\textwidth}
        \centering
        \includegraphics[width=\textwidth]{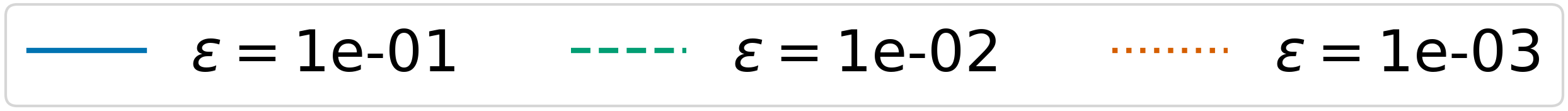}
        \includegraphics[width=\textwidth,height=3.3cm]{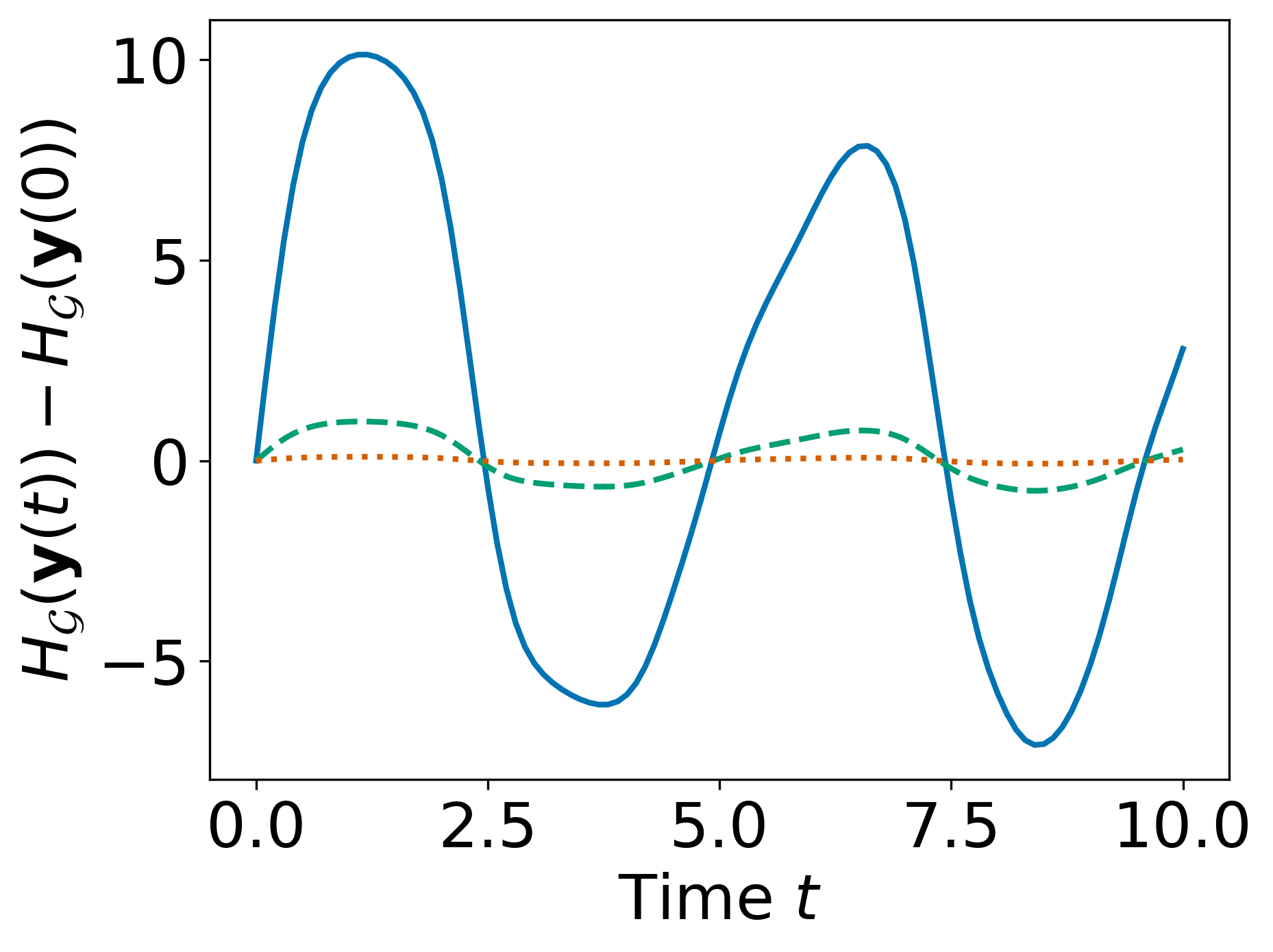}
        \caption{}\label{subfig:energy}
    \end{subfigure}%
    \hspace*{0.5em}
    \begin{subfigure}[b]{0.3\textwidth}
        \centering
        \includegraphics[width=\textwidth,height=3.3cm]{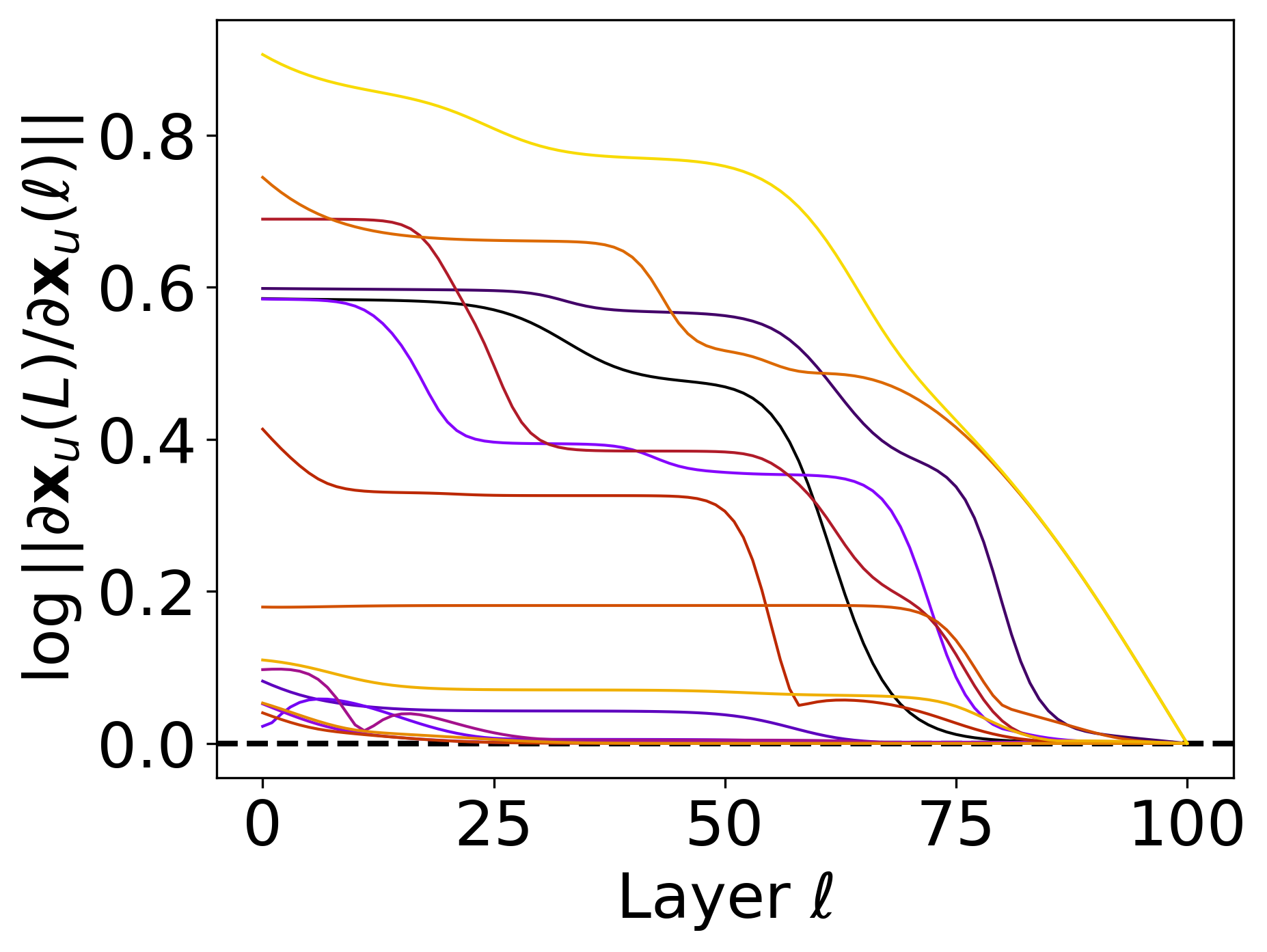}
        \caption{}\label{subfig:norms_a}
    \end{subfigure}%
    \hspace*{0.5em}
    \begin{subfigure}[b]{0.345\textwidth}
        \centering
        \includegraphics[width=\textwidth,height=3.3cm]{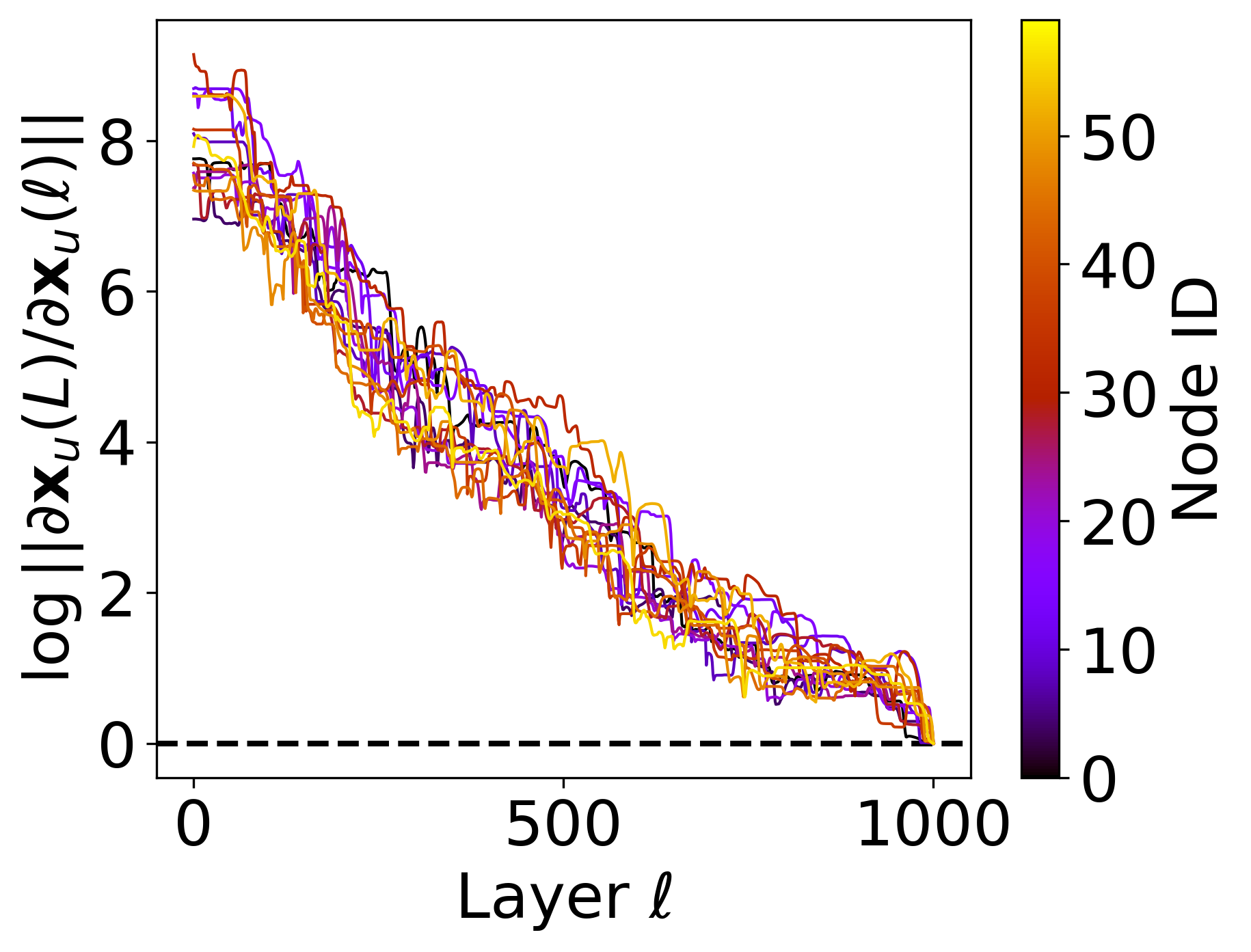}
        \caption{}\label{subfig:norms_b}
    \end{subfigure}%
    \caption{(a) Time evolution of the energy difference to the initial state $\mathbf{y}(0) =\mathbf{y}^{0}$ obtained from one forward pass of H-DGN with fixed random weights on the Carbon-60 graph with three different numbers of layers given by $T/\epsilon$. The sensitivity $\|{\partial\xk^{L}}/{\partial\xk^{\ell}}\|$ of 15 different node states to their final embedding obtained by backpropagation on the Carbon-60 graph after (b) $T=10$ and $\epsilon=0.1$ (\ie 100 layers) and (c) $T=300$ and $\epsilon=0.3$ (\ie 1000 layers). The log scale's horizontal line at $0$ indicates the theoretical lower bound.}\label{fig:grad_norms}
\end{figure}

\myparagraph{Results}
In Figure~\ref{subfig:energy}, we show the energy difference $H_{\mathcal{G}}(\bfy(\ell\epsilon))-H_{\mathcal{G}}(\bfy(0))$ for different step sizes. For a fixed time $T$, a smaller step size $\epsilon$ is related to a higher number of stacked layers.  
We note that the energy difference oscillates around zero, and the smaller the step size the more accurately the energy is preserved. This supports our intuition of H-DGN being a discretization of a divergence-free continuous Hamiltonian dynamic, that allows for non-dissipative forward propagation, as stated in Theorem~\ref{th:im_eig} and Theorem~\ref{th:divergence}. Even for larger step sizes, energy is neither gained nor lost.

Regarding the backward pass, Figures~\ref{subfig:norms_a},~\ref{subfig:norms_b} assert that the lower bound $\|\partial{\bfx(L)}/\partial{\bfx(\ell)}\|\ge 1$ stated in Theorem~\ref{th:lowerb} and its discrete version in Theorem~\ref{th:lowerb_discrete} leads to non-vanishing gradients. In particular, Figure~\ref{subfig:norms_b} shows a logarithmic-linear increase of sensitivity with respect to the distance to the final layer, hinting at the exponential upper bound derived in Theorem~\ref{th:upperb}. This growing  
behavior can be controlled by regularizing the weight matrices, or by use of normalized aggregation functions, as in GCN~\citep{GCN}.

\subsection{Graph Transfer}\label{sec:phdgn_graph_transfer_exp}
\myparagraph{Setup}
We address the task of propagating a label from a source node to a target node located at increasing distances $k$ in the graph as introduced in Section~\ref{sec:exp_transfer}.
Given the conservative nature of the task, we focus on assessing the purely Hamiltonian H-DGN model.

\begin{figure}[h]
    \begin{adjustbox}{center}
        \begin{subfigure}{0.45\textwidth}
            \centering \includegraphics[width=\linewidth]{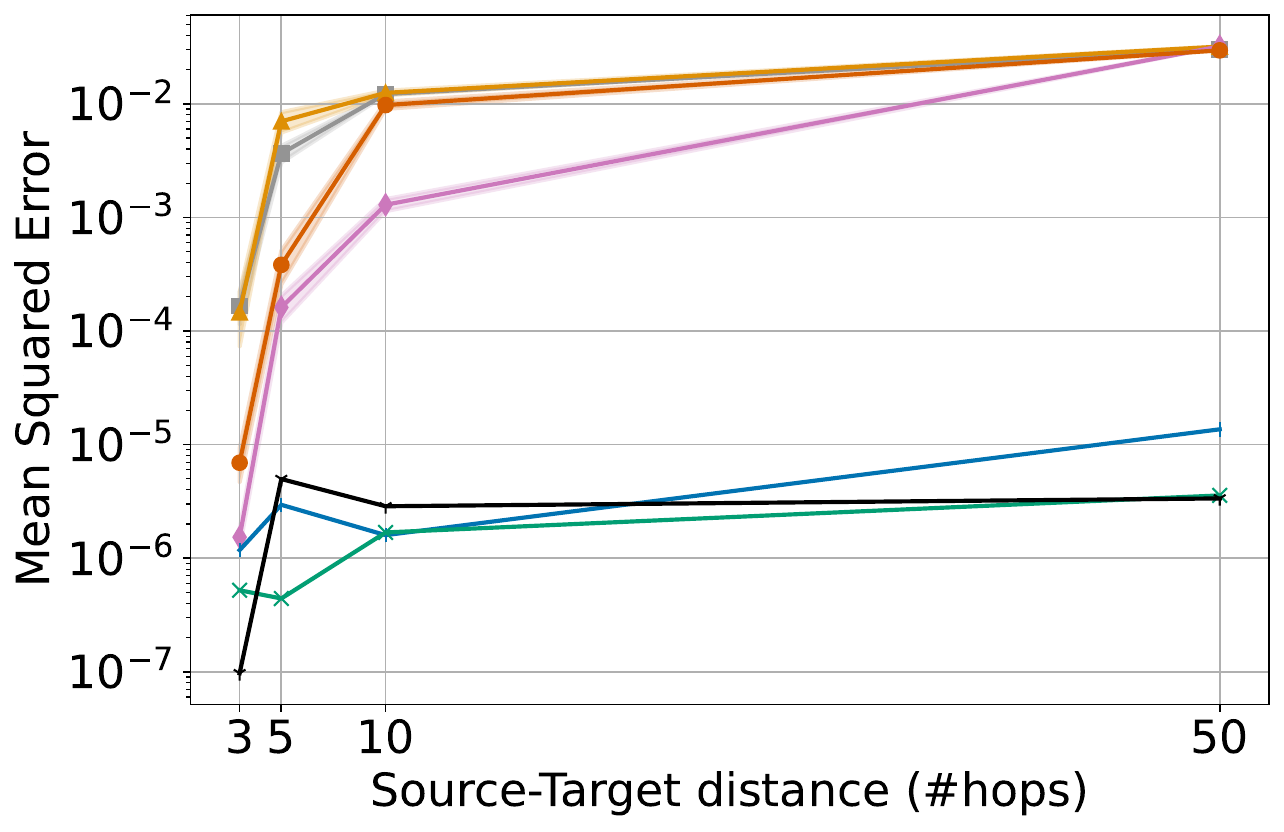}
            \caption{\small Line}
        \end{subfigure}\hspace{5mm}
        \begin{subfigure}{0.45\textwidth}
            \centering \includegraphics[width=\linewidth]{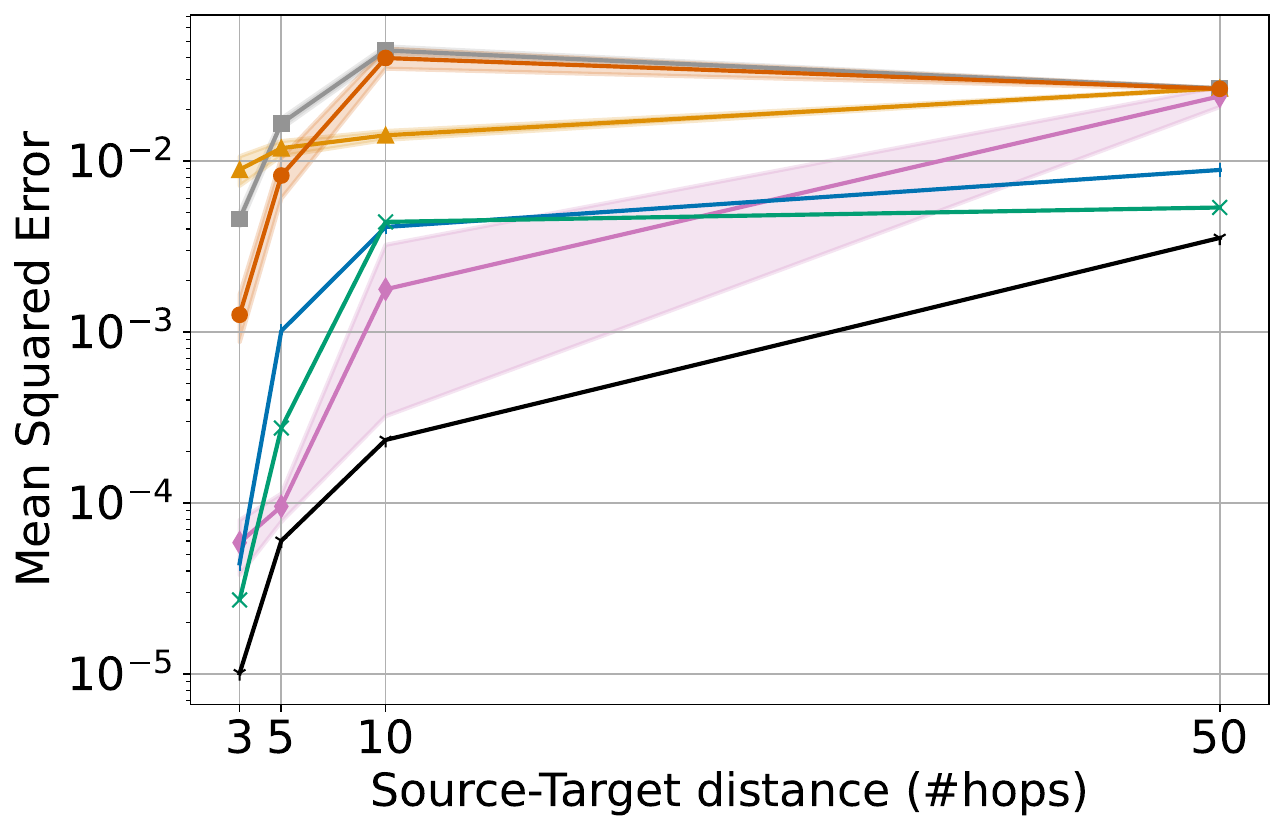}
            \caption{\small Ring}
        \end{subfigure}
    \end{adjustbox}\vspace{2mm}
    \begin{adjustbox}{center}
        \begin{subfigure}{\textwidth}
            \centering \includegraphics[width=0.45\linewidth]{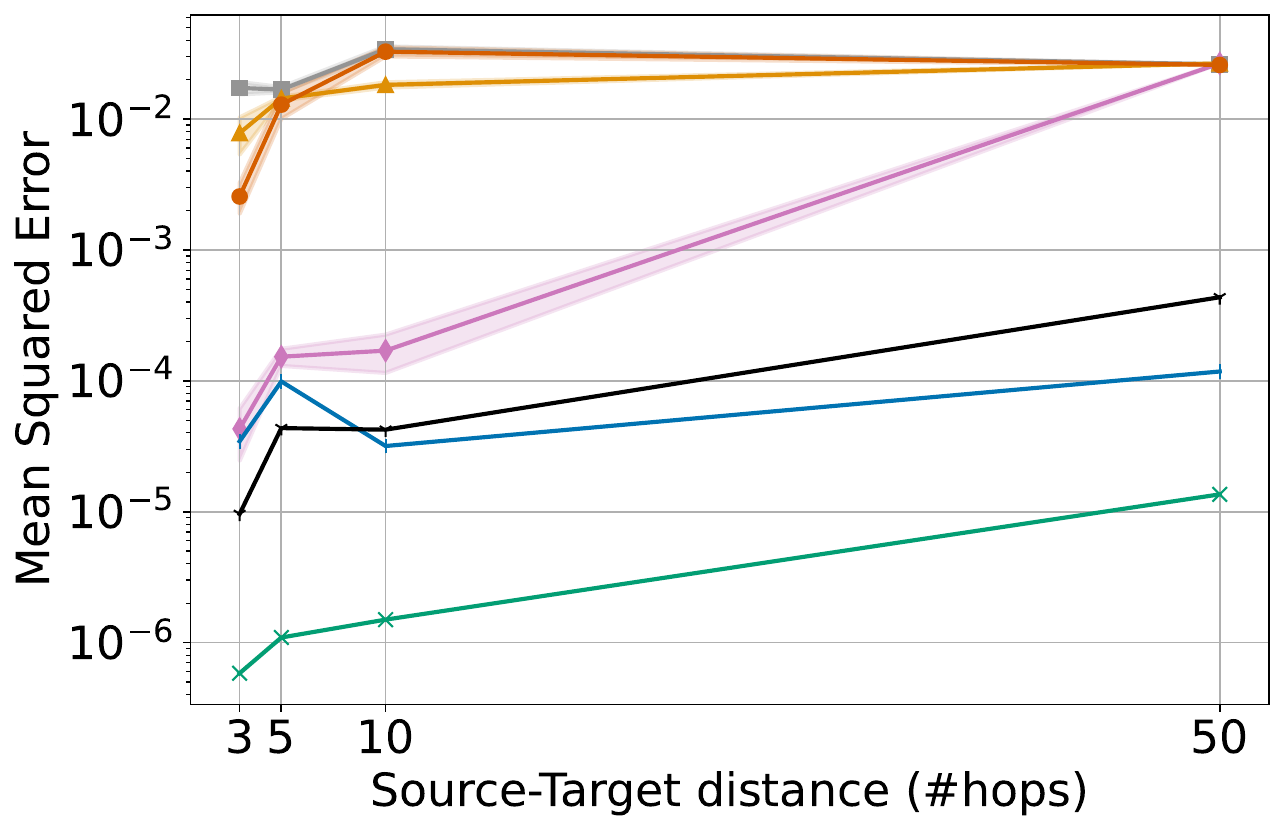}
            \caption{\small Crossed-Ring }
        \end{subfigure}
    \end{adjustbox}\vspace{2mm}
    \begin{subfigure}{1\textwidth}
        \begin{adjustbox}{center}
            \centering \includegraphics[width=1.18\linewidth]{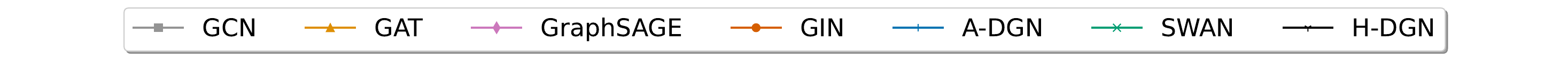}
        \end{adjustbox}
    \end{subfigure}
    \caption{Information transfer performance on (a) Line, (b) Ring, and (c) Crossed-Ring graphs. Overall, baseline approaches are not able to transfer the information accurately as distance increase, while non-dissipative methods like A-DGN, SWAN, and our H-DGN achieve low errors.}
    \label{fig:graph_transfer}
\end{figure}

\myparagraph{Results}
Figure~\ref{fig:graph_transfer} reports the test mean-squared error (and std) of H-DGN compared to literature models. It appears that classical MPNNs do not effectively propagate information across long ranges, as their performance decrease when $k$ increases. Differently, H-DGN achieves low errors even at higher distances, \ie $k\geq 10$. The only competitors to our H-DGN are A-DGN and SWAN, which are other non-dissipative methods. Overall, H-DGN outperforms all the classical MPNNs baseline while having on average better performance than A-DGN, thus empirically supporting our claim of long-range capabilities while introducing a new architectural bias. Moreover, our results highlight how our framework can push simple graph convolutional architectures to
state-of-the-art performance when imbuing them with dynamics capable of long-range message exchange.

\subsection{Graph Property Prediction}\label{sec:phdgn_gpp_exp}
\myparagraph{Setup}
We address tasks involving the prediction of three graph properties -  Diameter, Single-Source Shortest Paths (SSSP), and node Eccentricity on synthetic graphs as introduced in Section~\ref{sec:adgn_graph_prop_pred}. 
In this experiment, we investigate the performance of our complete port-Hamiltonian framework, PH-DGN, and present the pure Hamiltonian H-DGN as an ablation study. 

\myparagraph{Results}
We present the results on the graph property prediction tasks in Table~\ref{tab:phdgn_graphproppred}, reporting {\small$log_{10}(\text{MSE})$} as evaluation metric. 
We observe that both our H-DGN and PH-DGN show a strong improvement with respect to baseline methods, achieving new state-of-the-art performance on all the tasks.
Indeed, our ablation reveals that the purely conservative H-DGN model has, on average, a {\small$log_{10}(\text{MSE})$} that is 0.33 lower than the best baseline. 
Such gap is pushed to 0.81 when the full port-Hamiltonian bias (\ie PH-DGN) is employed, marking a significant decrease in the test loss. The largest gap is achieved by PH-DGN in the eccentricity task, where it improves the {\small$log_{10}(\text{MSE})$} performance of the best baseline by 1.36. Moreover, PH-DGN improves SWAN's performance by 0.52, on average.

\begin{table}[h]
\centering

\caption{Mean test $log_{10}(\text{MSE})$ and std average over 4 training seeds on the Graph Property Prediction. Our methods and DE-DGN baselines are implemented with \textbf{weight sharing}.
The \one{first}, \two{second}, and \three{third} best scores are colored.}
\label{tab:phdgn_graphproppred}
\scriptsize
\begin{tabular}{lccc}
\toprule
\textbf{Model} & \textbf{Diameter} & \textbf{SSSP} & \textbf{Eccentricity} \\
\midrule
\textbf{MPNNs} &&&\\
$\,$ GCN          & 0.7424$_{\pm0.0466}$ & 0.9499$_{\pm9.2 \cdot10^{-5}}$ & 0.8468$_{\pm0.0028}$\\
$\,$ GAT          & 0.8221$_{\pm0.0752}$ & 0.6951$_{\pm0.1499}$ & 0.7909$_{\pm0.0222}$\\
$\,$ GraphSAGE    & 0.8645$_{\pm0.0401}$ & 0.2863$_{\pm0.1843}$ & 0.7863$_{\pm0.0207}$\\
$\,$ GIN          & 0.6131$_{\pm0.0990}$ & -0.5408$_{\pm0.4193}$ & 0.9504$_{\pm0.0007}$\\
$\,$ GCNII        & 0.5287$_{\pm0.0570}$ & -1.1329$_{\pm0.0135}$ & 0.7640$_{\pm0.0355}$\\\midrule
\textbf{DE-DGNs} &&&\\
$\,$ DGC          &  0.6028$_{\pm0.0050}$ & -0.1483$_{\pm0.0231}$ & 0.8261$_{\pm0.0032}$\\
$\,$ GraphCON     &  0.0964$_{\pm0.0620}$ & -1.3836$_{\pm0.0092}$ & 0.6833$_{\pm0.0074}$\\
$\,$ GRAND        &  0.6715$_{\pm0.0490}$ & -0.0942$_{\pm0.3897}$ & 0.6602$_{\pm0.1393}$ \\\midrule
\textbf{Ours} &&&\\
$\,$ A-DGN & -0.5188$_{\pm0.1812}$ & \three{-3.2417$_{\pm0.0751}$} & 0.4296$_{\pm0.1003}$  \\
$\,$ SWAN & -0.5249$_{\pm0.0155}$ &  -3.2370$_{\pm0.0834}$ & 0.4094$_{\pm0.0764}$ \\

$\,$ SWAN-\textsc{learn} & \one{-0.5981$_{\pm0.1145}$}  & \two{-3.5425$_{\pm0.0830}$}  & \three{-0.0739$_{\pm0.2190}$} \\

$\,$ H-DGN   & \two{-0.5473$_{\pm0.1074}$} & -3.0467$_{\pm0.1615}$ & \two{-0.7248$_{\pm0.1068}$}\\
$\,$ PH-DGN  & \three{-0.5385$_{\pm0.0187}$} &  \one{-4.2993$_{\pm0.0721 }$} & \one{-0.9348$_{\pm0.2097}$}\\
\bottomrule
\end{tabular}
\end{table}

Although our purely conservative H-DGN shows improved performance with respect to all baselines, it appears that relaxing such bias via PH-DGN is more beneficial overall,
leading to even greater improvements in long-range information propagation. 
Our intuition is that such tasks do not require purely conservative behavior since nodes need to count distances while exchanging
more messages with other nodes, similar to standard algorithmic solutions such as \cite{Dijkstra}. Therefore, the energy may not be constant during the resolution of the task, hence benefiting from the non-purely conservative behavior of PH-DGN.

As for the graph transfer task, our results demonstrate that our PH-DGNs can effectively learn and exploit long-range information while pushing simple graph neural architectures to state-of-the-art performance when modeling dynamics capable of long-range propagation.

In Table~\ref{tab:phdgn_results_GraphProp_time} we report runtimes of both our H-DGN and PH-DGN as well as baseline methods on the graph property prediction task as in Section~\ref{sec:adgn_graph_benchmark}. Our results shows that both H-DGN and PH-DGN have improved or comparable runtimes compared to MPNNs. Notably, H-DGN is on average 5.92 seconds faster than GAT and 5.19 seconds faster than GCN. Compared to DE-DGN baselines, our methods show longer execution times, which are inherently caused by the sequential computation of both $\bfp$ and $\bfq$ explicited in Section~\ref{sec:phdgn_discretization} and non-conservative components (detailed in Appendix~\ref{app:phdgn_hyperparams}). 

\begin{table}[ht]
\centering
\caption{Average time per epoch (measured in seconds) and std, averaged over 4 random weight initializations. Each time is obtained by employing 20 layers and an embedding dimension equal to 30. Our methods and DE-DGN baselines are implemented with \textbf{weight sharing}. The evaluation was carried out on an AMD EPYC 7543 CPU @ 2.80GHz. \one{First}, \two{second}, and \three{third} best results. \label{tab:phdgn_results_GraphProp_time}}
\scriptsize
\begin{tabular}{lccc}
\toprule
\textbf{Model} &\textbf{Diameter} & \textbf{SSSP} & \textbf{Eccentricity} \\\midrule
\textbf{MPNNs} \\
$\,$ GCN &   32.45$_{\pm2.54}$ &  17.44$_{\pm3.85}$ & 11.78$_{\pm2.43}$\\
$\,$ GAT &  20.20$_{\pm5.18}$ & 26.41$_{\pm8.34}$ & 17.28$_{\pm1.92}$\\
$\,$ GraphSAGE & 13.12$_{\pm2.99}$ & 13.12$_{\pm2.99}$ & \three{8.20$_{\pm0.75}$}\\
$\,$ GIN & \one{6.63$_{\pm0.28}$}  &21.16$_{\pm2.33}$ & 14.22$_{\pm3.17}$\\
$\,$ GCNII & 13.13$_{\pm6.85}$ & 14.96$_{\pm7.17}$ & 15.70$_{\pm3.92}$\\

\midrule
\textbf{DE-GNNs} \\
$\,$ DGC & 8.97$_{\pm9.07}$ & 12.54$_{\pm1.62}$ & \two{7.21$_{\pm11.10}$}\\
$\,$ GRAND & 133.84$_{\pm42.57}$ & 109.15$_{\pm27.49}$ & 202.46$_{\pm85.01}$\\
$\,$ GraphCON & 9.26$_{\pm0.47}$ & 7.76$_{\pm0.05}$ & 7.80$_{\pm0.05}$ \\


\midrule
\multicolumn{4}{l}{\textbf{Ours}}\\
$\,$ A-DGN
& \three{8.42$_{\pm2.71}$} & \two{7.86$_{\pm2.11}$} & 13.18$_{\pm9.07}$\\
$\,$ H-DGN & 15.49$_{\pm0.05}$ & 15.28$_{\pm0.02}$ & 15.34$_{\pm0.04}$\\
$\,$ PH-DGN & 17.18$_{\pm0.04}$ & 17.12$_{\pm0.07}$ & 17.13$_{\pm0.06}$\\

\bottomrule

\end{tabular}
\end{table}

\subsection{Long-Range Graph Benchmark}\label{sec:phdgn_lrgb_exp}
\myparagraph{Setup}
We address the LRGB benchmark as introduced in Section~\ref{sec:exp_lrb}, focusing on the Peptide-func and Peptide-struct tasks. As in Section~\ref{sec:phdgn_gpp_exp}, we decouple our method into H-DGN and PH-DGN to provide an ablation study on the strictly conservative behavior. Acknowledging the results from \cite{tonshoff2023where}, we also report results with a 3-layer MLP readout.

\myparagraph{Results}
We report results on the LRGB tasks in Table~\ref{tab:phdgn_long_range_complete}. Our results show that both H-DGN and PH-DGN outperform classical MPNNs, graph transformers, most of the multi-hop DGNs, and recent DE-DGNs (which represent a direct competitor to our method). 
Overall, our (port-)Hamiltonian methods show great benefit in propagating long-range information without requiring additional strategies such as global position encoding, global attention mechanisms, or rewiring techniques that increase the overall complexity of the method. 
Consequently, our results reaffirm the effectiveness of our (port-)Hamiltonian framework in enabling efficient long-range propagation, even in simple DGNs characterized by purely local message exchanges.

\begin{table}[h]

\centering
\caption{Results for Peptides-func and Peptides-struct averaged over 3 training seeds. Baseline results are taken from \cite{LRGB} and \cite{drew}. Re-evaluated methods employ the 3-layer MLP readout proposed in \cite{tonshoff2023where}. Our methods and DE-DGN baselines are implemented with \textbf{weight sharing}. Note that all MPNN-based methods include structural and positional encoding.
The \one{first}, \two{second}, and \three{third} best scores are colored. 
}\label{tab:phdgn_long_range_complete}
\scriptsize
\begin{tabular}{@{}lcc@{}}
\toprule
\multirow{2}{*}{\textbf{Model}} & \textbf{Peptides-func}  & \textbf{Peptides-struct}              
\\
& \scriptsize{AP $\uparrow$} & \scriptsize{MAE $\downarrow$} 
\\ \midrule  
\textbf{MPNNs} \\
$\,$ GCN                        & 0.5930$_{\pm0.0023}$ & 0.3496$_{\pm0.0013}$ \\ 
$\,$ GINE                       & 0.5498$_{\pm0.0079}$ & 0.3547$_{\pm0.0045}$ \\ 
$\,$ GCNII                      & 0.5543$_{\pm0.0078}$ & 0.3471$_{\pm0.0010}$ \\ 
$\,$ GatedGCN                   & 0.5864$_{\pm0.0077}$ & 0.3420$_{\pm0.0013}$ \\ 

\midrule
\textbf{Multi-hop DGNs}\\
$\,$ DIGL+MPNN           & 0.6469$_{\pm0.0019}$         & 0.3173$_{\pm0.0007}$ \\ 
$\,$ DIGL+MPNN+LapPE     & 0.6830$_{\pm0.0026}$         & 0.2616$_{\pm0.0018}$ \\ 
$\,$ MixHop-GCN          & 0.6592$_{\pm0.0036}$         & 0.2921$_{\pm0.0023}$ \\ 
$\,$ MixHop-GCN+LapPE    & 0.6843$_{\pm0.0049}$         & 0.2614$_{\pm0.0023}$ \\ 
$\,$ DRew-GCN            & 0.6996$_{\pm0.0076}$         & 0.2781$_{\pm0.0028}$ \\ 
$\,$ DRew-GCN+LapPE             & \one{0.7150$_{\pm0.0044}$}   & 0.2536$_{\pm0.0015}$ \\ 
$\,$ DRew-GIN            & 0.6940$_{\pm0.0074}$         & 0.2799$_{\pm0.0016}$ \\ 
$\,$ DRew-GIN+LapPE      & \two{0.7126$_{\pm0.0045}$}   & 0.2606$_{\pm0.0014}$ \\ 
$\,$ DRew-GatedGCN       & 0.6733$_{\pm0.0094}$         & 0.2699$_{\pm0.0018}$ \\ 
$\,$ DRew-GatedGCN+LapPE & 0.6977$_{\pm0.0026}$         & 0.2539$_{\pm0.0007}$ \\ 

\midrule
\textbf{Transformers} \\
$\,$ Transformer+LapPE & 0.6326$_{\pm0.0126}$ & 0.2529$_{\pm0.0016}$           \\ 
$\,$ SAN+LapPE         & 0.6384$_{\pm0.0121}$ & 0.2683$_{\pm0.0043}$           \\ 
$\,$ GraphGPS+LapPE    & 0.6535$_{\pm0.0041}$ & 0.2500$_{\pm0.0005}$   \\ 
\midrule
\textbf{DE-DGNs} \\
$\,$ GRAND      & 0.5789$_{\pm0.0062}$ & 0.3418$_{\pm0.0015}$ \\ 
$\,$ GraphCON   & 0.6022$_{\pm0.0068}$ & 0.2778$_{\pm0.0018}$ \\ 
\midrule
\textbf{Re-evaluated}\\
$\,$ GCN
& 0.6860$_{\pm0.0050}$ & \one{0.2460$_{\pm0.0007}$}\\
$\,$ GINE
& 0.6621$_{\pm0.0067}$ & 0.2473$_{\pm0.0017}$\\
$\,$ GatedGCN
& 0.6765$_{\pm0.0047}$ & 0.2477$_{\pm0.0009}$\\
$\,$ DRew-GCN+LapPE
& 0.6945$_{\pm0.0021}$        & 0.2517$_{\pm0.0011}$\\
$\,$ GraphGPS+LapPE
& 0.6534$_{\pm0.0091}$ & 0.2509$_{\pm0.0014}$\\
\midrule
\textbf{Ours} \\ 
$\,$ A-DGN      & 0.5975$_{\pm0.0044}$ & 0.2874$_{\pm0.0021}$ \\
$\,$ SWAN                & 0.6313$_{\pm0.0046}$ & 0.2571$_{\pm0.0018}$       \\
$\,$ SWAN-\textsc{learn} & 0.6751$_{\pm0.0039}$ & \three{0.2485$_{\pm0.0009}$} \\

$\,$ H-DGN
& 0.6961$_{\pm0.0070}$         & 0.2581$_{\pm0.0020}$\\
$\,$ PH-DGN
& \three{0.7012$_{\pm0.0045}$} & \two{0.2465$_{\pm0.0020}$}\\
\bottomrule
\end{tabular}
\end{table}

\section{Related Work}
Recent advancements in the field of representation learning have introduced new architectures that establish a connection between neural networks and dynamical systems, as observed in Section~\ref{sec:ADGN_mpnn_comparison}.
Indeed, works like GDE~\citep{GDE}, GRAND~\citep{GRAND}, PDE-GCN~\citep{pdegcn}, DGC~\citep{DGC}, and GRAND++~\citep{grand++} propose to interpret DGNs as discretization of ODEs and PDEs. The conjoint of dynamical systems and DGNs have found favorable consensus, as these new methods exploit the intrinsic properties of differential equations to extend the characteristic of message passing within DGNs, as introduced in Section~\ref{sec:rel_work_adgn}. GRAND, GRAND++, and DGC bias the node representation trajectories to follow the heat diffusion process, thus performing a gradual \emph{smoothing} of the initial node states. On the contrary, GraphCON~\citep{graphcon} used oscillatory properties to enable linear dynamics that preserve the Dirichlet energy encoded in the node features; PDE-GCN${_{\text{M}}}$~\citep{pdegcn} uses an interpolation between anisotropic diffusion and conservative oscillatory properties; and HamGNN \citep{Ham_ICML} leverages Hamiltonian dynamics to encode node input features, which are then fed into classical DGNs to enhance their conservative properties. 
Similarly to HamGNN, HANG~\citep{HANG_Neurips} leverages Hamiltonian dynamics to improve robustness to adversarial attacks to the graph structure. Differently from HamGNN and HANG, our PH-DGN is the first port-Hamiltonian framework, which provides theoretical guarantees of both conservative and non-conservative behaviors while seamlessly incorporating the most suitable aggregation function for the task at hand, enabling long-range information propagation.

A-DGN (Section~\ref{sec:ADGN}) and SWAN (Section~\ref{sec:SWAN}) introduces antisymmetric constraint mechanisms that lead to non-dissipative dynamics. Differently from such approaches, PH-DGN designs the information flow within a (static) graph as a solution of a port-Hamiltonian system, thus allowing for non-dissipative propagation without relying on such architectural constraints.

As observed in Sections~\ref{sec:static_dgn_fundamentals} and \ref{sec:rel_work_adgn}, the graph representation learning community focused on graph rewiring and transformer-based method to effectively transfer information across distant nodes. Despite the success of these techniques in addressing oversquashing, a potential drawback is the increased complexity associated with propagating information at each update, often linked to denser graph shift operators. Similar to A-DGN and SWAN, our PH-DGN allows effective long-range propagation without densifying the original graph.
\section{Summary}
In this chapter, we have presented \textit{(port-)Hamiltonian Deep Graph Network} (PH-DGN), a general framework that gauges the equilibrium between non-dissipative long-range propagation and non-conservative behavior while seamlessly incorporating the most suitable neighborhood aggregation function.  We theoretically prove that, when pure Hamiltonian dynamic is employed, both the continuous and discretized versions of our framework allow for long-range propagation in the message passing flow since node states retain their past. To demonstrate the benefits of including (port-)Hamiltonian dynamics in DE-DGNs, we conducted several experiments on synthetic and real-world benchmarks requiring long-range interaction. Our results show that our method outperforms state-of-the-art models and that the inclusion of data-driven forces that deviate from a purely conservative behavior is often key to maximize efficacy of the approach on tasks requiring long-range propagation. 
Indeed, in practice, effective information propagation requires a balance between long-term memorization and propagation and the ability to selectively discard and forget information when necessary.

\part{ Space and Time Propagation for Dynamic Graphs}
\chapter{Learning irregularly-sampled D-TDGs}\label{ch:tgode}
Graph-based processing methods turned out to be extremely effective in processing the spatio-temporal evolution of dynamic graphs, as shown in Chapter~\ref{ch:learning_dyn_graphs}. 
However, real-world complex problems described as D-TDGs call for novel methods that can move beyond the common assumptions found in most of the methods proposed until now. Indeed, modern graph representation learning for D-TDGs works mostly under the assumption of dealing with \emph{regularly sampled} temporal graph snapshots, which is far from realistic. Such problems require dealing with mutable relational information, \textbf{irregularly} and \textbf{severely under-sampled} data. As an example, social networks and physical systems are characterized by continuous dynamics and sporadic observations. Indeed, the use of strategies that involve recording only changes of state are necessary to conserve sensor battery life or reduce storage needs, inherently producing sporadic data. For instance, temperature sensors in environmental monitoring may log data only when significant changes occur, even though the environment is constantly changing. Similarly, sensor failures can lead to intermittent data collection and inconsistencies.

As discussed in Section~\ref{sec:neural_de}, recent works propose to model input-output data relations as a continuous dynamic described by a learnable differential equation. Notably, relying on differential equations has shown promising for modeling complex temporal patterns from irregularly and sparsely sampled data \citep{neuralODE,ode-irregular,neuralCDE}.

Inspired by such findings, in this chapter we introduce \emph{Temporal Graph Ordinary Differential Equation} (\gls*{TG-ODE})\index{TG-ODE}, a general continuous-time modeling framework for D-TDGs, which learns both the temporal and spatial dynamics from graph streams where the intervals between observations are not regularly spaced. 
TG-ODE is designed through the lens of ODEs for effective learning of irregularly sampled D-TDGs, and its predictions are trajectories obtained by numerical integration of the learned ODE. 

The key contributions of this chapter can be summarized as follows: 
\begin{itemize}
    \item we introduce TG-ODE, a general modeling framework suited for handling irregularly sampled D-TDGs; 
    \item we introduce new benchmarks of synthetic and real-world scenarios for evaluating forecasting models on irregularly sampled D-TDGs;
    \item we conduct extensive experiments to demonstrate the benefits of our approach and show that TG-ODE outperforms state-of-the-art DGNs on all benchmarks.
\end{itemize}

Finally, we stress that, other than the outstanding empirical performance achieved by even simple TG-ODE instances, the framework allows us to reinterpret many state-of-the-art DGNs as a discretized solution of an ODE, thus facilitating their extension to handle graph streams with irregular sampling.
\\

This chapter has been developed during a three months visiting period at the Swiss AI Lab IDSIA (Istituto Dalle Molle di Studi sull'Intelligenza Artificiale) in Lugano, Switzerland. We base this chapter on \cite{gravina_tgode}. 

\section{Temporal Graph ODE}\label{sec:tgode_method}
We consider a dynamical system of interacting entities $u\in\mathcal{V}$ that is described by a Cauchy problem defined on an ODE of the form
\begin{equation}\label{eq:sys-model}
\frac{d\bfX(t)}{dt} = F(\bfX(t),  \bfE(t), \bfz(t)),
\end{equation}
with initial condition $\bfX(0)=\bfX_0$.
$\bfX(t)$ collects the node-level states (\ie $\bfx_u(t)$) associated with each node $u\in \mathcal{V}(t)$ and $\bfE(t)$ edge-level attributes (\ie $\bfe_{vu}(t)$), as described in Section~\ref{sec:dynamic_graph_notation}. The node set $\mathcal{V}(t)$ and the edge set $\mathcal{E}(t)$ are allowed to vary over time. 
The system can also be driven by vector $\bfz_u(t)\in\mathbb{R}^c$ accounting for exogenous variables relevant to the problem at hand, such as weather conditions, hour of the day, or day of the week.

Accordingly, for all $u\in\mathcal{V}(t)$, we write
\begin{equation}\label{eq:sys-model-node}
    \frac{d \bfx_u(t)}{d t} = F\left(\bfx_u(t), \bfz_u(t), \{\bfx_v(t)\}_{v\in\mathcal{N}_u(t)}, \{\bfe_{vu}(t)\}_{v\in\mathcal{N}_u(t)}\right),
\end{equation}
to emphasize the local dependencies of node state $\bfx_u(t)$ at a time $t$ from its neighboring nodes $v \in \mathcal N_u(t)$ at the corresponding time.

We express any solution of ODE in Equation~\ref{eq:sys-model} as the dynamic graph 
\begin{equation}
\mathcal G(t) = \left(\mathcal{V}(t), \mathcal{E}(t), \bfX(t), \bfE(t)\right)
\end{equation}
defined for $t\ge0$. 
However, we assume to observe the system in Equation~\ref{eq:sys-model} only as a (discrete) sequence of snapshot graphs 
\begin{equation}\label{eq:graph-seq}
\mathcal G=\{\mathcal G_{t_i}:i=0,1,2,\dots,T\}
\end{equation}
that arrive at irregular timestamps, i.e., the sampling is not uniform and, in general,  $t_i - t_{i-1} \neq t_{i+1} - t_i$.
Each snapshot $\mathcal{G}_t = (\mathcal{V}_t, \mathcal{E}_t, \overline{\bfX}_t, \bfE_t)$ corresponds to an observation of the system state at a specific timestamp $t\in\mathbb{R}$. 

In this section, we address the problem of learning a model of the differential equation underlying the observed data, which is subsequently exploited to provide estimates of unobserved system's node states and make forecasts.  
To ease readability, in the following of this chapter, we drop the time variable $t$.

To learn the function $F$ in Equation~\ref{eq:sys-model-node}, we consider a family of models 
\begin{equation}\label{eq:pred-model}
    f_\theta(\bfx_u,\bfz, \{\bfx_v\}_{v\in\mathcal N_u}, \{\mathbf{e}_{vu}\}_{v\in\mathcal N_u})
\end{equation}
parameterized by vector $\theta$, and optimized so that the solution $\hat{\bfx}$ of the differential equation
\begin{equation}\label{eq:pred-ode}
   \frac{d\bfx_u}{dt} = f_\theta\left(\bfx_u,\bfz_u, \{\bfx_v\}_{v\in\mathcal N_u}, \{\mathbf{e}_{vu}\}_{v\in\mathcal N_u}\right),\quad \forall u\in\mathcal{V}
\end{equation}
minimizes the discrepancy with the observed sequence of graphs in Equation~\ref{eq:graph-seq}.
We follow the message-passing paradigm introduced in Section~\ref{sec:static_dgn_fundamentals} and instantiate Equation~\ref{eq:pred-ode} as 
\begin{equation}\label{eq:tg-ode}
    \frac{d \bfx_u}{d t} = 
    \rho_U\left(\bfx_u, \bfz_u, \bigoplus_{v\in\mathcal{N}_u}\Bigl(\rho_M(\bfx_u,\bfx_v, \mathbf{e}_{vu})\Bigr)\right),
\end{equation}
with message function $\rho_M$, aggregation operator $\bigoplus$, and update function $\rho_U$ (as usual).
We refer to the above framework in Equation~\ref{eq:tg-ode} as \emph{Temporal Graph Ordinary Differential Equation} (TG-ODE). 

We observe that, since our framework relies on ODEs, it can naturally deal with snapshots that arrive at an arbitrary time. Indeed, the original Cauchy problem can be divided into multiple sub-problems, one per snapshot in the dynamic graph.
Here, the $i$-th sub-problem is defined for all $u\in \mathcal{V}_{t}$ as
\begin{equation}\label{eq:tg-ode-split}
    \begin{dcases}
    \frac{d \bfx_u}{d t} = 
    \rho_U\left(\bfx_u,  \bfz_u, \bigoplus_{v\in\mathcal{N}_u}\Bigl(\rho_M(\bfx_u,\bfx_v, \mathbf{e}_{vu})\Bigr) \right),
    \\
    \bfx_u(0) = \gls*{eta}(\overline{\bfx}^{t_{i-1}}_u, \hat{\bfx}_{u}(t_{i-1})) 
    \end{dcases}
\end{equation}
in the time span between the two consecutive timestamps, \ie $t \in [t_{i-1}, t_{i}]$, 
where $\eta$ is a function that combines the $i$-th observed state of the node $u$ related to the snapshot graph $\mathcal G_{t_{i-1}}$ in Equation~\ref{eq:graph-seq} (\ie $\overline{\bfx}^{t_{i-1}}_u$) and the prediction $\hat{\bfx}_u(t_{i-1})$ obtained by solving Equation~\ref{eq:tg-ode-split} at the previous step. 

When given, we consider the true -- potentially variable -- topology $\mathcal{E}(t)$ to define the neighborhoods for $t\in[t_{i-1},t_i]$, otherwise, we set $\mathcal{E}(t)\equiv\mathcal{E}_{t_{i-1}}$ for every $t$, i.e., equal to the last observed topology associated with $\mathcal G_{t_{i-1}}$.
Accordingly, we optimize $\theta$ in order to minimize the mean of some loss $\mathcal L$,
\begin{equation}
   \frac{1}{T} \sum_{i=1}^T \frac{1}{|\mathcal{V}_{t_i}|} \sum_{u\in\mathcal{V}_{t_i}}  \mathcal{L}(\overline{\bfx}^{t_i}_u, \hat{\bfx}_u(t_i))
\end{equation}
where prediction $\hat{\bfx}_u(t_i)$ at time $t_i$ is obtained by solving Equation~\ref{eq:tg-ode-split}.

\begin{figure}[ht]
    \centering
    \includegraphics[width=\linewidth]{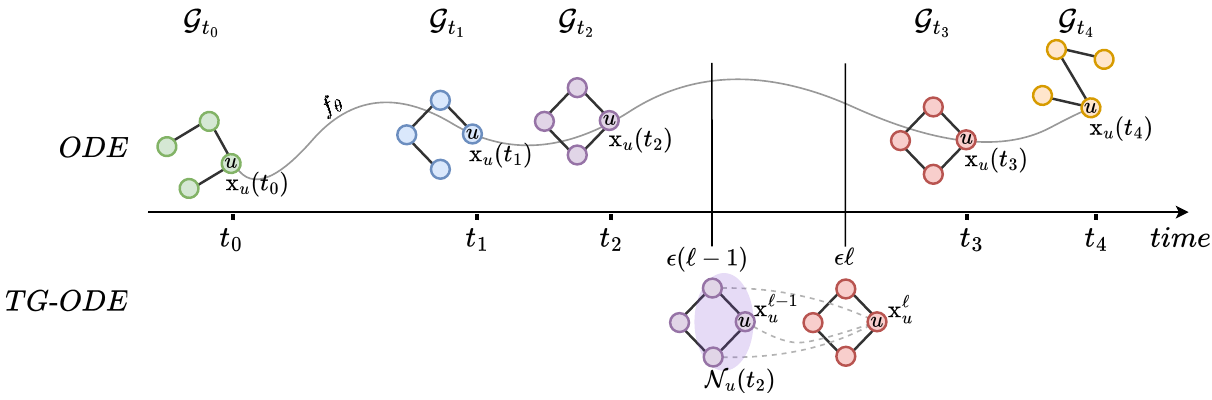}
    \caption{The continuous processing of node $u$'s state in a discrete-time dynamic graph with irregularly-sampled snapshots over a set of 4 nodes and fixed edge set. At the top, the node-wise ODE function $f_\theta$ defines the evolution of the states $\bfx_u(t)$. At the bottom, the discretized solution of the node-wise ODE, which corresponds to our framework TG-ODE. The node embedding $\bfx_u^\ell$ is computed iteratively over a discrete set of points by leveraging the temporal neighborhood and self-representation at the previous step.}
    \label{fig:tg-ode}
\end{figure}

As discussed in Section~\ref{sec:discretization_method_de}, for most ODEs, it is not possible to compute analytical solutions. 
For simplicity, here we employ the \textit{forward Euler's method}, thus the solution is computed through iterative applications of the method over a discrete set of points in the time interval. The solution to Equation~\ref{eq:tg-ode-split} is obtained by the recursion
\begin{equation}\label{eq:tg-ode_discretized}
    \bfx_u^{\ell+1} = \bfx_u^{\ell} +\epsilon \rho_U\left(\bfx_u^{\ell},  \bfz_u(t_{\ell}), \bigoplus_{v\in\mathcal{N}_u(t_{\ell})}\Bigl(\rho_M(\bfx_u^{\ell},\bfx_v^{\ell}, \mathbf{e}_{vu}^{\ell})\Bigr)\right),
\end{equation}
starting from initial condition $\bfx_u^{0}=\bfx_u(0)=\eta(\overline{\bfx}^{t_{i-1}}_u, \hat{\bfx}_{u}(t_{i-1}))$ and is reiterated until $\epsilon\ell\ge t_i-t_{i-1}$.
In Equation~\ref{eq:tg-ode_discretized}, $\epsilon \ll t_{i}-t_{i-i}$ is the step size, while $\ell$ indicates the generic iteration step, and $t_\ell=t_{i-1} + \epsilon\ell$. 
Finally, a solution $\hat{\bfx}(t)$ to Equation~\ref{eq:tg-ode-split} in the interval $[t_{i-1}, t_i]$ is provided by the discretization $\hat{\bfx}_u(t_{i-1}+\epsilon\ell)=\bfx_u^\ell$, for all $\ell$, and interpolated elsewhere. The process is visually summarized at the bottom of Figure~\ref{fig:tg-ode}.

As previously introduced in Chapter~\ref{ch:antisymmetry}, we acknowledge that not all resulting ODEs allow unique solutions and yield numerical stable problems. Generally, numerical stability is associated with the ODE to solve rather than the input data. Thus, a proper design of the ODE prevents stability issues.
Indeed, the solution of a Cauchy problem exists and is unique if the differential equation is uniformly Lipschitz continuous in its input and continuous in $t$, as states in the Picard–Lindel\"{o}f theorem~\citep{picard-theo}. Thus, different implementations of Equation~\ref{eq:tg-ode} should address the continuous behavior of the differential equation. We note that this theorem holds for our model if the underlying neural network has finite weights and uses Lipschitz non-linearities, \eg the tanh.

By Equation~\ref{eq:tg-ode_discretized} and the generality of the message passing in Equation~\ref{eq:tg-ode}, we observe that TG-ODE allows us to cast basically any standard DGN through the lens of an ODE for D-TDGs with irregular timestamps. Secondly, we stress that, even though TG-ODE is solved here by means of the forward Euler's method, other discretization methods can still be utilized. To conclude, our framework can be implemented using the aggregation function that is most suitable for the given task and the discretization method that best fits the computational resources and problem at hand. As a demonstration of this, in Section~\ref{sec:tgode_experiments} we explore the neighborhood aggregation scheme proposed in \cite{tagcn}. Thus, Equation~\ref{eq:tg-ode} can be reformulated as 
\begin{equation}
    \frac{d \bfx_u}{d t} 
    = \sigma\left(
        \sum_{k=0}^K \,\,\bfV_k\sum_{v\in\mathcal{N}_u^k \cup \{ u \}} 
           \alpha_{u,v}^{(k)}\,\bfx_v
    \right),
\end{equation}
where  $\sigma$ is an activation function, $K$ is the number of hops in the neighborhood, $\bfV_k$ is the $k$-th weight matrix, $\mathcal{N}_u^k$ is the $k$-hop neighborhood of $u$, and $\alpha_{u,v}^{(k)}$ is a normalization term. For instance, $\alpha_{u,v}^{(k)}=\left(\hat{d}_v^{(k)}\hat{d}_u^{(k)}\right)^{-1/2}$ weighs according to the degrees $\hat{d}_v^{(k)}$ and $\hat{d}_u^{(k)}$ of  nodes $v$ and $u$ in the $k$-hop graph; other choices can include edge attributes as well.
We note that $\theta_k$ is a parameter specific to the $k$-hop neighborhood of node $u$. Thus, it allows the model to learn different transformation patterns at different distances from the considered node $u$.

\section{Experiments}
\label{sec:tgode_experiments}
We provide an empirical assessment of our method against related temporal DGN models from the literature. 
First, we test the efficacy in handling dynamic graphs with irregularly sampled time series by evaluating the models on several heat diffusion scenarios (see Section~\ref{sec:heat_diffusion}). Afterward, we assess and discuss the performance on real graph benchmarks on traffic forecasting problems (see Section~\ref{sec:real_bench}).  
We report in Table~\ref{tab:tgode_general_configs} (Appendix~\ref{app:tgode_hyperparam}) the grid of hyperparameters employed in our experiments by each method.
We carried out the experiments on 7 nodes of a cluster with 96 CPUs per node
.
We release the code implementing our methodology and reproducing our empirical analysis at \url{https://github.com/gravins/TG-ODE}.

\subsection{Heat Diffusion}
\label{sec:heat_diffusion}
\myparagraph{Setup}
In this section, we focus on simulating the heat diffusion over time on a graph. The data is composed of irregularly sampled graph snapshots providing the temperature of the graph's nodes at the given timestamp, where the initial temperature profile located at some nodes was altered with hot and cold spikes. 
We address the task of predicting the nodes' temperature at future (irregular) timestamps. 
We considered two experimental scenarios. In the first, we altered the temperature of a single node, in the following referred to as \emph{single-spike}. In the second, we altered the temperature of one third of the graph's nodes, which we refer to as \emph{multi-spikes}.
In both scenarios, we simulated seven different diffusion functions. We report additional details about heat diffusion datasets in Appendix~\ref{app:tgode_data_stats}.

We explored the performance of TG-ODE leveraging the aggregation scheme in \cite{tagcn} and the forward Euler's method as discretization procedure, for simplicity. Thus, the nodes' states for the entire snapshot are updated as
\begin{equation}
    \bfX^\ell = \bfX^{\ell-1} +\epsilon\sigma\left(\sum_{k=0}^K \bfL^k \bfX^{\ell-1} \bfV_{k}\right),
\end{equation}
where $K$ corresponds to the number of neighborhood hops and $\bfV_{k}$ is the $k$-th weight matrix.
We recall that other choices of aggregation and discretization schemes are possible. We compared our method with six common DGNs for dynamic graphs: 
A3TGCN~\citep{a3tgcn}, DCRNN~\citep{DCRNN}, TGCN~\citep{T-GCN}, GCRN-GRU~\citep{GCRN}, GCRN-LSTM~\citep{GCRN}, and NDCN~\citep{ndcn}. We note that whenever we used the NDCN model with embedding dimension set to none (see Table~\ref{tab:tgode_general_configs}), the resulting model corresponds to DNND~\citep{ijcai2023p242}.

Moreover, we considered two additional baselines: NODE~\citep{neuralODE} and LB-baseline. 
NODE represents an instance of our approach that does not take into account node interactions. 
Instead, LB-baseline returns the same node states received as input (\ie,
the prediction of $\overline{\mathbf X}_{t_{i+1}}$ is $\hat{\bfX}(t_{i+1}) = \overline{\bfX}_{t_i}$) and
provides a lower bound on the performance we should expect from the learned models. 

We designed each model as a combination of three main components. The first is the encoder which maps the node input features into a latent hidden space; the second is the temporal graph convolution (\ie TG-ODE or the DGN baselines) or the NODE baseline; and the third is a readout that maps the output of the convolution into the output space. The encoder and the readout are MLPs that share the same architecture among all models in the experiments.

To allow all considered baseline models to handle irregular timestamps, we used a similar strategy employed for TG-ODE. Specifically, we selected the unit of time, $\tau$,  and then we iteratively applied the temporal graph convolution for a number of steps equal to the ratio between the time difference between two consecutive timestamps and the time unit, \ie $\#steps = (t_{i+1} - t_i)/ \tau$. 

We performed hyperparameter tuning via grid search, optimizing the Mean Absolute Error (MAE). We trained the models using the Adam optimizer for a maximum of 3000 epochs and early stopping with patience of 100 epochs on the validation error.

\myparagraph{Results}
We present the results on the heat diffusion tasks in Table~\ref{tab:single_spike} and Table~\ref{tab:multi_spike}, using the $log_{10}(\mathrm{MAE})$ as performance metric in both the single-spike and multi-spikes scenarios. The first observation is that TG-ODE has outstanding performance compared to literature models and the baseline. Despite its simpler architecture, our method produces an error that is significantly lower than the runner-up in each task. In the single-spike setting, TG-ODE achieves a $log_{10}(\mathrm{MAE})$ that is on average 308\% to 628\% 
better than the competing models in each task.
Interestingly, not all DGN-based models are capable of improving the results of the LB-baseline. This situation suggests that such approaches attempt to merely learn the mapping function between inputs and outputs rather than learning the actual latent dynamics of the system. Such behavior becomes more evident in the more complex multi-spike scenario. Here, our method achieves up to almost 2080\% 
better $log_{10}(\mathrm{MAE})$ score and more literature models fail in improving the performance with respect to the LB-baseline. These results indicate that capturing the latent dynamics is fundamental, in particular, when the time intervals between observations are not regular over time. We conclude that such methods from the literature might not be suitable for more realistic settings characterized by continuous dynamics and sporadic observations.

Finally, we observe that GCRN-GRU and GCRN-LSTM generate the highest error levels, while DCRNN, NDCN, and NODE are the best among the baselines. Since literature models use RNN architectures to learn temporal patterns, it is reasonable to assume that the poor performance might be due to the limited capacity of RNNs to handle non-uniform time gaps between observations. In contrast, ODE-based models (NODE, NCDN and ours) demonstrate enhanced learning capabilities in this scenario. The performance gap between our model and the considered baselines 
is an indication that the diverse spatial patterns learned by different DGN architectures can heavily impact the performance of the performed tasks.
\setlength{\tabcolsep}{3pt}
\begin{table}[h!]
\centering
\caption{Test $log_{10}(\mathrm{MAE})$ score and std in the single-spike heat diffusion experiments, averaged over 5 separate runs. The lower the better. \one{First}, \two{second}, and \three{third} best results for each task are color-coded.
}\label{tab:single_spike}
\scriptsize
\begin{adjustbox}{angle=90} 
\begin{tabular}{lccccccc}
\toprule
\textbf{Model} & $-\mathbf{L}\mathbf{X}(t)$   & $-\mathbf{L}^2\mathbf{X}(t)$ & $-\mathbf{L}^5\mathbf{X}(t)$ & $-\tanh(\mathbf{L})\mathbf{X}$& $-5\mathbf{L}\mathbf{X}(t)$  & $-0.05\mathbf{L}\mathbf{X}(t)$ & $-(\mathbf{L}+ \mathcal{N}_{0,1})\mathbf{X}(t)$ 
\\\midrule

\textbf{Baselines} \\
$\,$ LB-baseline & -0.557                     & -0.572 &      -0.562 &     -0.538 &       -0.337 &       -0.565 &               -0.837\\ 
$\,$ NODE       & \two{-2.828$_{\pm0.063}$}   & \two{-2.657$_{\pm0.053}$} & \two{-2.139$_{\pm0.005}$} & \two{-2.711$_{\pm0.136}$} & \two{-2.313$_{\pm0.016}$} & \two{-3.983$_{\pm0.003}$} & \two{-2.059$_{\pm0.005}$} \\\midrule

\multicolumn{4}{l}{\textbf{DGN for D-TDGs}} \\
$\,$ A3TGCN     & -0.834$_{\pm0.145}$         & -0.902$_{\pm0.093}$         & -0.819$_{\pm0.036}$         & -0.890$_{\pm0.035}$         & -1.084$_{\pm0.004}$         & -0.653$_{\pm0.001}$         & -0.781$_{\pm0.094}$\\
$\,$ DCRNN      & -1.320$_{\pm0.163}$         & -0.913$_{\pm0.242}$         & \three{-0.867$_{\pm0.305}$} & -1.273$_{\pm0.075}$         & \three{-1.098$_{\pm0.154}$} & -0.964$_{\pm0.366}$         & \three{-1.150$_{\pm0.375}$}\\
$\,$ GCRN-GRU   & -0.474$_{\pm0.232}$         & -0.633$_{\pm0.004}$         & -0.464$_{\pm0.064}$         & -0.621$_{\pm0.047}$         & -0.695$_{\pm0.002}$         & -0.640$_{\pm0.019}$         & -0.490$_{\pm0.094}$\\
$\,$ GCRN-LSTM  & -0.430$_{\pm0.140}$         & -0.323$_{\pm0.019}$         & -0.405$_{\pm0.053}$         & -0.351$_{\pm0.097}$         & -0.511$_{\pm0.157}$         & -0.428$_{\pm0.140}$         & -0.367$_{\pm0.790}$\\
$\,$ TGCN       & -0.825$_{\pm0.108}$         & -0.900$_{\pm0.143}$         & -0.804$_{\pm0.074}$         & -0.834$_{\pm0.149}$         & -1.051$_{\pm0.020}$         & -0.653$_{\pm0.001}$         & -0.781$_{\pm0.094}$ \\
\midrule
\multicolumn{4}{l}{\textbf{DE-DGN for D-TDGs}} \\
$\,$ NDCN       & \three{-1.497$_{\pm0.034}$} & \three{-1.337$_{\pm0.070}$} & -0.350$_{\pm0.328}$         & \three{-1.485$_{\pm0.075}$} & -1.097$_{\pm0.046}$         & \three{-2.408$_{\pm0.183}$} & -0.414$_{\pm0.155}$\\
\midrule 
\textbf{Our} \\
$\,$ TG-ODE & \one{-4.087$_{\pm0.171}$} & \one{-3.106$_{\pm0.181}$} & \one{-2.265$_{\pm0.053}$} & \one{-4.166$_{\pm0.140}$} & \one{-2.351$_{\pm0.036}$} & \one{-4.811$_{\pm0.198}$} & \one{-2.069$_{\pm0.001}$}\\

\bottomrule
\end{tabular}
\end{adjustbox}
\end{table}

\setlength{\tabcolsep}{3pt}
\begin{table}[h!]
\centering
\caption{Test $log_{10}(\mathrm{MAE})$ score and std in the multi-spikes heat diffusion experiments, averaged over 5 separate runs. The lower the better. \one{First}, \two{second}, and \three{third} best results for each task are color-coded.}\label{tab:multi_spike}

\scriptsize
\begin{adjustbox}{angle=90} 
\begin{tabular}{lccccccc}
\toprule
\textbf{Model}            & $-\mathbf{L}\mathbf{X}(t)$ & $-\mathbf{L}^2\mathbf{X}(t)$ & $-\mathbf{L}^5\mathbf{X}(t)$ & $-\tanh(\mathbf{L})\mathbf{X}(t)$ & $-5\mathbf{L}\mathbf{X}(t)$ & $-0.05\mathbf{L}\mathbf{X}(t)$ & $-(\mathbf{L}+ \mathcal{N}_{0,1})\mathbf{X}(t)$ \\ 
\midrule
\textbf{Baselines} \\
$\,$ LB-baseline &  0.490 &       0.517 &       0.552 &      0.523 &        0.256 &        0.561 &                0.666 \\
$\,$ NODE & \two{-1.708$_{\pm0.016}$} & \two{-1.426$_{\pm0.021}$} & \two{-1.093$_{\pm0.004}$} & \two{-1.671$_{\pm0.006}$} & \two{-1.198$_{\pm0.016}$} & \two{-2.749$_{\pm0.016}$} & \two{-0.979$_{\pm0.047}$}\\
\midrule
\multicolumn{4}{l}{\textbf{DGN for D-TDGs}} \\
$\,$ A3TGCN    & 0.443$_{\pm0.087}$          & 0.244$_{\pm0.124}$          & 0.174$_{\pm0.071}$          & 0.509$_{\pm0.058}$          & 0.187$_{\pm0.010}$          & 0.628$_{\pm0.023}$          & 0.328$_{\pm0.060}$\\
$\,$ DCRNN     & \three{-0.140$_{\pm0.092}$} & \three{-0.143$_{\pm0.111}$} & \three{-0.123$_{\pm0.132}$} & -0.122$_{\pm0.120}$         & \three{-0.421$_{\pm0.227}$} & -0.002$_{\pm0.125}$         & \three{-0.212$_{\pm0.333}$}\\
$\,$ GCRN-GRU  & 0.586$_{\pm0.003}$          & 0.614$_{\pm0.004}$          & 0.639$_{\pm0.002}$          & 0.610$_{\pm0.002}$          & 0.440$_{\pm0.003}$          & 0.629$_{\pm0.001}$          & 0.719$_{\pm0.003}$\\
$\,$ GCRN-LSTM & 0.584$_{\pm0.001}$          & 0.610$_{\pm0.002}$          & 0.637$_{\pm0.002}$          & 0.612$_{\pm0.003}$          & 0.440$_{\pm0.005}$          & 0.631$_{\pm0.002}$          & 0.705$_{\pm0.002}$\\
$\,$ TGCN      & 0.404$_{\pm0.236}$          & 0.313$_{\pm0.072}$          & 0.113$_{\pm0.071}$          & 0.493$_{\pm0.056}$          & 0.113$_{\pm0.086}$          & 0.615$_{\pm0.023}$          & 0.364$_{\pm0.134}$ \\
\midrule
\multicolumn{4}{l}{\textbf{DE-DGN for D-TDGs}} \\
$\,$ NDCN      & 0.120$_{\pm0.325}$          & -0.070$_{\pm0.056}$         & 0.315$_{\pm0.245}$          & \three{-0.128$_{\pm0.020}$} & 0.146$_{\pm0.107}$          & \three{-1.357$_{\pm0.053}$} & 0.384$_{\pm0.013}$\\
\midrule
\textbf{Our} \\
$\,$ TG-ODE & \one{-4.259$_{\pm0.037}$} & \one{-3.705$_{\pm0.143}$} & \one{-1.314$_{\pm0.249}$} & \one{-3.572$_{\pm0.010}$} & \one{-2.350$_{\pm0.083}$} & \one{-4.567$_{\pm0.109}$} & \one{-1.021$_{\pm0.002}$}\\
\bottomrule
\end{tabular}
\end{adjustbox}
\end{table}

\subsection{Traffic Benchmarks}\label{sec:real_bench}

\myparagraph{Setup}
This section introduces a set of graph benchmarks whose objective is to assess traffic forecasting performance from irregular time series; similar to the heat diffusion tasks, we predict the future node values given only the past history.
We considered six real-world graph benchmarks for traffic forecasting: MetrLA~\citep{DCRNN}, Montevideo~\citep{rozemberczki2021pytorch}, PeMS03~\citep{pems03-08}, PeMS04~\citep{pems03-08}, PeMS07~\citep{pems03-08}, and PeMS08~\citep{pems03-08}. We used a modified version of the original datasets where we employed irregularly sampled observations. We will refer to the datasets by using the subscript ``i'' -- e.g., \textbf{MetrLA}$_i$ -- to make apparent the difference from the original versions.
We report additional details about the datasets in Appendix~\ref{app:tgode_data_stats}.    
We considered a temporal data splitting in which 80\% of the previously selected snapshots are used as training set, 10\% as validation set, and the remaining as test set.
    
For these experiments, we considered the same models, baseline and architectural choices of the heat diffusion experiments. Since NODE does not take into account interactions between nodes for its predictions, we choose not to include it as a baseline in this scenario. Hyperparameter tuning has been performed by grid search, optimizing the MAE. Optimizer settings are the same as for the previous experiments. 

\begin{figure}[ht]
    \centering
    \includegraphics[width=0.95\textwidth]{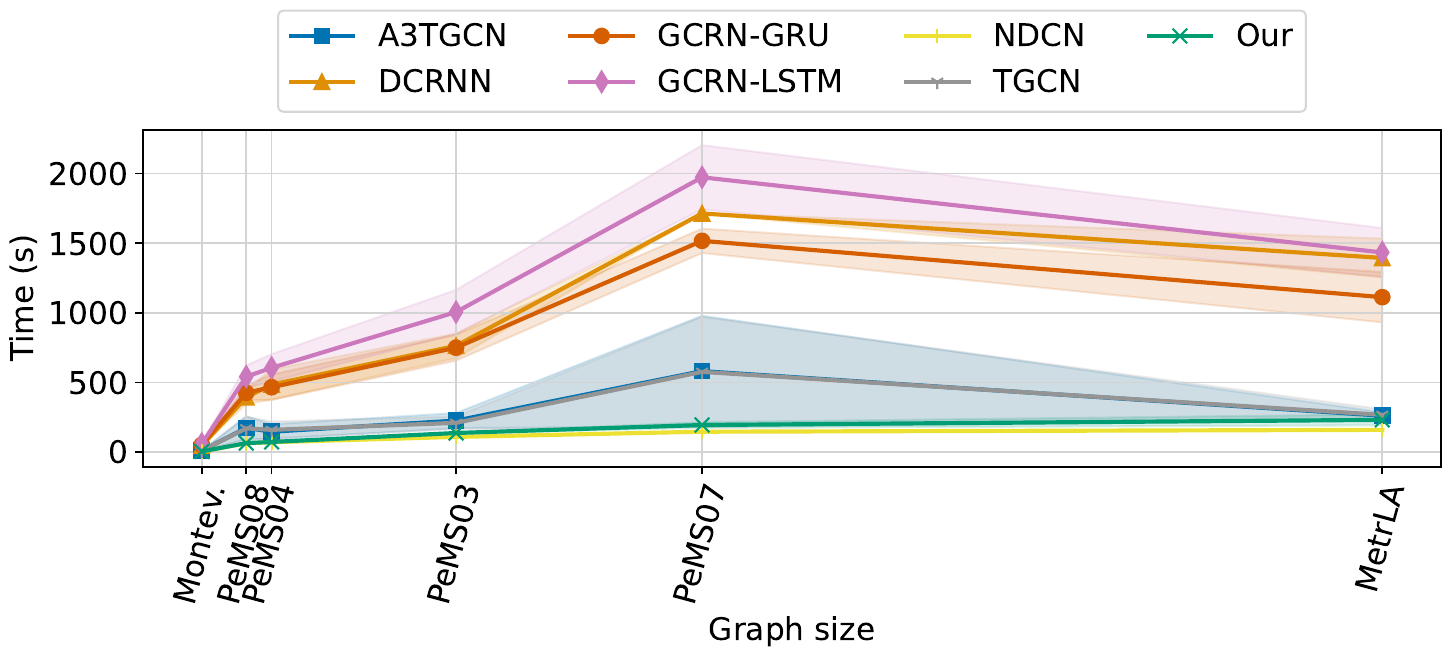}
    \caption{Average time per epoch (measured in seconds) and std computed using an Intel Xeon Gold 6240R CPU @ 2.40GHz. Each time is obtained using 5 neighbor hops (when possible) and embedding dimension equal to 64. The graph size is computed as $size=\#steps * \#edges$.}
    \label{fig:timing}
\end{figure}

\myparagraph{Results}
Table~\ref{tab:graph_benchmark} reports the traffic forecasting results in
terms of $\mathrm{MAE}$. Similarly to the heat diffusion scenario, TG-ODE shows a remarkable performance improvement compared to literature models, achieving an MAE that is up to 202\% 
better than the runner-up model. Moreover, as reported in Figure~\ref{fig:timing}, we observe that TG-ODE is $2\times$ to $13\times$ faster than the other approaches under test. NDCN is the sole method matching the speed of our approach. However, it's noteworthy that NDCN utilizes only one neighbor hop, thereby simplifying the final computation.
\setlength{\tabcolsep}{2pt}
\begin{table}[ht!]
\centering
\caption{Test $\mathrm{MAE}$ score and std in the traffic forecasting setting, averaged over 5 separate runs. $\dagger$ means gradient explosion. The \one{first}, \two{second}, and \three{third} best scores are colored.}\label{tab:graph_benchmark}
\scriptsize
\begin{adjustbox}{center}
    
\begin{tabular}{lccccccc}
\toprule
\textbf{Model}            & \textbf{MetrLA$_i$}         & 
\textbf{Montevid.$_i$} & \textbf{PeMS03$_i$}         & \textbf{PeMS04$_i$}          & \textbf{PeMS07$_i$}         & \textbf{PeMS08$_i$} \\ 
\midrule
\textbf{Baseline} \\
$\,$ LB-baseline    & 58.191                      & 0.442                              & 165.015                     & 211.230                      & 314.710                     & 227.380           \\ 
\midrule
\multicolumn{4}{l}{\textbf{DGN for D-TDGs}} \\
$\,$ A3TGCN      & \two{5.731$_{\pm0.011}$}    & 0.378$_{\pm4\cdot10^{-4}}$ & 28.897$_{\pm0.733}$         & \two{32.221$_{\pm1.355}$}   & \two{38.303$_{\pm0.795}$}   & \two{30.652$_{\pm0.995}$}   \\
$\,$ DCRNN       & $\dagger$                   & \two{0.332$_{\pm0.001}$}   & \two{18.652$_{\pm0.136}$}   & $\dagger$                   & $\dagger$                   & $\dagger$                   \\
$\,$ GCRN-GRU    & 8.438$_{\pm0.004}$          & \two{0.332$_{\pm0.001}$}   & 49.360$_{\pm18.619}$        & 53.389$_{\pm4.728}$         & 68.785$_{\pm5.787}$         & 51.787$_{\pm10.872}$        \\
$\,$ GCRN-LSTM   & 8.440$_{\pm0.009}$          & \three{0.333$_{\pm0.002}$} & 62.210$_{\pm0.923}$         & 52.427$_{\pm4.162}$         & 151.824$_{\pm17.654}$       & 80.567$_{\pm24.891}$        \\
$\,$ TGCN        & \three{5.832$_{\pm0.125}$}  & 0.380$_{\pm4\cdot10^{-4}}$ & \three{28.506$_{\pm0.332}$} & \three{33.059$_{\pm1.063}$} & \three{38.750$_{\pm1.429}$} & \three{33.114$_{\pm1.963}$} \\
\midrule
\multicolumn{4}{l}{\textbf{DE-DGN for D-TDGs}} \\
$\,$ NDCN & 8.471$_{\pm0.022}$ & 0.435$_{\pm0.021}$ & $\dagger$ & 127.202$_{\pm0.334}$ & $\dagger$ & 129.667$_{\pm44.385}$ \\
\midrule 
\textbf{Our} \\
$\,$ TG-ODE         & \one{2.828$_{\pm0.001}$} & \one{0.327$_{\pm6\cdot10^{-5}}$} & \one{17.423$_{\pm0.012}$} & \one{24.739$_{\pm0.014}$}  & \one{26.081$_{\pm0.004}$} & \one{18.818$_{\pm0.021}$}      \\
\bottomrule
\end{tabular}
\end{adjustbox}
\end{table}

We observe that the baseline performs poorly in these benchmarks, suggesting that such tasks are more complex than the ones based on heat diffusion. 
Despite all DGN models outperforming the LB-baseline, they still produce an error that is on average double
than that of TG-ODE, highlighting the added value of our approach when dealing with D-TDGs characterized by irregular sampling.

Finally, we comment that the DCRNN and NDCN suffered from gradient issues in most of the tasks. We believe this is due to
their inability to learn the latent dynamics of the system when the models' outputs are not computed over a regular time series. 

\subsubsection{Impact of the Sample Sparsity}\label{par:tgode_ablation}
To demonstrate the effectiveness of our approach, we study the prediction performance under different sparsity levels. We consider here the PeMS04 dataset. In this analysis, we systematically decreased the number of graph snapshots considered in the time series. This reduction makes the resulting task more challenging than the original one, as the snapshots become more sparse over time -- the expected difference $t_{i+1}-t_{i}$ gets larger. Additionally, the model has fewer data to learn the task, thereby amplifying the task complexity. We generated the irregular time series by randomly selecting 500, 1000, 2000, 4000, 8000, or 16000 graph snapshots from the original dataset (from 3\% to 94\% of the original data), resulting in varying degrees of sparsity. 
For each dataset size, we used an 80/10/10 temporal data split and performed hyperparameter tuning, as previously done in this section.

\begin{figure}[ht]
    \centering
    \includegraphics[width=0.9\textwidth]{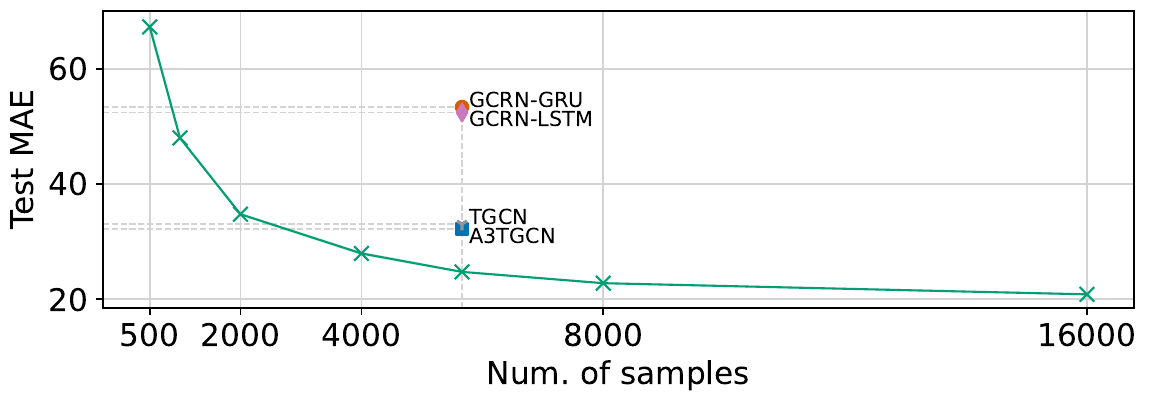}
    \caption{Test $\mathrm{MAE}$ scores and std of TG-ODE on PeMS04, averaged over 5 runs, for different sparsity levels.}
    \label{fig:ablation}
\end{figure}

Figure~\ref{fig:ablation} illustrates the performance of our method, TG-ODE, at various degrees of data sparsity. As expected, we observe that as the number of samples increases, the test MAE decreases. 
Notably, TG-ODE maintains robust performance even with higher degrees of sparsity, with a decrease in performance by only $\sim$7 points when reducing the size from 16000 to 4000 samples. 
While the prediction error is indeed relatively large when considering only 3-7\% of the original data, we comment that it is still substantially better than that of the LB-baseline and comparable to that of the other DGN's which used 33\% of the data. Overall, the model exhibits excellent performance even in situations of high to extreme sparsity (i.e., less than 8000 samples).
The observed outcome supports the effectiveness of TG-ODE, emphasizing its potential for real-world applications with irregularly sampled D-TDGs. 

\section{Related Work}
A natural choice for many methods in the D-TDG domain has
been to extend Recurrent Neural Networks to graph data. Indeed,
most of the models presented in the literature can be summarized as a combination of DGNs and RNNs. We refer the reader to Section~\ref{sec:D_TDG} for a deeper discussion on such methods. 
Differently from these approaches, which are intrinsically designed to deal with regular time series, TG-ODE can naturally handle arbitrary time gaps between observations. This makes our framework more suitable for realistic scenarios, in which data are irregular over time. 

More recently, neural differential equation approaches have also been developed for D-TDGs. TDE-GNN~\citep{eliasof2024temporal} employs higher-order ODEs to capture the D-TDG dynamic, while NDCN \citep{ndcn} extends GDE~\citep{GDE} to learn continuous-time dynamics on both static and D-TDGs by diffusing input features until the termination time. MTGODE~\citep{related_first} adopts an ODE-based approach to deduce missing graph topologies from the time-evolving node features in regularly sampled D-TDGs. Differently, \cite{related_fourth} and \cite{related_fifth} propose an ODE-based model in the form of a variational auto-encoder for learning latent dynamics from sampled initial states. To infer missing observations, the methods consider both past and future neighbors' information. This prevents them from being used in an online setting, where data becomes available in a sequential order. 
Lastly, STG-NCDE~\citep{related_second} employs a stacked architecture of two neural controlled differential equations to model temporal and spatial information, respectively. In the STG-NCDE's paper, irregular data are considered, yet they are handled by making them regular via interpolation.
In contrast to these approaches, in this chapter we explicitly address irregularly-sampled D-TDGs and we propose a simple model to showcase the effectiveness and efficiency of the TG-ODE framework in working with such data, eliminating the need for additional strategies, such as interpolation. It should be noted that while many of the ODE-based approaches mentioned earlier can be viewed as instances of the introduced TG-ODE framework, our model is specifically designed to demonstrate the benefits of this approach.

Finally, we note that the general formulation of TG-ODE allows extending the Message Passing Neural Network (MPNN)~\citep{MPNN} -- hence, all its specific instances, such as \cite{GCN,GAT,SAGE,GIN,chebnet,gine,tagcn} -- to the domain of D-TDGs by selecting appropriate operators in Equation~\ref{eq:tg-ode} and, in turn, it allows us to tailor our TG-ODE to exploit relational inductive biases and fulfill given application requirements.
For instance, by considering a graph attentional operator~\citep{GAT} in Equation~\ref{eq:tg-ode}, we can implement an anisotropic message passing within the D-TDG during the update of node states.

\section{Summary}
\label{sec:conclusions}
In this chapter, we have presented \textit{Temporal Graph Ordinary Differential Equation} (TG-ODE), a new general framework for effectively learning from irregularly sampled D-TDGs. 
Thanks to the connection between ODEs and neural architectures, TG-ODE can naturally handle arbitrary time gaps between observations, allowing to address a common limitation of DGNs for D-TDGs, \ie the restriction to work solely on regularly sampled data.

To demonstrate the benefits of our approach, we conducted extensive experiments on ad-hoc benchmarks that include several synthetic and real-world scenarios. The results of our experimental analysis show that our method outperforms state-of-the-art models for D-TDGs by a large margin. Furthermore, our method benefits from a faster training, thus suggesting scalability to large networks. 


\chapter{Non-Dissipative Propagation for C-TDGs}\label{ch:ctan}
In this chapter, we focus on dynamic graphs with a continuous evolution, \ie C-TDGs (see Section~\ref{sec:dynamic_graph_notation}). Real-world scenario examples include the continual activities and interactions between members of social as well as communication networks, recurrent purchases by users on e-commerce platforms, or evolving interactions of processes with files in an operating system.

As observed in Chapter~\ref{ch:learning_dyn_graphs}, recent works investigated models that can process the spatio-temporal dimension of a dynamic graph defined through irregularly sampled event streams.
However, such dynamic methods are based on static DGNs and RNNs as backbone architectures, thus retaining the limitations of their core components. Specifically, static DGNs suffer from the \textit{oversquashing} phenomenon (see Section~\ref{sec:dgn_plights}), which prevents the final network to learn and propagate long range information as demonstrated in Chapters~\ref{ch:antisymmetry} and \ref{ch:phdgn}. Similarly, RNNs often face similar challenges in propagating long-term dependencies, as evidenced by \citet{chang2018antisymmetricrnn}, mainly due to exploding or vanishing gradients. With growing evidence from the static and dynamic case~\citep{LRGB, yu2023towards} that long-range dependencies are necessary for effective learning, the ability to learn properties beyond the event's temporal and spatial locality remains an open challenge in the C-TDG domain.

In this chapter, we propose the \emph{Continuous-Time Graph Antisymmetric Network}\index{CTAN} (\gls*{CTAN}), a framework for learning of C-TDGs with \emph{scalable} long range propagation of information, thanks to properties inherited from stable and non-dissipative ODEs.

We establish theoretical conditions for achieving stability and non-dissipation in the CTAN ODE by employing antisymmetric weight matrices, which is the key factor for modeling long-range spatio-temporal interactions.
The CTAN layer is derived from the forward Euler discretization of the designed differential equation. 
The formulation of CTAN allows scaling the radius of propagation of information depending on the number of discretization steps, \ie the number of layers in the final architecture.
Remarkably, even with a limited number of layers, the non-dissipative behavior enables the transmission of information for a past event as new events occur, since node states are used to efficiently retain and propagate historical information. 
This mechanism permits scaling the single event propagation to cover a larger portion of the C-TDG.
The general formulation of the node update state function allows the implementation of the more appropriate dynamics to the problem at hand. Specifically, it allows the inclusion of static DGN dynamics, thus reinterpreting current state-of-the-art static DGNs as a discretized representation of non-dissipative ODEs tailored for C-TDGs, mirroring previous approaches in the static case (discussed in Section~\ref{sec:ADGN_mpnn_comparison}).
To the best of our knowledge, CTAN is the first framework to effectively address the problem of long-range propagation in C-TDGs and the first to bridge the gap between ODEs and C-TDGs.

The key contributions of this chapter can be summarized as follows:
\begin{itemize}
    \item We introduce the problem of long-range propagation (\ie non-dis\-si\-pa\-tive\-ness) within C-TDGs.
    \item We introduce CTAN, a new deep graph network for learning C-TDGs based on ODEs, which enables stable and non-dissipative propagation to preserve long term dependencies in the information flow, and it does so in a theoretically founded way.
    \item We present novel benchmark datasets specifically designed to assess the ability of DGNs to propagate information over long spatio-temporal distances within C-TDGs.
    \item We conduct extensive experiments to demonstrate the benefits of our method, showing that CTAN not only outperforms state-of-the-art DGNs on synthetic long-range tasks but also outperforms them on several real-world benchmark datasets.
\end{itemize}

This chapter has been developed during a six months
internship at Huawei Technologies, Munich Research Center, in Munich, Germany.
We base this chapter on \cite{gravina_ctan}.
\section{Continuous-Time Graph Antisymmetric Network}
Learning the dynamics of a C-TDG can be cast as the problem of learning information propagation following newly observed events in the system. This entails learning a diffusion function that updates the state of node $u$ as 
\begin{equation}\label{eq:propagation}
\mathbf{h}_u(t) = F\left(t, \mathbf{x}_u, \mathbf{h}_u(t), \{\mathbf{h}_v(t)\}, \{\mathbf{e}_{uvt^-}\}\right),
\end{equation}
where $(v,t^-)\in\mathcal{N}^t_u$, and $\mathcal{N}^t_u = \{(v, t^-) \, |\,  \{u,v,t^-\} \in \mathcal{E}(t) \}$ is the temporal neighborhood of a node $u$ at time $t$, which consists of all the historical neighbors of $u$ prior to current time $t$ (as usual). In the following, we omit the time subscript from the edge feature vector to enhance readability, since it refers to a time in the past in which the edge appeared. 

As evidenced in Section~\ref{sec:ADGN_mpnn_comparison}, Equation~\ref{eq:propagation} can be modeled through a dynamical system described by a learnable ODE. Differently from discrete models, neural-ODE-based approaches learn more effective latent dynamics and have shown the ability to learn complex temporal patterns from irregularly sampled timestamps \citep{neuralODE, ode-irregular, neuralCDE}, making them more suitable to address C-TDG problems.

Here, we leverage non-dissipative ODEs for the processing of C-TDGs. Thus, we propose a framework as a solution to a \textit{stable} and \textit{non-dissipative} ODE over a streamed graph. The main goal of this chapter is therefore achieving preservation of long-range information between nodes over a stream of events. We do so by first showing how a generic ODE can learn the hidden dynamics of a C-TDG and then by deriving the condition under which the ODE is constrained to the desired behavior.

\subsection{Modeling C-TDGs as Cauchy Problems}
\begin{figure}[ht]
\begin{center}
    \includegraphics[width=0.9\linewidth]{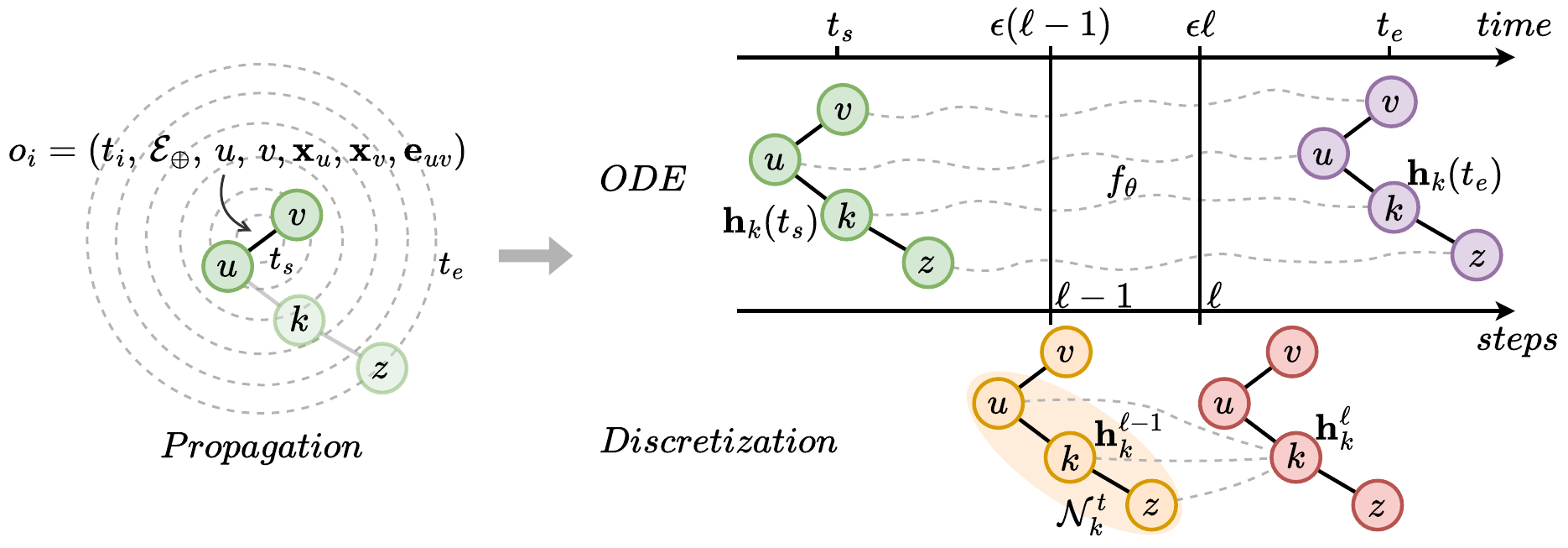}
\end{center}
\caption{A high-level overview of the proposed framework illustrated for the $i$-th Cauchy sub-problem. On the left, we depict the propagation of the information of event $o_i$ through the graph. $o_i$ is an interaction event ($\mathcal{E}_\oplus$) between nodes $u$ and $v$.  The faded portion of the graph corresponds to historical information, while the rest is the incoming event. On the right, we illustrate the evolution of node states given the propagation of the incoming event. Specifically, the top right shows the evolution as an ODE, $f_\theta$, that computes the node representation for a node $k$,  $\mathbf{h}_k(t)$. Such computation is subject to an initial condition $\mathbf{h}_k(t_s)=\eta(\mathbf{h}_k^{i-1}(t_e), \mathbf{x}_k(i))$ that includes the node representations computed in the previous sub-problem $\mathbf{h}_k^{i-1}(t_e)$ and the current node input state. In the bottom right, the discretized solution of the ODE is computed as iterative steps of the method over a discrete set of points in the time interval $[t_s, t_e]$.}
\label{fig:ctan_method}
\end{figure}

As in previous chapters, we first define a Cauchy problem in terms of the node-wise ODE defined in time $t\in[0,T]$
\begin{equation}\label{eq:ctan_ode}
    \frac{d\mathbf{h}_u(t)}{d t}=f_\theta\left(t, \mathbf{x}_u, \mathbf{h}_u(t), \{\mathbf{h}_v(t)\}_{v\in\mathcal{N}^t_u}, \{\mathbf{e}_{uv}\}_{v\in\mathcal{N}^t_u}\right)
\end{equation}
and subject to an initial condition $\mathbf{h}_u(0) \in \mathbb{R}^d$. The term $f_\theta$ is a function parametrized by the weights $\theta$ that describes the dynamics of node state. We observe that this framework can naturally deal with events that arrive at an arbitrary time. Indeed, the original Cauchy problem in Equation~\ref{eq:ctan_ode} can be divided into multiple sub-problems, one per each event in the C-TDG.
The $i$-th sub-problem, defined in the interval $t\in[t_s, t_e]$, is responsible for propagating only the information encoded by the $i$-th event. Overall, when a new event $o_i$ happens, the ODE in Equation~\ref{eq:ctan_ode} computes new nodes representations $\mathbf{h}_u^i(t_e)$, starting from the initial configurations $\mathbf{h}_u^i(t_s)$. In other words, $f_\theta$ evolves the state of each node given its initial condition. The top-right of Figure~\ref{fig:ctan_method} visually summarizes this concept, showing the nodes evolution given the propagation of an incoming event.
We observe that the knowledge of past events is preserved and propagated in the system thanks to an initial condition that includes not only the current node input states but also the node representations computed in the previous sub-problem, \ie $\mathbf{h}^i_u(t_s)=\eta(\mathbf{h}_u^{i-1}(t_e), \mathbf{x}_u(i))$. We notice that the terminal time $t_e$ (treated as an hyperparameter) is responsible for determining the extent of information propagation across the graph, since it limits the propagation to a constrained distance from the source. Consequently, smaller values of $t_e$ allow only for localized event propagation, whereas larger values enable the dissemination of information to a broader set of nodes. 

While this approach similar to the one in Chapter~\ref{ch:tgode} and it is applicable to all ODE-based DGNs for C-TDGs, we note that we are the first to introduce this truncated history propagation method in C-TDGs.

\subsection{Non-Dissipativeness in C-TDGs}
We now proceed to derive the condition under which the ODE is constrained to a stable and non-dissipative behavior, allowing for the propagation of long-range dependencies in the information flow. Non-Dissipativeness in C-TDGs can be dissected into two components: non-dissipativeness over space and over time. 
\begin{definition}[Non-dissipativeness over space]\label{def:non-dissip}\index{non-dissipativity over space|see {non-dissipativity}}\index{non-dissipativity!{over space, in C-TDGs}}
Let $u,v\in\mathcal{V}(t)$ be two nodes of the C-TDG at some time $t$, connected by a path of length $L$. If an event $o_i$ occurs at node $u$, then the information of $o_i$ is propagated from $u$ to $v$, $\forall L\geq0$.
\end{definition}

We start by instantiating Equation~\ref{eq:ctan_ode} as 
\begin{equation}
\label{eq:our_ode}
\frac{d \mathbf{h}_u(t)}{d t} = \sigma \Bigl(\mathbf{W}_t \mathbf{h}_u(t) 
     +\Phi\left(\{\mathbf{h}_v(t),\mathbf{e}_{uv}, t_v^-, t \}_{v\in\mathcal{N}_u^t}\right) \Bigr)
\end{equation}
where $\sigma$ is a monotonically non-decreasing activation function; $\Phi$ is the aggregation function that computes the representation of the neighborhood of the node $u$ considering node states and edge features; $t_v^-$ is the time point of the previous event for node $v$; and  $\mathbf{W}_t\in\mathbb{R}^{d\times d}$. Here and in the following, the bias term is omitted for simplicity. We notice that including $t_v^-$ in $\Phi$ encodes the time elapsed since the previous event involving node $v$. This inclusion allows for smooth updates of the node’s current state during the time interval to prevent the staleness problem (see Section~\ref{sec:survey_ctdg}).

As discussed in Chapter~\ref{ch:antisymmetry}, a non-dissipative propagation is directly linked to the sensitivity of the solution of the ODE to its initial condition, thus to the stability of the system. Such sensitivity is controlled by the Jacobian's eigenvalues of Equation~\ref{eq:our_ode}. Given $\lambda_i(\mathbf{J}(t))$ the $i$-th eigenvalue of the Jacobian, when 
$Re(\lambda_i(\mathbf{J}(t))) = 0$ for $i=1,...,d$ the initial condition is effectively propagated into the final node representation, making the system both stable and non-dissipative\footnote{This result holds also when the eigenvalues of the Jacobian are still bounded in a small neighborhood around the imaginary axis (see Section~\ref{sec:adgn}).}. 

\begin{definition}[Non-dissipativeness over time]\label{def:nd-time}\index{non-dissipativity over time|see {non-dissipativity}}\index{non-dissipativity!{over time, in C-TDGs}}
Let $u\in \mathcal{V}(t)$ be a node in the C-TDG at time $t$ and $o_t$ an event that occurs at node $u$ at time $t$. A DGN for C-TDGs is non-dissipative over time if, regardless of how many more events subsequently occur at $u$, the information of event $o_t$ will persist in $u$'s embedding.
\end{definition}
In essence, Definition~\ref{def:nd-time} captures the idea that the embedding computed by a DGN for a node in a C-TDG retains the information from a specific event indefinitely, ensuring that the historical context is preserved and not forgotten despite the occurrence of additional events at that node.

To show the property of non-dissipativity over time, we analyze the entire system defined in Equation~\ref{eq:our_ode} from a temporal perspective. Thus, Equation~\ref{eq:our_ode} can be reformulated as:
\begin{align}\label{eq:our_ode_time}
    \frac{d \mathbf{h}_u(t)}{d t} = \sigma \Bigl(&\mathbf{W}_t \eta(\mathbf{h}_u(t),\mathbf{x}_u(t))  \nonumber\\
    &+  \Phi\left(\{\eta(\mathbf{h}_v(t),\mathbf{x}_v(t)),\mathbf{e}_{uv}, t_v^-, t \}_{v\in\mathcal{N}_u^t}\right) \Bigr)
\end{align}
where $\eta$ is the function that computes the initial condition for the propagation of each event considering the node representations computed in the previous event propagation $\mathbf{h}_u(t)$ and the current node
input state $\mathbf{x}_u(t)$ (as before).

In this context, we can view the system as having $e\Delta t$ steps, where $e$ denotes the number of events and $\Delta t$ represents the propagation time of an event. Furthermore, the input state of the node $\mathbf{x}_u(t)$ is only present upon  occurrence of a new event, meaning that during the propagation of events, $\mathbf{x}_u(t)$ is set to 0. Therefore, its impact on information propagation is confined to the event's specific occurrence and does not affect each step of the propagation process.

The next proposition ensures that when the eigenvalues of the Jacobian matrix of Equation~\ref{eq:our_ode_time} are placed only on the imaginary axes, then the ODE in Equation~\ref{eq:our_ode_time} is non-dissipative in both space and time. Thus, we guarantee the preservation of historical context over time and the propagation of event information through the C-TDGs.
\begin{proposition}\label{prop:space_time_non_dissip}
Provided that the weight matrix $\mathbf{W_t}$ is antisymmetric and the aggregation function $\Phi$ does not depend on $\mathbf{h}_u(t)$, then the ODE in Equation~\ref{eq:our_ode_time} is stable and non-dissipative over space and time if the resulting Jacobian matrix has purely imaginary eigenvalues, \ie $$ Re(\lambda_i(\mathbf{J}(t))) = 0, \forall i=1, ..., d.$$
\end{proposition}
For the proof, we refer the reader to the proof of Proposition~\ref{prep:stableCondition} and substitute the Jacobian computed with respect to space with a Jacobian computed with respect to time.

By constraining weight matrix $\mathbf{W_t}$ to be antisymmetric we obtain that the ODE in Equation~\ref{eq:our_ode} is not-dissipative in both space and time, guaranteeing the preservation of historical node context over time while propagating event information ``spatially'' through the C-TDGs. We provide a more in-depth analysis of non-dissipativeness over time in the following paragraph where we show that varying the formulation of $\eta$ can yield to diverse behaviors.

\myparagraph{In-depth analysis of non-dissipativeness over time}
We note that the non-dissipative behavior of the system in Equation~\ref{eq:our_ode_time} is contingent on the specific definition of the function $\eta$. Varying the formulation of $\eta$ can yield to diverse behaviors, significantly impacting the system's ability to either preserve or dissipate information over time.
\begin{proposition}\label{prop:non-diss_space_and_time}
Provided that the aggregation function $\Phi$ does not depend on $\mathbf{h}_u(t)$, the Jacobian matrix resulting from the ODE in Equation~\ref{eq:our_ode_time} has purely imaginary eigenvalues, \ie $Re(\lambda_i(\mathbf{J}(t))) = 0, \forall i=1, ..., d$ if the function $\eta$ is implemented as one of the following functions:
\begin{itemize}
    \item addition, \ie $\eta=\mathbf{h}_u(t)+\mathbf{x}_u(t)$;
    \item concatenation, \ie $\eta=\mathbf{h}_u(t)\|\mathbf{x}_u(t)$;
    \item composition of \textit{tanh} and concatenation, i.e., $\eta=\textrm{tanh}(\mathbf{h}_u(t)\|\mathbf{x}_u(t))$.
\end{itemize}
\end{proposition}

\begin{boxedproof}
Let's consider $\eta=\mathbf{h}_u(t)+\mathbf{x}_u(t)$, \ie \textbf{addition}. In this case Equation~\ref{eq:our_ode_time} can be reformulated as 
\begin{equation}\label{eq:addition_stace_and_time}
    \frac{d \mathbf{h}_u(t)}{d t} = \sigma \Bigl(\mathbf{W}_t\mathbf{h}_u(t) + \mathbf{W}_t\mathbf{x}_u(t) +  \Phi\left(\{(\mathbf{h}_u(t)+\mathbf{x}_u(t)),\mathbf{e}_{uv}, t_v^-, t \}_{v\in\mathcal{N}_u^t}\right) \Bigr). 
\end{equation}
The Jacobian matrix of Equation~\ref{eq:addition_stace_and_time} is defined as 
\begin{multline}
    \mathbf{J}(t) = \text{diag}\Bigl[\sigma^\prime \Bigl(\mathbf{W}_t\mathbf{h}_u(t) + \mathbf{W}_t\mathbf{x}_u(t) \\
    +  \Phi\bigl(\{(\mathbf{h}_u(t) + \mathbf{x}_u(t)),\mathbf{e}_{uv}, t_v^-, t \}_{v\in\mathcal{N}_u^t}\bigr) \Bigr)\Bigr]\mathbf{W}_t.
\end{multline}
Thus, it is the result of a matrix multiplication between invertible diagonal matrix and a weight matrix. Imposing $\mathbf{A}=\text{diag}\left[\sigma^\prime \left(\mathbf{W}_t \mathbf{h}_u(t) +  \Phi\left(\{\mathbf{h}_v(t),\mathbf{e}_{uv}, t_v^-, t \}_{v\in\mathcal{N}_u^t}\right) + \mathbf{b}_t \right)\right]$, then the Jacobian can be rewritten as $\mathbf{J}(t)=\mathbf{A}\mathbf{W}_t$.

Let us now consider an eigenpair of $\mathbf{A} \mathbf{W}_t$, where the eigenvector is denoted by $\mathbf{v}$ and the eigenvalue by $\lambda$. Then:
\begin{align}
\label{eq:ctan_eigen}
    \mathbf{A}\mathbf{W}_t\mathbf{v} &= \lambda \mathbf{v},\notag \\
    \mathbf{W}_t\mathbf{v} &= \lambda \mathbf{A}^{-1}\mathbf{v},\notag \\
    \mathbf{v}^*\mathbf{W}_t\mathbf{v} &= \lambda (\mathbf{v}^*\mathbf{A}^{-1}\mathbf{v})
\end{align}
where $*$ represents the conjugate transpose.
On the right-hand side of Equation~\ref{eq:ctan_eigen}, we can notice that the $(\mathbf{v}^*\mathbf{A}^{-1}\mathbf{v})$ term  is a real number. 
If the weight matrix $\mathbf{W}_t$ is anti-symmetric (\ie skew-symmetric), then it is true that  $\mathbf{W}_t^* = \mathbf{W}_t^\top=-\mathbf{W}_t$. Therefore, 
$(\mathbf{v}^*\mathbf{W}_t\mathbf{v})^* = \mathbf{v}^*\mathbf{W}_t^*\mathbf{v} = -\mathbf{v}^*\mathbf{W}_t\mathbf{v}$. Hence, 
the $\mathbf{v}^*\mathbf{W}_t\mathbf{v}$ term on the left-hand side of Equation~\ref{eq:ctan_eigen} is an imaginary number.
Thereby, $\lambda$ needs to be purely imaginary, and, as a result, all eigenvalues of $\mathbf{J}(t)$ are purely imaginary.

Let's now consider $\eta=\mathbf{h}_u(t)\|\mathbf{x}_u(t)$, \ie \textbf{concatenation}. In this case, the product $\mathbf{W}_t(\mathbf{h}_u(t)\|\mathbf{x}_u(t))$ can be decomposed as $\mathbf{K}_t\mathbf{h}_u(t) + \mathbf{V}_t\mathbf{x}_u(t)$, with $\mathbf{K}_t$ and $\mathbf{V}_t$ weight matrices. Similarly to the addition case, the Jacobian has purely imaginary eigenvalues.

Lastly, we consider the case of $\eta=\textrm{tanh}(\mathbf{h}_u(t)\|\mathbf{x}_u(t))$, \ie the \textbf{composition of \textit{tanh} and concatenation}. Here, Equation~\ref{eq:our_ode_time} is
\begin{multline}\label{eq:our_ode_time_tanh}
\frac{d \mathbf{h}_u(t)}{d t} = \sigma \bigg(\mathbf{W}_t \textrm{tanh}(\mathbf{h}_u(t)) + \mathbf{V}_t \textrm{tanh}(\mathbf{x}_u(t)) \\
+  \Phi\Bigl(\{\textrm{tanh}(\mathbf{h}_u(t)\|\mathbf{x}_u(t)),\mathbf{e}_{uv}, t_v^-, t \}_{v\in\mathcal{N}_u^t}\Bigr) \bigg). 
\end{multline}
The Jacobian matrix is the results of the multiplication of three matrices, i.e., $\mathbf{J}(t)=\mathbf{A}\mathbf{B}\mathbf{W}_t$, with $\mathbf{A} = \mathrm{diag}\left[\sigma' \left(\mathbf{W}_t\textrm{tanh}(\mathbf{h}_u(t)) + \mathbf{V}_t\textrm{tanh}(\mathbf{x}_u(t)) +\Phi(...) + \mathbf{b}\right)\right]$ and $\mathbf{B} = \mathrm{diag}[1-\textrm{tanh}^2(\mathbf{h}_u(t))]$. Thanks to the associative property of multiplication $\mathbf{J}(t)=\mathbf{A}\mathbf{B}\mathbf{W}_t=(\mathbf{A}\mathbf{B})\mathbf{W}_t= \mathbf{D}\mathbf{W}_t$, where $\mathbf{D}$ is the result of the multiplication of two diagonal matrices, thus $\mathbf{D}$ is diagonal. As detailed for the addition case, we can conclude that the Jacobian matrix has purely imaginary eigenvalues.
\end{boxedproof}

As a counterexample, if $\eta=\mathbf{x}_u(t)$, Equation~\ref{eq:our_ode_time} can result in a dissipative behavior, leading to the loss of information over time and compromising the model's ability to preserve historical context, since past node information is always discarded between new events. As a result, the function $\eta$ can function as a parameter to control the balance between the dissipative and non-dissipative behavior of CTAN.

\subsection{Numerical Discretization} 
Now that we have defined the conditions under which the ODE in Equation~\ref{eq:our_ode} is stable and non-dissipative, \ie it can propagate long-range dependencies between nodes in the C-TDG, we 
rely on a discretization method to compute an approximate solution
. We employ the \textit{forward Euler's method} to discretize Equation~\ref{eq:our_ode} for the $i$-th Cauchy sub-problem, yielding the following node state update equation for the node $u$ at step $\ell$:
\begin{equation}\label{eq:ctan_discretization}
    \mathbf{h}^{\ell}_u = \mathbf{h}^{\ell-1}_u +\epsilon \sigma \Bigl((\mathbf{W}_\ell-\mathbf{W}_\ell^\top-\gamma\mathbf{I})\mathbf{h}^{\ell-1}_u +\Phi\left(\{\mathbf{h}_v(t),\mathbf{e}_{uv}, t_v^-, t \}_{v\in\mathcal{N}_u^t}\right) \Bigr),
\end{equation}
with $\epsilon>0$ being the discretization step size.
We notice that the antisymmetric weight matrix $(\mathbf{W}_\ell-\mathbf{W}_\ell^\top)$ is subtracted by the term $\gamma\mathbf{I}$ to preserve the stability of the forward Euler's method, see Section~\ref{sec:discretization_method_de} for a more in-depth analysis. We refer to $\mathbf{I}$ as the identity matrix and $\gamma$ to a hyperparameter that regulates the stability of the discretized diffusion. We note that the resulting neural architecture contains as many layers as the discretization steps, \ie $L = t_e/\epsilon$.

\subsection{Truncated Non-Dissipative Propagation}
As previously discussed, the number of iterations in the discretization (i.e., the terminal time $t_e$) plays a crucial role in the propagation. Specifically, few iterations result in a localized event propagation. Consequently, the non-dissipative event propagation does not reach each node in the graph, causing a \textit{truncated} non-dissipative propagation. This method allows scaling the radius of propagation of information depending on the number of discretization steps, thus allowing for a scalable long-range propagation in C-TDGs. Crucially, we notice that, even with few discretization steps, it is still possible to propagate information from a node $u$ to $z$ (if a path of length $P$ connects $u$ and $z$). As an example, consider the situation depicted in the left segment of Figure~\ref{fig:ctan_method}, where nodes $u$ and $v$ establish a connection at some time $t$, and our objective is to transmit this information to node $z$. In this scenario, we assume $L=1$, thus the propagation is truncated before $z$. Upon the arrival of the event at time $t$, this is initially relayed (due to the constraint of $L=1$) to node $k$, which then captures and retains this information. If a future event at time $t+\tau$ involving node $k$ occurs, its state is propagated, ultimately reaching node $z$. Consequently, the information originating from node $u$ successfully traverses the structure to reach node $z$. 
More formally, if it exists a sequence of (at least $P/L$) successive events, such that each future $i$-th event is propagated to an intermediate node at distance $iP/L$ from $u$, then $u$ is able to directly share its information with $z$. 
Therefore, even with a limited number of discretization steps, the non-dissipative behavior enables scaling the single event propagation to cover a larger portion of the C-TDG.
We also notice that if the number of iterations is at least equal to the longest shortest path in the C-TDG, then each event is always propagated throughout the whole graph. 

\subsection{The CTAN Framework}
We name the framework defined through the above sections as Continuous-Time Graph Antisymmetric Network (CTAN). Note that $\Phi$ in Equations~\ref{eq:our_ode} and \ref{eq:ctan_discretization} can be any function that aggregates nodes and edges states. Then, CTAN can leverage the aggregation function that is more adequate for the specific task. As an exemplification of this, in Section~\ref{sec:ctan_experiments} we leverage the aggregation scheme based on the one proposed by \cite{graphtransformer}:
\begin{equation}\label{eq:ctan_aggregation}
    \Phi\left(\{\mathbf{h}_v(t),\mathbf{e}_{uv}, t_v^-, t \}_{v\in\mathcal{N}_u^t}\right) =\sum_{v \in \mathcal{N}_u^t \cup \{ u \}} \alpha_{uv} \left(\mathbf{V}_n\mathbf{h}_{v}^{\ell-1} + \mathbf{V}_e\hat{\mathbf{e}}_{uv}\right)
\end{equation}
where $\hat{\mathbf{e}}_{uv} = \mathbf{e}_{uv} \| \left(\mathbf{V}(t-t_v^-)\right)$ is the new edge representation computed as the concatenation between the original edge attributes and a learned embedding of the elapsed time from the previous neighbor interaction, $\alpha_{uv}=\textrm{softmax} \left(\frac{\mathbf{q}^{\top}\mathbf{K}}{\sqrt{d}} \right)$ is the attention coefficient 
with $d$ the hidden size of each head, $\mathbf{q} = \mathbf{V}_q\mathbf{h}_u^{\ell-1}$, and $\mathbf{K} = \mathbf{V}_k\mathbf{h}_v^{\ell-1} + \mathbf{V}_e\hat{\mathbf{e}}_{uv}$.

Despite CTAN being designed from the general perspective of layer-dependent weights, it can be used with weight sharing between layers (as in Section~\ref{sec:ctan_experiments}).

\section{Experiments}\label{sec:ctan_experiments}

To evaluate the performance of CTAN, we design two novel temporal tasks which require propagation of long-range information by design, Section~\ref{sec:ctan_long_range} and Section~\ref{sec:pascal}.
Afterward, we assess the performance of the proposed CTAN approach on classical benchmarks for C-TDGs in Section~\ref{sec:exp_jodie}.
We complement these classical benchmarks with a larger evaluation on the TGB framework~\citep{huang2023temporal} in Section~\ref{sec:tgb_results}, showcasing the model capabilities in diverse settings, covering evaluations with (i) improved negative sampling techniques and (ii) transductive and inductive settings.

In Section~\ref{sec:ctan_ablation} we conduct an investigation on the scalability property of CTAN and computational efficiency.
In Appendix~\ref{app:ctan_datasets}, we present comprehensive descriptions and statistics of the datasets. 
We release the long-range benchmarks and the code implementing our methodology and reproducing our analysis at \url{https://github.com/gravins/non-dissipative-propagation-CTDGs}.

\myparagraph{Shared Experimental Settings}
In the following experiments, we consider weight sharing of CTAN parameters across the neural layers.  
We compare CTAN against four popular dynamic graph network methods (\ie DyRep~\citep{dyrep}, JODIE~\citep{jodie}, TGAT~\citep{TGAT}, and TGN \citep{tgn_rossi2020}) and include recent methods GraphMixer~\citep{cong2023we} and DyGFormer~\citep{yu2023towards} for evaluation in long-range tasks. In the TGB experiments we also consider CAWN~\citep{CAW} and TCL~\citep{tcl}. 
To ensure fair comparison and efficient implementation, we implement these methods in our framework.
With the same purpose, we reuse the graph convolution operators in the original literature, considering for all methods the aggregation function defined in Equation~\ref{eq:ctan_aggregation}.
We designed each model as a combination of two components: (i) the DGN (\ie CTAN or a baseline) which is responsible to compute the node representations; (ii) the readout that maps node embeddings into the output space.
The readout is a 2-layer MLP, used in all models with the same architecture. 
We perform hyperparameter tuning via grid search, considering a fixed parameter budget based on the number of graph convolutional layers (GCLs). Specifically, for the maximum number of GCL in the grid, we select the embedding dimension so that the total number of parameters matches the budget; such embedding dimension is used across every other configuration.
We report more detailed information on each task in their respective subsections. Detailed information about hyperparameter grids employed in our experiments are in Tables~\ref{tab:ctan_hyper_param} and \ref{tab:hyper_param_tgb} (Appendix~\ref{app:ctan_hyperparams}).
While we do not directly investigate the optimal terminal time $t_e$ within the hyperparameter space, we implicitly address this aspect through the choice of the step size $\epsilon$ and the maximum number of layers $L$, as they jointly determine the terminal time, \ie $t_e=\epsilon L$.

\subsection{Long Range Tasks}\label{sec:exp_non-diss}
Here, we introduce two temporal tasks which contain long-range interaction (LRI). The first is a \textit{Sequence Classification} task on path graphs (see Section~\ref{sec:graphs}) and the second an extension to the temporal domain of the classification task \textit{PascalVOC-SP} introduced in the Long Range Graph Benchmark~\citep{LRGB}.

\subsubsection{Sequence Classification on Temporal Path Graph}\label{sec:ctan_long_range}

\myparagraph{Setup}
Inspired by the tasks in~\cite{chang2018antisymmetricrnn}, we consider a sequence classification task requiring long-range information on a temporal interpretation of a path graph.
Here, the nodes
of the path graph appear sequentially over time from first to last, \ie each event in the C-TDG connects each node to the previous one in the path graph.
The task objective is to predict the feature observed at the source node in the first event after having traversed the entire temporal path graph, \ie after reaching the last event in the stream.
After the model processes the last event in the graph, the output prediction for the whole graph is computed by a readout that takes as input the updated embedding of the destination node of the last event in the C-TDG.
The task requires models to propagate the information seen at the first node through the entire sequence. Models that exhibit smoothing or dissipative behavior will fail to transmit relevant information to the destination node for longer sequences, resulting in poor performance.

When creating the dataset, we set the feature of the first source node to be either 1 or -1, and we use uniformly random sampled features for intermediate nodes and edges to ensure the only task-relevant information is on the earliest node.
We forward events one at a time to update neighboring nodes representations  (\ie batch size is 1).
We considered graphs of different sizes, from length 3 to 20, to test how long information is propagated, i.e., longer graphs force models to propagate information for longer.
During training, we optimize the binary cross-entropy loss over two classes corresponding to the two possible signals (1 or -1) placed on the initial node.

We performed hyperparameter tuning via grid search, optimizing the accuracy score. We trained the models using the Adam optimizer for a maximum of 20 epochs and early stopping with patience of 5 epochs on the validation Binary Cross Entropy loss. Each experimental run is repeated 10 times for different weight initializations. To give models a fair setting for comparison, the grid is computed considering a budget of $\sim$20k trainable parameters per model.

\myparagraph{Results}
The test accuracy on the sequence classification task is in Table~\ref{tab:results-classif-path-graph}.
CTAN exhibits exceptional performance in comparison to reference state-of-the-art methods.
This result highlights the capability of our method to propagate information seen on the first node throughout long paths. 
Meanwhile, several baseline models struggle in solving such a task because the information is lost through the time-steps:
in practice, informative gradients vanish over time.

Note that, memory-less methods such as TGAT, GraphMixer and DyGformer can not effectively propagate information past the number of layers (\ie hops) used in the neighbor aggregation. Note that while the latter two methods are designed for 1-hop aggregation, TGAT allows for variable number of GCLs aggregations, which we test up to 5. We notice TGAT can solve the task at distance 5, but fails for longer graphs.
JODIE and TGN are memory-based methods, which grants them the ability to solve tasks for longer distances, but being RNN-based methods inherently struggle to maintain long-term dependencies~\citep{279181,chang2018antisymmetricrnn}. TGN fails at distance 7, while JODIE at distance 15.
CTAN on the other hand, better propagates information for longer distances, solving the task even at length 20.
\begin{table}[h]
\centering
\caption{Results of the sequence classification on path graph long-range task, for increasing graph length~$n$. The performance metric is the mean test set accuracy score, averaged over 10 different random weights initializations for each model configuration. Models have a maximum budget of learnable parameters equal to $\sim$20k.}\label{tab:results-classif-path-graph}

\scriptsize
\setlength{\tabcolsep}{1.3pt} 
\begin{adjustbox}{center}
\begin{tabular}{lcccccccc}
\toprule 
 \textbf{Model} & $n$=3 & $n$=5 & $n$=7 & $n$=9 & $n$=11 & $n$=13 & $n$=15 & $n$=20 \\
\midrule
\multicolumn{4}{l}{\textbf{DGNs for C-TDGs}}\\
$\,$ DyGFormer    &      \onenobf{100.0$_{\pm0.0}$} &     42.55$_{\pm16.95}$ &       52.94$_{\pm7.3}$ &      53.02$_{\pm6.06}$ &        51.80$_{\pm9.52}$ &        51.70$_{\pm8.52}$ &       42.80$_{\pm16.25}$ &      42.79$_{\pm19.62}$  \\
$\,$ DyRep        &       \onenobf{100.0$_{\pm0.0}$} &        49.20$_{\pm2.10}$ &       51.00$_{\pm1.76}$ &      47.93$_{\pm2.73}$ &       44.87$_{\pm0.89}$ &       46.73$_{\pm1.55}$ &        48.60$_{\pm2.48}$ &       50.47$_{\pm2.88}$ \\
$\,$ GraphMixer   &       \onenobf{100.0$_{\pm0.0}$} &      42.58$_{\pm21.2}$ &       55.40$_{\pm6.44}$ &       52.80$_{\pm5.56}$ &      44.65$_{\pm19.42}$ &      43.77$_{\pm16.51}$ &       52.49$_{\pm5.36}$ &        52.04$_{\pm8.20}$ \\
$\,$ JODIE        &       \onenobf{100.0$_{\pm0.0}$} &       \onenobf{100.0$_{\pm0.0}$} &       \onenobf{100.0$_{\pm0.0}$} &       \onenobf{100.0$_{\pm0.0}$} &       98.53$_{\pm4.64}$ &        97.40$_{\pm7.99}$ &       60.00$_{\pm14.91}$ &       50.87$_{\pm2.46}$ \\
$\,$ TGAT         &       \onenobf{100.0$_{\pm0.0}$} &       \onenobf{100.0$_{\pm0.0}$} &      50.67$_{\pm4.12}$ &      47.87$_{\pm2.72}$ &       42.67$_{\pm2.15}$ &       43.53$_{\pm0.83}$ &       50.53$_{\pm2.15}$ &       49.07$_{\pm1.55}$ \\
$\,$ TGN          &       \onenobf{100.0$_{\pm0.0}$} &       \onenobf{100.0$_{\pm0.0}$} &       60.20$_{\pm13.2}$ &      48.13$_{\pm1.63}$ &       45.07$_{\pm1.64}$ &        44.40$_{\pm0.64}$ &       48.67$_{\pm2.76}$ &       50.13$_{\pm2.17}$ \\

\midrule
\multicolumn{4}{l}{\textbf{Our}}\\
$\,$ CTAN    &       \onenobf{100.0$_{\pm0.0}$} &      \onenobf{100.0$_{\pm0.0}$} &       \onenobf{100.0$_{\pm0.0}$} &      99.93$_{\pm0.21}$ &        \onenobf{99.6$_{\pm0.56}$} &       \onenobf{98.67$_{\pm1.89}$} &       \onenobf{93.47$_{\pm8.78}$} &      \onenobf{88.93$_{\pm12.06}$} \\
\bottomrule
\end{tabular}
\end{adjustbox}
\end{table}

\subsubsection{Classification on Temporal Pascal-VOC}\label{sec:pascal}
\myparagraph{Setup}
We consider edge classification on a temporal interpretation of the PascalVOC-SP dataset, which has been previously employed by \cite{LRGB} as a benchmark to show the efficacy of capturing LRI in static graphs. 
Here, we adapt the task to the C-TDG domain: we forward edges one at a time and predict the class of the destination node.
We generate temporal graphs by considering that nodes in each rag-boundary graph appear from the top-left to the bottom-right of the image, sequentially.
We consider two degrees of SLIC superpixels compactness, \ie 10 and 30. Larger compactness means more patches, with less information included in each patch and more to be propagated.

To benchmark the ability of models to propagate information through the graph, we test model performance for an increasing number of GCLs. 
Fewer GCLs require models to store and transmit relevant information along node embeddings rather than relying on effectively aggregating information from increasingly larger neighborhoods.

We performed hyperparameter tuning via grid search, optimizing the F1-score. We trained the models using the Adam optimizer for a maximum of 200 epochs and early stopping with patience of 20 epochs on the validation score. Each experimental run is repeated 5 times for different weight initializations. To give models a fair setting for comparison, the grid is computed considering a budget of $\sim$40k trainable parameters per model and the neighbor sampler size is set to 5.

\myparagraph{Results}
Table~\ref{tab:pascal_results} reports the average F1-score on the temporal PascalVOC-SP task. 
Note that DyRep, JODIE, GraphMixer and DyGFormer, in their original definition, do not support a variable number of GCLs, hence the results of such models are presented in the table under ``1 GCL'' for clarity.
CTAN largely outperforms reference methods. 
\begin{table}[htb]
\centering
\caption{Results of the classification on the Temporal PascalVOC task, for increasing number of GCLs. The performance metric is the mean test set F1-score, averaged over 5 different random weights initializations for each model configuration.}\label{tab:pascal_results}

\scriptsize

\begin{tabular}{l|ccc|ccc}
\toprule
 & \multicolumn{3}{c|}{\textbf{Temporal Pascal VOC (sc=10)}} & \multicolumn{3}{c}{\textbf{Temporal Pascal VOC (sc=30)}}\\
no. GCLs & 1 & 3 & 5 & 1 & 3 & 5\\
\midrule
\multicolumn{4}{l}{\textbf{DGNs for C-TDGs}}\\
$\,$ DyGFormer          & \one{8.45$_{\pm0.13}$} & $-$ & $-$ & 8.07$_{\pm0.27}$ & $-$  &   $-$  \\
$\,$ DyRep        & 5.29$_{\pm0.47}$ & $-$ & $-$ & 5.23$_{\pm0.11}$ & $-$  &  $-$   \\
$\,$ GraphMixer          & 6.60$_{\pm0.11}$ & $-$ & $-$ & 5.88$_{\pm0.08}$ & $-$  &   $-$  \\
$\,$ JODIE        & 6.33$_{\pm0.41}$ & $-$ & $-$ & 5.76$_{\pm0.35}$ & $-$ & $-$   \\
$\,$ TGAT         & 5.39$_{\pm0.19}$ & 6.53$_{\pm0.58}$ & 8.23$_{\pm0.73}$ & 6.04$_{\pm0.26}$ & 8.79$_{\pm0.29}$  &  10.38$_{\pm0.7}$   \\
$\,$ TGN          & 6.04$_{\pm0.27}$ & 6.55$_{\pm0.46}$ &  7.51$_{\pm0.80}$ & 5.59$_{\pm0.24}$ & 7.26$_{\pm0.82}$  &   7.90$_{\pm1.31}$  \\

\midrule
\multicolumn{4}{l}{\textbf{Our}}\\
$\,$ CTAN & 7.89$_{\pm0.33}$ & \one{8.53$_{\pm1.06}$} & \one{8.88$_{\pm0.98}$} & \one{9.98$_{\pm0.33}$} & \one{10.16$_{\pm0.52}$} & \one{10.41$_{\pm0.52}$} \\
\bottomrule

\end{tabular}
\end{table}

We observe that for SLIC compactness equal to 30, CTAN achieves a 65\% and 16\% improvement against the second best performing model (\ie TGAT), for one and three GCLs, respectively.
Interestingly, TGAT almost matches the performance of CTAN when considering five GCLs.
This is in line with the excellent results of computationally expensive Transformers-based models in the static case~\citep{LRGB}, corroborating the advantages of self-attention blocks in modeling long-range dependencies between far away nodes.
This result also suggests that the majority of the relevant information necessary to solve the temporal Pascal VOC task may lie within neighborhoods five hops away.
We note that at SLIC compactness 10, DyGFormer benefits from the shorter long-range propagation (when sc=10 the graph contains fewer patches, hence fewer nodes and spatially closer relevant information compared to sc=30), and from its deeper architecture compared to CTAN's single-layer design, when considering the same number of spatial hops. In fact, in this setting DyGFormer contains two transformer blocks, while CTAN does not. However, we observe that by including multiple layers of CTAN (i.e., $\text{no.GCLs}>1$), our method effectively propagates information and outperforms DyGFormer even in the sc=10 task.
Nevertheless, the results indicate how CTAN is capable of propagating relevant information across the time-steps to achieve accurate predictions, even when the model is only allowed to extract information from limited, very local neighborhoods.

\subsection{Future Link Prediction Tasks}\label{sec:exp_jodie}
\myparagraph{Setup} 
For the C-TDG benchmarks we consider four well-known datasets proposed by \cite{jodie} (\ie Wikipedia, Reddit, LastFM, and MOOC) to assess the model performance in real-world setting, with the task of future link prediction. We consider as additional baseline EdgeBank~\citep{edgebank} with the aim of showing the performance of a simple heuristic that merely stores previously observed interactions (without any learning), and then predicts stored links as positive. 

We performed hyperparameter tuning via grid search, optimizing the AUC score. We trained the models using the Adam optimizer for a maximum of 1000 epochs and early stopping with patience of 50 epochs on the validation score. Each experimental run is repeated 5 times for different weight initializations. To give models a fair setting for comparison, the grid is computed considering a budget of $\sim$140k trainable parameters per model and the neighbor sampler size is set to 5.

\myparagraph{Results}
Table~\ref{tab:fair_results_complete} reports the average test AUC on the C-TDG benchmarks. CTAN shows remarkable performance, ranking first across datasets.
Our method achieves a score that on average is 4.7\% better than other baselines. 
This finding shows the importance of a non-dissipative behavior of the method even on real-world tasks, since more information need to be retained and propagated from the past to improve the final performance.
Our results demonstrate that CTAN is able to better capture and exploit such information.
Nevertheless, note that not all real-world datasets inherently present long-range dependencies. To evaluate how CTAN fares against state-of-the-art methods on several datasets, we complement this analysis with an evaluation on the TGB Benchmark, see Section~\ref{sec:tgb_results}. In this setting, CTAN characterizes by the best performing behavior when considering the combination of TGB datasets. 

\begin{table}[ht]
\centering
\caption{Mean test set AUC and std in percent averaged over 5 random weight initializations. Each model have a maximum budget of learnable weights equal to $\sim$140k. The higher, the better. \one{First}, \two{second}, and \three{third} best results for each task are color-coded.}\label{tab:fair_results_complete}

\scriptsize
\begin{tabular}{lcccc}
\toprule
& \textbf{Wikipedia} & \textbf{Reddit} & \textbf{LastFM} & \textbf{MOOC}

\\
\midrule
\textbf{Baseline}\\
$\,$ EdgeBank$_{1\% \,tr\, set}$         & 71.03 
                            & 71.92 
                            & 77.59 
                            & 61.29 
                            \\
$\,$ EdgeBank$_{5\% \,tr\, set}$         & 81.65 
                            & 85.07 
                            & 86.75 
                            & 63.93 
                            \\
$\,$ EdgeBank$_{10\% \,tr\, set}$        & 85.26 
                            & 89.07 
                            & 89.87 
                            & 65.18 
                            \\
$\,$ EdgeBank$_{25\% \,tr\, set}$        & 88.31 
                            & 92.92 
                            & 92.74 
                            & 67.49 
                            \\
$\,$ EdgeBank$_{50\% \,tr\, set}$        & 90.29 
                            & 94.82 
                            & 94.06 
                            & 69.63 
                            \\
$\,$ EdgeBank$_{75\% \,tr\, set}$        & 91.11 
                            & 95.63 
                            & 94.55 
                            & 70.46 
                            \\
$\,$ EdgeBank$_{100\% \,tr\, set}$       & 91.52 
                            & 96.08 
                            & 94.69 
                            & 70.80 
                            \\
$\,$ EdgeBank$_\infty$           & 91.82 
                            & 96.42 
                            & \one{94.72} 
                            & 70.85 
                        \\
\midrule
\multicolumn{4}{l}{\textbf{DGNs for C-TDGs}}\\
$\,$ DyRep & 88.64$_{\pm0.15}$ 
      & 97.51$_{\pm0.10}$ 
      & 77.89$_{\pm1.39}$ 
      & 81.87$_{\pm2.47}$ 
      \\
$\,$ JODIE & 94.68$_{\pm1.05}$ 
      & 96.34$_{\pm0.83}$ 
      & 69.76$_{\pm2.74}$ 
      & 81.90$_{\pm9.03}$ 
      \\
$\,$ TGAT & \three{94.91$_{\pm0.25}$} 
     & \three{98.18$_{\pm0.05}$} 
     & \three{81.53$_{\pm0.34}$} 
     & \three{87.61$_{\pm0.15}$} 
     \\
$\,$ TGN & \two{95.60$_{\pm0.18}$} 
    & \two{98.23$_{\pm0.10}$} 
    & 79.18$_{\pm0.79}$ 
    & \two{90.74$_{\pm0.99}$}
\\
\midrule
\textbf{Our}\\
$\,$ CTAN & \one{97.55$_{\pm0.09}$} 
          & \one{98.61$_{\pm0.04}$} 
          & \two{83.81$_{\pm0.92}$} 
          & \one{92.47$_{\pm0.78}$} 
          \\
\bottomrule 
\end{tabular}
\end{table}

Lastly, we note that Table~\ref{tab:fair_results_complete} includes the performance of EdgeBank with different time window sizes. We recall that
EdgeBank is a memorization-based method without learning that simply stores previously observed edges from a fixed-size time-window from the immediate past, and predicts stored edges as positive. We evaluated EdgeBank with different time windows spanning from a size of 1\% of the training set to infinite size, \ie all observed edges are stored in memory.

In this scenario, EdgeBank is particularly good at capturing long-range information along the time dimension in the LastFM task, surpassing all the baselines and CTAN as the time window increases. We highlight that the experiments in this section are meant to outline how CTAN outperforms baselines under an \textit{even field} of number of trainable parameters (\ie 140k) and restricted range of hyperparameter values, \eg sampler size equal to 5. On the other hand, EdgeBank is a non-parametric method that at the time of inference accesses the entire temporal adjacency matrix. In LastFM, the median node degree after training is 903 (mean $1152\pm1722$), which is high compared to other datasets. At validation time, for the average node in LastFM, EdgeBank pools information from 903 node neighbors, while in our setting we allow DGN baselines to pool information from 5 randomly sampled neighbors. As nodes have larger degrees, sampling larger neighborhoods is fundamental to access and therefore retain information. To show that CTAN performance is limited by the considered range of hyperparameter values, we present, in Table~\ref{tab:lastfm-investigation}, the performance of CTAN by solely adjusting the neighbor sampler size, while maintaining a budget of $\sim$350k learnable parameters. The evaluation involves substituting various sampler size values into the optimal combination of hyperparameters obtained for CTAN on the LastFM dataset in Table~\ref{tab:fair_results_complete}, with the embedding dimension configured to achieve the target of $\sim$350k learnable parameters (\ie 192). The results indicate that CTAN performs better by adjusting the sampler size alone.
\begin{table}[h]
\centering
\caption{Mean test set AUC and std on LastFM (in percent) for increasing size of sampled neighbors, averaged over three different weights initializations. The model has a budget of learnable weights equal to $\sim$350k. When nodes have large degrees as in LastFM, accessing larger neighborhoods with the neighbor sampler is fundamental to access and retain important information.}\label{tab:lastfm-investigation}
\scriptsize
\begin{tabular}{lcccccc}
\toprule
\textbf{Sampler size} & 2 & 8 & 16 & 32 & 64 & 128 \\
\midrule
CTAN         &    82.64$_{\pm0.93}$   &  86.21$_{\pm0.58}$   &   86.16$_{\pm0.55}$   &   86.27$_{\pm0.55}$   &    86.32$_{\pm0.81}$ & \one{87.82$_{\pm0.42}$}   \\
\bottomrule
\end{tabular}
\end{table}

\subsection{TGB Benchmarks}\label{sec:tgb_results}

\myparagraph{Setup}
We evaluate CTAN on the Temporal Graph Benchmark (TGB)~\citep{huang2023temporal}. 
TGB contains a set of real-world small-to-large scale benchmark datasets with varying graph properties. 
We focus on dynamic link property prediction tasks, \ie tgbl-wiki-v2, tgbl-review-v2, tgbl-coin-v2, tgbl-comment.
To overcome the existing limitations on negative edge sampling, \ie where only one random negative edge is sampled per each positive edge, TGB provides pre-sampled negative edge sets with both \textit{random} and \textit{historical} negatives~\citep{edgebank}.
Here, for each positive edge, several negatives are sampled for the  evaluation~\citep{huang2023temporal}. Note that for computational efficiency, since validation passes are extremely costly given the large number of negative edges, we only do a validation pass every three training epochs.

We performed hyperparameter tuning via grid search, optimizing the Mean Reciprocal Rank. We trained our model on tgbl-wiki-v2 using the Adam optimizer for a maximum of 200 epochs and early stopping with patience of 20 epochs on the validation score, with each experimental run repeated 5 times for different weight initializations. On the other tasks, we trained our model for a maximum of 50 epochs, with early stopping equal to 3 and 3 different runs.

\myparagraph{Results}
In Table~\ref{tab:tgb_results}, we report the test Mean Reciprocal Rank (MRR) for the experiments.
We note that CTAN performs quite well in general: its average rank across the four datasets is 3.25 which is the highest, together with DyGFormer.
CTAN performs quite well on tgbl-review-v2, even significantly outperforming state-of-the-art methods DyGFormer and GraphMixer.
In such dataset, the surprise index (namely, the proportion of unseen edges at test time~\citep{edgebank}) is 0.987, meaning that nodes do not have large histories. In this case, it seems that CTAN  better propagates information from neighbors compared to methods focusing on first-hop information passing such as GraphMixer and DyGFormer.
On the other hand, it seems that DyGFormer is well suited in propagating long-range \textit{time} information by modeling a large number of previous node interactions within the transformer input sequence, given enough computational budget, particularly in tgbl-wiki-v2, where nodes have long histories.
Nevertheless, we notice that even with limited number of parameters, CTAN is extremely competitive within the leaderboard.

\begin{table}[ht]
\centering
\caption{Results of the Dynamic Link Property Prediction task on the TGB benchmark datasets~\citep{huang2023temporal}. The table reports the average MRR on the test split of the datasets over the considered weight initializations. For CTAN, the average is taken over a maximum of five runs with different random seeds for different weight initializations. All baselines' results are taken from~\cite{yu2023empirical}. The number of parameters is computed from the TGB Baselines repository~\citep{huang2023temporal} by loading the best performing model across the model selection search. \one{First}, \two{second}, and \three{third} best results for each task are color-coded.}
\label{tab:tgb_results}
\scriptsize
\begin{tabular}{lc|cccc|c}
\hline\toprule 
\multirow{2}{*}{\textbf{Model}}&\multirow{2}{*}{\textbf{N. params}} & \textbf{tgbl-}  & \textbf{tgbl-}  & \textbf{tgbl-} & \textbf{tgbl-}& \textbf{Avg.}\\
& & \textbf{wiki-v2} & \textbf{review-v2} & \textbf{coin-v2} &\textbf{comment} &\textbf{rank}\\
\midrule
\textbf{Baseline}\\
$\,$ EdgeBank$_\infty$         & $-$  & 52.50         &  2.29         & 35.90 & 10.87 & 11 \\
$\,$ EdgeBank$_{\text{tw-ts}}$ & $-$  & 63.25         &  2.94         & 57.36 & 12.44 & 8.25 \\
$\,$ EdgeBank$_{\text{re}}$    & $-$  & 65.88         &  2.84         & 59.15 & $-$  & 8.25 \\
$\,$ EdgeBank$_{\text{th}}$    & $-$  & 52.81         &  1.97         & 43.36 & $-$ & 11.33 \\
\midrule
\multicolumn{7}{l}{\textbf{DGNs for C-TDGs}}\\
$\,$ CAWN & 4M & \three{73.04$_{ \pm 0.60}$}  & 19.30$_{ \pm 0.10}$ & $-$  & $-$ & 5.50\\
$\,$ DyRep & 700k & 51.91$_{ \pm 1.95}$  & \three{40.06$_{ \pm 0.59}$} & 45.20$_{ \pm 4.60}$ & 28.90$_{ \pm 3.30}$ & 8.00 \\
$\,$ GraphMixer & 600k & 59.75$_{ \pm 0.39}$  & 36.89$_{ \pm 1.50}$ & \one{75.57$_{ \pm 0.27}$} & \one{76.17$_{ \pm 0.17}$} & 4.25\\
$\,$ DyGFormer & 1.1M & \one{79.83$_{ \pm 0.42}$}  & 22.39$_{ \pm 1.52}$ & \two{75.17$_{ \pm 0.38}$} & 67.03$_{ \pm 0.14}$ & \one{3.25} \\
$\,$ JODIE & 200k & 63.05$_{ \pm 1.69}$  & \one{41.43$_{ \pm 0.15}$} & $-$ & $-$  & 4.50 \\
$\,$ TCL & 900k & \two{78.11$_{ \pm 0.20}$}  & 16.51$_{ \pm 1.85}$ & 68.66$_{ \pm 0.30}$ & \two{70.11$_{ \pm 0.83}$} & 4.25 \\
$\,$ TGAT & 1.1M & 59.94$_{ \pm 1.63}$  & 19.64$_{ \pm 0.23}$ & 60.92$_{ \pm 0.57}$ & 56.20$_{ \pm 2.11}$ & 6.50 \\
$\,$ TGN & 1M & 68.93$_{ \pm 0.53}$ &  37.48$_{ \pm 0.23}$ & 58.60$_{ \pm 3.70}$ & 37.90$_{ \pm 2.00}$  & 5.25  \\
\midrule
\textbf{Our}\\
$\,$ CTAN & 600k & 66.76$_{ \pm 0.74}$  & \two{40.52$_{ \pm 0.41}$} & \three{74.82$_{ \pm 0.42}$} & \three{67.10$_{ \pm 6.72}$} & \one{3.25} \\

\bottomrule\hline

\end{tabular}

\end{table}

\subsection{Ablation Study}\label{sec:ctan_ablation}
In this section, we investigate the scalability property of CTAN and its computational efficiency.

\begin{figure}[h]
\begin{center}
    \includegraphics[width=\linewidth]{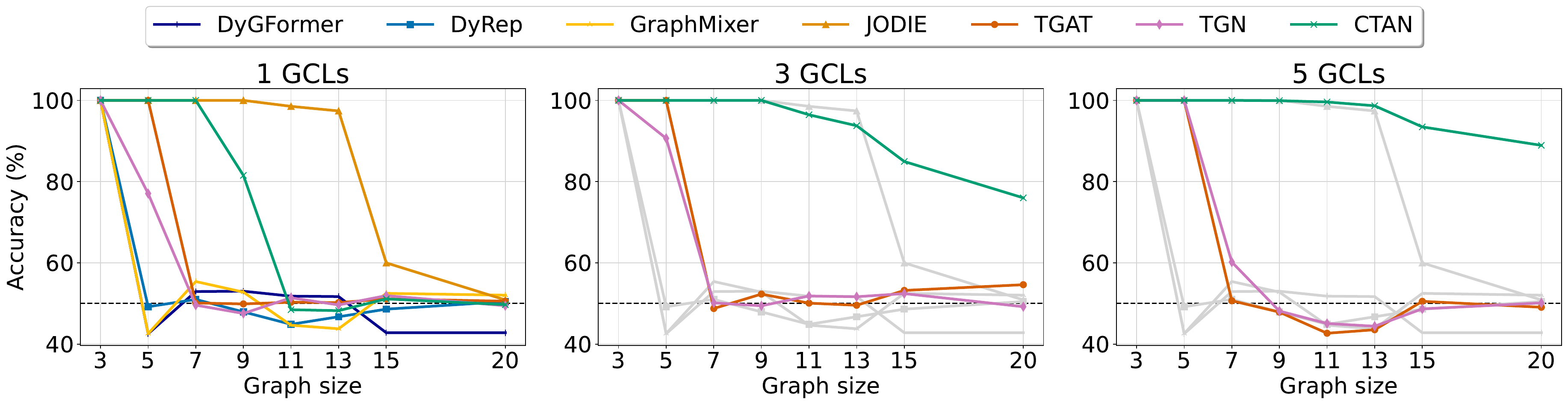}
\end{center}
\caption{Mean accuracy on the T-PathGraph task on the experiment of Section~\ref{sec:ctan_long_range}, with distinction between the performance at different number of GCLs (whenever possible). With 3 and 5 GCLs we report in grey the results of DyGFormer, DyRep, GraphMixer, and JODIE, which are designed for 1-hop aggregation only. The plots show that not only CTAN can better retain information at low number of GCLs, but also that increasing the number of GCL enables solving the T-PathGraph task on longer graphs, where the task is harder because information needs to be propagated further away. The number of GCL allows CTAN to \textit{scale} up the range of information propagation.}
\label{fig:scalability-of-ctan}
\end{figure}

\myparagraph{Scalability of CTAN}\label{sec:ctan_scalability}
We conduct an investigation on the scalability property of CTAN. 
Note that while in some related works the term scalable refers to the computational complexity of methods, here we use scalable to refer to how the range of information propagation can be controlled by increasing the number of graph convolutions in CTAN.
To show this property, we assess the task in Section~\ref{sec:ctan_long_range} for different values of GCLs (when possible).
We report the results in Figure~\ref{fig:scalability-of-ctan}, which shows how for increasing GCLs, CTAN is capable of conveying information further away in the graph compared to other graph convolutional based models.
In addition, we observe that both DyGFormer and GraphMixer may have increased capability to capture long-range dependencies, however, this is only applicable to \emph{time}-only dependencies, and not spatial ones. Indeed, DyGFormer and GraphMixer model long-range time dependencies on node representations by fetching previous interactions for a node, both only relying on first-hop neighbors information and not considering spatial propagation of higher-order node information, which is in fact mentioned as a limitation of DyGFormer. Comparably, CTAN remains a graph convolution-based model, hence capable of propagating information in a non-dissipative way over time as well as over the spatial dimension of the graph, scaling the range of propagation with the number of discretization steps (equivalently, the termination time $t_e$). This property enables propagating information to neighbors beyond first-hop ones, which in turns allows solving tasks such as those in Sections~\ref{sec:ctan_long_range} and \ref{sec:pascal}.

\begin{figure}[h]
\centering
    \includegraphics[width=0.9\linewidth]{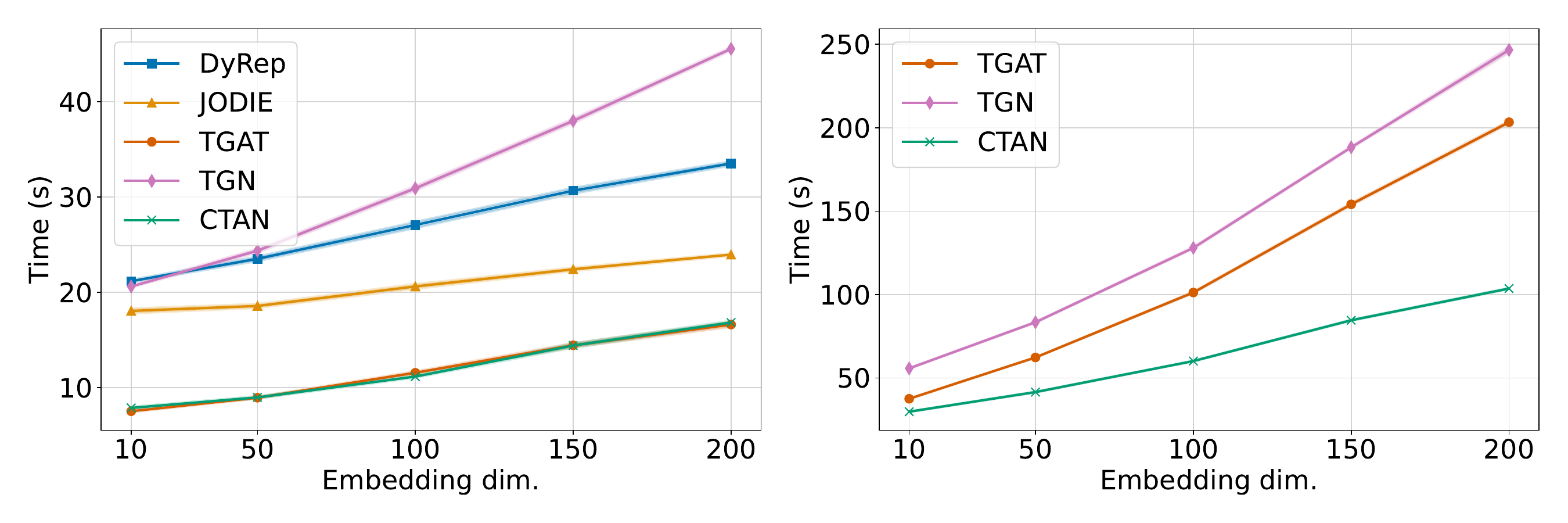}
    {\\\footnotesize\hspace{0.4cm}(a)\hspace{6cm}(b)}
\caption{Average time per epoch (measured in seconds) and std with respect to the embedding size computed on the Wikipedia dataset, averaged over 10 epochs. The experiments were carried out on an Intel(R) Xeon(R) Gold 6278C CPU @ 2.60GHz. On the left (a), each model has 1 DGN layer (when possible), while on the right (b) the models have 5 GCLs.}
\label{fig:ctan_time}
\end{figure}

\myparagraph{Runtimes}
To show the computational advantage of our CTAN, we report in Table~\ref{tab:ctan_time} the average time per epoch (measured in seconds) for each model on the four considered link prediction datasets in Section~\ref{sec:exp_jodie}. In this evaluation, each model has the same embedding dimension and number of GCLs. Similarly, Figure~\ref{fig:ctan_time} shows the average time per epoch of each model on the Wikipedia dataset. Here, the time is reported with respect to a varying embedding size and similar number of GCLs. We observe that our method has a speedup on average of $1.3\times$ to $2.2\times$ on the four benchmarks when one layer of graph convolutions is considered, and $1.5\times$ to $1.9\times$ when five layers are used.

\begin{table}[h!]
\centering
\caption{Mean time (in seconds) and std averaged over 10 epochs. Each model is run with an embedding dimension equal to 100 on an Intel(R) Xeon(R) Gold 6278C CPU @ 2.60GHz.}
\label{tab:ctan_time}
\scriptsize
\begin{tabular}{rlcccc}
\toprule
&   \textbf{Model} &    \textbf{Wikipedia}   &     \textbf{Reddit}     &       \textbf{LastFM}     &     \textbf{MOOC} \\\midrule
\multirow{7}{*}{1 layer} & \multicolumn{4}{l}{\textbf{DGNs for C-TDGs}}\\
& $\,$ DyRep     & 27.07$_{\pm0.32}$       & 161.43$_{\pm0.96}$      & 216.88$_{\pm2.83}$        & 53.32$_{\pm0.56}$ \\
& $\,$ JODIE     & 20.62$_{\pm0.24}$       & 131.71$_{\pm0.85}$      & 176.61$_{\pm3.02}$        & 43.92$_{\pm0.68}$ \\
& $\,$ TGAT      & 11.56$_{\pm0.14}$       & 67.83$_{\pm0.64}$       & 139.79$_{\pm20.78}$       & \one{33.92$_{\pm0.50}$} \\
& $\,$ TGN       & 30.92$_{\pm0.25}$       & 196.87$_{\pm1.35}$      & 289.22$_{\pm30.38}$       & 53.46$_{\pm0.62}$\\
\cmidrule{2-6}
&\multicolumn{4}{l}{\textbf{Our}}\\
& $\,$ CTAN       & \one{11.16$_{\pm0.11}$} & \one{64.48$_{\pm0.56}$} & \one{123.19$_{\pm11.33}$} & 34.42$_{\pm0.50}$ \\
\midrule
\multirow{5}{*}{5 layer} & \multicolumn{4}{l}{\textbf{DGNs for C-TDGs}}\\
& $\,$ TGAT        & 101.26$_{\pm0.46}$      & 895.35$_{\pm5.46}$       & 862.47$_{\pm217.38}$       & 73.77$_{\pm1.29}$ \\ 
& $\,$ TGN       & 127.99$_{\pm0.60}$      & 1099.19$_{\pm3.91}$      & 1034.24$_{\pm221.04}$      & 95.45$_{\pm1.07}$\\
\cmidrule{2-6}
&\multicolumn{4}{l}{\textbf{Our}}\\
& $\,$ CTAN       & \one{60.16$_{\pm0.20}$} & \one{532.36$_{\pm9.87}$} & \one{495.18$_{\pm111.13}$} & \one{56.19$_{\pm0.63}$} \\

\bottomrule
\end{tabular}
\end{table}

\section{Related Work}
Nowadays, most of the DGNs tailored for learning C-TDGs can be generalized within the Temporal Graph Network (TGN) framework \citep{tgn_rossi2020}\footnote{We refer the reader to Section~\ref{sec:survey_ctdg} for a deeper discussion on the TGN framework.}.
Many state-of-the-art architectures~\citep{jodie, dyrep, TGAT, streamgnn, pint} fit this framework, with later methods outperforming earlier ones thanks to advances in the \emph{local} message passing or even in the encoding of positional features.
Two recent methods \cite{cong2023we} and \cite{yu2023towards} focus on modeling long-range (temporal) dependencies by including longer node histories in the context while not relying on memory modules, as in the TGN framework.
While recent methods often provide improved results, none of them explicitly models long-range \textit{temporal and spatial} dependencies between nodes or events in the C-TDG. 
As increasingly evidenced both in sequence-model architectures~\citep{chang2018antisymmetricrnn}, and in the static graph case~\citep{LRGB}, propagating information across various time steps is extremely beneficial for learning.
CTAN, instead, provably enables effective long-range propagation by design.
Note that our approach does not require the co-existence of memory \textit{and} graph propagation module, as in the TGN framework. CTAN stores all necessary information within the node embeddings themselves as computed by the graph convolution, while achieving non-dissipative propagation by design.
This makes CTAN more lightweight. 
Lastly, as TGN allows for different graph propagation modules, the general formulation of the aggregation function $\Phi$ in Equation~\ref{eq:ctan_discretization} allows extending state-of-the-art DGNs for static graphs to the domain of C-TDGs through the lens of non-dissipative and stable ODEs.   

Lastly, we compare CTAN with A-DGN (see Section~\ref{sec:ADGN}), an ODE-based model achieving non-dissipative propagation through \emph{static} graphs, \ie in the time-unaware spatial domain. We note that time-aware nodes and edges combined with possibly irregularly sampled repetitive edges between the same pair of nodes natively render A-DGN (as well as other methods designed for static graphs) inapplicable to C-TDGs. 
Less trivially, non-dissipative propagation in C-TDGs cannot be achieved through mere non-dissipative propagation through space. On the contrary, non-dissipative propagation of information through time is a property unique to DGNs designed for C-TDG, necessary for their overall non-dissipativeness.

To the best of our knowledge, we are the first to propose an ODE-based architecture suitable for C-TDGs that can effectively propagate long-range information between nodes.

\section{Summary}
In this chapter, we have presented \emph{Continuous-Time Graph Antisymmetric Network} (CTAN), a new framework based on stable and non-dissipative ODEs for learning long-range interactions in Continuous-Time Dynamic Graphs (C-TDGs). Differently from previous approaches, CTAN's formulation allows scaling the radius of effective propagation of information in C-TDGs (\ie allowing for a scalable long-range propagation in C-TDGs) and reimagines state-of-the-art static DGNs as a discretization of non-dissipative ODEs for C-TDGs. To the best of our knowledge, CTAN is the first framework to address the long-range propagation problem in C-TDGs, while bridging the gap between ODEs and C-TDGs.

Our experimental investigation reveals, at first, that when it comes to capturing long-range dependencies in a task, our framework significantly surpasses state-of-the-art DGNs for C-TDGs. Our experiments indicate that CTAN is capable of propagating relevant information incrementally across time to achieve accurate predictions, even when the model is only allowed to extract information from very local neighborhoods, \ie by using only a single or few layers. Thus, CTAN enables scaling the extent of information propagation in C-TDG data structures without increasing the number of layers nor incurring in dissipative behaviors.
Moreover, our results indicate that CTAN is effective across various graph benchmarks in both real and synthetic scenarios. In essence, CTAN showcased its ability to explore long-range dependencies (even with limited resources), suggesting its potential in mitigating over-squashing in C-TDGs.

We believe that CTAN lays down the basis for further investigations of the problem of oversquashing and long-range interaction learning in the C-TDG domain. 


\chapter{Conclusions}\label{ch:conclusions}
In this thesis, we have investigated novel information propagation dynamics in DGNs for both static and dynamic graphs, integrating concepts from dynamical systems. 
Despite the 
progress of recent years, learning effective information propagation patterns remains a critical challenge that heavily influences the DGNs' capabilities.
With this aim, we highlighted the pivotal role of differential-equations-inspired DGNs (DE-DGNs) in addressing the challenges of learning \emph{long-term dependencies} in graphs and complex spatio-temporal patterns from \emph{irregular and sparsely sampled} data. 

We started this dissertation with the broader objective of fostering the research in the graph representation learning domain by reviewing the principles underlying DGNs and their limitations in information propagation. We followed with a survey that focused on recent representation learning techniques for dynamic graphs under a unified formalism 
and a fair performance comparison among the most popular methods, thus filling the fragmented and scattered literature in terms of model formalization, empirical setups, and performance benchmarks.

Afterward, we moved our focus to the main objective of this thesis. We presented multiple frameworks achieved from the study of ordinary differential equation (ODE) representing a continuous process of information propagation on both static and dynamic graphs. 

Starting from static graphs, we studied how non-dissipative dynamical systems can provide a general design principle for introducing non-dissipativity as an inductive bias in any DE-DGN. Therefore, we explored the benefits of a DE-DGN with antisymmetric weight parametrization, theoretically proving that the differential equation corresponding to our framework is stable, non-dissipative as well as has improved capacities of propagating information over long radii. Consequently, typical problems of systems with unstable and lossy dynamics do not occur. 

We improved the node-wise non-dissipative behavior introduced with our first framework with two methods. The first includes antisymmetric constraints on both the space domain, \ie the neighborhood aggregation function, and the weight domain, with the goal of achieving graph- and node-wise non-dissipative behavior, so that more information is preserved and conveyed during propagation. The second approach explores a new message-passing scheme based on the principles of physics-inspired dynamical systems, \ie port-Hamiltonian systems.
Therefore, leveraging the connection with such systems, we provide theoretical guarantees that information is conserved, thus our method allows the preservation and propagation of long-range information by obeying the conservation laws. Moreover, we showed how additional forces can be used to deviate from this purely conservative behavior, potentially increasing effectiveness in the downstream task. 

Afterward, we moved our focus to space and time propagation in dynamic graphs. We started from the domain of D-TDGs, whose approaches are usually restricted to work solely on regularly sampled data. Therefore, we addressed the problem of learning complex information propagation  patterns from irregular and sparsely sampled data, typical of real-world complex scenarios. In this case, it is fundamental for novel approaches to thoroughly comprehend the underlying dynamics of information flow to effectively solve the task. With this aim, we showed that thanks to the connection between ODEs and neural architectures, we can naturally handle arbitrary time gaps between observations, allowing to address the most common limitation of DGNs for D-TDGs. 
Our last contribution of this dissertation fuses the aforementioned problems and learns complex spatio-temporal information patterns from an irregular domain (\ie that of C-TDGs) while exhibiting a non-dissipative behavior that allows long-range spatio-temporal propagation. We leveraged antisymmetric weight parametrization and established theoretical conditions for achieving spatio-temporal non-dissipation enabling the transmission of information of
a past event as new events occur, since node states are used to efficiently retain and propagate historical information.

We formulate each of the proposed frameworks from a general perspective, allowing the incorporation of the most appropriate neighborhood aggregation function for the task at hand. Hence, our methods can be used to reinterpret and extend most of the classical DGNs as neural ODEs for graph, enabling for long-range propagation and/or learning from irregular and sparsely sampled data.
Overall, our experimental investigations reveal that when it comes to capturing long-range dependencies within graphs (either static or dynamic), our frameworks significantly surpass state-of-the-art DGNs, indicating that our methods are capable of propagating relevant information incrementally across space and time to achieve accurate predictions. At the same time, our experiments show that thanks to the connection between ODEs and neural architectures, we can naturally handle arbitrary time gaps between observations and improve literature approaches by a large margin on real-world problems.

Furthermore, our analysis shows that our methods remain within the computational complexity bounds of other fast, standard literature methods while setting new state-of-the-art standards. We note that, in the static case, introducing antisymmetric aggregation adds complexity compared to using antisymmetry solely in the weight domain, especially in the case of learned aggregation terms. Similarly, the port-Hamiltonian framework enhances long-range propagation capabilities even more, albeit at a computational cost of decoupling the state computation into two components and accounting not only for the self-node evolution but also for the neighbor's evolution influence. This requires additional aggregation steps, increasing the computational overhead. Therefore, tasks with resource constraints may benefit from our first framework, which is constrained only to the weight domain, offering a balance between performance and efficiency.
Importantly, while the port-Hamiltonian framework is more beneficial for capturing intricate long-range dependencies, its added complexity may be less suitable for low-resource environments. By highlighting this performance-cost tradeoff, we emphasize that the choice between these approaches should be guided by the specific resource limitations and accuracy requirements of the task.

%
\vfill

\section{Future Directions}
Looking ahead to future developments, there are a number of potential directions to be investigated in the future, which broaden across multiple areas.

All the discussed frameworks employ Euler's discretization schemes for the ease of simplicity. However, as discusses in Section~\ref{sec:discretization_method_de}, there are several discretization techniques that can be employed. Looking at more sophisticated numerical methods, adaptive multistep schemes~\citep{Ascher1998, dormand1996numerical} allows us to dynamically adjust the step size based on local error estimates. As a result, both the accuracy and efficiency of the solution are enhanced. In the static domain, the adaptive step size enables the adaptation of the number of layers to the specific requirements of each task, while in the temporal domain, it enables ad-hoc event propagation tailored to diverse and complex data patterns, improving propagation flexibility. 
Additionally, these schemes reduce the need to manually tune the step size hyperparameter, simplifying the model selection process. 

Another interesting future direction is that of designing DE-DGNs by leveraging theory from state space models. State space models (SSMs)~\citep{decarlo1989linear, slotine1991applied} are mathematical models used to describe the behavior of dynamic systems and are widely employed in control theory and signal processing. Recently, SSM-based neural architectures have demonstrated state-of-the-art performance in learning time-series data~\citep{NEURIPS2019_952285b9, s4, s4d}, particularly excelling in handling sequences with very long (potentially unbounded) dependencies while being highly efficient and requiring fewer training parameters compared to current literature methods. Thus, we believe that integrating SSM architectures into DE-DGNs can be beneficial, especially for propagating long-range spatio-temporal dependencies.

Another promising avenue for future research is the use of adaptive message passing schemes to improve the non-dissipative behavior of the resulting DE-DGN. Recent approaches~\citep{finkelshtein2024cooperative, errica2024adaptive} propose to learn a generalization of message passing by allowing each node to decide how to propagate information from or to its neighbors, allowing for a more flexible flow of information. The interplay between node actions and the ability to change them locally and dynamically makes the overall approach richer than standard message passing. This adaptability can lead DE-DGNs to improved efficiency in capturing complex dependencies from larger radii, ultimately enhancing the model's capability to represent intricate patterns transcending the classical task-agnostic message-passing scheme.

Lastly, we foresee neuromorphic implementations of our proposed frameworks allowing for efficient and effective DGNs that preserve long-range dependencies between nodes and learn complex spatio-temporal propagation patterns from irregular and undersampled data. By leveraging the energy efficiency and parallel processing capabilities inherent in neuromorphic hardware~\citep{TANAKA2019100, Marković2020}, such implementations can significantly enhance the performance and scalability of DGNs. Additionally, this approach could enable real-time processing and learning in resource-constrained environments, ultimately paving the way for advanced applications in areas such sensor networks, and autonomous systems. 


\newpage
\renewcommand\bibname{References}
\bibliographystyle{icml24_ref_style}
\bibliography{refs_bib}


\newpage
\appendix
\chapter{Supplementary materials of Chapter~\ref{ch:learning_dyn_graphs}}\label{app:suppl_ch3}
\section{Datasets, models, and previous studies}\label{app:benchmark_data_and_models}
In Table~\ref{tab:benchmark_datasets_survey} we provide the community with a selection of datasets useful for benchmarking future works. In Table~\ref{tab:benchmark_methods_wrt_events} we report an overview of the examined models with respect to the specific changes in the graph structure that each model was designed to address, \ie node/edge additions/deletions. We observe that each method is designed to address changes in node/edge features. Furthermore, the methods developed for D-TDGs that are not specifically designed to address changes in the node set can still be applied to tasks involving an evolving node set by treating nodes not in the current snapshot as isolated entities. Lastly, in Table~\ref{tab:benchmark_comparison_survey} we show a comparative analysis with previous benchmarking studies and surveys, assessing the provision of datasets and benchmarks and delineating the types of analyzed graphs.

\begin{landscape}
\tiny
\setlength{\tabcolsep}{5pt}
\begin{longtable}{p{2cm}|c|c|c|c|c|c|l}
\caption{A selected list of datasets used in dynamic graphs representation learning field. The ``$\mathcal{C}$'' in the type column means C-TDG, ``$\mathcal{D}$'' corresponds to D-TDG, and ``$\mathcal{ST}$'' to spatio-temporal graph.\label{tab:benchmark_datasets_survey}}
\\\toprule
\textbf{Name} &
  \textbf{\#Nodes} &
  \textbf{\#Edges} &
  \textbf{Seq. len.} &
  \makecell{\textbf{Snapshot sizes}\\\textbf{(nodes/edges)}} & \textbf{Granularity}& \textbf{Type}&
  \textbf{Link}\\\midrule
  
\makecell[l]{Autonomous\\systems}  & 
  7,716 & 13,895 & 733 & 103-6,474 / 243-13,233 & daily & $\mathcal{D}$ & \href{http://snap.stanford.edu/data/as-733.html}{\makecell[l]{http://snap.stanford.edu/data/as-733.html}}  \\ \midrule
  
Bitcoin-$\alpha$ &
  3,783 &  24,186 &  24,186 & $-$  & seconds & $\mathcal{C}$ & \href{http://snap.stanford.edu/data/soc-sign-bitcoin-alpha.html}{\makecell[l]{http://snap.stanford.edu/data/\\soc-sign-bitcoin-alpha.html}} \\ \midrule
  
Bitcoin-OTC &
  5,881 &  35,592 &  35,592 & $-$ & seconds & $\mathcal{C}$ & \href{http://snap.stanford.edu/data/soc-sign-bitcoin-otc.html}{\makecell[l]{http://snap.stanford.edu/data/\\soc-sign-bitcoin-otc.html}} \\ \midrule

CONTACT &
  274 &  2,712 &  28,244 & $-$  & $-$ & $\mathcal{C}$ &\href{https://networkrepository.com/ia-contact.php}{\makecell[l]{https://networkrepository.com/ia-contact.php}}\\ \midrule
  
ENRON &
  151 & 2,227 & 50,572 & $-$ & unix timestamp & $\mathcal{C}$ &\href{https://networkrepository.com/ia-enron-employees.php}{\makecell[l]{https://networkrepository.com/\\ia-enron-employees.php}} \\ \midrule
  
Elliptic &
  203,769 &  234,355 &  49 & 1,552-12,856 / 1,168-9,164 &  49 steps & $\mathcal{D}$ &\href{https://www.kaggle.com/ellipticco/elliptic-data-set}{\makecell[l]{https://www.kaggle.com/ellipticco/elliptic-data-set}} \\ \midrule
  
FB-Forum &  
  899 & 7,089 & 33,700 & $-$  &  $-$ &  $\mathcal{C}$ &\href{https://networkrepository.com/fb-forum.php}{\makecell[l]{https://networkrepository.com/fb-forum.php}} \\ \midrule
  
FB-Covid19 &
   \makecell{152(ENG)\\104(ITA)\\95(FRA)\\53(ESP)} & \makecell{2,347(ENG)\\771(ITA)\\864(FRA)\\145(ESP)} &
   \makecell{61(ENG)\\105(ITA)\\78(FRA)\\122(ESP)}& 
   \makecell{152 / 2,347(ENG)\\104 / 771(ITA)\\95 / 864(FRA)\\53 / 145(ESP)} &  daily &  $\mathcal{D}$ &\href{https://github.com/geopanag/pandemic_tgnn}{\makecell[l]{https://github.com/geopanag/pandemic\_tgnn}} \\ \midrule
   
Github &
  284 & 1,420 & 20,726 & $-$ &  $-$ & $\mathcal{C}$ & \href{https://github.com/uoguelph-mlrg/LDG}{\makecell[l]{https://github.com/uoguelph-mlrg/LDG}} \\ \midrule
  
HEP-TH&
   27,770 & 352,807 & 3487 & 1-650 / 0-688 &  montly &  $\mathcal{D}$ &\href{https://snap.stanford.edu/data/cit-HepTh.html}{\makecell[l]{https://snap.stanford.edu/data/cit-HepTh.html}} \\ \midrule

\makecell[l]{HYPER-\\TEXT09} &
  113 & 2,498 & 20,819 & $-$ &  seconds & $\mathcal{C}$ &\href{https://networkrepository.com/ia-contacts-hypertext2009.php}{\makecell[l]{https://networkrepository.com/\\ia-contacts-hypertext2009.php}} \\ \midrule
  
IA-Email-EU &
  986 & 24,929 &  332,334 & $-$ &  seconds & $\mathcal{C}$ & \href{https://snap.stanford.edu/data/email-Eu-core-temporal.html}{\makecell[l]{https://snap.stanford.edu/data/\\email-Eu-core-temporal.html}}\\ \midrule
  
LastFM & 
    2,000 & 154,993 & 1,293,103 & $-$ &  unix timestamp & $\mathcal{C}$ & \href{http://snap.stanford.edu/jodie/lastfm.csv}{\makecell[l]{http://snap.stanford.edu/jodie/lastfm.csv}} \\ \midrule

Los-loop &
   207  & 2,833 & 2,017 & 207 / 2,833 &  5 mins & $\mathcal{ST}$ &\href{https://github.com/lehaifeng/T-GCN/tree/master/data}{\makecell[l]{https://github.com/lehaifeng/T-GCN/tree/\\master/data}}\\ \midrule
   
METR-LA &
  207 &  1,515 &  34,272 &  207 / 1515 & 5 mins & $\mathcal{ST}$ &\href{https://github.com/liyaguang/DCRNN}{\makecell[l]{https://github.com/liyaguang/DCRNN}}\\ \midrule

Montevideo & 675 & 690 & 740 & 675 / 690 & hourly & $\mathcal{ST}$ &\href{https://pytorch-geometric-temporal.readthedocs.io/en/latest/modules/dataset.html}{\makecell[l]{https://pytorch-geometric-temporal.readthedocs.io/\\en/latest/modules/dataset.html}}\\\midrule
    
MOOC &
   7,144 & 411,749 & 178,443 & $-$ & unix timestamp & $\mathcal{C}$ &\href{http://snap.stanford.edu/data/act-mooc.html}{\makecell[l]{http://snap.stanford.edu/data/act-mooc.html}} \\ \midrule

PeMS03 & 358 & 442 & 26208 & 358 / 442 & 5 mins & $\mathcal{ST}$ &\href{https://torch-spatiotemporal.readthedocs.io/en/latest/modules/datasets.html}{\makecell[l]{https://torch-spatiotemporal.readthedocs.io/en/\\latest/modules/datasets.html}}\\ \midrule

PeMS04 & 307 & 209 & 16992 & 307 / 209 &5 mins & $\mathcal{ST}$ &\href{https://torch-spatiotemporal.readthedocs.io/en/latest/modules/datasets.html}{\makecell[l]{https://torch-spatiotemporal.readthedocs.io/en/\\latest/modules/datasets.html}}\\ \midrule

PeMS07 & 883 & 790 & 28225 & 883 / 790 & 5 mins & $\mathcal{ST}$ &\href{https://torch-spatiotemporal.readthedocs.io/en/latest/modules/datasets.html}{\makecell[l]{https://torch-spatiotemporal.readthedocs.io/en/\\latest/modules/datasets.html}}\\ \midrule

PeMS08 & 170 & 137 & 17856 & 170 / 137 & 5 mins & $\mathcal{ST}$ &\href{https://torch-spatiotemporal.readthedocs.io/en/latest/modules/datasets.html}{\makecell[l]{https://torch-spatiotemporal.readthedocs.io/en/\\latest/modules/datasets.html}}\\ \midrule

PeMSBay &
    325 & 2,369 &  52,116 & 325 / 2,369 & 5 mins & $\mathcal{ST}$ &\href{https://github.com/liyaguang/DCRNN}{\makecell[l]{https://github.com/liyaguang/DCRNN}} \\ \midrule
    
PeMSD7 &
  228 &  19,118 &  1,989 &  228 / 19,118 &5 mins & $\mathcal{ST}$ &\href{https://github.com/hazdzz/STGCN/tree/main/data/pemsd7-m}{\makecell[l]{https://github.com/hazdzz/STGCN/tree/main/\\data/pemsd7-m}} \\ \midrule

RADOSLAW & 
    167 &  5,509 &   82,927 & $-$ & seconds &$\mathcal{C}$ &\href{https://networkrepository.com/ia-radoslaw-email.php}{\makecell[l]{https://networkrepository.com/\\ia-radoslaw-email.php}}\\ \midrule
    
Reddit & 
    11,000 & 78,516 & 672,447 & $-$ & unix timestamp &$\mathcal{C}$ &\href{http://snap.stanford.edu/jodie/reddit.csv}{\makecell[l]{http://snap.stanford.edu/jodie/reddit.csv}} \\ \midrule
  
Reddit Hyperlink Network &
  55,863 & 339,643 & 858,490 & $-$ & seconds &$\mathcal{C}$ &\href{http://snap.stanford.edu/data/soc-RedditHyperlinks.html}{\makecell[l]{http://snap.stanford.edu/data/\\soc-RedditHyperlinks.html}}\\ \midrule
  
SBM-synthetic &
  1,000 & 130,415 & 50 & 1000 / 93,835-105,358 & 50 steps & $\mathcal{D}$ &\href{https://github.com/IBM/EvolveGCN/tree/master/data}{\makecell[l]{https://github.com/IBM/EvolveGCN/tree/\\master/data}}\\ \midrule

SOC-Wiki-Elec &
  7,118 &  103,673 & 107,071 & $-$ & $-$ &$\mathcal{C}$ & \href{https://networkrepository.com/soc-wiki-elec.php}{\makecell[l]{https://networkrepository.com/soc-wiki-elec.php}} \\ \midrule

SZ-taxi &
   156 & 532  & 2,977  & 156 / 532 & 15 mins & $\mathcal{ST}$ & \href{https://github.com/lehaifeng/T-GCN/tree/master/data}{\makecell[l]{https://github.com/lehaifeng/T-GCN/tree/\\master/data}} \\ \midrule

Traffic &
   4,438 & 8,996 & 2,160 & 4,438 / 8,996 & hourly & $\mathcal{ST}$ &\href{https://github.com/chocolates/Predicting-Path-Failure-In-Time-Evolving-Graphs}{\makecell[l]{https://github.com/chocolates/\\Predicting-Path-Failure-In-Time-Evolving-Graphs}} \\ \midrule

Twitter-Tennis & 
    1000 & 40,839 & 120 & 1000 / 41-936 & hourly & $\mathcal{D}$ & \href{https://pytorch-geometric-temporal.readthedocs.io/en/latest/modules/dataset.html}{\makecell[l]{https://pytorch-geometric-temporal.readthedocs.io/\\en/latest/modules/dataset.html}}\\ \midrule

UCI messages &
  1,899 & 20,296 & 59,835 & $-$ & unix timestamp &$\mathcal{C}$ &\href{https://snap.stanford.edu/data/CollegeMsg.html}{\makecell[l]{https://snap.stanford.edu/data/CollegeMsg.html}} \\ \midrule

Wikipedia &
   9,227 & 18,257 & 157,474 & $-$ & unix timestamp &$\mathcal{C}$ &\href{http://snap.stanford.edu/jodie/wikipedia.csv}{\makecell[l]{http://snap.stanford.edu/jodie/wikipedia.csv}}\\ 

\bottomrule
\end{longtable}
\end{landscape}
\begin{table}[ht]
\centering
\scriptsize
\caption{An overview of the examined models and the specific changes in the graph structure that each model was designed to address, \ie node additions/deletions and edge addition/deletions.
The ``$\mathcal{C}$'' in the type column means C-TDG, ``$\mathcal{D}$'' corresponds to D-TDG, and ``$\mathcal{ST}$'' to spatio-temporal graph.\label{tab:benchmark_methods_wrt_events}}

\begin{tabular}{l|c|c|cc|cc}
\toprule
               \multicolumn{3}{c}{}                   & \multicolumn{2}{c}{\textbf{Node}}     & \multicolumn{2}{c}{\textbf{Edge}}\\
\textbf{Name}  & \textbf{Cit.}             & \textbf{Type}  & \textbf{Add} & \textbf{Del} & \textbf{Add} & \textbf{Del}\\\midrule
A3TGCN         & \cite{a3tgcn}             & $\mathcal{ST}$       & \xmark            & \xmark            & \xmark            & \xmark           \\
ASTGCN         & \cite{astgcn}             & $\mathcal{ST}$       & \xmark            & \xmark            & \xmark            & \xmark           \\
CAW            & \cite{CAW}                & $\mathcal{C}$        & \cmark            & \cmark            & \cmark            & \cmark           \\
CTDNG          & \cite{temporal_node2vec}  & $\mathcal{C}$        & \cmark            & \cmark            & \cmark            & \cmark           \\
DCRNN          & \cite{DCRNN}              & $\mathcal{ST}$       & \xmark            & \xmark            & \xmark            & \xmark           \\
DyGrAE         & \cite{dygrae}             & $\mathcal{D}$        & \cmark            & \cmark            & \cmark            & \cmark           \\
DyRep          & \cite{dyrep}              & $\mathcal{C}$        & \cmark            & \cmark            & \cmark            & \cmark           \\
DynGEM         & \cite{dyngem}             & $\mathcal{D}$        & \cmark            & \cmark            & \cmark            & \cmark           \\
DynGESN        & \cite{dyngesn}            & $\mathcal{D}$        & \cmark            & \cmark            & \cmark            & \cmark           \\
DynGraph2Vec   & \cite{dyngraph2vec}       & $\mathcal{D}$        & \cmark            & \cmark            & \cmark            & \cmark           \\
E-GCN          & \cite{egcn}               & $\mathcal{D}$        & \cmark            & \cmark            & \cmark            & \cmark           \\
Evolve2Vec     & \cite{evolve2vec}         & $\mathcal{D}$        & \cmark            & \cmark            & \cmark            & \cmark           \\
GC-LSTM        & \cite{GC-LSTM}            & $\mathcal{D}$        & \xmark            & \xmark            & \cmark            & \cmark           \\
GCRN           & \cite{GCRN}               & $\mathcal{ST}$       & \xmark            & \xmark            & \xmark            & \xmark           \\
JODIE          & \cite{jodie}              & $\mathcal{C}$        & \cmark            & \cmark            & \cmark            & \cmark           \\
LRGCN          & \cite{LRGCN}              & $\mathcal{D}$        & \xmark            & \xmark            & \cmark            & \cmark           \\
MPNN-LSTM      & \cite{mpnn_lsltm}         & $\mathcal{D}$        & \xmark            & \xmark            & \cmark            & \cmark           \\
NeurTW        & \cite{neurtws}            & $\mathcal{C}$        & \cmark            & \cmark            & \cmark            & \cmark           \\
PINT           & \cite{pint}               & $\mathcal{C}$        & \cmark            & \cmark            & \cmark            & \cmark           \\
ROLAND         & \cite{roland}             & $\mathcal{D}$        & \cmark            & \cmark            & \cmark            & \cmark           \\
SGP            & \cite{cini2023scalable}   & $\mathcal{D}$        & \xmark            & \xmark            & \cmark            & \cmark           \\
STGCN          & \cite{STGCN}              & $\mathcal{ST}$       & \xmark            & \xmark            & \xmark            & \xmark           \\
StreamGNN      & \cite{streamgnn}          & $\mathcal{C}$        & \cmark            & \cmark            & \cmark            & \cmark           \\
T-GCN          & \cite{T-GCN}              & $\mathcal{ST}$       & \xmark            & \xmark            & \xmark            & \xmark           \\
TGAT           & \cite{TGAT}               & $\mathcal{C}$        & \cmark            & \cmark            & \cmark            & \cmark           \\
TGN            & \cite{tgn_rossi2020}      & $\mathcal{C}$        & \cmark            & \cmark            & \cmark            & \cmark           \\
\bottomrule

\end{tabular}
\end{table}
\begin{table}[h!]
\setlength{\tabcolsep}{4pt}
    \centering
    \caption{Comparative analysis with previous benchmarking studies and surveys, assessing the provision of datasets and benchmarks and delineating the types of analyzed graphs. The ``$\mathcal{C}$'' in the columns means C-TDG, ``$\mathcal{D}$'' corresponds to D-TDG, and ``$\mathcal{ST}$'' to spatio-temporal graph.\label{tab:benchmark_comparison_survey}\\}
    \scriptsize
\begin{tabular}{l|cccc|c|cc|ccc}
    \toprule
     & \multicolumn{4}{c|}{\multirow{3}{*}{\bf Survey}}             & \multirow{3}{*}{\bf\makecell{Year of \\the last\\surveyed \\method}}                & \multicolumn{2}{c|}{\multirow{3}{*}{\bf Datasets}} & \multicolumn{3}{c}{\multirow{3}{*}{\bf \makecell{Dyn. graph\\benchmark}}}\\
    & & & & & & & & & &\\
    & & & & & & & & & &\\
  \textbf{Study}  & Static & $\mathcal{ST}$ & $\mathcal{D}$ & $\mathcal{C}$ & & static & dynamic              & $\mathcal{ST}$ & $\mathcal{D}$ & $\mathcal{C}$\\

    \midrule
    \citet{hamilton} & \cmark & \xmark & \xmark & \xmark & 2017 & \xmark & \xmark & \xmark & \xmark & \xmark \\
    \citet{BACCIU2020203}   & \cmark & \xmark & \xmark & \xmark & 2020 & \xmark & \xmark & \xmark & \xmark & \xmark \\
    \citet{GNNsurvey}       & \cmark & \cmark & \xmark & \xmark & 2019 & \cmark & \xmark & \xmark &\xmark &\xmark \\
    \citet{dynamicgraph_survey}    & \cmark & \cmark & \cmark & \cmark & 2020 & \xmark & \cmark & \xmark &\xmark & \xmark \\
    \citet{traffic_forecasting_survey}    & \cmark & \cmark & \xmark & \xmark & 2022 & \xmark & \cmark{\tiny($\mathcal{ST}$ only)} & \xmark & \xmark & \xmark \vspace{5pt}\\
    Our                 & \cmark & \cmark & \cmark & \cmark & 2023 & \xmark & \cmark & \cmark & \cmark & \cmark \\
    \bottomrule
\end{tabular}
\end{table}

\section{Explored hyperparameter space}\label{app:benchmark_hyperparam}
In Table~\ref{tab:benchmark_st_grid}, Table~\ref{tab:benchmark_dtdg_grid}, and Table~\ref{tab:benchmark_ctdg_grid} we report the grids of hyperparameters employed during model selection for the spatio-temporal, D-TDG, and C-TDG experiments by each method. 
\begin{table}[h]
\centering
\caption{The grid of hyperparameters employed during model selection for the spatio-temporal tasks. \label{tab:benchmark_st_grid}}
\scriptsize
\begin{tabular}{ll}
\toprule 
\textbf{Hyperparameter} & \textbf{Values}\\\midrule
learning rate & $10^{-2}$, $10^{-3}$, $10^{-4}$\\
weight decay & $10^{-3}$, $10^{-4}$\\
embedding dim & 1, 2, 4, 8\\
$\sigma$ & ReLU\\
Chebishev poly. filter size & 1, 2, 3\vspace{2pt}\\
normalization scheme for $\mathbf{L}$ & \makecell[l]{$\mathbf{L} = \mathbf{D} - \mathbf{A}$,\\$\mathbf{L}^{sym} = \mathbf{I} - \mathbf{D}^{-1/2} \mathbf{A} \mathbf{D}^{-1/2}$,\\$\mathbf{L}^{rw} = \mathbf{I} - \mathbf{D}^{-1} \mathbf{A}$} \\
\bottomrule 
\end{tabular}
\end{table}
\begin{table}[h]
\centering
\caption{The grid of hyperparameters employed during model selection for the D-TDG tasks. The ``$*$'' value refer only to LRGCN model, while ``$\diamond$'' to DynGESN. \label{tab:benchmark_dtdg_grid}}
\scriptsize
\begin{tabular}{ll}
\toprule
 \textbf{Hyperparameter} & \textbf{Values}\\\midrule
learning rate & $10^{-2}$, $10^{-3}$, $10^{-4}$\\
weight decay & $10^{-3}$, $10^{-4}$\\
embedding dim & 8, 16, 32\\
$\sigma$ & ReLU\\
Chebishev poly. filter size & 1, 2, 3\vspace{2pt}\\
normalization scheme for $\mathbf{L}$ & \makecell[l]{$\mathbf{L} = \mathbf{D} - \mathbf{A}$,\\$\mathbf{L}^{sym} = \mathbf{I} - \mathbf{D}^{-1/2} \mathbf{A} \mathbf{D}^{-1/2}$,\\$\mathbf{L}^{rw} = \mathbf{I} - \mathbf{D}^{-1} \mathbf{A}$} \vspace{1pt}\\
\makecell[l]{n. bases in the basis-decomposition\\regularization scheme$^*$} & None, 10, 30\\
leakage (\ie $\gamma$)$^\diamond$  & 0.1, 0.5, 0.9\\
random weight init. value$^\diamond$  & 0.1, 0.5, 0.9\\
\bottomrule
\end{tabular}
\end{table}
\begin{table}[h]
\centering
\caption{The grid of hyperparameters employed during model selection for the C-TDG tasks.} 
\label{tab:benchmark_ctdg_grid}
\scriptsize
\begin{tabular}{ll}
\toprule
 \textbf{Hyperparameter} & \textbf{Values}\\\midrule
learning rate & $10^{-3}$,$10^{-4}$\\
weight decay & $10^{-4}$,$10^{-5}$\\
n. DGN layers & 1, 3 \\
embedding dim & 32, 64, 96\\
DGN dim & emb dim, emb dim / 2 \\
$\sigma$ & tanh\\
Neighborhood sampler size & 5\\
\bottomrule
\end{tabular}
\end{table}

To favor reproducibility of the results and transparency of our benchmarking analysis, in Table~\ref{tab:benchmark_hyper_st}, Table~\ref{tab:benchmark_hyper_dt}, and Table~\ref{tab:benchmark_hyper_ct} we report the hyperparameters selected for benchmarking task and model assessed in our empirical analysis.

\begin{table}[h]
\renewcommand{\arraystretch}{1.1}
\centering
\caption{The table of selected hyper-parameters for the spatio-temporal tasks.}\label{tab:benchmark_hyper_st}
\scriptsize
\begin{tabular}{llccccc}
\toprule
& &  \textbf{emb dim} &    \textbf{lr} &  \textbf{weight decay} & \textbf{filter size} &    \textbf{norm.} \\
\midrule
\multirow{5}{*}{\rotatebox[origin=c]{90}{\textbf{Montevideo}}} &
 A3TGCN    &              8 &  $10^{-2}$ &        $10^{-3}$ &           - &                - \\
& DCRNN     &              8 &  $10^{-2}$ &        $10^{-3}$ &         1 &                 - \\
& GCRN-GRU  &              8 &  $10^{-2}$ &        $10^{-3}$ &           3 &             $\bfL$ \\
& GCRN-LSTM &              8 &  $10^{-2}$ &        $10^{-4}$ &           3 &             $\bfL$ \\
& TGCN      &              8 &  $10^{-2}$ &        $10^{-3}$ &               - &             - \\
\midrule
\multirow{5}{*}{\rotatebox[origin=c]{90}{\textbf{Metr-LA}}} &
A3TGCN    &              8 &  $10^{-3}$ &        $10^{-3}$ &               - &             - \\
& DCRNN     &              8 &  $10^{-3}$ &        $10^{-3}$ &         2 &                - \\
& GCRN-GRU  &              8 &  $10^{-3}$ &        $10^{-3}$ &             3 &             $\bfL$ \\
& GCRN-LSTM &              8 &  $10^{-3}$ &        $10^{-3}$ &             3 &           $\bfL^{sym}$ \\
& TGCN      &              8 &  $10^{-3}$ &        $10^{-4}$ &               - &             - \\
\midrule
\multirow{5}{*}{\rotatebox[origin=c]{90}{\textbf{PeMSBay}}} &
A3TGCN    &              8 &  $10^{-3}$ &        $10^{-4}$ &               - &             - \\
& DCRNN     &              8 &  $10^{-4}$ &        $10^{-4}$ &         2 &                 - \\
& GCRN-GRU  &              8 &  $10^{-4}$ &        $10^{-4}$ &             2 &             $\bfL$ \\
& GCRN-LSTM &              8 &  $10^{-4}$ &        $10^{-4}$ &             3 &            $\bfL^{rw}$ \\
& TGCN      &              8 &  $10^{-4}$ &        $10^{-4}$ &               - &             - \\

\midrule
\multirow{5}{*}{\rotatebox[origin=c]{90}{\textbf{Traffic}}} &
A3TGCN    &              8 &  $10^{-2}$ &        $10^{-4}$ &               - &             - \\
& DCRNN     &        8       &  $10^{-3}$ &        $10^{-3}$ &         3 &                 - \\
& GCRN-GRU  &              8 &  $10^{-2}$ &        $10^{-4}$ &             2 &             $\bfL^{sym}$ \\
& GCRN-LSTM &              8 &  $10^{-2}$ &        $10^{-3}$ &             2 &            $\bfL^{sym}$ \\
& TGCN      &              8 &  $10^{-2}$ &        $10^{-4}$ &               - &             - \\
\bottomrule
\end{tabular}
\end{table}


\begin{table}[h]
\renewcommand{\arraystretch}{1.3}
    \centering
    \caption{The table of selected hyper-parameters for the D-TDG tasks.}\label{tab:benchmark_hyper_dt}
    \scriptsize
\begin{tabular}{llcccccccc}
\toprule
&           &  \makecell[c]{\textbf{emb}\vspace{-3pt}\\\textbf{dim}} &           \textbf{lr} &  \makecell[c]{\textbf{weight}\\\textbf{decay}} &    {\bf \makecell[c]{n.\\bases}} &    \textbf{K} & \textbf{norm.} & {\bf \makecell[c]{random\\weight\\init.\\value}} & \textbf{$\gamma$} \\
\midrule
\multirow{5}{*}{\rotatebox[origin=c]{90}{\textbf{Twitter tennis}}} &
DynGESN      &             32 &        $10^{-2}$ &        $10^{-3}$ &             - &    - &             - &   0.9 &     0.5 \\
&EvolveGCN-H  &              8 &        $10^{-2}$ &        $10^{-3}$ &             - &    - &          $\bfL$ &     - &       - \\
&EvolveGCN-O  &             32 &        $10^{-3}$ &        $10^{-3}$ &             - &    - &          $\bfL$ &     - &       - \\
&GCLSTM       &             32 &        $10^{-2}$ &        $10^{-4}$ &             - &    2 &             $\bfL$ &     - &       - \\
&LRGCN        &             32 &        $10^{-2}$ &        $10^{-4}$ &             None &    - &             - &     - &       - \\

\midrule

\multirow{5}{*}{\rotatebox[origin=c]{90}{\textbf{Elliptic}}} &
DynGESN     &              8 &         $10^{-2}$ &        $10^{-4}$ &             - &    - &             - &   0.1 &     0.9 \\
&EvolveGCN-H &             32 &         $10^{-4}$ &        $10^{-4}$ &            - &    - &             $\bfL^{sym}$ &     - &       - \\
&EvolveGCN-O &             16 &         $10^{-4}$ &        $10^{-3}$ &            - &    - &             $\bfL^{sym}$ &     - &       - \\
&GCLSTM     &              8 &         $10^{-4}$ &        $10^{-4}$ &             - &    1 &             $\bfL$ &     - &       - \\
&LRGCN      &              8 &         $10^{-4}$ &        $10^{-4}$ &            10 &    - &             - &     - &       - \\

\midrule
\multirow{5}{*}{\rotatebox[origin=c]{90}{\textbf{AS-733}}} &
DynGESN     &             32 &         $10^{-2}$ &        $10^{-4}$ &            - &  - &             - &   0.9 &     0.5 \\
&EvolveGCN-H &             32 &         $10^{-3}$ &        $10^{-3}$ &           - &  - &             $\bfL$ &     - &       - \\
&EvolveGCN-O &             16 &         $10^{-3}$ &        $10^{-4}$ &           - &  - &             $\bfL$ &     - &       - \\
&GCLSTM     &             32 &         $10^{-2}$ &        $10^{-3}$ &            - &  2 &           $\bfL^{sym}$ &     - &       - \\
&LRGCN      &             32 &         $10^{-3}$ &        $10^{-3}$ &           10 &  - &             - &     - &       - \\

\midrule

\multirow{5}{*}{\rotatebox[origin=c]{90}{\textbf{Bitcoin $\alpha$}}} &
DynGESN      & 32 & $10^{-2}$& $10^{-4}$ & - &    - &  - & 0.9 &     0.1\\
&EvolveGCN-H &             32 &          $10^{-4}$ &        $10^{-3}$ &                 - &    - &             $\bfL$ &     - &       - \\
&EvolveGCN-O &             16 &          $10^{-3}$ &        $10^{-4}$ &                 - &    - &             $\bfL$ &     - &       - \\
&GCLSTM     &             16 &          $10^{-2}$ &        $10^{-4}$ &                 - &  3 &             $\bfL$ &     - &       - \\
&LRGCN      &             32 &          $10^{-2}$ &        $10^{-4}$ &              10 &    - &             - &     - &       - \\

\bottomrule
\end{tabular}
\end{table}


\begin{table}[h]
\renewcommand{\arraystretch}{1.2}
    \centering
    \caption{The table of selected hyper-parameters for the C-TDG tasks.}\label{tab:benchmark_hyper_ct}
    \scriptsize
\begin{tabular}{llccccc}
\toprule
&           &  \textbf{emb dim} &           \textbf{lr} &  \textbf{weight decay} &    \textbf{n. DGN layers} & \textbf{DGN dim} \\
\midrule
\multirow{4}{*}{\rotatebox[origin=c]{90}{\textbf{Wikipedia}}} &
DyRep &                      96 &           $10^{-3}$ &          $10^{-4}$ &             -  & -\\
&JODIE &                      96 &           $10^{-4}$ &          $10^{-5}$ &             -  & -\\
&TGAT  &                      96 &           $10^{-3}$ &          $10^{-4}$ &           3 & 96\\
&TGN   &                      96 &           $10^{-3}$ &          $10^{-5}$ &           3 & 48\\

\midrule
\multirow{4}{*}{\rotatebox[origin=c]{90}{\textbf{Reddit}}} &
DyRep &                      96 &           $10^{-4}$ &          $10^{-4}$ &             - & -\\
&JODIE &                      96 &           $10^{-4}$ &          $10^{-5}$ &             - & -\\
&TGAT  &                      96 &           $10^{-3}$ &          $10^{-4}$ &           3 & 96\\
&TGN   &                      96 &           $10^{-3}$ &          $10^{-5}$ &           1 & 96\\

\midrule
\multirow{4}{*}{\rotatebox[origin=c]{90}{\textbf{LastFM}}} &
DyRep &                      96 &           $10^{-4}$ &          $10^{-4}$ &             - & -\\
&JODIE &                      32 &           $10^{-4}$ &          $10^{-5}$ &             - & -\\
&TGAT  &                      48 &           $10^{-4}$ &          $10^{-4}$ &           3 & 48\\
&TGN   &                      32 &           $10^{-3}$ &          $10^{-5}$ &           3 & 16\\

\bottomrule

\end{tabular}
\end{table}

\section{Stability of training under various hyperparameters}\label{app:benchmark_training stability}
Table~\ref{tab:benchmark_robustness_st}, Table~\ref{tab:benchmark_robustness_dt}, and Table~\ref{tab:benchmark_robustness_ct}  report the minimum and maximum standard deviation (std) of validation scores obtained by fixing individual hyperparameter values in spatio-temporal, D-TDGs, and C-TDGs, respectively. Thus, we study the stability of training under various hyperparameters. We observe that models typically exhibit stability across diverse hyperparameter settings, and that weight decay is among the hyperparameters with less influence on the final score, while the embedding dimension and learning rate are the most affecting ones. 

\begin{table}[h]
    \centering
    \caption{The \textbf{min}imum and \textbf{max}imum standard deviation of validation scores obtained by fixing individual hyperparameter values in spatio-temporal models. The hyperparameter names corresponding to these values are also provided for reference. ``wd'' means weight decay, ``ed'' embedding dimension, ``lr'' learning rate, and ``fs'' filter size.}\label{tab:benchmark_robustness_st}
    \scriptsize
\begin{tabular}{lll|ll}
    \toprule
& \multicolumn{2}{c|}{\textbf{Montevideo}} & \multicolumn{2}{c}{\textbf{MetrLA}}   \\
&      \multicolumn{1}{c}{\textbf{min}} & \multicolumn{1}{c|}{\textbf{max}}
    &      \multicolumn{1}{c}{\textbf{min}} & \multicolumn{1}{c}{\textbf{max}}\\

\midrule
A3TGCN    & ed: 0.009 &  lr: 0.015 & wd: 0.099 &  ed: 0.127 \\
DCRNN     & ed: 0.016 &  lr: 0.025 & wd: 0.019 &  ed: 0.034 \\
GCRN-GRU  & lr: 0.038 &   K: 0.043 & wd: 0.057 &   K: 0.071 \\
GCRN-LSTM & wd: 0.031 &  ed: 0.039 & wd: 0.056 &  ed: 0.082 \\
TGCN      & ed: 0.013 &  lr: 0.018 & wd: 0.099 &  ed: 0.120 \\\midrule

& \multicolumn{2}{c|}{\textbf{PeMSBay}} & \multicolumn{2}{c}{\textbf{Traffic}}\\
    &      \multicolumn{1}{c}{\textbf{min}} & \multicolumn{1}{c|}{\textbf{max}}
    &      \multicolumn{1}{c}{\textbf{min}} & \multicolumn{1}{c}{\textbf{max}}\\\midrule
 A3TGCN    & ed: 0.039 &  lr: 0.081 &  ed: 0.006 &  lr: 0.009 \\
 DCRNN     & ed: 0.050 &  fs: 0.131 &  wd: 0.017 &  ed: 0.020 \\
 GCRN-GRU  & ed: 0.086 &   K: 0.150 &  wd: 0.040 &   K: 0.047 \\
 GCRN-LSTM & ed: 0.072 &   K: 0.155 &  wd: 0.024 &  lr: 0.031 \\
 TGCN      & ed: 0.041 &  lr: 0.087 &  ed: 0.006 &  lr: 0.009 \\
    \bottomrule
\end{tabular}
\end{table}

\begin{table}[h]
    \centering
    \caption{The \textbf{min}imum and \textbf{max}imum standard deviation of validation scores obtained by fixing individual hyperparameter values in D-TDGs models. The hyperparameter names corresponding to these values are also provided for reference. ``wd'' means weight decay, ``ed'' embedding dimension, ``lr'' learning rate, ``ns'' normalization scheme, and ``nb'' number of bases.}\label{tab:benchmark_robustness_dt}
    \scriptsize
\begin{tabular}{lll|ll}
    \toprule
     & \multicolumn{2}{c|}{\textbf{Twitter tennis}} & \multicolumn{2}{c}{\textbf{Elliptic}} \\
     &      \multicolumn{1}{c}{\textbf{min}} & \multicolumn{1}{c|}{\textbf{max}}
         &      \multicolumn{1}{c}{\textbf{min}} & \multicolumn{1}{c}{\textbf{max}}\\
     \midrule
 DynGESN      &  lr: 0.004 &  $\sigma$: 0.006  &   lr: 0.004 &  ed: 0.005 \\
 EvolveGCN-H  &  wd: 0.011 &  lr: 0.012        &   lr: 0.003 &  ed: 0.005 \\
 EvolveGCN-O  &  ed: 0.010 &  lr: 0.012        &   lr: 0.005 &  ed: 0.007 \\
 GCLSTM       &  lr: 0.012 &  ed: 0.016        &   wd: 0.004 &  ed: 0.005 \\
 LRGCN        &  nb: 0.008 &  ed: 0.01         &   wd: 0.005 &  nb: 0.005 \\\midrule

& \multicolumn{2}{c|}{\textbf{AS-733}} & \multicolumn{2}{c}{\textbf{Bitcoin-$\alpha$}}  \\
         &      \multicolumn{1}{c}{\textbf{min}} & \multicolumn{1}{c|}{\textbf{max}}
         &      \multicolumn{1}{c}{\textbf{min}} & \multicolumn{1}{c}{\textbf{max}}\\\midrule
 DynGESN     & ed: 0.003 &  $\sigma$: 0.005 &  $\gamma$: 0.009 &  lr: 0.052  \\
EvolveGCN-H & wd: 0.082 &  ed: 0.109       &  ns: 0.018       &  ed: 0.020  \\
EvolveGCN-O & ed: 0.095 &  wd: 0.105       &  lr: 0.022       &  wd: 0.024  \\
GCLSTM      & K: 0.201  &  ed: 0.226       &  K: 0.025        &  lr: 0.207  \\
LRGCN       & nb: 0.036 &  lr: 0.057       &  nb: 0.003       &  ed: 0.004  \\
    \bottomrule
\end{tabular}
\end{table}

\begin{table}[h]
    \centering
    \caption{The \textbf{min}imum and \textbf{max}imum standard deviation of validation scores obtained by fixing individual hyperparameter values in C-TDGs models. The hyperparameter names corresponding to these values are also provided for reference. ``wd'' means weight decay, ``ed'' embedding dimension, ``lr'' learning rate, ``gl'' number of DGN layers, ``ge'' DGN dimension, and ``re'' readout embedding dim.}\label{tab:benchmark_robustness_ct}
    \scriptsize
\begin{tabular}{lll|ll|ll}
    \toprule
&  \multicolumn{2}{c|}{\textbf{Wikipedia}} & \multicolumn{2}{c|}{\textbf{Reddit}} & \multicolumn{2}{c}{\textbf{LastFM}}  \\
&      \multicolumn{1}{c}{\textbf{min}} & \multicolumn{1}{c|}{\textbf{max}}
    &      \multicolumn{1}{c}{\textbf{min}} & \multicolumn{1}{c|}{\textbf{max}}
    &      \multicolumn{1}{c}{\textbf{min}} & \multicolumn{1}{c}{\textbf{max}}\\

\midrule
DyRep  & ed: 0.004 & wd: 0.009 &   lr: 0.002 & wd: 0.003   &   lr: 0.004 &     ed: 0.007  \\
JODIE  & lr: 0.023 & ed: 0.027 &   wd: 0.007 & ed: 0.007   &   wd: 0.004 &     ed: 0.006  \\
TGAT   & gl: 0.004 & re: 0.023 &   ge: 0.002 & re: 0.011   &   gl: 0.007 &     wd: 0.030  \\
TGN    & ed: 0.004 & wd: 0.005 &   wd: 0.004 & ge: 0.006   &   gl: 0.012 &     lr: 0.033  \\
    \bottomrule
\end{tabular}
\end{table}

\chapter{Supplementary materials of Chapter~\ref{ch:antisymmetry}}\label{app:suppl_ch4}
\section{Supplementary materials of Section~\ref{sec:ADGN}}
\subsection{Continuity of layer-dependent weights}\label{app:ldw}
To ensure stability of the forward propagation, the continuous interpretation of the DGN should not change significantly in time. In other words, the weights of the model should not drastically change between layers. This can be easily achieved by sharing weights across layers, thereby implementing weight sharing, as done in previous approaches \citep{neuralODE, chang2018antisymmetricrnn, GDE, GRAND, graphcon}. However, when layer-dependent weights are used, maintaining such smoothness becomes more challenging. To address this, \cite{Ruthotto2020} introduced a regularization term that penalizes abrupt changes, thus promoting smooth weight transitions across layers.

\begin{figure}[h]
\centering
\begin{adjustbox}{angle=90}
\begin{minipage}{0.001\textwidth}
\small
    \begin{equation*}
\hspace{-10mm}
\frac{|\bar{\mathbf{W}}_{\ell}-\bar{\mathbf{W}}_{\ell-1}|}{|\bar{\mathbf{W}}_{\ell-1}|}
    \end{equation*}
  \end{minipage}%
    \end{adjustbox}\hspace{-3.5mm}
\begin{minipage}{.6\textwidth}
    \centering
    \includegraphics[width=0.9\textwidth]{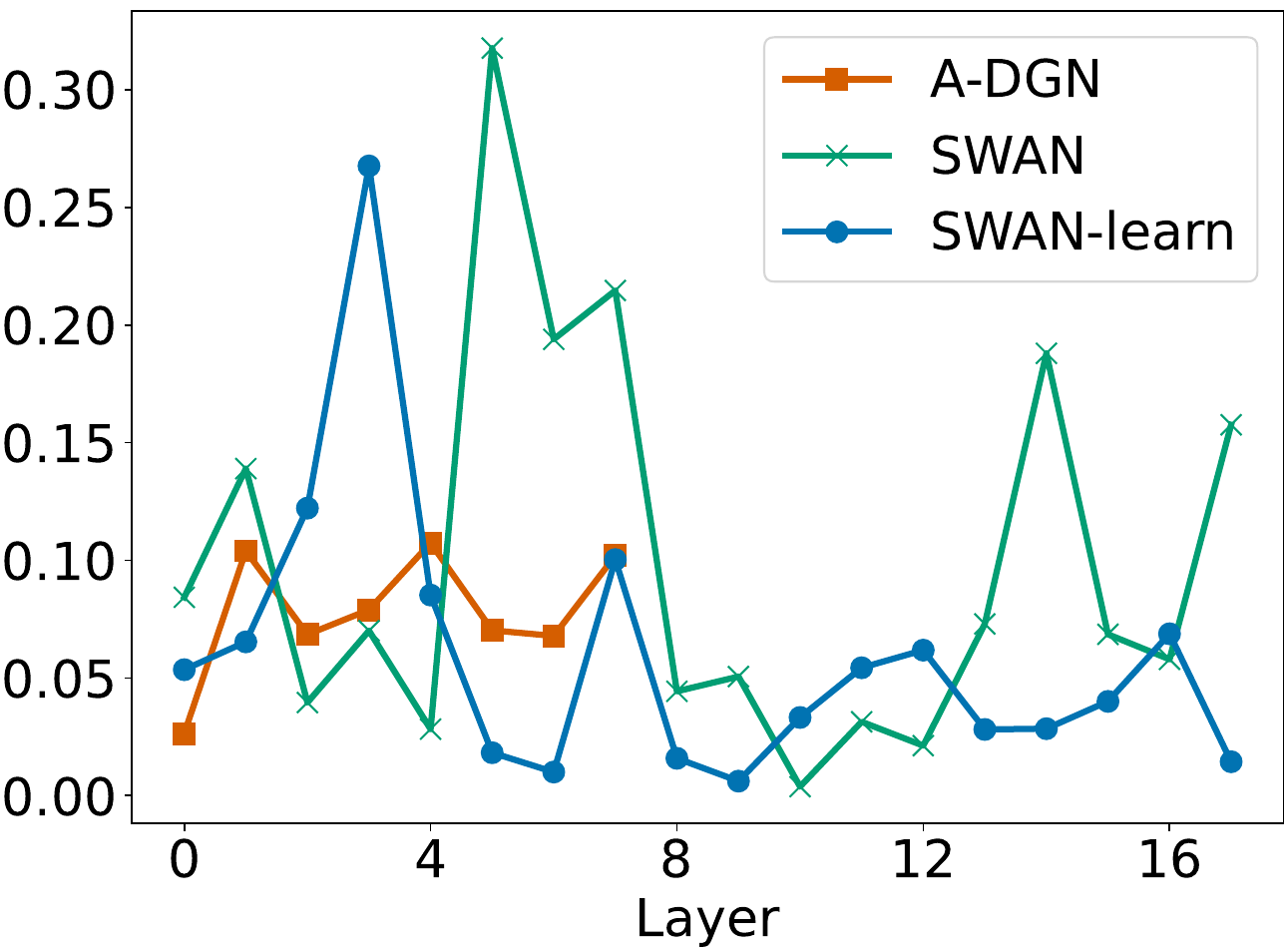}
\end{minipage}
\caption{
The relative change in the weights ($\bar{\mathbf{W}}$) across layers for A-DGN, SWAN, and SWAN-\textsc{learn} models, measured on the Diameter task of the Graph Property Prediction experiment. Here, $\bar{\mathbf{W}}$ represents the norm of the weights of the entire graph convolutional layer, \ie $\bar{\mathbf{W}}_\ell = norm\left(concat(\bfW_\ell, \bfV_\ell, \bfZ_\ell)\right)$, with $\bfV_\ell$ the weights of the aggregation function $\Phi$ and $\bfZ_\ell$ the weights of the antisymmetric aggregation function $\Psi$ (when possible). The encoder and readout components' weights are excluded.}
\label{fig:continuity}
\end{figure}

Building on this observation, we provide an empirical measurement to study the continuity (\ie smoothness) of the weights in the case of layer-dependent weights for both our A-DGN (Section~\ref{sec:ADGN}) and SWAN (Section~\ref{sec:SWAN}), which is crucial for ensuring the smooth and accurate evolution of the learned dynamics. 
We consider the best configurations of A-DGN and SWAN on the Diameter task of the Graph Property Prediction experiment. Figure~\ref{fig:continuity} shows that weight changes between consecutive layers are minimal, with averages of 0.08 for A-DGN, 0.10 for SWAN, and 0.06 for SWAN-\textsc{learn}. This confirms the smooth transitions of weights across layers, supporting the model's validity even with layer-dependent weights.

Lastly, we observe that the layer-dependent weight setting can also be obtained by stacking multiple instances of the A-DGN (or SWAN) model. In this scenario, the resulting framework can be interpreted as a composition of dynamical systems, where each layer (\ie the A-DGN or SWAN instance) evolves the state computed by the previous one, ensuring the non-dissipative properties discussed in Chapter~\ref{ch:antisymmetry}.
Specifically, each layer operates with its hyperparameter values (\eg the step size or the number of iterations) and evolves the system in a weight sharing setting before passing its output to the next layer.
The resulting prediction then serves as the initial condition for the subsequent dynamical system, ensuring that each layer builds upon the state produced by its predecessor, thereby capturing complex dependencies across the network.


\subsection{Datasets description and statistics}\label{app:adgn_data_stats}
In the graph property prediction (GPP) experiments, we employed the same generation procedure as in \citet{PNA}. Graphs are randomly sampled from several graph distributions, such as Erd\H{o}s–R\'{e}nyi, Barabasi-Albert, and grid. Each node have random identifiers as input features. Target values represent single source shortest path, node eccentricity, and graph diameter. 

PubMed is a citation network where each node represents a paper and each edge indicates that one paper cites another one. Each publication in the dataset is described by a 0/1-valued word vector indicating the absence/presence of the corresponding word from the dictionary. The class labels represent the papers categories.

Amazon Computers and Amazon Photo are portions of the Amazon co-purchase graph, where nodes represent goods and edges indicate that two goods are frequently bought together. Node features are bag-of-words encoded product reviews, and class labels are given by the product category.

Coauthor CS and Coauthor Physics are co-authorship graphs extracted from the Microsoft Academic Graph\footnote{https://www.kdd.org/kdd-cup/view/kdd-cup-2016/Data} where nodes are authors, that are connected by an edge if they co-authored a paper. Node features represent paper keywords for each author’s papers, and class labels indicate most active fields of study for each author.

Cornell, Texas, and Wisconsin are subgraphs of the WebKB 
dataset\footnote{http://www.cs.cmu.edu/afs/cs.cmu.edu/project/theo-11/www/wwkb}, where nodes represent web pages, and edges are hyperlinks between them. Node features are the bag-of-words representation of web pages. The web pages are manually classified into the five categories, \ie student, project, course, staff, and faculty.

The Actor dataset is a film-directoractor-writer network. Each node corresponds to an actor, and the edge between two nodes denotes co-occurrence on the same Wikipedia page. Node features correspond to some keywords in the Wikipedia pages. Nodes are classified into five categories in terms of words of actor’s Wikipedia.

Chameleon and Squirrel are two page-page networks on specific Wikipedia topics. In those datasets, nodes represent web pages and edges are mutual links between pages. Node features correspond to the presence of specific nouns in the Wikipedia pages. Nodes are classified into five categories in terms of the number of the average monthly traffic of the web page.

Table~\ref{tab:data_stats} contains the statistics of the employed datasets, sorted by graph density. The density of a graph is computed as the ratio between the number of edges and the number of possible edges, \ie $d =\frac{|\mathcal{E}|}{|\mathcal{V}|(|\mathcal{V}|-1)}$.

\begin{table}[h]
\centering
\caption{Datasets statistics ordered by graph density.\label{tab:data_stats}}
\scriptsize
\begin{tabular}{lccccc}
\toprule
& \textbf{Nodes} & \textbf{Edges} & \textbf{Features} & \textbf{Classes} & \textbf{Density} \\\midrule
GPP & 25 - 35 & 22 - 553 & 2 & --- & 0.0275 - 0.5\\
Texas & 183 & 309 & 1703 & 5 & 9.3$e^{-3}$\\
Cornell & 183 & 295 & 1703 & 5 & 8.9$e^{-3}$\\
Squirrel & 5201 & 217073 & 2089 & 5 & 8.0$e^{-3}$\\
Wisconsin & 251 & 499 &  1703 & 5 & 8.0$e^{-3}$\\
Chameleon & 2277 & 36101 & 2325 & 5 & 7.0$e^{-3}$\\
Amazon Computers & 13,752 & 491,722 & 767 & 10 & 2.6$e^{-3}$\\
Amazon Photo & 7,650 & 238,162 & 745 & 8 & 4.1$e^{-3}$\\
Actor & 7600 & 33544 & 931 & 5& 5.8$e^{-4}$\\
Coauthor CS & 18,333 & 163,788 & 6,805 & 15 & 4.9$e^{-4}$\\
Coauthor Physics & 34,493 & 495,924 & 8,415 & 5 & 4.2$e^{-4}$\\
PubMed & 19,717 & 88,648 & 500 & 3 & 2.3$e^{-4}$\\
\bottomrule
\end{tabular}
\end{table}

\subsection{Explored hyperparameter space}\label{app:adgn_hyperparam}
In Table~\ref{tab:adgn_configs} we report the grids of hyperparameters employed in our experiments by each method. We recall that the hyperparameters $\epsilon$ and $\gamma$ refer only to our method.
\begin{table}[h!]
\centering
\caption{The grid of hyperparameters employed during model selection for the graph property prediction tasks (GraphProp), graph benchmarks (Bench), and graph heterophilic benchmarks (H-Bench).\label{tab:adgn_configs}}
\scriptsize
\begin{tabular}{l|l|l|l}
\toprule 
\multirow{2}{*}{\textbf{Hyperparameter}} & \multicolumn{3}{c}{\textbf{Values}}\\\cmidrule{2-4}
& \textbf{GraphProp} & \textbf{Bench} & \textbf{H-Bench}\\\midrule
optimizer & Adam                     &  AdamW                                        &  Adam\\
learning rate & 0.003                     &  $10^{-2}$, $10^{-3}$, $10^{-4}$                                        &  $10^{-1}$, $10^{-2}$, $10^{-4}$\\
weight decay & $10^{-6}$                     &  0.1                                        &  $10^{-2}$\\
n. layers & 1, 5, 10, 20                      &  1, 2, 3, 5 ,10, 20, 30                                         &  8, 16, 32, 64 \\
embedding dim & 10, 20, 30                     &  32, 64, 128                                        &  128, 256, 512, 1024\\
$\sigma$ & tanh                     &  tanh                                        &  tanh, relu\\
$\epsilon$ & 1, $10^{-1}$, $10^{-2}$, $10^{-3}$                     &  1, $10^{-1}$, $10^{-2}$, $10^{-3}$, $10^{-4}$                                        &  $10^{-1}$, $10^{-2}$\\
$\gamma$ & 1, $10^{-1}$, $10^{-2}$, $10^{-3}$                     &  1, $10^{-1}$, $10^{-2}$, $10^{-3}$, $10^{-4}$                                        &  $10^{-1}$, $10^{-2}$\\
dropout & --- & --- & 0, 0.2, 0.4, 0.6\\
\bottomrule
\end{tabular}
\end{table}
\clearpage
\section{Supplementary materials of Section~\ref{sec:SWAN}}
\subsection{The stability of the Jacobian}\label{appendix:jacobian_stable}
\begin{figure}[h]
\centering
\begin{adjustbox}{angle=90}
\begin{minipage}{0.001\textwidth}
\small
    \begin{equation*}
\hspace{-3mm}
\frac{|\mathbf{J}^{\ell}-\mathbf{J}^{\ell-1}|}{|\mathbf{J}^{\ell-1}|}
    \end{equation*}
  \end{minipage}%
    \end{adjustbox}\hspace{-2mm}
\begin{minipage}{.6\textwidth}
    \centering
    \includegraphics[width=\textwidth]{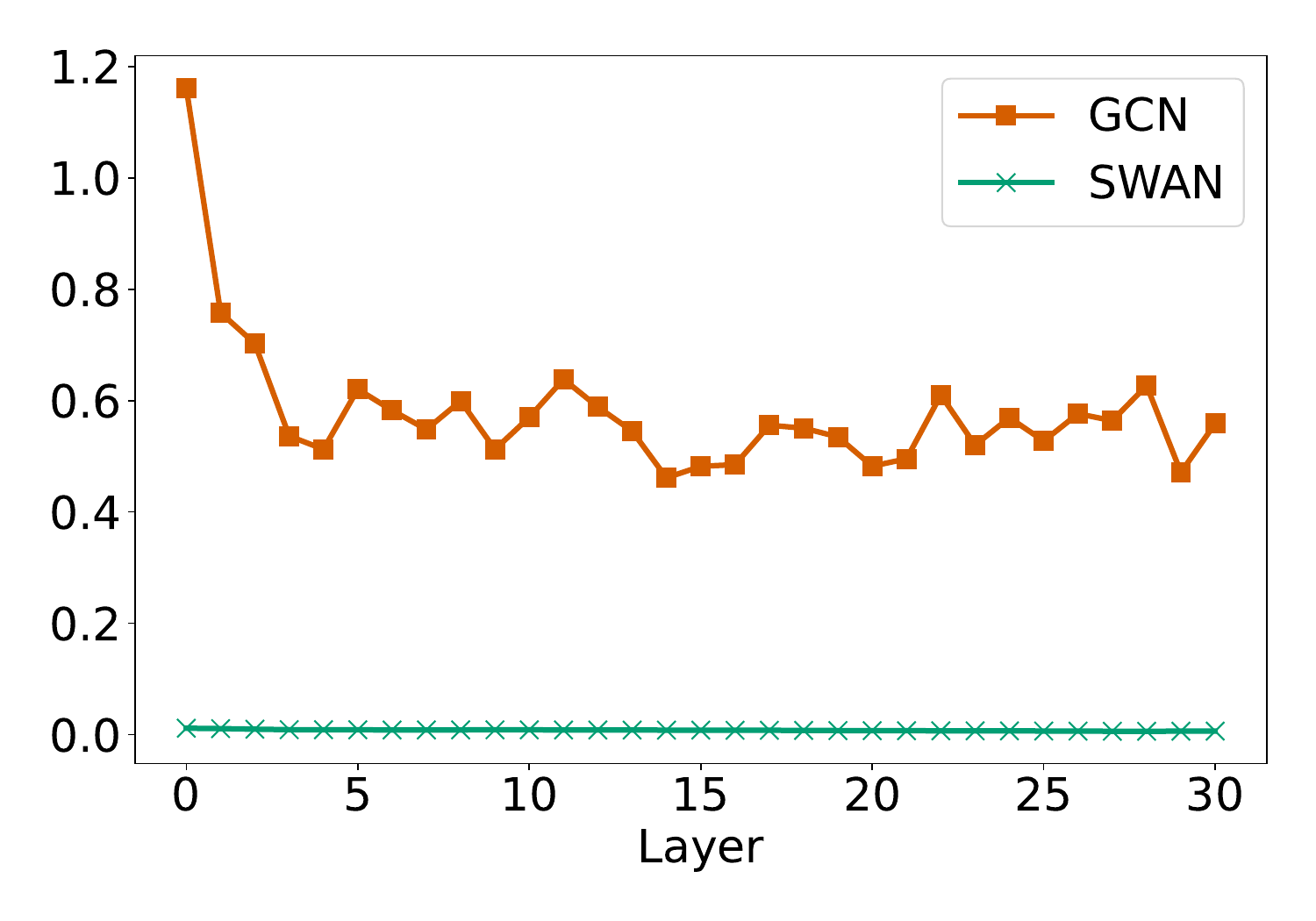}
    \caption{Arithmetic scale}
\end{minipage}\\
\begin{adjustbox}{angle=90}
\begin{minipage}{.001\textwidth}
    \small
    \begin{equation*}
    \hspace{-3mm}
\frac{|\mathbf{J}^{\ell}-\mathbf{J}^{\ell-1}|}{|\mathbf{J}^{\ell-1}|}
    \end{equation*}
  \end{minipage}%
\end{adjustbox}\hspace{-2mm}
\begin{minipage}{.6\textwidth}
    \centering
    \includegraphics[width=\textwidth]{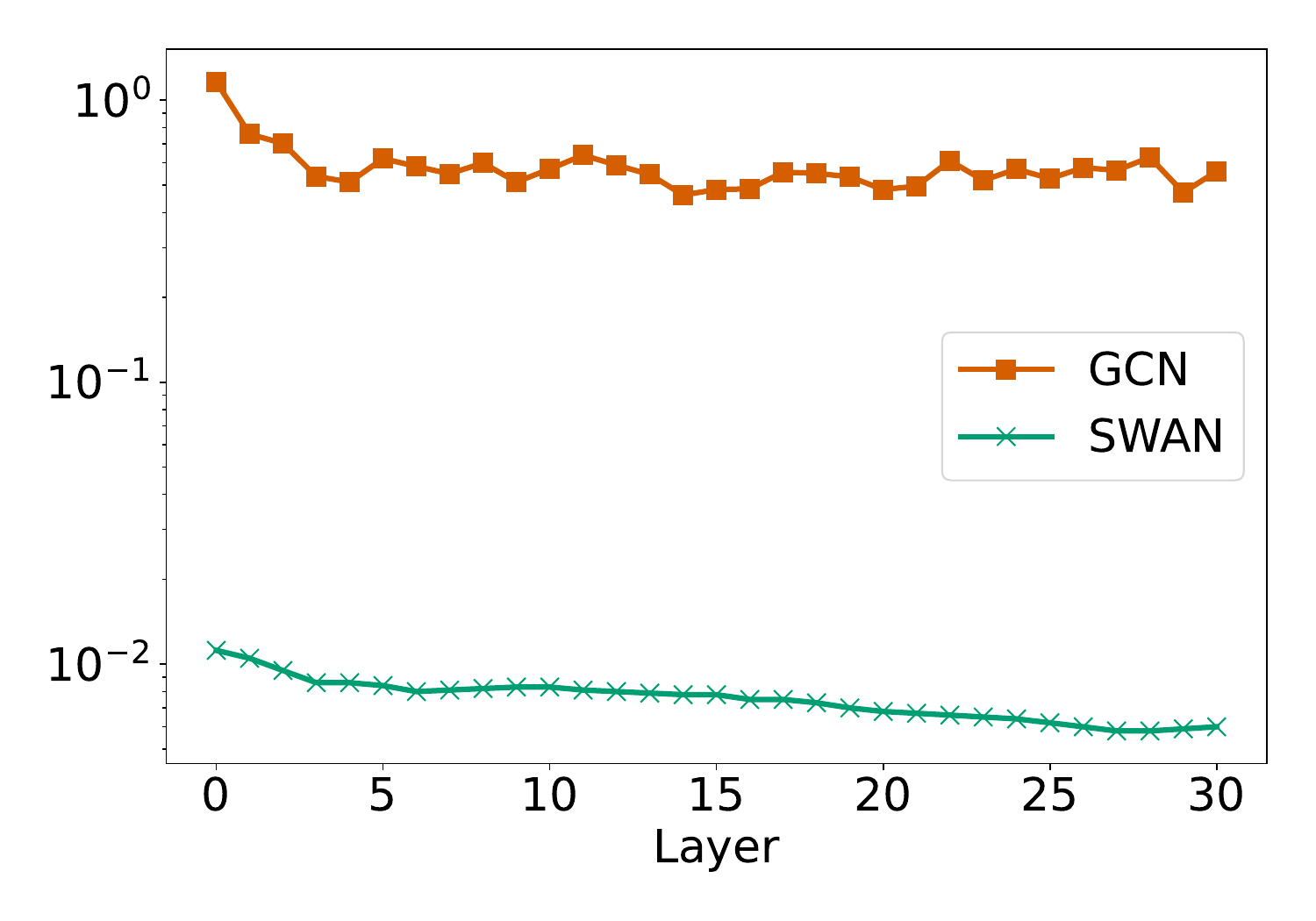}
    \caption{Logarithmic scale}
\end{minipage}
\caption{The relative change in Jacobian of GCN and our SWAN over layers, measured on the peptides-func data.}
\label{fig:jacobian_over_time}
\end{figure}

As discussed in Section \ref{sec:ADGN} and \ref{sec:SWAN}, assuming that the Jacobian of the underlying system does not change significantly over time allows us to analyze the system from an autonomous system perspective~\citet{Ascher1998} 
and mirrors prior approaches, such as \cite{Ruthotto2020, neuralODE,chang2018antisymmetricrnn, GRAND}. In addition to building on existing literature, below, we provide an empirical measurement on a real-world dataset (pepties-func) of the Jacobian of our SWAN over time (layers). For reference, we compare it with the Jacobian of GCN. As can be seen from Figure \ref{fig:jacobian_over_time}, the Jacobian of SWAN has a minimal Jacobian change over time with an average of 0.6\% between layers, while the change in the Jacobian over time in GCN is 40\% on average.

\subsection{Derivation of the graph-wise Jacobian}\label{app:jacobian}
Recall the ODE that defines SWAN  in Equation~\ref{eq:new_adgn_graphwise}:
\begin{align}
 \label{eq:new_adgn_graphwise_APP}
     \frac{\partial\mathbf{X}(t)}{\partial t} = \sigma \Bigl(\mathbf{X}(t){(\mathbf{W}-\mathbf{W}^\top)} 
     &+{(\hat{\mathbf{A}}+\hat{\mathbf{A}}^\top)\mathbf{X}(t)(\mathbf{V}-\mathbf{V}^\top)}+ \nonumber\\
     &+ \beta{(\tilde{\mathbf{{A}}}-\tilde{\mathbf{{A}}}^\top)\mathbf{X}(t)(\mathbf{Z}+\mathbf{Z}^\top)}\Bigr).
 \end{align}

 The Jacobian of Equation~\ref{eq:new_adgn_graphwise_APP} with respect to $\mathbf{X}(t)$ is:
 \begin{align}
 \mathbf{M}_1 &= \sigma'\Bigl(\mathbf{X}(t){(\mathbf{W}-\mathbf{W}^\top)} 
     +{(\hat{\mathbf{A}}+\hat{\mathbf{A}}^\top)\mathbf{X}(t)(\mathbf{V}-\mathbf{V}^\top)}+\nonumber\\
     &\hspace{5cm}+ \beta{(\tilde{\mathbf{{A}}}-\tilde{\mathbf{{A}}}^\top)\mathbf{X}(t)(\mathbf{Z}+\mathbf{Z}^\top)} \Bigr)\\
 \mathbf{M}_2 &=
     {(\mathbf{W}-\mathbf{W}^\top)}  
     +{(\mathbf{V}-\mathbf{V}^\top)^\top \otimes (\hat{\mathbf{A}}+\hat{\mathbf{A}}^\top) }+\nonumber\\
     &\hspace{5cm}+ \beta(\mathbf{Z}+\mathbf{Z}^\top)^\top \otimes  (\tilde{\mathbf{{A}}}-\tilde{\mathbf{{A}}}^\top)
 \end{align}
 where $\otimes$ is the Kronecker product.
 To analyze $\textbf{M}_1$, we use the following identity for arbitrary matrices $\mathbf{A}, \mathbf{X}, \mathbf{B}$ with appropriate dimensions: 
 \begin{equation}    
 \label{eq:identity}
 \rm{vec}(\mathbf{AXB}) = \left( \mathbf{B}^\top \otimes \mathbf{A} \right) {\rm{vec}}(\mathbf{X})
 \end{equation}
 Using the identity from Equation~\ref{eq:identity}, we can rewrite $\mathbf{M}_1$ as follows:
 \begin{align}
\nonumber \mathbf{M}_1 &= \sigma'\Bigl(\mathbf{I}\mathbf{X} (\mathbf{W}-\mathbf{W}^\top) + (\hat{\mathbf{A}}+\hat{\mathbf{A}}^\top)\mathbf{X}(t)(\mathbf{V}-\mathbf{V}^\top)+\\\nonumber&\hspace{3cm}+ \beta(\tilde{\mathbf{{A}}}-\tilde{\mathbf{{A}}}^\top)\mathbf{X}(t)(\mathbf{Z}+\mathbf{Z}^\top) \Bigr) \\
\nonumber & ={\rm{diag}}\Bigl(\rm{vec}(\sigma'\Bigl(\mathbf{I}\mathbf{X} (\mathbf{W}-\mathbf{W}^\top) +  (\hat{\mathbf{A}}+\hat{\mathbf{A}}^\top)\mathbf{X}(t)(\mathbf{V}-\mathbf{V}^\top)  +\nonumber\\\nonumber&\hspace{3cm}+\beta(\tilde{\mathbf{{A}}}-\tilde{\mathbf{{A}}}^\top)\mathbf{X}(t)(\mathbf{Z}+\mathbf{Z}^\top) ))\Bigr) \\
\nonumber &={\rm{diag}}\Bigl(\sigma'\Bigl(((\mathbf{W}-\mathbf{W}^\top)^\top \otimes \mathbf{I)} \rm{vec}(\mathbf{X})  +\\\nonumber&\hspace{3cm}+ ((\mathbf{V}-\mathbf{V}^\top)^\top \otimes (\hat{\mathbf{A}}+\hat{\mathbf{A}}^\top)) \rm{vec}(\mathbf{X}(t)) +\\&\hspace{3cm}+\beta ((\mathbf{Z}+\mathbf{Z}^\top)^\top \otimes (\tilde{\mathbf{{A}}}-\tilde{\mathbf{{A}}}^\top)) \rm{vec}(\mathbf{X}(t)) \Bigr)\Bigr),
 \end{align}
 where $\mathbf{I}$ is the identity matrix. Therefore $\mathbf{M}_1$ is a diagonal matrix.
We note that Equation~\ref{eq:new_adgn_graphwise_APP} is the result of the composite function $\sigma(g(\mathbf{X}(t)))$, where $g(\mathbf{X}(t))=\mathbf{X}(t){(\mathbf{W}-\mathbf{W}^\top)} +{(\hat{\mathbf{A}}+\hat{\mathbf{A}}^\top)\mathbf{X}(t)(\mathbf{V}-\mathbf{V}^\top)} + \beta{(\tilde{\mathbf{{A}}}-\tilde{\mathbf{{A}}}^\top)\mathbf{X}(t)(\mathbf{Z}+\mathbf{Z}^\top)}$ and $\sigma$ is the activation function.

Therefore, $\textbf{M}_2$ results from the derivative of $g$ with respect to $\mathbf{X}(t)$. Considering the $\rm{vec}$ operator, we have
\begin{align}
\nonumber\textbf{M}_2 &= \rm{vec}(g'(\mathbf{X}(t)))\\
\nonumber&= g'(\rm{vec}(\mathbf{X}(t)))\\
&=
     {(\mathbf{W}-\mathbf{W}^\top)}  
     +{(\mathbf{V}-\mathbf{V}^\top)^\top \otimes (\hat{\mathbf{A}}+\hat{\mathbf{A}}^\top) } +\nonumber\\&\hspace{6cm}
     + \beta(\mathbf{Z}+\mathbf{Z}^\top)^\top \otimes  (\tilde{\mathbf{{A}}}-\tilde{\mathbf{{A}}}^\top)
\end{align}

\subsection{Datasets description}\label{app:swan_data}
\begin{figure}[h]
\centering
\begin{subfigure}{0.29\textwidth}
    \includegraphics[width=\textwidth]{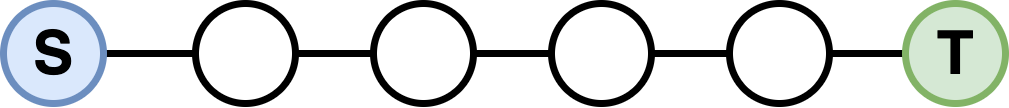}
    \vspace{-0.5mm}
    \caption{Line}
\end{subfigure}
\hspace{0.8cm}
\begin{subfigure}{0.26\textwidth}
    \includegraphics[width=\textwidth]{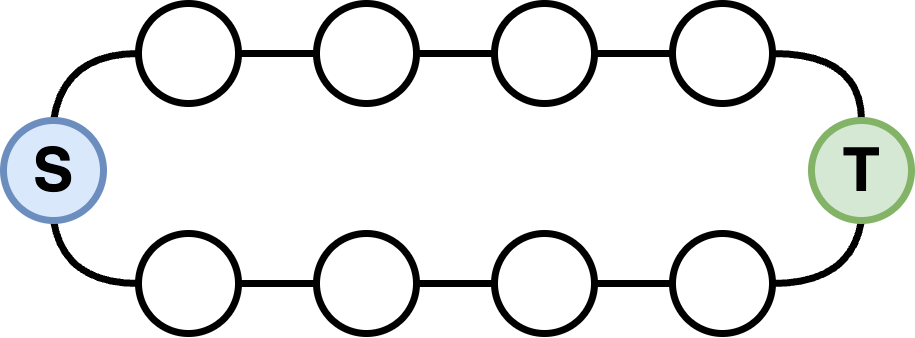}
    \caption{Ring}
\end{subfigure}
\hspace{0.8cm}
    \begin{subfigure}{0.26\textwidth}
    \includegraphics[width=\textwidth]{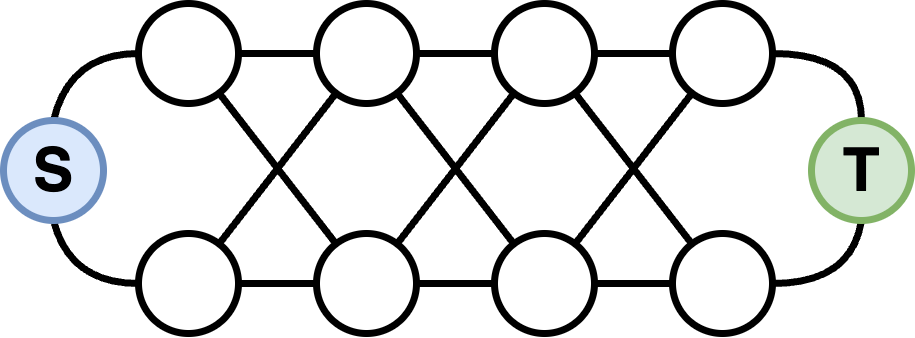}
\caption{Crossed-Ring}
\end{subfigure}
\caption{Line, ring, and crossed-ring graphs where the distance between source and target nodes is equal to 5. Nodes marked with ``S'' are source nodes, while the nodes with a ``T'' are target nodes.}
\label{fig:graph_transfer_example}
\end{figure}

\myparagraph{Graph Transfer Dataset} \label{app:datasets_graphtransfer}
We built the graph transfer datasets upon \cite{diGiovanniOversquashing}. In each task, graphs use identical topology, but, differently from the original work, nodes are initialized with random input features sampled from a uniform distribution in the interval $[0, 0.5)$. In each graph, we selected a source node and target node and initialized them with labels of value ``1'' and ``0'', respectively. We sampled graphs from three graph distributions, \ie line, ring, and crossed-ring. 
 Figure~\ref{fig:graph_transfer_example} shows a visual exemplification of the three types of graphs when the distance between the source and target nodes is 5. Specifically, ring graphs are cycles of size $n$, in which the target and source nodes are placed at a distance of $\lfloor n/2 \rfloor$ from each other. Crossed-ring graphs are also cycles of size $n$, but include crosses between intermediate nodes. Even in this case, the distance between source and target nodes remains $\lfloor n/2 \rfloor$. Lastly, the line graph contains a path of length $n$ between the source and target node. We refer the reader to Section~\ref{sec:static_graph_notation} for additional details about these graph distributions. 
 In our experiments, we consider a regression task, whose aim is to swap source and target node labels while maintaining intermediate nodes unchanged. We use an input dimension of 1, and the distance between source and target nodes is equal to 3, 5, 10, and 50. We generated 1000 graphs for training, 100 for validation, and 100 for testing.

\myparagraph{Graph Property Prediction}\label{app:datasets_graphproppred}
In our experiments on graph property prediction, we used the same data outlined in Appendix~\ref{app:adgn_data_stats}, thus graphs are randomly selected from various graph distributions and each node is assigned with random identifiers while target values represented single-source shortest paths, node eccentricity, and graph diameter. The dataset comprised a total of 7040 graphs, with 5120 used for training, 640 for validation, and 1280 for testing.

\myparagraph{Long Range Graph Benchmark}\label{app:datasets_lrgb}
In the Long Range Graph Benchmark section, we considered the ``Peptides-func'', ``Peptides-struct'', and ``PascalVOC-sp'' datasets~\cite{LRGB}. In the first two datasets, the graphs correspond to 1D amino acid chains, and they are derived such that the nodes correspond to the heavy (non-hydrogen) atoms of the peptides while the edges represent the bonds between them. Peptides-func is a multi-label graph classification dataset, with a total of 10 classes based on the peptide function, e.g., Antibacterial, Antiviral, cell-cell communication, and others. Peptides-struct is a multi-label graph regression dataset based on the 3D structure of the peptides. Specifically, the task consists of the prediction of the inertia of the molecules using the mass and valence of the atoms, the maximum distance between each atom-pairs, sphericity, and the average distance of all heavy atoms from the plane of best fit. Both Peptides-func and Peptides-struct consist of 15,535 graphs with a total of 2.3 million nodes. 
PascalVOC-sp is a node classification dataset composed of graphs created from the images in the Pascal VOC 2011 dataset~\citep{everingham2015pascal}. 
A graph is derived from each image by extracting superpixel nodes using the SLIC algorithm~\citep{achanta2012slic} and constructing a rag-boundary graph to interconnect these nodes.
Each node represents a region of an image that belongs to a specific class. The dataset consists of 11,355 graphs for a total of 5.4 million nodes. The task involves predicting the semantic segmentation label for each superpixel node across 21 different classes.

We applied stratified splitting to Peptides-func and Peptides-struct to generate balanced train–valid–test dataset splits, using the ratio of 70\%–15\%–15\%, respectively.
On PascalVOC-sp we consider 8,498 graphs for training, 1,428 for validation and 1,429 test.

\subsection{Explored hyperparameter space}\label{app:swan_hyperparams}
In Table~\ref{tab:hyperparams} we report the grids of hyperparameters employed in our experiments by each method. We recall that the hyperparameters $\epsilon$, $\gamma$, and $\hat{\bfA}$ refer to SWAN and A-DGN, while $\beta$ only to SWAN. Moreover, we note that, in each graph transfer task, we use a number of layers that isequal to the distance between the source and target nodes.
\begin{table}[h]
\centering
\caption{The grid of hyperparameters employed during model selection for the 
graph transfer tasks (\emph{Transfer}), graph property prediction tasks (\emph{GraphProp}), and Long Range Graph Benchmark (\emph{LRGB}). We observe that, in each graph transfer task, we use a number of layers that is equal to the distance between the source and target nodes. \label{tab:hyperparams}}

\scriptsize
\begin{tabular}{l|l|l|l}
\toprule
\multirow{2}{*}{\textbf{Hyperparameters}}  & \multicolumn{3}{c}{\textbf{Values}}\\\cmidrule{2-4}
& {\bf {Transfer}} & {\bf {GraphProp}} & {\bf {LRGB}}\\\midrule
optimizer       & Adam            & Adam                     & AdamW\\
learning rate   & 0.001           & 0.003                    & 0.001, 0.0005\\
weight decay    & 0              & $10^{-6}$                & 0, 0.0001\\
n. layers       & 3, 5, 10, 50    & 1, 5, 10, 20             & 5, 8, 16, 32\\
embedding dim   & 64              & 10, 20, 30               & 64, 128\\
$\hat{\bfA}$    & $\bfA$, $\bfD^{-1/2}\bfA \bfD^{-1/2}$ & $\bfA$, $\bfD^{-1/2}\bfA \bfD^{-1/2}$ & $\bfA$, $\bfD^{-1/2}\bfA \bfD^{-1/2}$\\ 
$\sigma$        & tanh            & tanh                     & tanh\\
$\epsilon$      & 0.5, 0.1        & 1, 0.1, 0.01             & 1, 0.01\\
$\gamma$        & 0.1             & 1, 0.1                   & 1, 0.1\\
$\beta$         & 1, 0.1, 0.01, -1           & 2, 1, 0.5, 0.1, -0.5, -1 & 1, 0.1\\
\bottomrule
\end{tabular}
\end{table}

\chapter{Supplementary materials of Chapter~\ref{ch:phdgn}}\label{app:suppl_ch5}
\section{Explored hyperparameter space}\label{app:phdgn_hyperparams}
Before providing the grid of hyperparameters employed in our experiments, we first provide architectural details of the dissipative components in our PH-DGN.

As typically employed, we follow physics-informed approaches that learn how much dissipation and external control is necessary to model the observations \citep{desai2021port}. In particular, we consider these deep (graph) network architectures for the dampening term $D(\mathbf{q})$ and external force term $F(\mathbf{q},t)$, assuming for simplicity $\mathbf{q}_u \in \mathbb{R}^{\frac{d}{2}}$.

\textbf{Dampening} $D(\mathbf{q})$: it is a square $\frac{d}{2}\times \frac{d}{2}$ matrix block with only diagonal entries being non-zero and defined as:
\begin{itemize}
    \item \emph{param}: a learnable vector $\mathbf{w}\in \mathbb{R}^{\frac{d}{2}}$.
    \item \emph{param+}: a learnable vector followed by a ReLU activation, \ie $\text{ReLU}(\mathbf{w})\in \mathbb{R}^{\frac{d}{2}}$.
    \item \emph{MLP4-ReLU}: a $4$-layer MLP with ReLU activation and all layers of dimension $\frac{d}{2}$.
    \item \emph{DGN-ReLU}: a DGN node-wise aggregation layer from Equation~\ref{eq:simple_aggregation} with ReLU activation.
\end{itemize}

\textbf{External forcing} $F(\mathbf{q},t)$: it is a $\frac{d}{2}$ dimensional vector where each component is the force on a component of the system. Since it takes as input $\frac{d}{2}+1$ components it is defined as:
\begin{itemize}
    \item \emph{MLP4-Sin}: 3 linear layers of $\frac{d}{2}+1$ units with sin activation followed by a last layer with $\frac{d}{2}$ units.
    \item \emph{DGN-tanh}: a single node-wise DGN aggregation from Equation~\ref{eq:simple_aggregation} followed by a tanh activation.
\end{itemize}

Note that dampening, \ie energy loss, is only given when $D(\mathbf{q})$ represents a positive semi-definite matrix. Hence, we used ReLU-activation, except for \emph{param}, which offers a flexible trade-off between dampening and acceleration learned by backpropagation.

In Table~\ref{tab:phdgn_hyperparams} we report the grid of hyperparameters employed in our experiments by each method.
\begin{table}[h]
\centering
\caption{The grid of hyperparameters employed during model selection for the graph transfer tasks (\emph{Transfer}), graph property prediction tasks (\emph{GraphProp}), and Long Range Graph Benchmark (\emph{LRGB}).
We refer to Appendix~\ref{app:phdgn_hyperparams} for more details about dampening and external force implementations.}\label{tab:phdgn_hyperparams}
\scriptsize 
\begin{tabular}{l|l|l|l}
\toprule
\multirow{2}{*}{\textbf{Hyperparameters}}  & \multicolumn{3}{c}{\textbf{Values}}\\\cmidrule{2-4}
& {\bf \emph{Transfer}} & {\bf \emph{GraphProp}} & {\bf \emph{LRGB}}\\\midrule
optimizer       & Adam            & Adam                     & AdamW\\
learning rate   & 0.001           & 0.003                    & 0.001, \ 0.0005\\
weight decay    & 0              & $10^{-6}$                 & 0\\
embedding dim   & 64              & 10, 20, 30               & 195, \ 300\\
n. layers ($L$)       & 3, \ 5, \ 10, \ 50,    & 1, \ 5, \ 10, \ 20, \ 30             & 32, \ 64\\
                      & 100, \ 150 & &\\
termination time ($T$) & $L\epsilon$             & 0.1, \ 1, \ 2, \ 3, \ 4               & 3, \ 5, \ 6\\
$\epsilon$      & 0.5, \ 0.2, \ 0.1, & $T/L$             & $T/L$\\
                & 0.05, \ 0.01, \ $10^{-4}$ & & \\
$\Phi_\bfp$     & Eq.~\ref{eq:simple_aggregation}, \ GCN & Eq.~\ref{eq:simple_aggregation}, \ GCN & Eq.~\ref{eq:simple_aggregation}, \ GCN\\
$\Phi_\bfq$     & Eq.~\ref{eq:simple_aggregation}, \ GCN & Eq.~\ref{eq:simple_aggregation}, \ GCN & GCN\\
readout input   & $\bfp$, $\bfq$, $\bfp\|\bfq$ & $\bfp$, $\bfq$, $\bfp\|\bfq$ & $\bfp$, $\bfq$, $\bfp\|\bfq$ \\
$\sigma$        & tanh            & tanh                     & tanh\\
dampening       & -- & \emph{param}, \emph{param+},  & \emph{param} \\
                &    & \emph{MLP4-ReLU}, \emph{DGN-ReLU} & \\
external Force  & --  & \emph{MLP4-Sin}, \emph{DGN-tanh} & \emph{DGN-tanh} \\
n. readout layers & 1 & 1 & 1, \ 3\\
\bottomrule
\end{tabular}
\end{table}

\chapter{Supplementary materials of Chapter~\ref{ch:tgode}}\label{app:suppl_ch6}
\section{Datasets description and statistics}\label{app:tgode_data_stats}
In the heat diffusion experiments, we consider a grid graph consisting of 70 nodes, each of which is characterized by an initial temperature $\mathbf x_u(0)$ randomly sampled in the range between $0$ and $0.2$. We randomly alter the initial temperature profile by generating hot and cold spikes located at some nodes. A hot spike is characterized by a temperature between $10$ and $15$, while a cold spike is between $-15$ and $-10$. 
Each altered node has a 40\% chance of being associated with a cold spike and 60\% with a hot spike.
We considered two different experimental scenarios depending on the number of altered nodes. In the first scenario, we alter the temperature of a single node. In the second one, we alter the temperature of one third of the graph's nodes. We will refer to these settings as \textit{single-spike} and \textit{multi-spikes}, respectively.

\begin{figure}
    \centering
    \subfloat[]{\includegraphics[width=0.45\textwidth]{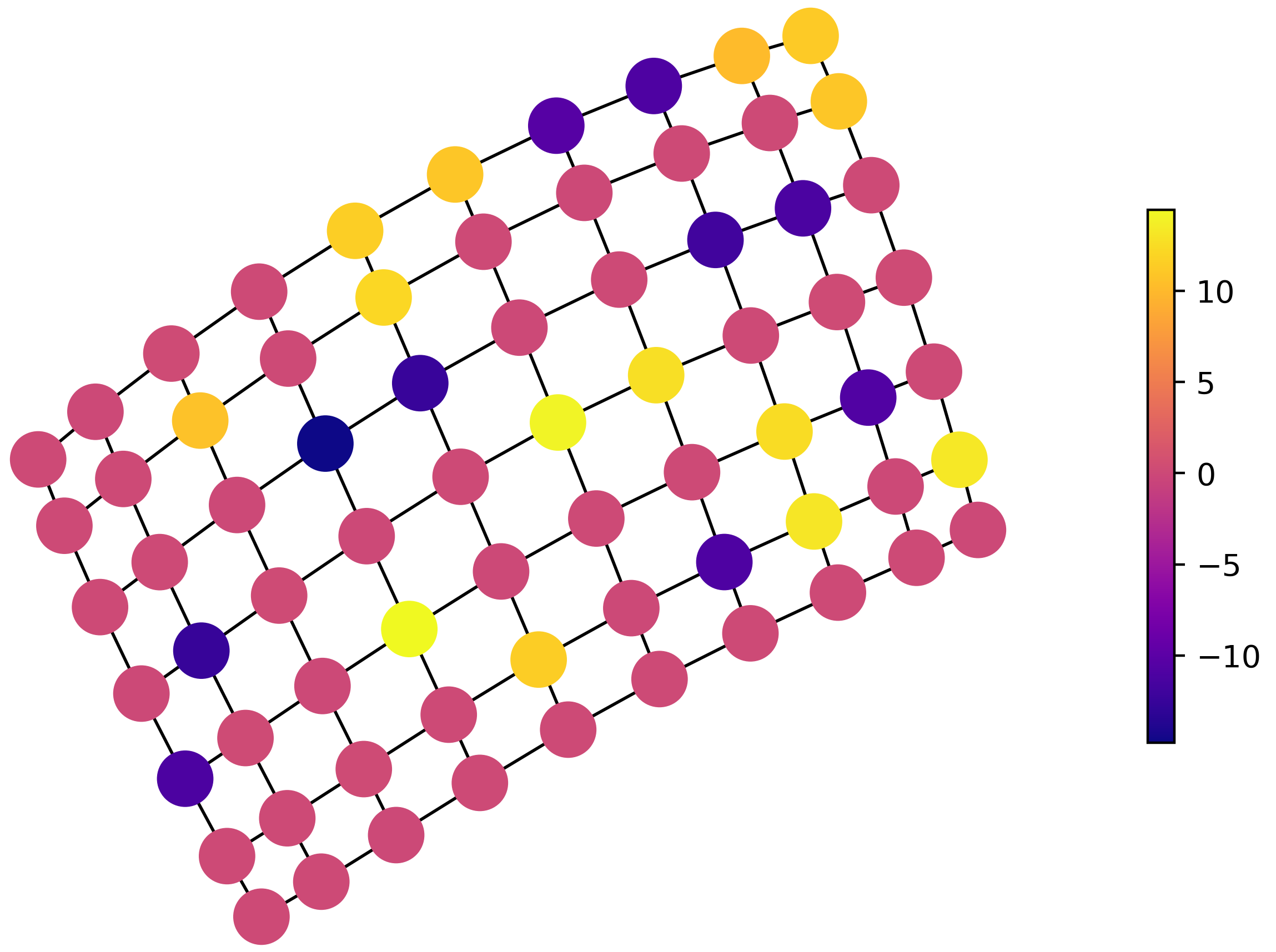}}
    \subfloat[]{\includegraphics[width=0.45\textwidth]{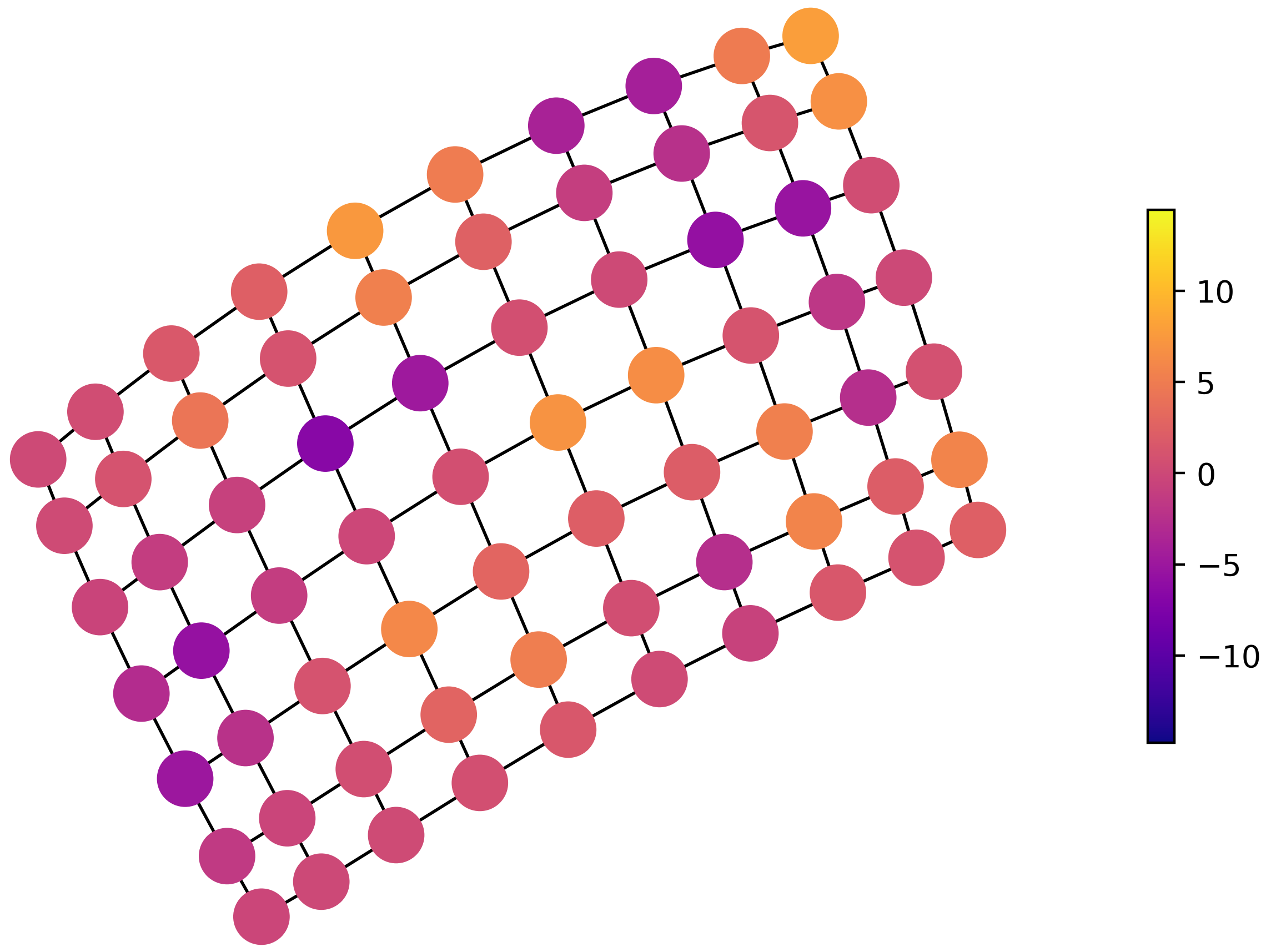}}
    \caption{(a) A grid graph consisting of 70 nodes in which each node is characterized by an initial temperature. Darker colors correspond to colder temperatures, while brighter colors mean warmer temperatures. (b) The heat diffusion simulation is computed through 1000 steps forward Euler's method leveraging $-\mathbf{L}\mathbf{X}(t)$ as diffusion.}
    \label{fig:heat_diffusion}
\end{figure}

We collected the ground truth by simulating the heat diffusion equation through the forward Euler's method with step size $\epsilon=10^{-3}$. Figure~\ref{fig:heat_diffusion} illustrates two snapshot graphs from the simulated heat diffusion.
The training set consists of 100 randomly selected timestamps over the 1000 steps used to simulate the diffusion process. The validation and test sets are generated from two different simulations similar to the one used for building the training set. However, validation and test 
sets are obtained through 500-step simulations, and only 50 of them are kept as validation/test sets. We simulated seven different diffusion functions, \ie $-\mathbf{L}\mathbf{X}(t)$, $-\mathbf{L}^2\mathbf{X}(t)$, $-\mathbf{L}^5\mathbf{X}(t)$, $-\tanh(\mathbf{L})\mathbf{X}(t)$, $-5\mathbf{L}\mathbf{X}(t)$, $-0.05\mathbf{L}\mathbf{X}(t)$, and $-(\mathbf{L}+\mathcal{N}_{0,1})\mathbf{X}(t)$. Here, $\mathcal{N}_{0,1}$ stands for a noise sampled from a standard normal distribution, and $\mathbf{L}$ is the normalized graph Laplacian. 

In the traffic benchmarks, we considered six real-world graph benchmarks for traffic forecasting: MetrLA, Montevideo, PeMS03, PeMS04, PeMS07, and PeMS08; we report additional details about the datasets in Table~\ref{tab:graph_benchmark_stats}. 
We used a modified version of the original datasets where we employed irregularly sampled observations. 
We generated irregular time series by randomly selecting a third of the original graph snapshots for most of the experiments; ratios from 3\% to 94\% are studied in Section~\ref{par:tgode_ablation}. 

\begin{table}[ht]
\centering
\caption{Statistics of the original version of the datasets.}\label{tab:graph_benchmark_stats}
\scriptsize
\begin{tabular}{lcccc}
\toprule
           & \textbf{\# Steps} & \textbf{\# Nodes} & \textbf{\# Edges} & \textbf{Timespan}\\\midrule 
MetrLA     & 34,272 & 207 & 1,515 & \nth{1} Mar. - \nth{30} Jun. 2012 \\ 
Montevideo &   739  & 675 & 690   & \nth{1} Oct. - \nth{31} Oct. 2020\\ 
PeMS03     & 26,208 & 358 & 442   & \nth{1} Sep. - \nth{30} Nov. 2018 \\ 
PeMS04     & 16,992 & 307 & 209   & \nth{1} Jan. - \nth{28} Feb. 2018\\ 
PeMS07     & 28,225 & 883 & 790   & \nth{1} May - \nth{31} Aug. 2017\\ 
PeMS08     & 17,856 & 170 & 137   & \nth{1} Jul. - \nth{31} Aug. 2016\\ 
\bottomrule
\end{tabular}
\end{table}

\section{Explored hyperparameter space}\label{app:tgode_hyperparam}
In Table~\ref{tab:tgode_general_configs} we report the grids of hyperparameters employed in our experiments by each method. We recall that the hyperparameter $\epsilon$ refers only to our method.
\begin{table}[h]
\centering
\caption{The grid of hyperparameters employed during model selection for the \textcolor{first}{heat diffusion tasks} (\textcolor{first}{Heat}) and \textcolor{second}{graph benchmark tasks} (\textcolor{second}{Bench}). 
The $\epsilon$ hyperparameter is only used by our method (\ie TG-ODE), and \emph{embedding dim} equal to \emph{None} means that no encoder and readout are employed.\label{tab:tgode_general_configs}}
\scriptsize
\begin{tabular}{l|c|c}
\toprule
\multirow{2}{*}{\textbf{Hyperparameters}} & \multicolumn{2}{c}{\textbf{Values}}\\\cmidrule{2-3}
 &  \textcolor{first}{\textbf{Heat}} & \textcolor{second}{\textbf{Bench}} \\\midrule
learning rate & \multicolumn{2}{c}{$10^{-2}$, $10^{-3}$, $10^{-4}$}\\
weight decay &  \multicolumn{2}{c}{$10^{-2}$, $10^{-3}$}\\
$\eta$ & \multicolumn{2}{c}{concat, sum, $\eta(\overline{\mathbf x}, \hat{\mathbf x})=\overline{\mathbf x}$}\\
activation fun. & \multicolumn{2}{c}{tanh, relu, identity}\\
embedding dim. & \textcolor{first}{None, 8} & \textcolor{second}{64, 32}\\
$\epsilon$ & \textcolor{first}{$10^{-3}$} & \textcolor{second}{1, 0.5, $10^{-1}$, $10^{-2}$, $10^{-3}$}\\
n. hops & \textcolor{first}{5} &  \textcolor{second}{1, 2, 5} \\
\bottomrule
\end{tabular}
\end{table}

\chapter{Supplementary materials of Chapter~\ref{ch:ctan}}\label{app:suppl_ch7}
\section{Datasets description and statistics}\label{app:ctan_datasets}
Table~\ref{tab:ctan_stats} contains the statistics of the employed datasets. In the following, we describe the datasets and their generation.  

\begin{table}[ht]
\centering
\caption{Statistics of the datasets used in our experiments. We report the total number of nodes and edges in the dataset for the temporal path graph (\ie T-PathGraph) and temporal Pascal VOC (\ie T-PascalVOC). ``Chron.'' means that chronological order is preserved.}\label{tab:ctan_stats}

\scriptsize
\setlength{\tabcolsep}{4pt} 
\begin{tabular}{lccccc}
\toprule
& \textbf{\# Nodes} & \textbf{\# Edges}  & \textbf{\# Edge ft.}    & \textbf{Split} & \textbf{Surprise Index} \\\midrule

T-PathGraph & 3,000-20,000 & 2,000-19,000 & 1 & 70/15/15 & 1.0\\
T-PascalVOC$_{10}$ & 2,671,704 & 2,660,352 & 14 & 70/15/15 & 1.0\\
T-PascalVOC$_{30}$ & 2,990,466 & 2,906,113 & 14 & 70/15/15 & 1.0\\
Wikipedia & 9,227 & 157,474 & 172 & 70/15/15, Chron. & 0.42\\
Reddit & 11,000 & 672,447 & 172 & 70/15/15, Chron. & 0.18 \\
LastFM & 2,000 & 1,293,103 & 2 & 70/15/15, Chron. & 0.35 \\
MOOC & 7,144 & 411,749 & 4 & 70/15/15, Chron. & 0.79\\
tgbl-wiki-v2 & 9,227 & 157,474  & 172 & 70/15/15, Chron. & 0.108\\
tgbl-review-v2 & 352,637 & 4,873,540 & - & 70/15/15, Chron. & 0.987\\
tgbl-coin-v2 & 638,486 	 & 22,809,486 & - & 70/15/15, Chron. & 0.120\\
tgbl-comment & 994,790 	 	 & 44,314,507 & - & 70/15/15, Chron. & 0.823\\
\bottomrule
\end{tabular}
\end{table}

\begin{figure}[t]
\begin{center}
    \includegraphics[width=0.4\linewidth]{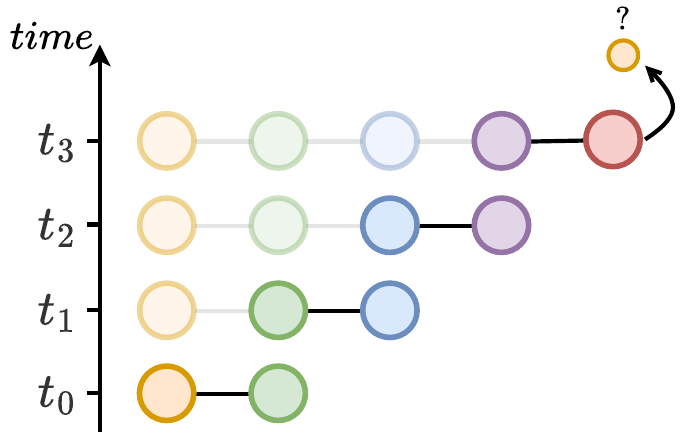}
\end{center}
\caption{The illustration of the sequence classification task on a temporal path graph consisting of 5 nodes. The first node (colored in orange) has an initial feature that can be either $1$ or $-1$. All the other nodes and edges have a feature set to random value sampled uniformly in $[-1, 1]$. At the end of the sequence, the representation computed for the last node (colored in red) is used to predict the original value of the first node. At each timestamp, the faded portion of the graph corresponds to historical information.}
\label{fig:path-graph}
\end{figure}

\myparagraph{Sequence classification on temporal path graphs} To craft a temporal long-range problem, we first introduced a sequence classification problem on path graphs (see Section~\ref{sec:static_graph_notation}), which is a simple linear graph consisting of a sequence of nodes where each node is connected to the previous one. In the temporal domain, the nodes of the path graph appear sequentially over time from first to last (e.g., bottom-to-top in Figure~\ref{fig:path-graph}).

We define the task objective as the prediction of the feature seen in the first node (colored in orange in Figure~\ref{fig:path-graph}) by making the prediction leveraging only the last node representation (colored in red in Figure~\ref{fig:path-graph}) computed at the end of the sequence, i.e., when the last event appears. Note that this task is akin to the sequence classification task designed in~\cite{chang2018antisymmetricrnn}, with the addition of a graph convolution.
We set the feature of the first node to be either $1$ or $-1$, while we set every other node and edge feature to be sampled uniformly in the range $[-1, 1]$. 
In other words, the feature $\mathbf{x}_{u_0}$ of the first node $u_0$ contains a signal to be remembered as noise is added through the propagations steps along the graph.
Formally, we create a C-TDG: $\mathcal{G} = \{o_t\, | \, t\in[t_0, t_n]\}$, such that $$o_t=(t, \,\mathcal{E}_\oplus,\,u_t,\,u_{t+1}, \mathbf{x}_{u_t}, \mathbf{x}_{u_{t+1}}, \mathbf{e}_{u_t, u_{t+1}}),$$ 
where $\mathcal{E}_\oplus$ corresponds to an edge addition event; $\mathbf{x}_{u_0} \sim \text{Bernoulli}(0.5)$\footnote{Note that we sample 1 or -1 rather than 0 or 1 to make the problem balanced around zero.};  $\mathbf{x}_{u_j} \sim \mathcal{U}_{[-1, 1]}, \forall j > t_0$; and $\mathbf{e}_{u_t,u_{t+1}} \sim \mathcal{U}_{[-1, 1]}, \forall t$.

For this task we considered 8 temporal graph path datasets with different sizes, ranging from $n=3$ to $n=20$, with $n$ the number of nodes. For every graph size we generate 1,000 different graphs, and we split the dataset into train/val/test with the ratios 70\%-15\%-15\%.

\begin{figure}[t]
\begin{center}
    \includegraphics[width=\linewidth]{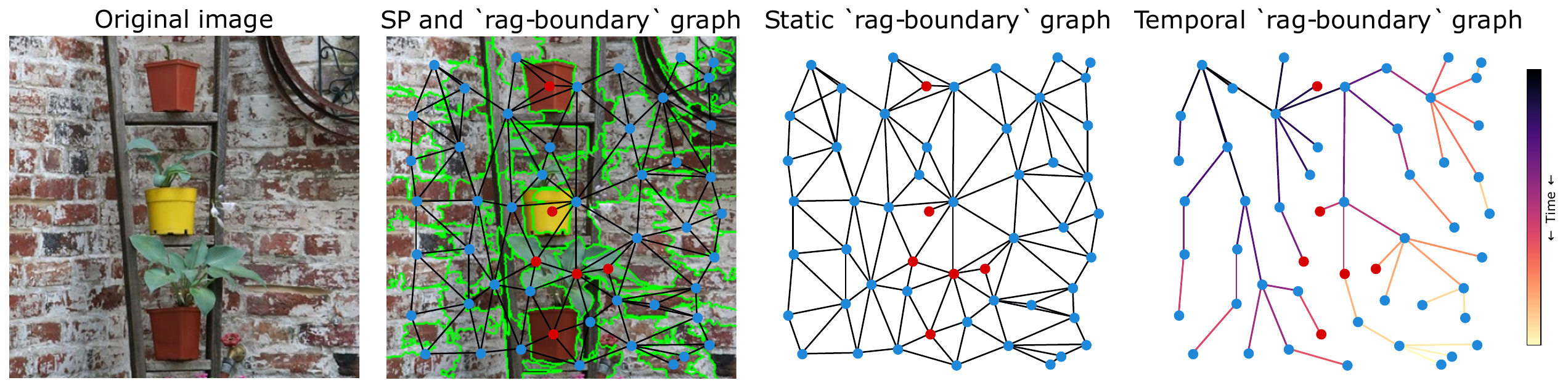}
\end{center}
\caption{Construction of the Temporal PascalVOC-SP dataset. The SLIC algorithm extracts patches from an image. We create the rag-boundary graph connecting neighboring patches based on spatial closeness. We construct a temporal graph by traversing from the topleftmost node with BFS. The goal of the task is to predict the class of the destination node at each visited edge - in the figure, either 'potted plant' (red) or 'background' (blue). For clarity in this visualization, the compactness of the SLIC algorithm is low.}
\label{fig:pascalvoc-vis}
\end{figure}

\myparagraph{Temporal Pascal-VOC}
We use the PascalVOC-SP dataset introduced in \cite{LRGB} and discussed in Appendix~\ref{app:swan_data} to design a new temporal long-range task for edge classification. 
PascalVOC-SP is a node classification dataset composed of graphs  derived from each image in the Pascal VOC 2011 dataset~\citep{everingham2015pascal} by extracting superpixel nodes using the SLIC algorithm~\citep{achanta2012slic} and constructing a rag-boundary graph to interconnect these nodes.
Each node in a graph corresponds to one region of the image belonging to a particular class, see Figure~\ref{fig:pascalvoc-vis} for an example. 
PascalVOC-SP contains long-range interactions between spatially distant image patches, evidenced by its average shortest path length of 10.74 and average diameter of 27.62~\citep{LRGB}.

To craft a temporal task, we consider that nodes in a rag-boundary graph appear from the top-left to the bottom-right of the image, sequentially. 
We do so by selecting the top-leftmost node, \ie the one closest (by means of $L_1$ norm) to the origin in image coordinates.
From this node, we traverse the graph with a Breadth-First-Search, visiting each node exactly once. 
The order of edge traversal corresponds to the timestamp of edge appearance in the temporal task.
We set the task's objective to be the prediction of the class of the node that is being visited by the current edge. 
Note that the traversal removes a large number of edges from the initial graph, making the propagation of class information more difficult, see Figure~\ref{fig:pascalvoc-vis}.

Neighborhoods are constructed based on coordinates, connecting a node with its 8 spatially closest neighbors.
Nodes have 12 features extracted by channel-wise statistics on the image (mean, std, max, min) and 2 features defining the spatial location of the superpixel; we normalize these spatial features in the $[0, 1]$ range.
We consider two SLIC superpixels compactness of 10 and 30 (smaller compactness means fewer patches).
To allow for batching, we fix the number of nodes in each graph, allowing batching of edges that occur at the same timestep across different graphs together.
To do so, we discard rag-boundary  graphs with fewer nodes than the limit, and discard excess nodes on graphs with more nodes than the limit, according to time (i.e., the most recent nodes are dropped).
This removes a small number of nodes corresponding to image patches on the bottom-right of the image.
In practice, for the two compactness levels 10 and 30, we set the number of minimum nodes per graph to be 434 and 474, which gives us 6,156 and 6,309 temporal graphs (out of the total 11,355 images in the dataset).
The resulting temporal datasets have 2,660,352 and 2,906,113 edges respectively.

\myparagraph{C-TDG benchmarks}
For the C-TDG benchmarks on future link prediction, we consider four well-known datasets proposed by \cite{jodie}:
\begin{itemize}
    \item \textbf{Wikipedia}: one month of interactions (\ie 157,474 interactions) between user and Wikipedia pages. Specifically, it corresponds to the edits made by 8,227 users on the 1,000 most edited Wikipedia pages;
    \item \textbf{Reddit}: one month of posts (\ie interactions) made by 10,000 most active users on 1,000 most active subreddits, resulting in a total of 672,447 interactions;
    \item \textbf{LastFM}: one month of who-listens-to-which song information. The dataset contains 
    1000 users and the 1000 most listened songs, resulting in 1,293,103 interactions.
    \item \textbf{MOOC}: it consists of actions done by students on a MOOC online course. The dataset contains 7,047 students (\ie users) and 98 items (\eg videos and answers), resulting in 411,749 interactions.
\end{itemize}
Since the datasets do not contain negative instances, we perform negative sampling by randomly sampling non-occurring links in the graph, as follows: (i) during training we sample negative destinations only from nodes that appear in the training set, (ii) during validation we sample them from nodes that appear in training set or validation set and (iii) during testing we sample them from the entire node set. 

For all the datasets, we considered the same chronological split into train/val/test with the ratios 70\%-15\%-15\% as proposed by \cite{TGAT}.

\myparagraph{Transductive vs Inductive Settings}
In Section~\ref{sec:exp_jodie} we employed transductive setting and random negative sampling as in~\cite{jodie, TGAT, tgn_rossi2020, yu2023towards, cong2023we}.
We chose not to employ an inductive setting as it is not easily applicable to C-TDGs. Specifically, there is no clear consensus in the literature regarding the definition of inductive settings, making it difficult to identify the nodes considered for assessing this experimental setup (e.g. \cite{TGAT} differs from \cite{tgn_rossi2020}). Some definitions of inductive settings lead to the number of sampled inductive nodes to be not statistically relevant for evaluation.
Other interpretations of inductive settings disrupt the true dynamics of the graph, i.e., in~\cite{tgn_rossi2020}, certain nodes and their associated edges are removed from the training set with the purpose of isolating an inductive set of nodes.
Thanks to the analysis performed in~\cite{yu2023towards}, we can also observe that among all the considered datasets in Chapter~\ref{ch:ctan} there is mix of inductive and transductive edges, which can be measured with the \textit{surprise index} from~\cite{yu2023towards}, measuring the proportion of unseen edges at test time; reported in Table~\ref{tab:ctan_stats}.
Hence, achieving strong performance on tasks with a high surprise index offers valuable insights into the model's capability to address the inductive setting. Comparing CTAN performance to the surprise index, it is clear that CTAN can cope reasonably well even in fully inductive tasks, such as those in Section~\ref{sec:exp_non-diss} where it generally ranks first among other baselines.

\myparagraph{Temporal Graph Benchmark}
We consider four well-known datasets proposed in the Temporal Graph Benchmark (TGB)~\citep{huang2023temporal}:
\begin{itemize}
    \item \textbf{tgbl-wiki-v2}: This dataset stores the co-editing network on Wikipedia pages over one month introduced in \citep{jodie};
    \item \textbf{tgbl-review-v2}: Amazon product review network collected from 1997 to 2018 where users rate different products in the electronics category from a scale from one to five. Both users and products are nodes and each edge represents a particular review from a user to a product at a given time. Only users with a minimum of 10 reviews. The task consists in predicting which product a user will review at a given time.
    \item \textbf{tgbl-coin-v2}: cryptocurrency transaction network based on the Stablecoin ERC20 transactions
dataset~\citep{NEURIPS2022_e245189a}, collecting transaction data of 5 stablecoins and 1 wrapped token from April 1st, 2022 to November 1st, 2022. Each node is an address and each edge represents the transfer of funds between two addresses at a time. The task consists in predicting with which destination a given address will interact at a given time;
    \item \textbf{tgbl-comment}: directed reply network of Reddit where users reply to each other’s
threads, collected from 2005 to 2010. Each node is a user and each interaction is a reply from one user to another. The task consists in predicting if a given user will reply to another one at a given time.
\end{itemize}

TGB provides pre-sampled negative edge sets with both \textit{random} and \textit{historical} negatives~\citep{edgebank}. For all the datasets, we considered the same chronological split into train/val/test with the ratios 70\%-15\%-15\% as proposed by \cite{huang2023temporal}.

\section{Explored hyperparameter space}\label{app:ctan_hyperparams}
Table~\ref{tab:ctan_hyper_param} reports the grids of hyperparameters employed in our experiments by each method in Sections~\ref{sec:exp_non-diss} and \ref{sec:exp_jodie}, and Table~\ref{tab:hyper_param_tgb} lists the hyperparameters used in our experiment in Section~\ref{sec:tgb_results}. We recall that the hyperparameters $\epsilon$, $\gamma$, and $\eta$ refer only to our method.
We used dropout only for GraphMixer and DyGFormer, where the values are loosely based on best-performing values in~\citet{yu2023empirical}.

\begin{table}[h]
\centering
\caption{The grid of hyperparameters employed during model selection for the following three tasks: \textcolor{seqcolor}{Sequence classification on temporal path graphs} (\textcolor{seqcolor}{Seq}), \textcolor{pascalcolor}{Temporal Pascal-VOC} (\textcolor{pascalcolor}{Pasc}), and \textcolor{linkcolor}{Link Prediction} (\textcolor{linkcolor}{Link})
. 
For \emph{Seq} and \emph{Pasc}, we conducted \textcolor{seqcolor}{10 runs} and \textcolor{pascalcolor}{5 runs} with different random seeds for different weight initializations \emph{for each configuration}, whereas for \emph{Link}, we conducted \textcolor{linkcolor}{5 runs} only for the configuration that resulted in the best performance in the initial run. 
For the three tasks, the models were configured to have a maximum number of learnable parameters of \textcolor{seqcolor}{$\sim$20k}, \textcolor{pascalcolor}{$\sim$40k}, and \textcolor{linkcolor}{$\sim$140k}, respectively. 
Training was conducted for \textcolor{seqcolor}{20 epochs}, \textcolor{pascalcolor}{200 epochs}, and \textcolor{linkcolor}{1000 epochs}, respectively. For \emph{Seq} and \emph{Pasc}, we employed a scheduler \emph{halving the learning rate} with a patience of \textcolor{seqcolor}{5 epochs}, \textcolor{pascalcolor}{20 epochs}, respectively, whereas for \emph{Link} we used \textcolor{linkcolor}{early stopping} with a patience of \textcolor{linkcolor}{50 epochs}. 
For all tasks, the neighbor sampler size was set to 5. The batch size was set to \textcolor{seqcolor}{128}, \textcolor{pascalcolor}{256}, and \textcolor{linkcolor}{256}, respectively.
We used the \textcolor{seqcolor}{loss}, \textcolor{pascalcolor}{F1-score}, and \textcolor{linkcolor}{AUC} on the validation set to optimize for the hyperparameters.
We used dropout only for GraphMixer and DyGFormer, where the values are loosely based on best-performing values in~\cite{yu2023empirical}.}
%
\label{tab:ctan_hyper_param}
\scriptsize
\begin{tabular}{l|l|ccc}
\toprule
\multirow{2}{*}{\textbf{Hyperparameters}} & \multirow{2}{*}{\textbf{Method}} & \multicolumn{3}{c}{\textbf{Values}}\\\cmidrule{3-5}
&& \textcolor{seqcolor}{\bf Seq} & \textcolor{pascalcolor}{\bf Pasc} & \textcolor{linkcolor}{\bf Link} \\\midrule
\multicolumn{2}{l|}{optimizer} & \multicolumn{3}{c}{Adam} \\
\multicolumn{2}{l|}{learning rate} & \textcolor{seqcolor}{$3\cdot10^{-4}$} & \textcolor{pascalcolor}{$3\cdot10^{-4}$} & \textcolor{linkcolor}{$10^{-4}$, $10^{-5}$} \\
\multicolumn{2}{l|}{weight decay} & \textcolor{seqcolor}{$10^{-7}$} & \textcolor{pascalcolor}{$10^{-5}$} & \textcolor{linkcolor}{$10^{-6}$} \\
\multicolumn{2}{l|}{n. GCLs} & \multicolumn{3}{c}{1, 3, 5} \\
\multicolumn{2}{l|}{$\sigma$} & \multicolumn{3}{c}{tanh} \\
\multicolumn{2}{l|}{$\epsilon$} & \textcolor{seqcolor}{1, 0.5, $10^{-1}$, $10^{-2}$} & \textcolor{pascalcolor}{1, 0.5, $10^{-1}$, $10^{-2}$} & \textcolor{linkcolor}{0.5, $10^{-1}$, $10^{-2}$, $10^{-3}$} \\
\multicolumn{2}{l|}{$\gamma$} & \textcolor{seqcolor}{1, 0.5, $10^{-1}$, $10^{-2}$} & \textcolor{pascalcolor}{1, 0.5, $10^{-1}$, $10^{-2}$} & \textcolor{linkcolor}{0.5, $10^{-1}$, $10^{-2}$, $10^{-3}$} \\
\multicolumn{2}{l|}{$\eta$} & \multicolumn{3}{c}{concat, $\eta = \text{tanh}(\textbf{h}^{i-1}(t_e)||\mathbf{x}(i))$} \\
\multicolumn{2}{l|}{dropout} & \textcolor{seqcolor}{0.1, 0.2} & \textcolor{pascalcolor}{0.1, 0.2} & $-$ \\
\multicolumn{2}{l|}{time dim} & \textcolor{seqcolor}{1} & \textcolor{pascalcolor}{1} & \textcolor{linkcolor}{16} \\\midrule
\multirow{7}{*}{\shortstack{memory dim \\(= DGN dim)}}& DyGFormer & \textcolor{seqcolor}{10, 5} & \textcolor{pascalcolor}{14, 7} & $-$ \\
& DyRep & \textcolor{seqcolor}{53, 26} & \textcolor{pascalcolor}{74, 37} & \textcolor{linkcolor}{118, 87} \\
& GraphMixer & \textcolor{seqcolor}{30, 15} & \textcolor{pascalcolor}{24, 12} & $-$ \\
& JODIE & \textcolor{seqcolor}{69, 34} & \textcolor{pascalcolor}{97, 48} & \textcolor{linkcolor}{164, 122} \\
& TGAT & \textcolor{seqcolor}{24, 12} & \textcolor{pascalcolor}{24, 12} & \textcolor{linkcolor}{33, 23} \\
 & TGN & \textcolor{seqcolor}{19, 9} & \textcolor{pascalcolor}{21, 10} & \textcolor{linkcolor}{33, 20} \\
& CTAN & \textcolor{seqcolor}{53, 26} & \textcolor{pascalcolor}{74, 37} & \textcolor{linkcolor}{128, 96} \\

\bottomrule
\end{tabular}
\end{table}
\begin{table}[h]
\centering
\caption{The grid of hyperparameters employed during model selection for CTAN on the Dynamic Link Property Prediction task on the three TGB benchmark datasets considered: \textcolor{seqcolor}{tgbl-wiki-v2}, \textcolor{pascalcolor}{tgbl-review-v2}, \textcolor{linkcolor}{tgbl-coin-v2}, \textcolor{comcolor}{tgbl-comment}. For tgbl-wiki-v2 we conducted five runs with different random seeds for different weight initializations for each configuration, whereas for the other datasets we conducted three different runs.
The rest of the training configuration is taken from the TGB codebase: batch size is 200, weight decay penalty was 0, the optimized metric is Mean Reciprocal Rank and is evaluated with the TGB evaluator.}
\label{tab:hyper_param_tgb}
\scriptsize
\begin{tabular}{l|cccc}
\toprule
\multirow{2}{*}{\textbf{Hyperparameters}} & \multicolumn{4}{c}{\textbf{Values}}\\\cmidrule{2-5}
& \textcolor{seqcolor}{\textbf{tgbl-wiki-v2}} & \textcolor{pascalcolor}{\textbf{tgbl-review-v2}} & \textcolor{linkcolor}{\textbf{tgbl-coin-v2}} & \textcolor{comcolor}{\textbf{tgbl-comment}} \\\midrule
optimizer & \multicolumn{4}{c}{Adam} \\
$\sigma$ & \multicolumn{4}{c}{tanh} \\
$\gamma$ & \textcolor{seqcolor}{0.1} & \textcolor{pascalcolor}{0.1, 0.01} & \textcolor{linkcolor}{0.1, 0.01} & \textcolor{comcolor}{0.1} \\
$\eta$ & \multicolumn{4}{c}{concat, $\eta = \text{tanh}(\textbf{h}^{i-1}(t_e)||\mathbf{x}(i))$} \\
n. GCLs & \textcolor{seqcolor}{1, 2, 3} & \textcolor{pascalcolor}{1,2} & \textcolor{linkcolor}{1} & \textcolor{comcolor}{1}\\
$\epsilon$ & \textcolor{seqcolor}{1.0} & \textcolor{pascalcolor}{0.5, 1.0} & \textcolor{linkcolor}{0.5, 1.0} & \textcolor{comcolor}{1.0} \\
embedding dim & \multicolumn{4}{c}{256} \\
sampler size & \multicolumn{4}{c}{32} \\
learning rate &  \textcolor{seqcolor}{$10^{-3}$, $10^{-4}$,}  & \textcolor{pascalcolor}{$3\cdot10^{-6}$} & \textcolor{linkcolor}{$10^{-4}$} & \textcolor{comcolor}{$10^{-4}$, $3\cdot10^{-4}$,}   \\
& \textcolor{seqcolor}{$3\cdot10^{-4}$, $3\cdot10^{-5}$} & & & \textcolor{comcolor}{$10^{-5}$, $3\cdot10^{-5}$}\\
epochs & \textcolor{seqcolor}{200} & \textcolor{pascalcolor}{50} & \textcolor{linkcolor}{50} & \textcolor{comcolor}{50} \\
LR scheduler patience & \textcolor{seqcolor}{20} & \textcolor{pascalcolor}{3} & \textcolor{linkcolor}{3} & \textcolor{comcolor}{3} \\
\bottomrule
\end{tabular}
\end{table}

\chapter{Additional contribution}\label{ch:hmm4g}\label{app:suppl_ch8}
\section{Hidden Markov Models for dynamic graphs}\label{sec:hmm4g}
In Chapter~\ref{ch:learning_dyn_graphs}, we presented literature approaches for learning dynamic graphs. An attentive reader may have noticed that most of these approaches are confined to the class of neural networks, thus making probabilistic methods for dynamic graphs widely understudied despite their potential.
For instance, a probabilistic model can easily capture the multimodality of the data distribution, which is useful in the context of stochastic processes \citep{errica_graph_2021}. Additionally, they can deal with missing data and exploit large amounts of unlabeled data to build rich unsupervised embeddings \citep{bacciu_probabilistic_2020}. 
Therefore, in this chapter, we propose the \emph{Hidden Markov Model for Dynamic Graphs} (\gls*{HMM4G})\index{Hidden Markov Model for Dynamic Graphs}, a deep and purely probabilistic model for sequences of graph snapshots. 
HMM4G extends hidden Markov models for sequences to the D-TDG domain by stacking probabilistic layers that perform efficient message passing and learn representations for the individual nodes.

The key contributions of this chapter can be summarized as follows:
\begin{itemize}
    \item We introduce a new probabilistic framework for learning D-TDGs, named HMM4G. Our method combines the sequential processing of HMMs with message passing
to deal with topologically varying structures over time.
    \item We conduct experiments to demonstrate the benefits of our method on temporal node prediction tasks, showing competitive performance with neural network counterparts.
\end{itemize}

This contribution has been developed in collaboration with NEC Laboratories Europe, Heidelberg, Germany.
We base this chapter on \cite{gravina_hmm4g}.

\subsection{Basic Concepts of Probability, HMM, and IO-HMM}\label{sec:background_prob}
Before delving into the description of how to deal with D-TDG representation learning in a purely probabilistic fashion, we provide basic concepts that will be used in the following section. The reader can refer to \cite{bishop}, \cite{Bruni2017}, and \cite{pml1Book} for a complete treatment of this topic. 

We start by introducing basic concepts of probability theory. Consider the set of all possible outcomes of an experiment with the symbol $\Omega$, and the set of events of interest that may occur as $\mathscr{A}\subseteq \mathscr{P}(\Omega)$, where $\mathscr{P}(\cdot)$ represent the powerset operator. A \textbf{probability}\index{probability} is a function $P: \mathscr{A} \rightarrow [0,+\infty]$, such that 
\begin{enumerate}
    \item $P(\emptyset)=0$;
    \item $P(\Omega)=1$;
    \item $\forall A\in\mathscr{A} \Rightarrow P(A) \geq 0$;
    \item for any countable collection $\{A_i\}_{i\in\mathbb{N}} \subseteq \mathscr{A}$ of disjoints sets it holds that $P(\bigcup_{i\in\mathbb{N}}A_i)=\sum_{i\in\mathbb{N}}P(A_i)$.
\end{enumerate}
To better understand these concepts, let's consider the example of a coin toss. In such a scenario, $\Omega=\{\text{head, tail}\}$ and $\mathscr{A} = \{\{\emptyset\}, \{\text{head}\}, \{\text{tail}\}, \{\text{head, tail}\}\}$. In a fair coin toss, the probabilities are $P(\{\emptyset\})=0$, $P(\{\text{head}\})=1/2$, $P(\{\text{tail}\})=1/2$, and $P(\{\text{head, tail}\})=1$.

Another useful concept is that of \textbf{random variables}\index{random variable}. A random variable is a function describing the outcome of a random process by assigning unique values to all possible outcomes, \ie $X:\Omega\rightarrow E$ such that 
$\{\omega\in\Omega\,|\,X(\omega)\in E\}\in\mathscr{A}$. 
The values associated to a random variable are called \emph{states}.
Depending on the definition of the image $E$, we can distinguish between \emph{discrete} and \emph{continuous} random variables. 
Recalling the coin toss example, we can define a discrete random variable such that $P(X=\text{head})=1/2$ and $P(X=\text{tail})=1/2$. In the following, we refer to $P(x)$ instead of $P(X=x)$ for brevity.

In many scenarios, we seek to determine the probability of multiple events occurring simultaneously. These scenarios can be formalized by employing a set of random variables, each representing the occurrence of a specific event. Therefore, we can define the \textbf{joint probability distribution} $P(x_1, \dots, x_n)$ to represent this particular case. If the $n$ random variables are \textbf{mutually independent}, then it is true that $P(x_1, \dots, x_n)=\prod_{i=1}^nP(x_i)$. On the contrary, when an event $Y=y$ has an effect on the other random variables, then we talk about \textbf{conditional probabilities}, $P(x_1, \dots, x_n|y)$. In this case, random variables can be \textbf{conditionally independent} if $P(x_1, \dots, x_n|y)=\prod_{i=1}^nP(x_i|y)$. 
Applying the \textbf{Bayes' theorem}\index{Bayes' theorem} the conditional probability $P(x|y)$ (also referred to as \textbf{posterior} probability) can be equated to
\begin{equation}
    P(x|y)=\frac{P(y|x)P(y)}{P(x)}
\end{equation}
where $P(y)$ is the \textbf{prior} probability; $P(y|x)$ is the conditional probability of $Y=y$ given $X=x$ (and it is also referred to as \textbf{likelihood}); and $P(x)$ is the \textbf{marginal} probability.

We now turn to define the \textbf{distribution}\index{random variable!{distribution}}\index{random variable distribution|see {random variable}} of a random variable, which is the probability measure on the set of all possible states of the random variable. To provide the reader with a better understanding, we introduce the categorical distribution and the Dirichlet distribution.

The \textbf{categorical distribution}\index{random variable distribution!{categorical}}\index{categorical distribution |see {random variable distribution}} is a discrete probability distribution defined over a finite set of $C$ values. It is typically used to describe the possible results of a discrete random variable that can take on one of $C$ possible categories, which is usually represented as a real vector of size $C$ with entries that sum to 1. More formally, 
\begin{equation}
    P(X=i) = p_i, \; \text{for}\,\, i=1, \dots, C. 
\end{equation}

The \textbf{Dirichlet distribution}\index{random variable distribution!{Dirichlet}}\index{Dirichlet distribution |see {random variable distribution}} is a continuous multivariate probability distribution parameterized by a vector of $C$ positive reals, \ie $\bm{\alpha}\in\mathbb{R}^C$. The Dirichlet distribution is defined as
\begin{equation}
\begin{split}
    &P(x_1, \dots, x_C| \alpha_1, \dots, \alpha_C) = \frac{1}{B({\bm\alpha})} \prod_{i=1}^C x_i^{\alpha_i-1} \\
    &\text{where}\; B({\bm\alpha}) = \frac{\prod_{i=1}^C \Gamma(\alpha_i)}{\Gamma(\sum_{i=1}^C \alpha_i)} \;\,\text{and}\;\, \Gamma(\alpha_i)=(\alpha_i-1)!.
\end{split}
\end{equation}

\begin{figure}
    \centering
    \begin{subfigure}{0.45\textwidth}
            \centering \includegraphics[width=0.24\linewidth]{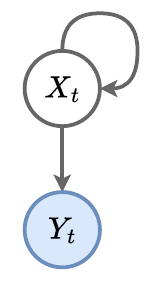}
            \caption{}
        \end{subfigure}
        \begin{subfigure}{0.45\textwidth}
            \centering \includegraphics[width=\linewidth]{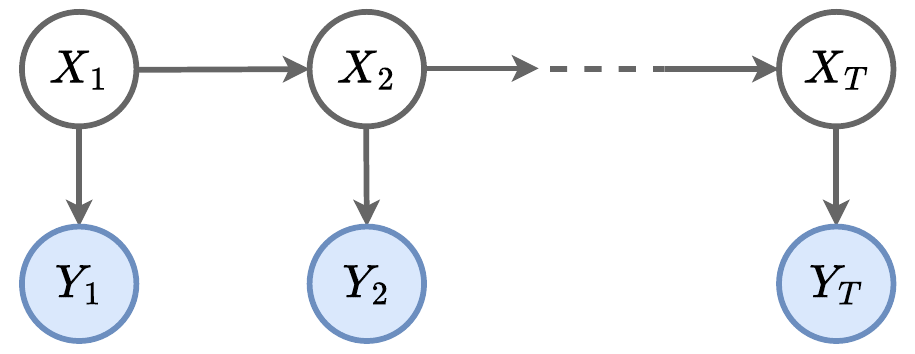}
            \caption{}
        \end{subfigure}
    \caption{Graphical model of an HMM, with its
    recurrent view (a) and its unfolded view (b).}
    \label{fig:hmm}
\end{figure}

We now move our focus to the \textbf{hidden Markov model} (HMM)\index{hidden Markov model}, which is a probabilistic model used to represent systems that transition between a series of latent (hidden) states over time. 
A HMM is specified by a random variable that recursively encode information 
over a set of $C$ discrete latent states, where the transition at time $t$ from one state ($X_t=i$) to another ($X_{t+1}=j$) is governed by the \textbf{transition probability matrix}\index{transition probability matrix}, where each element $(i,j)$ defines the likelihood of transitioning from state $i$ to $j$, \ie $P(X_{t+1}=j|X_t=i)$. The initial latent state $X_1$ is obtained by an \textbf{initial probability distribution}\index{initial probability distribution} that defines the probability that the HMM starts in state $i$, \ie $P(X_1=i)$. 
The specification of the probabilistic model is completed by defining the observations $Y_1, \dots, Y_t$, which are observed signals depended on the states. Observations are governed by an \textbf{emission probability}\index{emission probability}, which identify the likelihood of an observation being generated from a particular state, \ie $P(y_t|x_t, \theta)$ with $\theta$ a set of parameters governing the distribution, and $y_t$ and $x_t$ the realizations of $Y_t$ and $X_t$, respectively. Figure~\ref{fig:hmm} visually represent an HMM. 

The main task in HMM is to compute the joint probability of hidden states and observations, \ie
\begin{equation}
    P(Y|X) = \prod_{t=1}^T P(x_t|x_{t-1})P(y_t|x_t).
\end{equation}

Input-Output HMM (IO-HMM) for sequences \citep{iohmm} extends the HMM framework to support recurrent networks processing style, thus allowing input states to further drive the transition among states. Therefore, IO-HMM can be used to learn to map input sequences to output sequences.
Since states are influenced by the inputs, we can rewrite the transition distribution as 
\begin{equation}
    P(x_{t+1}|x_t, u_{1:t+1}) = \sum_{j=1}^C P(x_{t+1}|x_{t}, u_t)P(y_{t}|u_{1:t})
\end{equation}
where $C$ is the number of states, and $u_{1:t}$ is the subsequence of inputs from time 1 to $t$ and $u_t$ is the $t$-th input.
\subsection{The HMM4G Model}\label{sec:hmm4g_model}
In order to facilitate differentiation from neural architectures, we introduce a slightly modified notation and symbols. We introduce random variable (r.v.) $X_u^t$ with realization $\bm{x}_u$ to model the distribution of node $u$ attributes at time-step $t$. Similarly, we model the latent state of a node $u$ at time-step $t$ with a categorical r.v. $Q_u^t$ with $C$ possible states and discrete realization $q_u^t$. The posterior distribution of $Q_u^t$ conditioned on the evidence is another categorical distribution, and we refer to its parametrization with a vector $\bm{h}_u^t \in \mathbb{R}^C$ belonging to the $C$-$1$-simplex. Generally speaking, such a parametrization can be seen as the realization of a Dirichlet distribution of order $C$ that we denote with the letter $H_u^t$. \\

\begin{figure}[h]
    \centering
    \includegraphics[width=0.5\columnwidth]{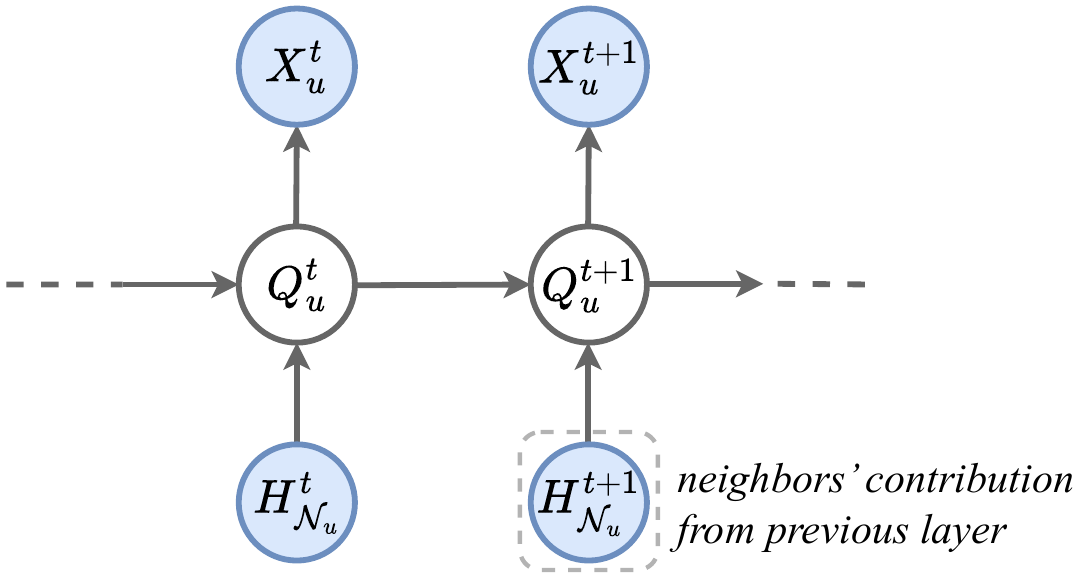}
    \caption{Graphical model of HMM4G at layer $\ell$ for node $u$ in the graph. Observed r.v.s (values estimated at layer $\ell-1$) are in blue and latent ones in white.}
    \label{fig:hmm4g-layer}
\end{figure}

In HMM4G we mirror the same message passing mechanism of DGNs for D-TDGs by stacking layers of temporal graph convolutions on top of each other. The key difference is that we implement each layer as a special case of an IO-HMM for sequences 
(introduced in Section~\ref{sec:background_prob}). We present HMM4G's graphical model for a generic node $u$ and layer $\ell$ in Figure \ref{fig:hmm4g-layer}, abstracting from the layer to ease the exposition. Compared to an HMM, in a classical IO-HMM the prior distribution of the latent variable $Q^t$ is replaced by the conditional distribution of $Q^t$ given the input evidence at time-step $t$. Similarly, in a generic layer $\ell$ of HMM4G, we define a similar conditional distribution that takes into account the ``messages'' of the neighbors of $u$ computed at a previous layer $\ell-1$.

Formally, the input evidence for node $u$ at time $t$ is
modeled by a Dirichlet r.v. $H^t_{\mathcal{N}_u}$ of order $C$, whose realization $\bm{h}^t_{\mathcal{N}_u} \in \mathbb{R}^C$ is computed as follows:
\begin{align}
    \bm{h}^t_{\mathcal{N}_u} = \frac{1}{|\mathcal{N}_u|}\sum_{v \in \mathcal{N}_u} \bm{h}^t_v,
\end{align}
where we have aggregated the parameters of the posterior distributions of the neighbors computed at the previous layer. This is akin to what happens in deep graph neural networks, where the latent representations of neighboring nodes are combined by a permutation invariant function. We use $\bm{h}^t_{\mathcal{N}_u} $ to explicitly parametrize the (categorical) transition distribution for node $u$ at layer $\ell$:
\begin{equation}
    P_{\bm{\theta}}(Q_u^t=i\mid Q_u^{t-1}=q_u^{t-1},H^t_{\mathcal{N}_u} = \bm{h}^t_{\mathcal{N}_u}) = \sum_{j=1}^{C}P_{\bm{\theta}_j}(Q_u^{t}=i\mid q_u^{t-1})\bm{h}^t_{\mathcal{N}_u}(j),
\end{equation}
where $\bm{\theta}=(\bm{\theta}_1,\dots,\bm{\theta}_C)$ are the transition parameters to be learned and $\bm{h}^t_{\mathcal{N}_u}(j)$ denotes the $j$-th component of a vector. The parameters $\bm{\theta}$ are \emph{shared} across all nodes to generalize to unseen graphs of arbitrary topology. The only other distribution of the model, that is $P(X_u^t|Q_u^t)$, is learned as in standard IO-HMMs and also shared across all nodes, e.g., a Gaussian for continuous attributes.

\begin{figure}[h]
    \centering
    \includegraphics[width=0.6\textwidth]{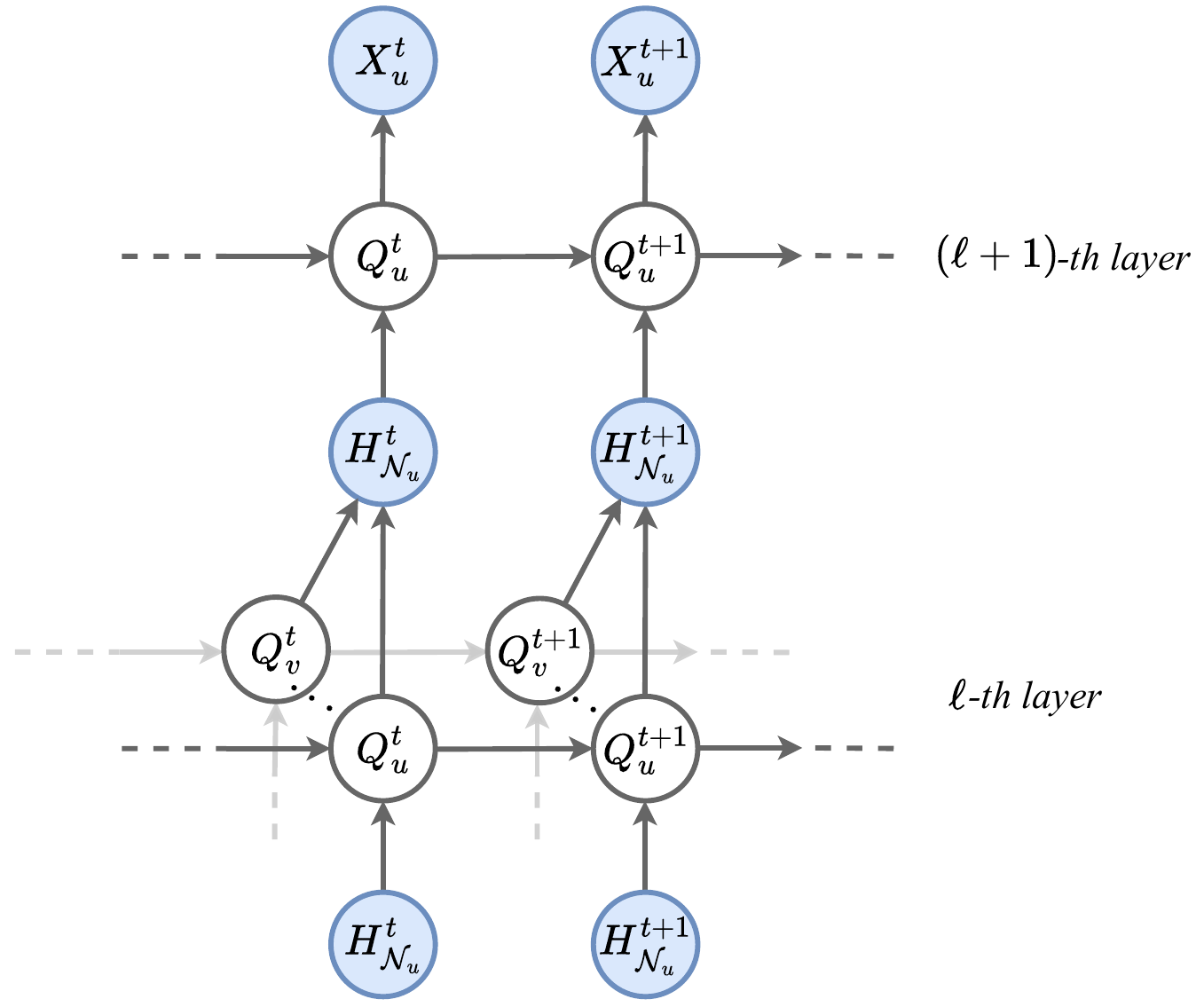}
    \caption{Computation process for node $u$ in two consecutive layers of our HMM4G. In layer $\ell$, Expectation Maximization is applied to infer the parameters of the posterior distribution of the variable $Q_u^t$ for all nodes and time-steps in the graph sequence. This information is then used to compute $\bm{h}^t_{\mathcal{N}u}$ in the subsequent layer $\ell+1$. Finally, the computed $\bm{h}^t{\mathcal{N}_u}$ is used to transition to the new state.}
    \label{fig:hmm4g-2layers}
\end{figure}

At each layer, due to the presence of cycles in the graphs, we break the mutual dependencies between the node variables as a product of conditional probabilities, and we maximize the following pseudo-log-likelihood with respect to the parameters $\bm{\Theta}$:
\begin{align}
    \log \prod_{u \in \mathcal{V}_g} P_{\bm{\Theta}}(X^1_u,\dots,X^T_u | \bm{h}^1_{\mathcal{N}_u},\dots,\bm{h}^T_{\mathcal{N}_u}).
\end{align}
Therefore, sequences of node attributes can be processed in parallel as it happens for temporal deep graph networks, meaning that the inference phase has the same linear complexity in the number of edges when processing the graph. We train HMM4G incrementally: we apply Expectation Maximization~\citep{bishop} to layer $\ell$, and we infer the parameters of the posterior distribution of the variable $Q_u^t$ for all nodes and time-steps in the graph sequence. 
We use this information to compute $\bm{h}^t_{\mathcal{N}_u}$ in the subsequent layer $\ell+1$. Figure~\ref{fig:hmm4g-2layers} shows how information is computed for node $u$ in two consecutive layers of our HMM4G.

When $\ell=0$, the layer reduces to an HMM. We can compute closed-form update equations for the M-step by extending the classical HMM derivation; we do not show them here in the interest of space. The final latent representation of each node $u$ at time $t$, which is eventually used to make predictions about the nodes, is the concatenation of the realizations $\bm{h}_u^t$ across the layers of the architecture. In particular, we apply the same \emph{unibigram} technique of \cite{bacciu_probabilistic_2020} to the learned representations, which modifies each $\bm{h}_u^t$ of layer $\ell$ to take into account some neighboring statistics as well. Indeed, it was shown that unibigrams improve the quality of the learned representation for static graphs.
Training complexity is also similar to other literature models in Section~\ref{sec:D_TDG}
. 

We could extend the model to account for discrete edge types and multiple prior layers using the techniques in \cite{bacciu_probabilistic_2020}, but we leave these extensions to future works. In addition, we do not assume a static graph structure over time, \ie limiting our model to spatio-temporal graphs only,  but rather we can let both the nodes' attributes and their interactions freely change. 
\subsection{Experiments}\label{sec:hmm4g_experiments}
In this section, we provide an empirical assessment of our method against related DGN models for dynamic graphs from the literature. Specifically, we consider the task of temporal node regression (see Section~\ref{sec:benchmark} for more details). 
We release the code implementing our methodology and reproducing our analysis at \href{https://github.com/nec-research/hidden_markov_model_temporal_graphs}{github.com/nec-research/hidden\_markov\_model\_temporal\_graphs}.

\subsubsection{Temporal Node Regression Tasks}
\myparagraph{Setup}
We consider four known graph datasets for node regression in the dynamic graph domain
:
\begin{itemize}
    \item \textbf{Twitter Tennis}~\citep{twitter_tennis}: Twitter mention graphs of
major tennis tournaments from 2017. Nodes are Twitter accounts and edges are mentions between them. Node features encode the number of mentions received and other structural properties. The target is the number of received mentions. 

\item \textbf{Chickenpox}~\citep{chickenpox}: this dataset comprises county-level chicken pox cases in Hungary between 2004 and 2014. Nodes represent counties, and edges denote neighboring relationships. Node features are lagged weekly counts of the chickenpox cases. 

\item \textbf{Pedalme}~\citep{pyg_temporal}:  this dataset consists of the number of weekly
bicycle package deliveries by Pedal Me in London between 2020 and 2021. Nodes are localities and edges are spatial connections. Node features are lagged weekly counts of the delivery demands. 

\item \textbf{Wikimath}~\citep{pyg_temporal}: this dataset comprises the number of daily visits of Wikipedia pages between 2019 and 2021. The nodes represent Wikipedia pages about popular mathematics topics, and edges describe the links between the pages. Node features are the daily visit counts.
\end{itemize}

In Twitter Tennis the underlying topology is dynamic, \ie changes over time. In contrast, the remaining three datasets have a static underlying topology in which only node features change over time. The task consists of predicting future node labels given the previous evolution of the graph (also known as graph snapshots). 
We report in Table~\ref{tab:hmm4g_data_stats} the statistics of the employed datasets.
\begin{table}[ht]
\centering
\caption{Statistics of the datasets used in our experiments.}\label{tab:hmm4g_data_stats}

\scriptsize
\setlength{\tabcolsep}{4pt} 
\begin{tabular}{lcccc}
\toprule
           & \textbf{\# Nodes} & \textbf{Seq. len.}  & \textbf{Frequencey} &  \textbf{Split}  \\\midrule
Chickenpox & 20                & 522                 & Weekly & 80/10/10 \\
Pedalme    & 15                & 30                  & Weekly & 80/10/10 \\
Wikimath   & 1,068             & 731                 & Daily  & 80/10/10 \\
Twitter tennis    & 1,000       & 120                 & Hourly & 80/10/10 \\
\bottomrule
\end{tabular}
\end{table}


We compare our method against 11 state-of-the-art DGNs for D-TDGs: DCRNN~\citep{DCRNN}, GCRN-GRU~\citep{GCRN}, GCRN-LSTM~\citep{GCRN}, GCLSTM~\citep{GC-LSTM}, DyGrAE~\citep{dygrae}, EvolveGCN-H~\citep{egcn}, EvolveGCN-O\citep{egcn}, A3TGCN~\citep{a3tgcn}, TGCN~\citep{T-GCN}, MPNN LSTM~\citep{mpnn_lsltm}, and DynGESN~\citep{dyngesn}. Such baselines differ in the learning strategy, attention mechanisms, and employed temporal and graph neural network layers (as detailed in Section~\ref{sec:D_TDG}).

Each model is designed as a combination of two main components. The first is the recurrent graph encoder which maps each node's input features into a latent representation. The second is the readout, which maps the output of the first component into the output space. The readout is a  Multi-Layer Perceptron for almost all models in the experiments (DynGESN uses ridge regression). For each timestamp of the sequence, we first obtain the latent node representations of the corresponding graph snapshot using the recurrent encoder, and then we feed them into the readout to obtain a prediction for each node. 

We leverage the same experimental setting and data splits reported in \cite{dyngesn}. Specifically, we performed hyperparameter tuning via grid search, optimizing the Mean Square Error (MSE). We train using the Adam optimizer for a maximum of 1000 epochs. We employ an early stopping criterion that stops the training if the validation error does not decrease for 100 epochs. We report in Table~\ref{tab:hmm4g_hyper-params} 
the grid of hyperparameters explored in our experiments.
\begin{table}[h]
\centering
\caption{The grid of hyperparameters employed during model selection for our experiments.} \label{tab:hmm4g_hyper-params}

\scriptsize

\begin{tabular}{l|cc}
\toprule
\textbf{Hyperparameters} & \textbf{HMM4G} & \textbf{Readout}\\\midrule
n. layers & 1, 2, 3, 4, 5 & 1, 2, 3 \\
C & 5, 10 & -- \\
epochs & 10, 20, 40 & 1000\\ 
optimizer & -- & Adam \\ 
learning rate & -- &  $10^{-3}$, $10^{-2}$ \\
weight decay & -- & 0, 0.0005, 0.005 \\
embedding dim & -- & $2^2, 2^3, 2^5, 2^6, 2^7$\\

\bottomrule
\end{tabular}
\end{table}


\myparagraph{Results}
We present the MSE test results of our experiments in Table~\ref{tab:hmm4g_results}. The first observation is that HMM4G has promising performances compared to the baselines employed in the experiments, ranking first or second in three out of four tasks. Indeed, HMM4G achieves an error score that is on average 16\% better than the other baselines. The larger gain is achieved on Tennis and Wikimath datasets, where HMM4G is on average 39\% and 28\% better than the baselines, respectively. It is worth noting that Tennis and Wikimath datasets are more challenging than the other tasks in our experiments, as they contain two orders of magnitude more nodes than the others.
Moreover, in the Twitter task, models also have to capture the evolution of temporal edges. This highlight that our method learns representative embeddings even in challenging scenarios.
In general, our method performs comparably with DynGESN. The worst performance is achieved on the Pedalme dataset, which consists of only 36 timestamps and 15 nodes, making it the smallest temporal graph in our experiments. The amount of nodes is so small that the initialization of the Gaussian emission distributions is problematic (probably connected to the high standard deviation), meaning there is not enough incentive for the model to differentiate the embeddings based on the structural information when maximizing the likelihood. On the other hand, HMM4G achieves the best performance on Wikimath, which is also the largest of the datasets considered, and it improves over DynGESN and the other baselines by a large margin. This suggests that, with enough data, our probabilistic model can learn good representations of the temporal graph dynamics.


\begin{table}[t]
\renewcommand{\arraystretch}{1.2}
\centering
\caption{Test MSE with standard deviation averaged over 10 final runs. \one{First}, \two{second}, and \three{third} best results for each task are color-coded. Baselines taken from \cite{dyngesn}.} \label{tab:hmm4g_results}

\scriptsize
\begin{tabular}{lcccc}
\toprule
\textbf{Model} & \textbf{Chickenpox} & \textbf{Tennis} & \textbf{Pedalme} & \textbf{Wikimath}\\
\midrule
\textbf{Baseline} \\
$\,$ Mean baseline   & 1.117                      & 0.482                      & 1.484                       & 0.843\\
$\,$ Linear baseline & \three{0.952}              & \three{0.356}              & 1.499                       & 0.663\\
\midrule
\multicolumn{4}{l}{\textbf{DGN for D-TDGs}} \\
$\,$ DCRNN           & 1.097$_{\pm 0.006}$        & 0.478$_{\pm 0.004}$        & 1.454$_{\pm 0.050}$         & 0.679$_{\pm 0.007}$\\
$\,$ GCRN-GRU        & 1.103$_{\pm 0.004}$        & 0.477$_{\pm 0.007}$        & \one{1.420$_{\pm 0.054}$}   & 0.680$_{\pm 0.021}$\\
$\,$ GCRN-LSTM       & 1.097$_{\pm 0.006}$        & 0.477$_{\pm 0.006}$        & 1.453$_{\pm 0.085}$         & 0.678$_{\pm 0.008}$\\
$\,$ GCLSTM         & 1.095$_{\pm 0.005}$        & 0.475$_{\pm 0.010}$        & 1.490$_{\pm 0.088}$         & 0.677$_{\pm 0.009}$\\
$\,$ DyGrAE          & 1.102$_{\pm 0.013}$        & 0.480$_{\pm 0.005}$        & \two{1.426$_{\pm 0.089}$}   & 0.621$_{\pm 0.012}$\\
$\,$ EvolveGCN-H          & 1.137$_{\pm 0.026}$        & 0.481$_{\pm 0.003}$        & \three{1.446$_{\pm 0.168}$} & 0.779$_{\pm 0.031}$\\
$\,$ EvolveGCN-O          & 1.135$_{\pm 0.011}$        & 0.484$_{\pm 0.002}$        & 1.469$_{\pm 0.137}$         & 0.807$_{\pm 0.047}$\\
$\,$ A3TGCN         & 1.078$_{\pm 0.009}$        & 0.477$_{\pm 0.005}$        & 1.494$_{\pm 0.049}$         & 0.618$_{\pm 0.008}$\\
$\,$ TGCN           & 1.083$_{\pm 0.011}$        & 0.478$_{\pm 0.004}$        & 1.515$_{\pm 0.059}$         & 0\three{.616$_{\pm 0.011}$}\\
$\,$ MPNN LSTM       & 1.125$_{\pm 0.005}$        & 0.482$_{\pm 0.001}$        & 1.580$_{\pm 0.102}$         & 0.856$_{\pm 0.021}$\\
$\,$ DynGESN         & \one{0.907$_{\pm 0.007}$}  & \one{0.300$_{\pm 0.003}$}  & 1.528$_{\pm0.063}$          & \two{0.610$_{\pm 0.003}$}\\
\midrule
\multicolumn{4}{l}{\textbf{Our}} \\
$\,$ HMM4G           & \two{0.939$_{\pm 0.013}$}  & \two{0.333$_{\pm 0.004}$}  & 1.769$_{\pm0.370}$          & \one{0.542$_{\pm 0.008}$} \\ 
\bottomrule
\end{tabular}
\end{table}

\subsection{Related Work}
HMM4G is inspired by two different lines of research. The first is the one of deep probabilistic models for static graphs \citep{bacciu_probabilistic_2020} that rely on message passing mechanisms \citep{MPNN}. These methods are trained incrementally, one layer after another, and the depth of the architecture is functional to the spreading of messages between nodes of the graph. The second is the one of dynamic graph representation learning (see Chapter~\ref{ch:learning_dyn_graphs}), which develops ad-hoc approaches to deal with the different technical and methodological challenges that the temporal extension, such as the varying topology of graphs across time, the sudden (dis)appearance of nodes, and the memory capacity of temporal models. HMM4G lies at the intersection between these two fields, by proposing a purely probabilistic method for graph representation learning. It also profoundly differs from \cite{kayaalp_hidden_2022}, where a multi-agent filtering algorithm is proposed to determine an underlying state of the graph, but it is orthogonal to the topic of graph representation learning.
\subsection{Summary}
We introduced a new probabilistic framework for learning D-TDGs. Our method combines the sequential processing of HMMs with message passing to deal with topologically varying structures over time. We showed how the learned representations are useful for temporal node prediction tasks, especially on larger datasets. We believe that our contribution is one of the first attempts at bridging the gap between probabilistic models and dynamic graph learning.

\newpage
\printindex

\end{document}